\documentclass[13pt,a4paper,twoside,openright]{authesis}

\input{Preamble/pkg_and_fmt}
\graphicspath{{./Images/}} 
\DeclareGraphicsExtensions{.pdf} 

\author{Ebenezer R. H. P. Isaac}
\fullname{Ebenezer Rajiv Hinston Prabhu I}
\title{Robust Analytics for Video-Based Gait Biometrics}
\date{February 2018}

\begin{document}


\pagestyle{fancy}
\fancyhf{}
\renewcommand{\headrulewidth}{0pt}
\renewcommand{\footrulewidth}{0pt}
\rhead{\thepage}
\rhead{}

\makeatletter
\hypersetup{
  pdfauthor={\@author},
  pdftitle={\@title},
  pdfsubject={PhD Thesis}
}
\let\Title\@title
\let\Author\@author
\let\Date\@date
\let\Fullname\@fullname
\makeatother

\begin{center}
  \LARGE
  \textbf{\MakeUppercase{\Title}} \\
  \vspace{1.5\baselineskip}
  \bigsize{\textbf{A THESIS}}\\
 \vspace{1\baselineskip}
  \normalsize{\textit{\textbf{Submitted by}}}\\
  \vspace{0.6\baselineskip}
  {\Large \textbf{\MakeUppercase{\Fullname}}}\\
  \vspace{1.6\baselineskip}
  \normalsize{\textit{\textbf{in partial fulfillment of the requirements for the degree of}}}\\
\vspace{.3\baselineskip}
  \bigsize{{\textbf{DOCTOR OF PHILOSOPHY}}}\\
\end{center}
\vspace{1\baselineskip}
  \begin{center}
   \includegraphics[width=26mm,height=25mm]{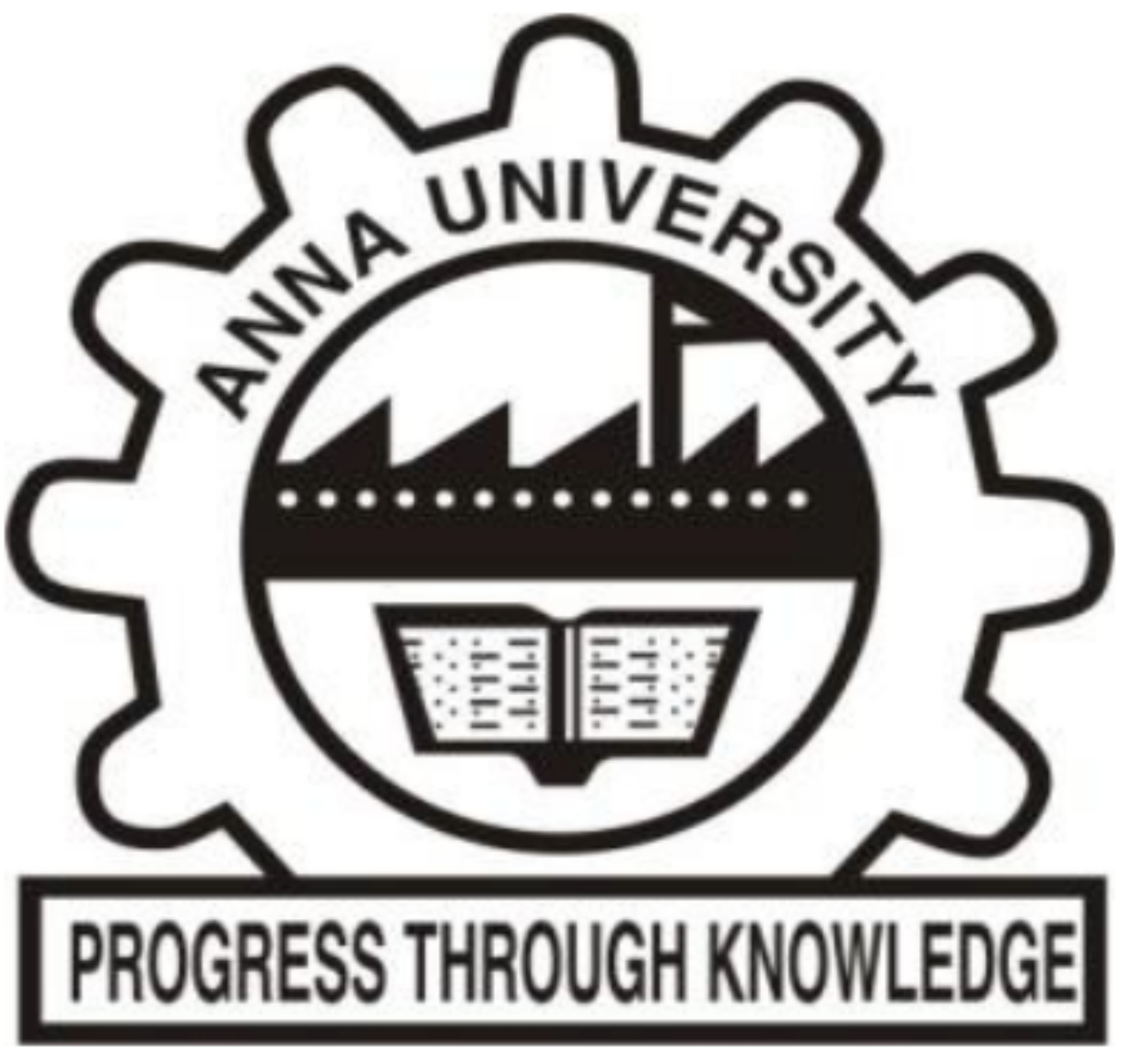}   \\
  \vspace{.3\baselineskip}
  \bigsize{\textbf{FACULTY OF INFORMATION AND }}\\
  \bigsize{\textbf{COMMUNICATION ENGINEERING}}\\
  \bigsize{\textbf{ANNA UNIVERSITY}}\\
  \bigsize{\textbf{CHENNAI  600 025}}\\ 
  \vspace{0.8\baselineskip}
  \large{\textbf{\MakeUppercase{\Date}}} 
 \end{center}
 \pagebreak

 \setboolean{@twoside}{false}
 
\rhead{\thepage}
\chapter*{ANNA UNIVERSITY\\
CHENNAI  600 025\\
\vspace{0.6\baselineskip}
CERTIFICATE}
\newlength{\aulength}
  \settowidth{\aulength}{Anna University
 Chennai}
\newlength{\datewidth}
\settowidth{\datewidth}{Chennai 600 025}
\begin{spacing}{1.8}
  \begin{sloppypar}
  \fontsize{13}{14.6}\selectfont 
    The research work embodied in the present
    thesis entitled ``\textbf{\MakeUppercase{\Title}}'' has been carried out in the
    Department of Computer Science and Engineering, Anna University,
    C.E.G. Campus, Chennai. The work reported herein is original and does not
    form part of any other thesis or dissertation on the basis of which a degree
    or award was conferred on an earlier occasion or to any other scholar.
    
    I understand the University's policy on plagiarism and declare that the
    thesis and publications are my own work, except where specifically
    acknowledged and has not been copied from other sources or been previously
    submitted for award or assessment.
  \end{sloppypar}
\end{spacing}

\newcommand*\leftright[2]{%
  \leavevmode
  \rlap{#1} %
  \hspace{0.6\linewidth}
  #2}
  \begin{flushleft}
    \leftright{\textbf{\MakeUppercase{\Author}}}{\textbf{K. S. EASWARAKUMAR}} \\
    \leftright{RESEARCH SCHOLAR}{SUPERVISOR} \\
    \leftright{}{Professor} \\
    \leftright{}{Department of Computer}\\
    \leftright{}{Science and Engineering} \\
    \leftright{}{Anna University} \\
    \leftright{}{Chennai 600025.} 
  \end{flushleft}


\DeclareRobustCommand*{\tocheading}{%
    {{\normalfont\large\bfseries\centering CHAPTER NO. \hfill TITLE \hfill PAGE
        NO.\endgraf\bigskip}}}

\DeclareRobustCommand*{\toctabheading}{%
  {{\normalfont\large\bfseries\centering TABLE NO. \hfill TITLE \hfill PAGE
      NO.\endgraf\bigskip}}}
		
\DeclareRobustCommand*{\tocfigheading}{%
  {{\normalfont\large\bfseries\centering FIGURE NO. \hfill TITLE \hfill PAGE
      NO.\endgraf\bigskip}}}

\addtocontents{toc}{\protect\flushleft\protect\bfseries~CHAPTER
  NO.\hfill~TITLE\hfill~PAGE NO.\endgraf}
\addtocontents{toc}{\protect\flushleft \protect\bfseries
 ~ \hfill ~ \hfill ~\endgraf}


\addtocontents{lof}{\protect\flushleft
\protect\bfseries FIGURE NO. \hfill
 TITLE \hfill  PAGE NO.\endgraf}%
\addtocontents{lot}{\protect\flushleft \protect\bfseries TABLE NO. \hfill
  TITLE \hfill  PAGE NO.\endgraf}%

\chapter*{ABSTRACT}
\addcontentsline{toc}{section}{\bfseries \uppercase{Abstract}}
\begin{spacing}{1.5}
  \begin{sloppypar}

    Gait analysis is the study of the systematic methods that assess and quantify animal locomotion. The research on gait analysis has considerably evolved through time. It was an ancient art, and it still finds its application today in modern science and medicine. Gait finds a unique importance among the many state-of-the-art biometric systems since it does not require the subject's cooperation to the extent required by other modalities. Hence by nature, it is an \textit{unobtrusive biometric}. Gait is associated with three types of signals, viz., kinetic, kinematic and EMG. Kinesiological EMG is mostly used for clinical purposes and kinetic measurement instruments are confined to a limited space. The applicability of kinetic and EMG data for the purpose of biometrics is thus constrained. Kinematic observation of gait is much more efficient and successful in literature. Also, the cost of kinematic measurement instruments such as a video camera and mobile accelerometer are much more cost effective than the apparatus required for EMG and kinetic observation.

    There are two aspects to biometrics: hard and soft. \textit{Soft biometrics} include determination of height, weight, gender, or ethnicity with gender recognition being the most common form associated with gait research. The method proposed in this thesis, \textit{Pose-Based Voting} employs a scheme which delineates the gait instance as a sequence of poses which is used to predict the gender of the respective subject through a voting scheme.

    \textit{Hard biometrics} associate people with their innate traits that identifies them. This data can be used either for identification or for authentication. Gait identification, widely known as gait recognition, is the process of mapping a given gait instance to a trained identity. On the other hand, gait authentication shows how closely the given gait instance matches to identity claimed. Recent advancements show that the more favourable biometric accuracy is attained through the use of gait templates.
A \textit{gait template} is a collation of silhouettes extracted from a gait video sequence. Many templates are proposed till date aimed to decrease the error of gait recognition. However, they still suffer from the effect of covariate factors such as variation in clothing and carrying conditions. Masking the templates to include only the covariate-resilient regions improve the overall performance of the recognition system. The proposed framework, the \textit{Genetic Template Segmentation} (GTS), automates the process of fitting the masking boundaries to select the regions that correspond to the optimum performance.

    The gait templates are also used for authentication that follows a similar procedure for feature extraction. The test instance and the stored gallery instance is separated by a distance in a hyperplane. In traditional gait authentication, if this Euclidean distance falls within a permissible threshold, then the test sequence is said to be authorized as claimed identity. However, Euclidean threshold-based methods trade off a set amount of False Accept Rate (FAR) for an acceptable False Reject Rate (FRR). This thesis proposes an intuitive technique called \textit{Multiperson Signature Mapping} (MSM) for gait authentication, which converts a gait recognition system to a powerful authentication system. The salient feature of this method allows to decrease the FAR in proportion to the system population while still attaining an FRR equal to the error rate of the base recognition system.

    Though the MSM framework enhances the performance of any existing gait authentication method, when the system population decreases, its greatest strength becomes its weakness. To circumvent this problem another framework, the \textit{Bayesian Thresholding}, is designed. This method measures the distance between the test and gallery instances in the form of posterior probabilities. This drastically decreases the overall error and can boost the power of any authentication system in general.
    
\end{sloppypar}
\end{spacing}

 \chapter*{ACKNOWLEDGEMENT}
\begin{sloppypar}
\begin{spacing}{1.5}
  I wish to record my deep sense of gratitude and profound thanks to my research
  supervisor \textbf{Dr. K. S. Easwarakumar} for his well-organized approach and
  inspiring guidance which helped me bring this thesis to fruition. I also thank
  \textbf{Dr. D. Manjula}, Professor and Head, Department of Computer Science
  and Engineering, Anna University, CEG Campus for the amenities provided to
  carry out my research. Further, I express my gratidude to my \textbf{Doctoral
    Committee} members for their valuable suggestions during the course of my
  research work.

  I like to convey my sincere thanks to my mentors \textbf{Dr.~Susan Elias}, Associate Professor,
  School of Electronics Engineering, VIT University, Chennai Campus, and
  \textbf{Dr. Srinivasan Rajagopalan}, Department of Physiology and Biomedical
  Engineering, Mayo Clinic, Rochester, Minnesota, USA, for providing a great
  support by spending their priceless time for the research discussions with me.
  
  I like to extend my thanks to the faculty and non-teaching staff members of
  the Department of Computer Science and Engineering, Anna University, CEG
  Campus, for their constant support during the whole of my time in the
  institution.

  I am extremely indebted to my parents for all their help and for their
  boundless encouragement and sacrifice throughout the course of my education.

  \vfill

\hfill \textbf{EBENEZER R. H. P. ISAAC}
\end{spacing}
\end{sloppypar}


\setlength\cftparskip{3pt}
\setlength{\cftbeforetoctitleskip}{-3em}
\setlength{\cftbeforelottitleskip}{-3em}
\setlength{\cftbeforeloftitleskip}{-3em}
\cftsetindents{chapter}{15 mm}{10 mm}
\cftsetindents{section}{25 mm}{10 mm}
\cftsetindents{subsection}{35 mm}{12 mm} 
\cftsetindents{subsubsection}{49 mm}{15 mm}
\cftsetindents{paragraph}{40 mm}{20 mm}
\cftsetindents{subparagraph}{50 mm} {20  mm}
\cftsetindents{table}{10 mm}{10 mm}
\cftsetindents{figure}{10 mm}{10 mm}


\newcommand*{\SetupNextPage}
  {\afterpage{\ifintoc\PrintChapterTocHeadline\fi}}
\begin{onehalfspacing}

  \begingroup
  \intoctrue
  \AtBeginShipout{\tocheading}
  \rhead{\thepage}
  \tableofcontents
  \AtBeginShipoutClear
  \endgroup
  \newpage 

  \addcontentsline{toc}{section}{\bfseries LIST OF TABLES}
  \rhead{\thepage}
  \AtBeginShipout{\toctabheading}
  \listoftables
  \AtBeginShipoutClear
  \clearpage

  \rhead{\thepage}
  \addcontentsline{toc}{section}{\bfseries
    LIST OF FIGURES}
  \AtBeginShipout{\tocfigheading}
  \listoffigures
  \AtBeginShipoutClear
\end{onehalfspacing}

\rhead{\thepage}
\fancyfoot[C]{\thepage}

\chapter*{LIST OF SYMBOLS AND ABBREVIATIONS}
\addcontentsline{toc}{section}{\bfseries LIST OF SYMBOLS AND ABBREVIATIONS}

\setlongtables
\begin{longtable}
  {>{\PBS\raggedright\hspace{0pt}}p{2.5cm}@{}%
   c%
   >{\PBS\raggedright\hspace{0pt}}p{11.5cm}@{}}


     \begin{filecontents*}{Data/Abbreviation.csv}
Key,Description
$\cap$,Intersection operator
$\cup$,Union operator
$\mu_k$,Mean vector of the $k^\text{th}$ class
$\pi$,Pi (approximately 3.14159)
$\Sigma_k$,Covariance matrix of the $k^\text{th}$ class
$\theta_d$,Distance threshold
$\theta_p$,Probability threshold
AEI,Active Energy Image
ASM,Active Shape Model
$B_t$,Binarized normalized silhouette at frame $t$
BT,Bayesian Thresholding
CASIA,Chinese Academy of Sciences -- Instrumentation and Automation sector
CCR,Correct Classification Rate
CCR$_\text{K}$,Correct Classification Rate of covariate K.
CDA,Canonical Discriminant Analysis
CMC,Cumulative Match Characteristic
CMS$_r$,Cumulative Match Score of the $r^\text{th}$ rank
CNN,Convolutional Neural Network
CNS,Central Nervous System
DARPA,Defense Advanced Research Projects Agency
DTW,Dynamic Time Warping
EER,Equal Error Rate
EFD,Elliptic Fourier Descriptors
EMG,Electromyography
FAR,False Acceptance Rate
FRR,False Rejection Rate
FSR,Force-Sensing Resistors
$G_\text{T}$,Gait template T, where T can be GEI, AEI, etc.
GA,Genetic Algorithm
GEI,Gait Energy Image
GEnI,Gait Entropy Image
GLM,Group Lasso Motion
GRF,Ground Reaction Force
GTS,Genetic Template Segmentation
$h$,Hypothesis of a genetic algorithm
HMM,Hidden Markov Model
IR,Infrared
$k$NN,$k$-Nearest Neighbour
KEMG,Kinesiological Electromyography
LDA,Linear Discriminant Analysis
$M_\text{GTS}$,Mask generated by GTS algorithm
MDA,Multiple Discriminant Analysis
MoCap,Motion Capture
MSM,Multiperson Signature Matching
NN,Nearest Neighbour
PCA,Principal Component Analysis
PDM,Point Distribution Model
PGR,Panoramic Gait Recognition
PNS,Peripheral Nervous System
Pr,Probability 
RCS,Row-Column Sum
RGB,Red Green Blue color coding
ROC,Receiver Operating Characteristic
SVM,Support Vector Machine
VI-MGR,View-Invariant Multiscale Gait Recognition
VPM,View Preserving Model
VTM,View Transformation Model
\end{filecontents*}
\csvreader[late after line=\\]{Data/Abbreviation.csv}
    {Key=\key, Description=\descr}
    {\key &-\hspace{1cm} &\descr}
 
  \end{longtable}

\clearpage
\newpage

\setboolean{@twoside}{true}


\titlespacing\section{0pt}{0pt plus 0pt minus 2pt}{0pt plus 0pt minus 2pt}
\titlespacing\subsection{0pt}{0pt plus 0pt minus 2pt}{0pt plus 0pt minus 2pt}
\titlespacing\subsubsection{0pt}{0pt plus 0pt minus 2pt}{0pt plus 0pt minus 2pt}

\pagenumbering{arabic}
\pagestyle{fancy}
\fancyhf{}
\renewcommand{\headrulewidth}{0pt}
\renewcommand{\footrulewidth}{0pt}

\rhead{\thepage}

\fancyhead{} 
\fancyhead[LE,RO]{\thepage} 


\rfoot{\includegraphics[width=13.7cm]{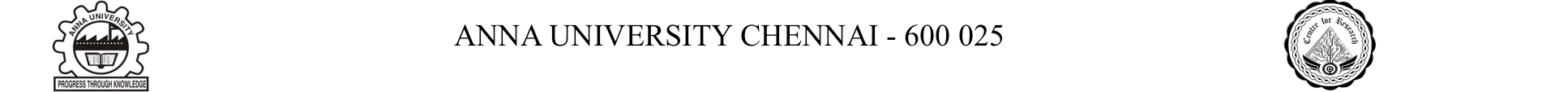}}
\ausection

\sloppy

\cleardoublepage
\pdfcatalog{%
  /PageLabels<<%
    /Nums[%
      0<</S/R>>%
      \the\numexpr\value{page}-1\relax<</S/D>>%
    ]%
  >>%
}
\pagenumbering{arabic}

\setlist[itemize]{topsep=0.0\parskip}



  
\chapter{INTRODUCTION}
\label{ch:intro} 

\section{Taxonomy of Gait Analysis}
\label{sec:gait-analysis}

The simplest definition of gait states that it is the manner and style of walking \citep{whittlegait}. This can refer to any animal that can walk whether bipedal or quadrupedal. It can be more sophisticatedly defined as the coordinated cyclic combination of movements that result in locomotion \citep{boyd2005biometric}. This definition can be equally applicable to any form of activity that is repetitive and coordinated so as to cause motion of the living being originating it. Gait can vary from walking, running, climbing and descending the stairs, swimming, hopping and so on; all of which follows the same principle by this definition. In the context of this thesis, the word `gait' generally refers to the `walking gait' of a human.

Out of all routine activities, walking (running or jogging) is by far the activity that makes use of most of our resources within a short span of time. The activity includes the work of the cardiorespiratory system \citep{van2013effect}, the musculoskeletal system \citep{foster2013pgals}, the nervous system \citep{kubo2011gait}, the renal system \citep{Kutner2015297} and metabolism \citep{Caldwell2013649}. This level of association is why the majority of the abnormal activities that can occur in the body reflect in the gait of the individual experiencing it. Hence, gait analysis plays a significant role as a clinical diagnostic. When applied in clinical practice, gait analysis proves to be more efficient when compared to a medical diagnosis that would otherwise be expensive and consume much more time.

\begin{figure}[t]
  \centering
  \includegraphics[width=1.0\linewidth]{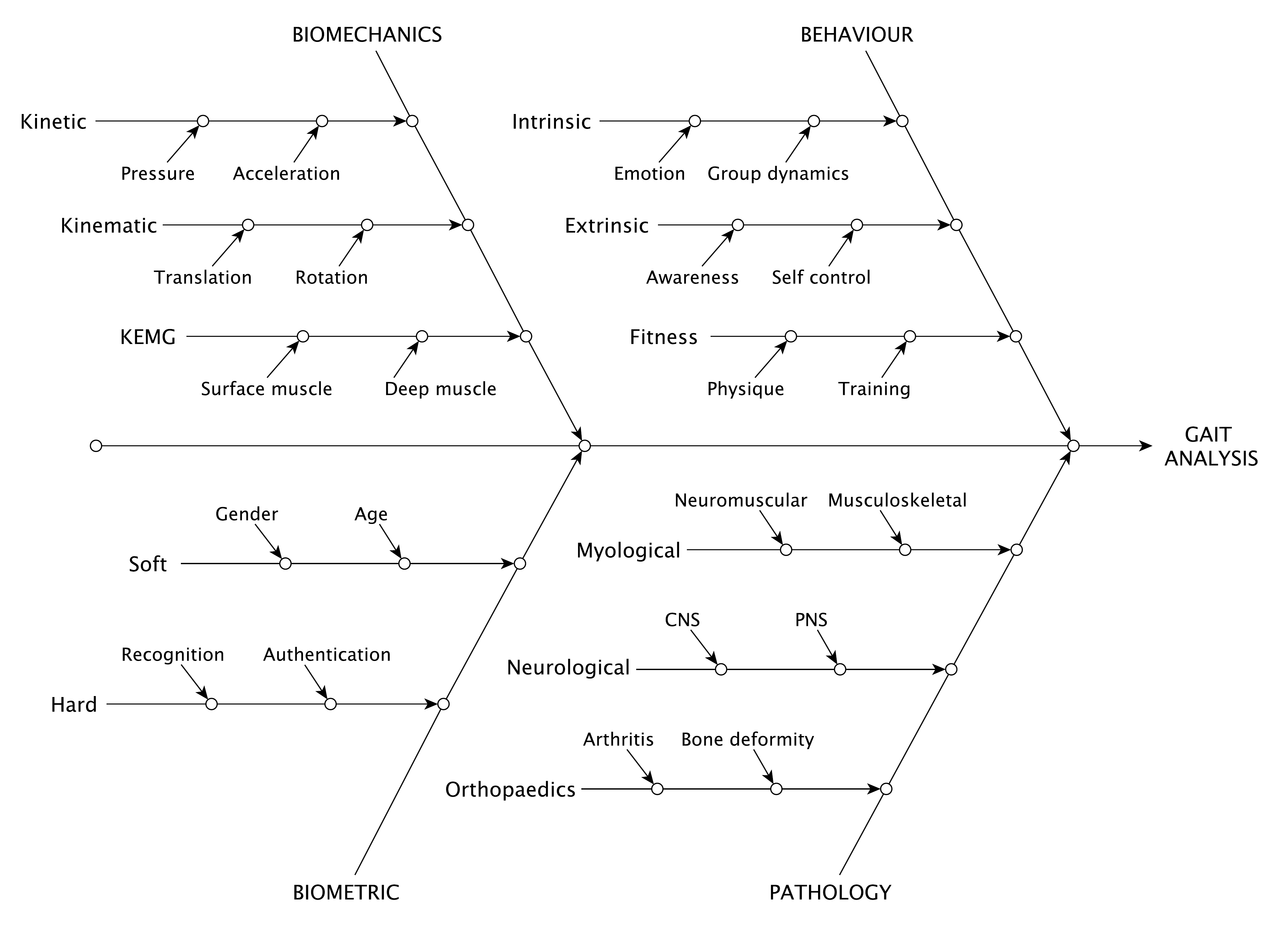} 
  \caption{Taxonomy of Gait Analysis}
  \label{fig:taxonomy}
\end{figure}

The broad spectrum of research on gait analysis can be classified based on their use or their modes of operation as illustrated in Figure~\ref{fig:taxonomy} as a fishbone (or Ishikawa) diagram. The application of gait analysis can be broadly classified into four areas, namely, pathology assessment, gait biometrics, biomechanics and behavioural analysis.

Clinical gait analysis focuses gait-related pathologies in depth. The analyses include measures taken to assess the level of progress associated with a particular pathology and detecting the onset of the disease that affects gait. Pathologies that are frequently connected to clinical gait analysis are Parkinson's disease, cerebral palsy, stroke and arthritis \citep{kirtley2006clinical}. A separate study addresses the fall of the elderly \citep{MonizPereira2012S229}. It helps in predicting the fall and to devise an effective treatment for the medical condition. Further studies revolve around post-operative study on gait -- one that focuses on how surgical treatments affect gait \citep{Tomasz_S}.

The attempts on recognizing people with their gait started back at the late 1990s \citep{little1998}. The research showed that the nuances in the natural gait of an individual can be considered an innate biometric that can be used to identify them. Apart from the possible application in authentication, gait biometrics has become a topic of serious value in forensic science. This value became of interest to the US Defense Department at the birth of the second millennium \citep{mcgrath2003}.

The mass of data that can be extracted from gait can also be of statistical relevance to provide inferences of a person's behaviour. Statistical gait analysis is inherently an interdisciplinary area that can aid the other areas of application. These include the design of a simple mobile app that can estimate units of calories burnt to complicated emotional inference systems \citep{thoresen2012first}. Statistical inferences in gait can also be utilized in demographics study which can be used to associate people into groups and study their group behaviour that can be influenced from gait.

There are two ways in which gait features can be classified in biomechanics -- kinetic and kinematic \citep{whittlegait}. The tools that are used for the observation limit the type of features that can be extracted. Kinetic gait analysis assesses the forces exerted and absorbed, moment generated and accelerations involved, but without the details that cover positional significance or orientation of objects concerned. A good example would be a simple foot switch, it provides the location and magnitudes concerning force but does not account the spatial positioning of the articulation points. Kinematic gait analysis is the converse of kinetic gait analysis. It describes the gait parameters that involve motion, position and orientation without considering the forces which cause the motion. The best example of a kinematic observation tool would be the simple RGB camera.

A relatively new mode of gait analysis uses kinesiological EMG (electromyography). An EMG is employed for recording the electrical activity produced by the musculoskeletal system. An in-depth analysis of gait can be done using this method. With an EMG, doctors can precisely diagnose which muscle tendons are damaged or under-performing from an individual's gait. The assessment of gait through a kinesiological electromyography (KEMG) was first lead by Inman to study both normal individuals and amputees. According to \cite{Sutherland200161}, the advent of the KEMG has revolutionized clinical gait analysis from the understanding of the knee-ankle functional link to the structure of treatment of cerebral palsy. EMG are not adapted for
biometric feature extraction due to their invasive nature.

\section{Historical Advents}
\label{sec:historical-advents}

\subsection{Inception of Gait Analysis}
\label{sec:inception}

The first study on gait dates back before 300\textsc{ bc} by Aristotle in Greece \citep{aristotle} which depicted the elements of movements of animals. A more theorized version was published by the Renaissance Italian physiologist, G.~A.~Borelli in 1680 \citep[][ English translation by P. Maquet]{borelli}. His work illustrated the notion of balance and how it was conserved in locomotion by the movement of the feet and the area of support which it provides. It was at that time that the concept of contraction in muscles was put forward for the cause of locomotion. The phases of a gait cycle was first depicted by the Weber brothers in 1836 \citep{weber}. The major advancements of gait analysis over time is provided in Table~\ref{tab:history} in chronological order.

\begin{table}
  \renewcommand{\arraystretch}{1.4}
  \centering
  \caption{Historical Advents of Human Gait Analysis}
  \begin{tabular}{|p{0.07\linewidth}|p{0.6\linewidth}|>{\raggedright\arraybackslash}p{0.22\linewidth}|}
    \hline
    Year & Event & Researchers \\
    \hline
    $\sim$300\textsc{ bc} & First work on \textit{De Motu Animalium} -- On the Movement of
                   Animals \citep{aristotle} & Aristotle \\
    1682 & A more theorized work on De Motu Animalium including the
           contractile movement of muscles \citep{borelli} & G.~A.~Borelli  \\
    1836 & Clear illustration of the gait cycle and the first
           published book on gait \citep{weber} & W.~Weber and
                                                 E.~Weber \\
    1873 & Analysis of gait through single plate chronophotography  \citep{mareyFr,mareyEn}
                 & E.~J.~Marey \\
    1878 & Gait analysis through chronophotography using a 24-camera
           array \citep{muybridge1,muybridge2} & E.~Muybridge \\
    1881 & Observed footing pattern through ink spray shoes \citep{vierordt} & K. H. Vierordt \\ 
    1895 & The effect of loads in the biomechanics of gait. The first
           3-D gait analysis \citep{braune} & Braune and Fischer \\
    1938 & The first operational force platform for gait analysis
           \citep{elftman1938} & H.~Elftman \\
    1947 & Theory of motor control and motor learning for posture and
           movement \citep{bernstein1947,bernstein1967} & N.~A.~Bernstein \\
    1950 & Extensive mechanical analysis of gait from hip, knee and
           ankle joints \citep{bresler} & B.~Bresler and J.~P.~Frankel \\
    1953 & Mechanisms involved in natural energy conservation during
           human gait \citep{saunders1953} & Saunders et al.\\
    1981 & Human Walking: the definitive guide for the normal gait of
           humans  \citep{inman} & Inman et al. \\
    \hline
  \end{tabular}
  \label{tab:history}
\end{table}

The first kinematic analysis of gait was done by two scientists, Marey and Muybridge in the 1870s through chronophotography.
French scientist E. J. \cite{mareyFr} visualized the phases of gait by taking multiple exposures of the moving subject on the same photographic plate \citep[translated into English in][]{mareyEn}. His initial study involved the motion of birds and insects. He later created the first visible spatiotemporal imaging of gait by capturing a human subject while walking with the prominent skeletal structure illuminated with stripes. 
English photographer, Eadweard Muybridge captured chronophotographs by using an array of 24 cameras that were made to shoot at a specified sequence of time within a fixed period interval. He captured many different animals in the same way \citep[work made available at][]{muybridge1}. Once such chronophotograph precisely interpreted the gait of horses proving the fact that at a certain phase of a galloping horse has all of its hooves off the ground. His chronophotographs of humans were taken while doing a wide range of activities \citep{muybridge2}.

Profound progress in the scientific study of gait was observed at the end of the 19\textsuperscript{th} century by German anatomist Wilhelm Braune and physiologist Otto Fischer in \textit{Der Gang des Menschen} \citep{braune}. Inspired by the E. J. Marey's chronophotographs, they applied scientific reasoning to explain the biomechanics of gait under the effect of load and without load. The study involved measuring the trajectories, velocities and acceleration of articulation points along the three-dimensional space. With the knowledge of the masses of each segment associated with the respective articulation points, they were able to provide further inferences to the forces that are involved in each phase of the gait cycle. Nikolai Bernstein, a neurophysiologist of Moscow, extended the work of Braune and Fischer in the late 1930s. His notable achievement was his work on the construction of movements in 1947 \citep{bernstein1947} by applying his theory of motor control and motor learning to discuss the role of the central nervous system (CNS) in limb coordination and postural control. His work was later translated into English in 1967 \citep{bernstein1967}.

\subsection{Clinical Relevance}
\label{sec:clinical-relevance}

K. H. \cite{vierordt} conducted a study to analyse human gait concerning the spatial pattern of footsteps. His work involves attaching small-scale ink spray nozzles to the shoes of the subjects and make them to walk on a moderately long surface in which the imprints of the footsteps due to the spraying of ink can be readily analysed.  The first force plate reported was used by Jules Amar in 1924 as a tool to help diagnose rehabilitation of patients who were injured during the First World War \citep{Baker2007331}. The apparatus includes the use of pneumatic three-component force plates. Its application was of clinical relevance but was not scientifically validated for its accuracy. The first successful force platform was introduced by \cite{elftman1938}. It consisted of a set of springs attached to force plates spread across the stepping region. The force is measured optically using a high frame-rate camera from the vertical up and down movement of the force plates measuring both point of force as well as its magnitude. This apparatus succeeded his previous experimental setup proposed in 1934 \citep{elftman1934} which only gave the point of application of the force but not the magnitude. All these early force plate designs were purely mechanical in their operation.

Much of the advancements in modern gait analysis were reported from California. This includes the detailed mechanical analysis of gait from the hip, knee and ankle joints by \cite{bresler}. Most were specifically from the Biomechanics Laboratory founded by V. T. Inman and H. D. Eberhart during the 1940s to 1960s \citep{historyww2}. Once such significant research includes the mechanisms in which the body conserves energy during gait \citep{saunders1953}. The first computerized video analysis of gait was done in the 1970's by David H. Sutherland who was mentored by Inman \citep{Sutherland2002}. He reinvented the art of EMG to be used for analysing gait. D. H. Sutherland was known to be the founder father of clinical gait analysis \citep{kaufman2006david}. The widely accepted definitive guide for normal gait was published by \cite{inman}.

The history of researchers who shaped gait analysis before World War II is detailed in \cite{historyenlight}, after its period in \cite{historyww2} and the modern era in \cite{historymodern}. Richard \cite{Baker2007331} provided a detailed account on the history of gait analysis focusing on the period before the application of computerized observation.

\subsection{Interest in Biometrics}
\label{sec:interest-biometrics}

\cite{murray} concluded that gait could pose a behavioural property which is unique to any human individual provided that all possible gait parameters are taken into account. Despite this statement, gait biometric research has proved that a subset of the vast features of gait is sufficient to produce a system that yields a performance of practical relevance.

Gait can be used as a non-obtrusive form of biometric \citep{boyd2005biometric}. This unique trait sets it apart from the other modalities. The other established forms of biometrics include face, fingerprint, handwritten signature and iris; all of which requires a closer observation of the subject to process the images at a sufficiently high resolution. However, gait can be observed from a person even from a lower resolution surveillance camera with minimal level of cooperation from the subject who is to be identified.
 
The Defense Advanced Research Projects Agency (DARPA) launched the Human Identification at a Distance (HumanID) programme in the year 2000 (ended in 2004). This research programme focused on human identification through the face, gait and the application of new technologies. The intention of the programme was to use the state of the art unobtrusive distance biometric techniques on real world data. Gait-based recognition played a significant role in here. Major institutions which took part in this programme were Massachusetts Institute of Technology, Carnegie Mellon University, University of South Florida (USF), University of Maryland, Georgia Institute of Technology, and the University of Southampton \citep{hidgait}. The datasets compiled by most of these institutions are publicly available. Section~\ref{sec:datasets} discusses more on those datasets.

\section{Nomenclature}
\label{sec:nomenclature}

\begin{figure}[b!]
  \centering
  \includegraphics[width=0.99\linewidth]{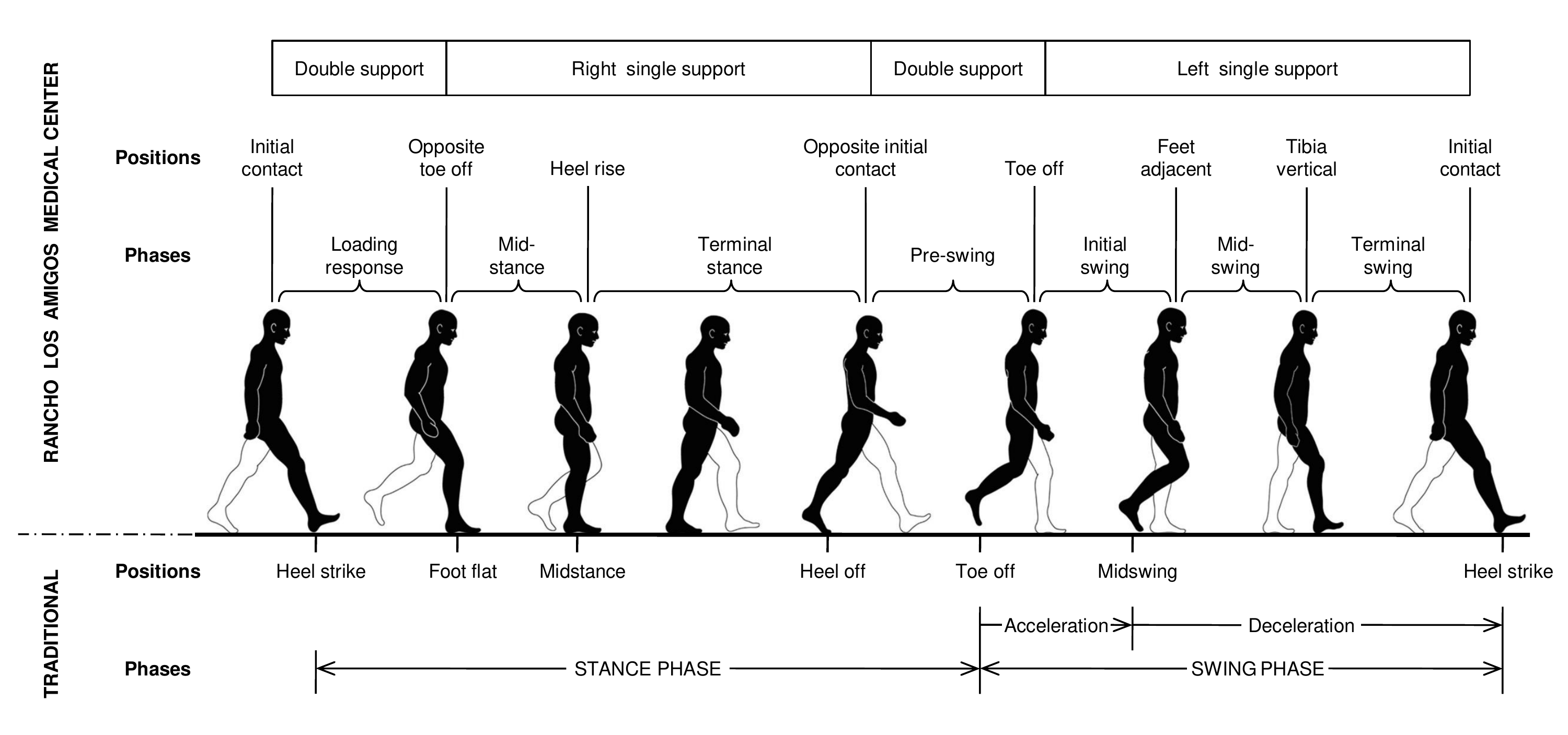}
  \caption{The human gait cycle}
  \label{fig:gaitcycle}
\end{figure}

The knowledge of the technical terms associated with the study of gait analysis is required for a deeper understanding in gait research. The stages of the human (normal) gait cycle are illustrated in Figure~\ref{fig:gaitcycle}.

\subsection{Phases of Gait}
\label{sec:phases-gait}

Gait phases and positions vary between two terminologies: the traditional terms and the standard of Rancho Los Amigos set by J.~\cite{perry1992gait} to handle both normal and pathological gait. Although both of which are used interchangeably, the knowledge of both system of terms is required to comprehend the literature. A standard human gait cycle consists of eight key positions, with the initial contact added twice to make the cycle complete. Note that the sequence of positions is accounted only for one foot. Hence, a complete gait cycle or a stride is the activity that concerns a sequence of events between two consecutive heel strikes (or heelstrikes) of the same foot. Whereas the activity that occurs from one foot's heel strike to the other is called a step. The time it takes for completing a stride is called cycle time. In a normal gait, the gait cycle of one foot is displaced from that of the other foot by a fixed phase shift.

The periods between consecutive gait positions are called the phases of the gait cycle. The load response phase depicts the response made by the body to absorb the impact of heelstrike by flexing the knee, contracting the quadriceps. In the mid-stance, the knee extends, and the associated foot is kept flat on the ground, i.e., both toe and heel are maintained in ground contact. During the terminal stance, the legs extend further while keeping its contact with the ground to advance the body forward. The legs then start to flex again to get ready to swing at the pre-swing phase. Then, at the initial swing, the foot accelerates to provide a forward momentum and controls its speed at midswing. The foot decelerates to reach its position, the next initial contact.

The period in which the foot is on the ground is called the stance phase, and it composes 60-62\% of the gait cycle \citep{kirtley2006clinical}. The remaining period, when the foot is off the ground, is called the swing phase. The proportions, however, differ depending on the individual and the speed of the walk. The stance phase encompasses the first four phases of the gait cycle from initial contact to toe off. The swing phases cover the phases between the toe off and the successive initial contact. The stance and swing phases alternate to produce a locomotion. While one foot is in swing phase, the other would be on stance phase for balance. 

\newpage
\subsection{Gait Parameters}
\label{sec:gait-phases}
Most of the following terminology is taken concerning the ones prescribed by \cite{whittlegait}.

\begin{description}[topsep=0pt]
\item[Cycle time] is the time taken to complete a single gait cycle. It is also known as a gait period.
\item[Double support] is the interval in which both feet are in contact with the ground when the stance phase of both legs overlap.
\item[Step length] is the length covered between initial contact of one foot to that of the other feet.
\item[Stride-length] is the distance displaced by the foot during its stride (one complete gait cycle); consists of two step lengths.
\item[Cadence] is the number of steps taken in a given time usually measured in steps per minute
\item[Walking base] is the perpendicular distance defined by the space between the track of both feet.
\item[Walking speed] is usually measured in metres per second, but can also be measured in statures (height) per second in clinical practice. 
\end{description}

The double support period would define the difference between walking and running. As long this period is positive, the person is walking, and her/his speed is inversely proportional to the magnitude of this interval; otherwise, the person is considered to be running (or jogging). The double support period is an important measure during a walking race (or race walking) where a negative double support time would indicate that the candidate is running instead of walking.

The stride length of both feet is the same regardless of the effect of the pathology. However, the step lengths commonly differ for a pathological gait due to differing stance and swing phase proportions between the left and right leg. For instance, a pathologically affected leg would tend to spend lesser stance time leading to a lesser swing in the other leg. Hence, a shorter step length on one side would infer antalgic conditions on the other side.

\subsection{Reference Planes}
\label{sec:reference-planes}

\begin{figure}[t]
  \centering
  \includegraphics[width=0.75\linewidth]{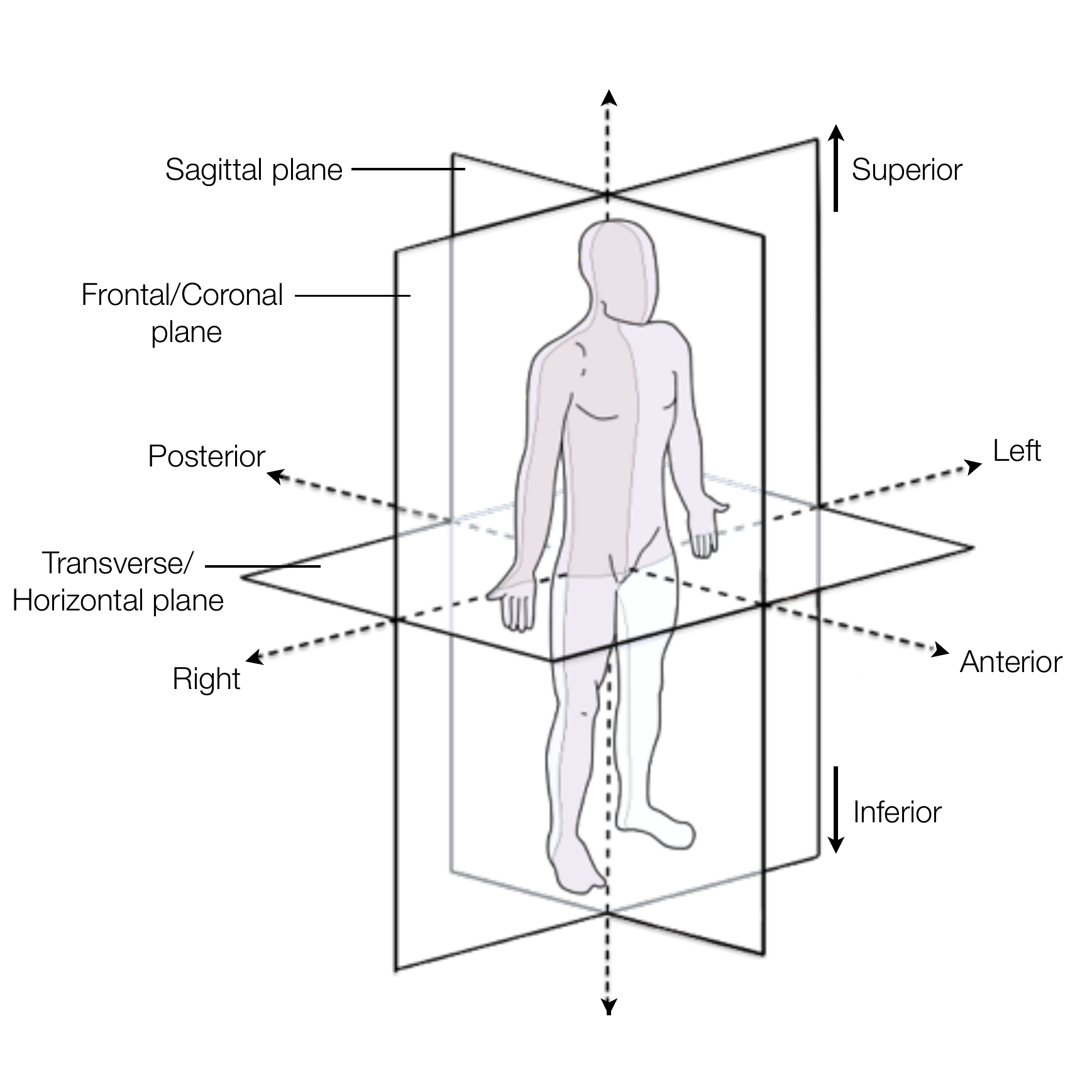}
  \caption{Anatomical positions and planes of reference}
  \label{fig:planes}
\end{figure}

Gait is usually observed in one of three planes of reference: sagittal, transverse, or frontal. A clear depiction of the reference planes is as provided in Figure~\ref{fig:planes}. The sagittal plane divides the body into right and left proportions. Most gait research in literature is based on the sagittal angle as it is perceived to be the simplest angle to obtain the salient features. When observed at a sagittal angle, the walking subject tends to move from left to right or vice versa, i.e., the person is said to walk frontoparallel to the viewpoint. Along the frontal or coronal plane, where the body is divided into front and back halves, a walking subject moving forward in motion is perceived to move towards the observer. The transverse or horizontal plane divides the body into upper and lower halves. There is not much literature that analyses gait in this angle due to the difficulty of extracting information from it.

\begin{figure}[t]
  \centering
  \includegraphics[width=0.8\linewidth]{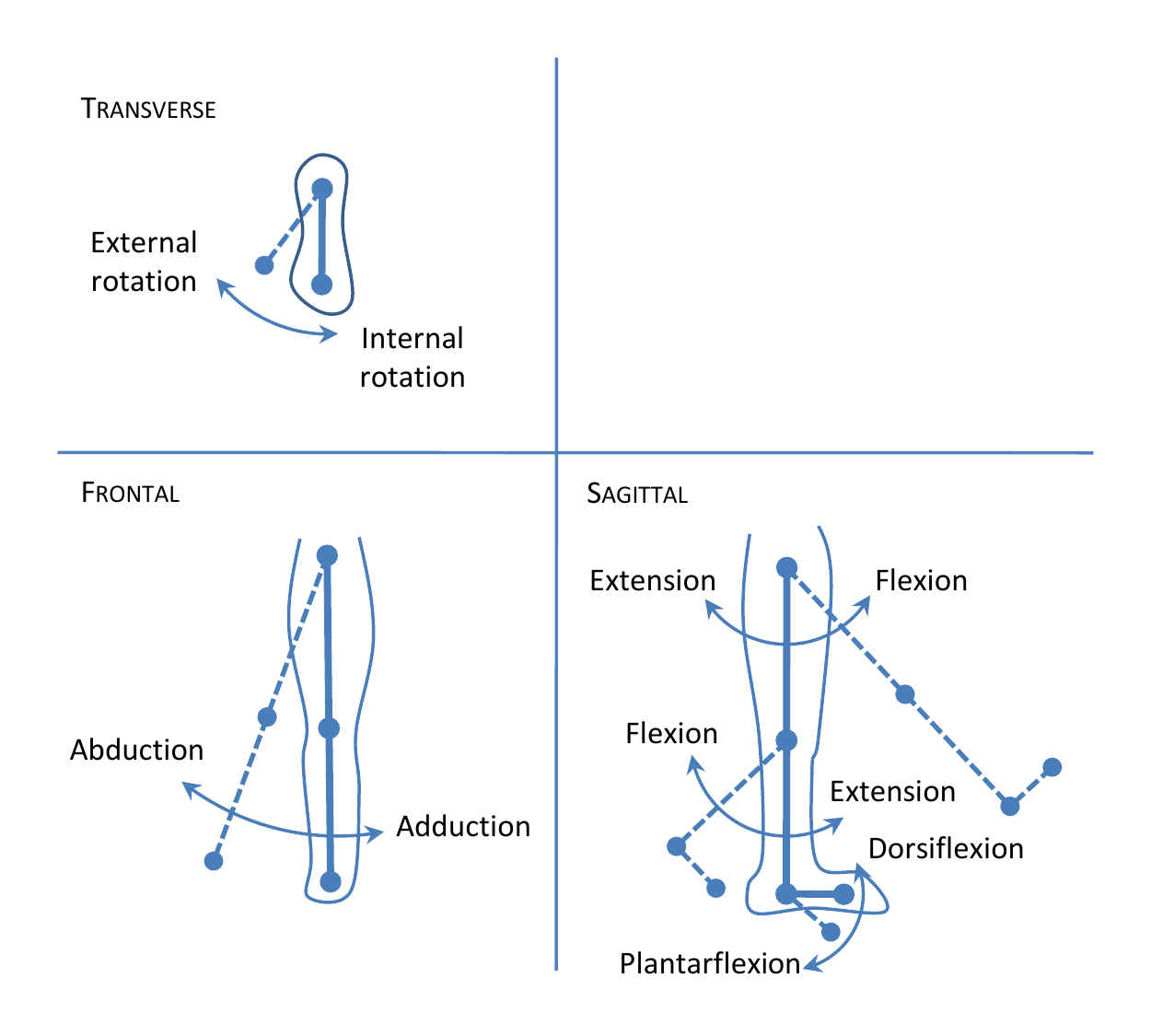}
  \caption{Axes of movement}
  \label{fig:movement}
\end{figure}

Three pairs of rotatory motion are required for the body to make any means of movement: internal and external rotations, abduction and adduction, extension and flexion. They are illustrated in Figure~\ref{fig:movement}. In general, flexion and extension occur in the sagittal plane, adduction and abduction take place in the frontal plane, and other rotatory motions occur in the transverse plane. Note that in Figure~\ref{fig:movement}, the internal and external rotation shown is performed by the knee and the thighs, not by the ankle. If done at the ankle, the similar result is said to be caused by the forefoot's adduction/abduction. Flexion at the ankle is known as dorsiflexion whereas its extension is known as plantarflexion.

\subsection{Performance Metrics}
\label{sec:biometrics}

The following terms are used when evaluating the performance of  biometric systems.

\begin{description}[topsep=0pt]
\item[FRR] (False Reject Rate) is the rate at which an authentic subject is rejected by the system under observation.
\item[FAR] (False Accept Rate) is the rate at which an impostor/intruder is accepted as an authentic subject by the system under observation.
\item[AER] (Average Error Rate) is the average of FRR and FAR, \\i.e, $\text{AER} = (\text{FRR}+\text{FAR})/2$.
\item[EER]  (Equal Error Rate) is a measure specific for threshold-based authentication systems. At a certain threshold, both FRR and FAR becomes equal; this error rate is termed as the EER.
\item[Rank] is a measure specific to distance-based identification systems. Accuracy at rank $k$ infers the probability of the true identity being present in the top $k$ closest identities returned by the system.
\item[CCR] (Correct Classification Rate) is simply the accuracy of the classification system used for identification (recognition).
\item[ROC] (Receiver Operating Characteristic) is the curve generated by plotting the verification rate (sensitivity; $1 - \text{FRR}$) against the FAR in the case of authentication.
\item[CMS$_r$] (Cumulative Match Score) is the recognition rate at rank $r$.
\item[CMC] (Cumulative Match Characteristic) is the curve generated by plotting the rank against the CCR for identification.
\end{description}

\section{Research Directions}
\label{sec:research-directions}

Gait analysis is an interesting study and is proven to be applicable for diverse domains of applications. There are many open areas of research from which one can choose in the field. They can be segregated into two groups -- active areas and those that are least explored.

\subsection{Active Areas}
\begin{itemize}[leftmargin=*]\itemsep4pt
\item \textit{Reliable recognition independent of clothing style.}  Several
  algorithms have been outlined to combat this problem
  \citep[e.g.,][]{yu2006framework,Yogarajah20153} and a handful of datasets to
  help with this regard (CASIA-B, OU-ISIR-B, TUM-GAID, and so on). However,
  state of the art gait biometrics algorithms do suffer a significant
  depreciation when the clothing style changes.

\item \textit{A comparative study of gait pathologies.}  Various papers are
  published classifying diseases concerning their nuances in gait
  \citep[e.g.,][]{Kohle1997, Zeng2015246, Geroin2015736}. The clinical attention
  to this segment is growing as there are still so many gait-related pathologies
  that are still not compared as of yet.

\item \textit{Gait recognition at different walking speed.}  With datasets like
  CASIA-C and OU-ISIR-A, gait recognition at different speeds is becoming more
  of interest
  \citep{tanawongsuwan2003study,iwashita2015gait,tanawongsuwan2003perf}. There
  are, however, further improvements that can be made.

\item \textit{Correlation between walking and running gait.} When an
  individual switches from walking to running, the double support time
  turns negative while the arm swing increases. To achieve balance in
  this state, the body makes many modifications to the posture.

\item \textit{View-independent gait recognition.} Numerous studies that propose
  recognition models that could cope with multiple views. Steady progress is
  still being made in this area.

\item \textit{A comparative study between model-based and model-free
    recognition methods.} Though studies contrast on the merits and
  demerits of both approaches, there is still space for an in-depth
  comparative analysis.

\item \textit{Unbiased comparison of state of the art gait recognition.} The
  methods in literature show conflicting results of correct classification rates
  when performances of established algorithms are compared, even with the same
  dataset. The existing algorithms are to be compared with standard datasets
  without bias.

\end{itemize}
 
\subsection{Least Explored}
\begin{itemize}[leftmargin=*]\itemsep4pt
\item \textit{Correlation between kinetic and kinematic features.}
  Studies show how kinetic and kinematic features can be assessed
  together \citep{eemcs15745}. Some  utilize both for gait assessment
  \citep{Viton199847,tao2012gait,zeni2008two}. However, relationship
  between kinetic and kinematic features of gait is still blurred.

\item \textit{Characteristic differences in gait based on ethnicity.}
  People from different regions do have a characteristic gait. CASIA
  datasets are composed of Chinese individuals, while the USF
  gait dataset are largely American. More datasets are to be created
  across various ethnicity to study the associated differences between
  them in terms of their gait.

\item \textit{Ranking of factors that inhibit gait.} The confounding factors of
  gait are as listed in Section~\ref{sec:covariate-factors}. Though known, they
  are not compared in a way that would determine to which extent they inhibit
  gait. For instance, to what instance could training affect biometric gait?
  This has been an open question to which no reports are currently available.

\item \textit{Different machine learning methods for gait recognition.} The
  novelty of the recognition algorithms lies on the technique in which features
  are extracted. Commonly used classifiers are Artificial Neural Networks (ANN),
  $k$-Nearest Neighbours ($k$NN), and Support Vector Machine (SVM). A detailed
  analysis on how different classifiers effect the prediction rate through state
  of the art gait feature extraction algorithms is yet to be reported.

\item \textit{Stratification of gait patterns.} People can be grouped
  by means of their gait. This grouping can reveal certain insights to
  the behaviour or traits that members of the same group share. In
  theory, we could find some means to cluster and associate them to
  different strata.

\item \textit{Gait in demographics.} Deeper statistical inferences can
  be drawn from gait within social groups. That is, the similarities
  of gait can be assessed within a demographic group, and differences
  can be studied across the groups in terms of gait. This is the
  converse of stratification wherein people are grouped based on their
  gait similarities.

\item \textit{More statistical gait analysis.} There are limited amount of
  studies reported in this area. Gait has been statistically analysed
  extensively for assessing pathology and studying gait biometrics. However,
  there is untapped potential lying within the broader statistical analysis of
  gait.

\end{itemize}

\section{Contributions}
\label{sec:contributions}

Four gait-based biometric assessment methods are proposed in this study: a gender classification algorithm, a gait recognition method, and two authentication paradigms.


Existing methods for gait-based gender classification extract spatiotemporal relationship between the frames of a gait sequence to make a single prediction on the subject's gender. The method proposed in this thesis, however, makes a prediction for every pose assumed by the subject during gait. The prediction of the gait instance itself is taken through a majority vote. Two sets of features are explored for this purpose, viz., EFD and RCS.

Gait recognition (or identification) is the process of identifying a person from her/his gait alone. Not all features extracted from gait may correlate with the identity of the subject, especially in the case of covariate factors that affect a person's walking style. Though gait recognition is well-developed in literature, the algorithm proposed in this thesis optimizes the efficiency of template-based recognition systems. The resulting template would include only features that are least affected by the covariate factors.

Gait authentication (or verification) is a process in which the system ascertains whether the subject is whom she/he claims to be based only on the gait observed. The literature available for this topic is limited as most adopt the features that are used in identification for authentication as well while defining closeness to the claimed identity through a Euclidean threshold. Two completely new paradigms proposed in this thesis greatly reduce the average error rates of gait authentication systems. 

This study will also attempt to answer the following research questions.
\begin{enumerate}[topsep=0pt, noitemsep, leftmargin=*, label=(\alph*)]
\item Can gender classification be improved to cope with occlusions?
\item Can chain encoding help with gait-based gender classification?
\item How to optimize feature selection for video-based gait recognition?
\item How to enhance the performance of gait authentication systems?
\item Is the sagittal angle really the best angle for gait biometrics?
\end{enumerate}

\section{Organization of the Thesis}
\label{sec:organization}

The rest of the thesis is organized as follows. Chapter 2 details the relevant
literature in the field of gait biometrics. The research contributions are
presented in chapters 3 through 6. These chapters elucidate the state of the
art and highlights their limitations followed by methods proposed to overcome
them. Algorithms that are included are experimentally
verified through established benchmark procedures and quantitatively compared
with the existing methods.

\begin{description}[topsep=0pt, leftmargin=2em]\itemsep=0.2em
\item[Chapter 2:] Provides a brief interpretation of the literature closely associated
  with the research work along with the description of the available datasets
  for the evaluation of gait biometrics. 
\item[Chapter 3:] This chapter focuses on the most popular soft biometric aspect
  of gait biometrics -- gender classification. Though the existing method produces
  close to perfect identification, the proposed method, \textit{pose-based
    voting}, allows for better recognition even in partially occluded gait sequences.
\item[Chapter 4:] Gait recognition makes up for a major share in gait
  literature. This chapter presents the \textit{genetic template segmentation}
  technique for gait recognition.
\item[Chapter 5:] This chapter presents an innovative framework called \textit{
    multiperson signature matching} to outperform the existing threshold-based
  gait authentication framework. This scalable method improves its performance
  in proportion to the system population. 
\item[Chapter 6:] When the system population is considerably low the
  effectiveness of the method proposed in chapter 5 is suppressed. To overcome
  its limitation, this chapter presents a novel framework using the
  \textit{multivariate Gaussian Bayes} posterior probability. 
\item[Chapter 7:] This chapter concludes the findings of the thesis and presents
  the work planned for the future.
  
\end{description}


\chapter{LITERATURE SURVEY} 
\label{ch:literature} 

Gait analysis is composed of three components, viz., kinematics, kinetics, and electromyographic data \citep{gage1995gait}. Each of these components can be separately analysed and studied. There is a large variety of gait analysis tools that are commercially available as shown in Table~\ref{tab:tools}. A detailed account of the different ways to measure gait analysis is provided by \cite{clayton2001measurement}.

\begin{table}
  \centering
  \caption{Commercially Available Tools for Gait Analysis}
  \label{tab:tools}
  \resizebox{\linewidth}{!}{
  \renewcommand{\arraystretch}{1.2}
  \begin{tabularx}{1.3\linewidth}{%
    r 
    p{0.4\linewidth} 
    l 
    X 
    } 
    \toprule
    S.No. & Tool & Mode & Description \\
    \midrule

    \begin{filecontents*}{Data/GaitTools.csv}
Tool,Mode,Description
3D Gait\newline Running Injury Clinic,Kinematic,VICON camera system designed for clinical gait analysis \citep{watari2016determination}
APAS\newline Ariel Dynamics,Kinematic,Ariel Performance Analysis System - 3D biomechanical analysis through multi-video feed \citep{kaczmarczyk2009gait}
CLIMA\newline STT Systems,Kinematic,IR marker-based 3D motion capture system \citep{fuentes2012clinical}
Codamotion\newline Charnwood Dynamics Ltd.,Kinematic,3D motion analysis through markers observed from a CX1 sensor unit \citep{OMalley2007149}
Electrogoniometer,Kinematic,A system of attachable goniometers to measure angle between joints \citep{kikuchi2016bioinspired}
Emerald\newline MIT Labs,Kinematic,Applied WiTrack technology for fall detection of the elderly \citep{Adib_F3}
Footswitch FSR sensors\newline Cometa,Kinetic,Force-sensing resistor -- used with footswitches to measure foot pressure \citep{taborri2014novel}
Force plate\newline  AMTI; Bertec Corp,Kinetic,Multicomponent pressure sensitive tablet used for measuring force when stepped on \citep{Needham2015}
GAITRite\newline CIR Systems Inc.,Kinetic,Extensible force platform walkway \citep{McIntosh2015}
Instrumented treadmill\newline Bertec Corporation,Kinetic,Split-belt treadmill with dual integrated force plates \citep{edgintoninstrumented}
Instrumented treadmill\newline Force Link,Kinetic,Large treadmill for variable movement integrated with force plates \citep{Bosmans20152116}
Kinect \newline Microsoft Corporation,Kinematic,A depth camera and a webcam in a single unit \citep{kinect}
Kinesiological fine-wire\newline electromyography,EMG,EMG based on fine nylon-coated wires placed by inserting hypodermic needles into the muscle \citep{kfinewireEMG}
Kinesiological Surface\newline Electromyography,EMG,EMG based on active electrodes placed in the surface of the skin over the muscle region \citep{konrad2005abc}
Opal \newline APDM Wearable Technologies,Kinematic,Multiple body-worn intertial sensors (max. 24) for biomechanical analysis \citep{opal}
Stride Analyzer\newline B\&L Engineering,Kinetic,Footswitch and analysing software couple to assess stride \citep{Sankar2009}
Telemetry device,Misc.,{Remote measurement collection device to track signals from kinetic, kinematic or EMG  measurements \citep{bamberg2008gait}}
TEMPLO motion analysis\newline Contemplas,Kinematic,Marker-based mocap capture and analysis software \citep{chang2015changes}
Triaxial accelerometer ,Kinematic,An accelerometer to measure all three spatial axes with an optional adjustable frequency response \citep{Millecamps2015164}
VICON\newline Vicon Motion Systems Ltd.,Kinematic,Multi-cam mocap system that captures reflective strobe markers \citep{Needham2015}
\end{filecontents*}
\csvreader[late after line=\\]{Data/GaitTools.csv}
    {Tool=\tool, Mode=\mode, Description=\descr}
    {\thecsvrow. & \tool & \mode & \descr}
                                            
    \bottomrule
  \end{tabularx}
}
\end{table}

\section{Approaches for Gait Biometrics}
\label{sec:approaches}

The biometric features of gait can be extracted through a variety of ways. According to \cite{gafurov}, the approaches for gait biometrics can be classified into three types, viz., wearable sensor-based, floor sensor-based and machine vision-based methods. Research over the past decade has vastly developed the field of gait biometrics since then. The methods can be grouped by the tools employed for feature extraction as follows:
\begin{enumerate}[topsep=0pt, noitemsep, leftmargin=*, label=(\alph*)]
\item Inertial sensors
\item Footstep analysis
\item Acoustic data
\item WiFi signals
\item 2D and 3D camera (computer vision)
\end{enumerate}

Each of these tools has its pros and cons that are assessed and addressed in ongoing research.

\subsection{Inertial Sensors}
\label{sec:intertial}

An inertial sensor is a device used to measure the motion of the attached subject based on inertia. The ones used for gait analysis are usually capable of measuring motion across three dimensions (triaxial). They can be either worn or carried by the person who is to be identified. When wearable sensors are used, especially when the system consists of multiple devices, there might be a problem of inconsistent sensor attachment \citep{forster2009evolving}. Sensor placement during testing phase should be the same as in the training phase. Inconsistent attachment can introduce an additional error factor to the biometric system. There is also a greater problem sensor orientation \citep{ouisir_sa} which can completely change the pattern of readings observed in the three-dimensional space.

An accelerometer is used to measure the acceleration exerted along a given direction. A triaxial accelerometer would give readings for accelerations along all three spatial axes. They are by far the most common form of kinematic gait measuring equipment as they are the least expensive of the alternatives and the easiest to calibrate and process. The frequency response of an accelerometer shows the maximum deviation of its sensitivity \citep{accel}. The triaxial accelerometer is the most common form of inertial sensor applied for gait recognition. 

The largest inertial sensor database till date was presented by \cite{ouisir_is}. It consists of sensor data from 744 subjects. Each subject wore a gait capture belt around the waist. The belt was fitted with three triaxial accelerometer-gyroscope couple and a smartphone. The triaxial accelerometer in the smartphone was also considered for the data collection. \cite{rong2007wearable} and \cite{derawi2010improved} both report an EER of 14.3\% using this dataset. 


The main challenge to mobile phone-based gait authentication approaches is that the subject should bear the phone in a specific location -- usually the trouser pocket for the most successful operation. So this constraint refrains the user from using any other attire that does not include a pocket. Other locations for inertial sensor placement include hip, back of the waist, arm, and ankle. A detailed account of gait verification through wearable sensors was given by \cite{nickel2012accelerometer} and \cite{sprager2015inertial}.

\subsection{Footstep Analysis}
\label{sec:footswitches}

\begin{figure}[t]
  \centering
  \includegraphics[width=0.5\linewidth]{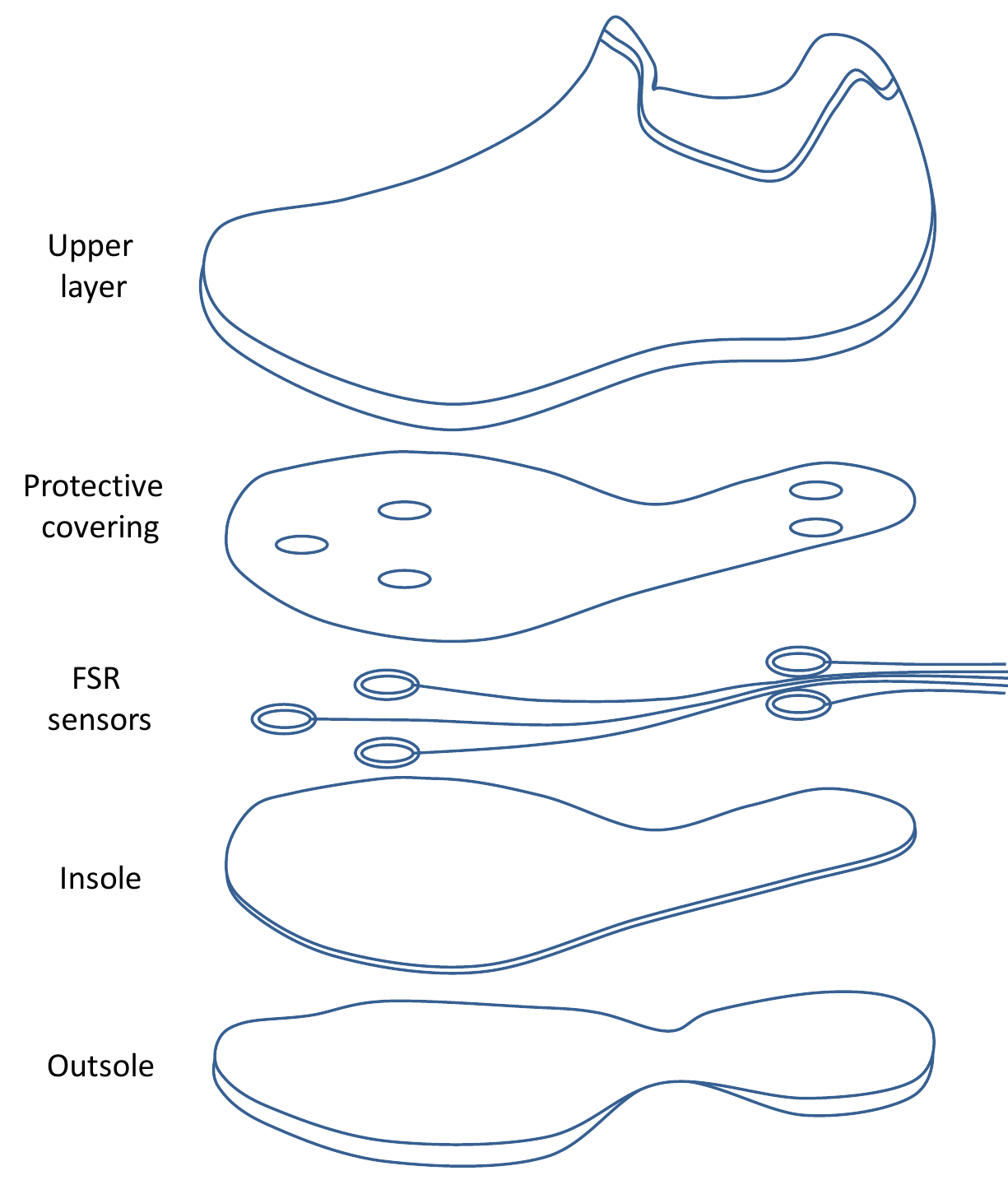}
  \caption{Schematics of a typical FSR footswitch}
  \label{fig:footswitch}
\end{figure}

A person's footstep can be analysed either by a wearable sensor, like a footswitch, or by a floor mounted sensor like a force platform. A footswitch measures the pressure applied by the foot on the ground. It can extract temporal data from swing-phase and stance-phase, and cadence and velocity. However, it is unable to provide spatial data such as step/stride length. The early form of footswitches used for gait analysis was designed through a conductive material placed on the inner soles of a pair of shoes \citep{Minns1982328}. The current technology allows more efficient sensors to be embedded in the soles of the shoe such as advanced force-sensing resistors (FSR). A simple way to implement this design is as shown in Figure~\ref{fig:footswitch}. It works on the principle that when force is applied to a conductive polymer surface, the resistance across varies inversely to the force applied -- a method patented by F. N. \cite{eventoff1984electronic}.

S. J. \cite{morrisThesis} designed an apparatus that combines both the footswitch and inertial sensors and later developed it \citep{bamberg2008gait}. It includes FSR sensors along with accelerometers and gyroscopes to a shoe thus forming a `GaitShoe'. This apparatus was used to assess heel strike and toe off as well as to approximate foot orientation and position. Although the footswitch is useful in clinical gait analysis, the practical relevance of this device is still questionable from a biometric perspective. 

To enable gait authentication in closed spaces and further avoiding the privacy concerns of visual surveillance systems, floor mounted systems became an option to extract gait biometric features \citep{mason2016gait}. These instruments allow the ground reaction force (GRF) to be measured directly from the floor on which the individual walks. The terms force plate and force platforms are used interchangeably since a platform can be assembled through multiple force plates, but in this context, they are described as two different instruments. A force plate is a metal plate coated with a non-slip substance beneath which transducers are attached that converts pressure to electrical signals which can be amplified and processed \citep{clayton2001measurement}. A force platform is the same in definition but can be extended to a larger surface which can facilitate efficient tracking of the foot placement during gait such as a walkway, e.g., GAITRite \citep{gaitrite}.

A detailed review of footstep recognition is given by \cite{ruben2009footstep}. Recently, \cite{mason2016gait} conducted detailed experiments on the biometric performance of the GRF using various machine learning algorithms. They show that the EER can be reduced to below 5\% using the GRF data alone. The literature in footstep analysis is not as vast as in computer vision-based and inertial sensor-based techniques. The main reason is due to the lack of a publicly available dataset. Force platforms are expensive instruments and are difficult to install, maintain and analyse. It would be one of the most expensive gait biometric assessment tools explained in this section.


\subsection{Acoustic Data}
\label{sec:accoustic}

The biometric features that can be extracted through footsteps can also be derived from the sound generated by them. The temporal analysis of the acoustic response of footsteps can be a highly unobtrusive gait biometric. The use of acoustic footstep analysis for gait identification is relatively new in literature. The sounds that can be heard from the same individual can vary across different footwear and surfaces. The challenge of acoustic gait recognition is the extraction of features that are invariant to these conditions. \cite{shoji2004personal} attempted to overcome this issue through mel-cepstrum analysis. This system was succeeded by \cite{itai2006footstep} using psycho-acoustics parameters, viz., loudness, fluctuation strength, sharpness, and roughness.

The apparatus of \cite{bland2006acoustic} consists of three microphones to hear the sound and three sensitive accelerometers as seismic sensors in the vicinity in which the subjects are to be identified. The sound that is generated from the footsteps is composed of a mixture of waves of both audible and inaudible frequencies. The components from the higher frequency bands are first converted to the audible range before the application of predictive modelling. \cite{kalgaonkar2007acoustic} used the acoustic Doppler sensors as an inexpensive alternative to the customized apparatus. \cite{itai2008footstep} approached the acoustic gait recognition problem in a similar way to speech recognition through Dynamic Time Warping (DTW). \cite{anguera2012acoustic} applied acoustic gait analysis for person identification, gender classification and shoe-type identification where SVM was the classifier used for the prediction. \cite{geiger2014acoustic} used only the audio data from the TUM-GAID dataset \citep{tum_gaid_paper} with Hidden Markov Models (HMM) for prediction.

\subsection{WiFi Signals}
\label{sec:wi-fi}
Groundbreaking research at Massachusetts Institute of Technology resulted in a novel technique to track gait using only WiFi signals. Fadel Abib and Dina Katabi first proposed a system called Wi-Vi -- WiFi Vision -- to detect and locate (to a degree) objects in motion through walls \citep{Adib_F1}. The system was also able to recognize primitive gestures. All it requires is a WiFi transmitter and two or more receivers; it requires no wearables or attachable items for its functioning. It can approximate the location of moving subjects behind concrete walls, hollow walls and wooden doors by computing the Doppler shift and spatial angle of reception. It uses a process called nulling to eliminate static objects from detection.

Katabi's research group generalized this concept for 3D tracking of humans \citep{Adib_F2}. Their new system, WiTrack, is a wireless device that can track a person within a 3D space which even works through occlusion and not limited to line-of-sight. Body parts can be coarsely tracked more efficiently than Wi-Vi but not so in detail, yet its accuracy was claimed to exceed that of RF localization systems at its time. WiTrack was more of an intermediate between RF-based localization systems and human-computer interaction systems like Kinect. It uses a single transmitter and three receivers to triangulate the motion of the subject. WiTrack is tuned only to pick out objects in motion and neutralize signals associated with static objects through subsequent reception cancellation. This process is termed to be background subtraction.  The tracking is not disrupted by small moving objects. It is supposedly the first 3D tracking device with an error within the order of centimetres; median of deviation along $(x,y,z)$ is $(10,13,21)$. While the first design \citep{Adib_F2} was only intended to track one human at a time, it was developed further \citep{Adib_F2} for multi-person localization. 
Recent developments of WiFi-based gait recognition includes WiWho by \cite{zeng2016wiwho} and WiFi-ID by \cite{zhang2016wifi}.

\subsection{Computer Vision}
\label{sec:video-based}

Computer vision is by far the oldest and the most researched approach for gait biometrics. While MoCap systems are used extensively in clinical aspect, methods that require just regular 2D video can be easily employed in legacy surveillance systems. The rest of the methods discussed in this thesis employ only computer vision.

\subsubsection*{MoCap Systems}
\label{sec:motion-capture}

Motion capture (MoCap) provides the most accurate technique to measure gait analysis. This approach consists of attaching reflective markers to strategic locations on the subject's body. Special cameras emit infrared (IR) strobe lights which reflect off the markers and back into the video camera. The camera can then pick up point-light motion of the reflectors that are visible from its viewpoint. An array of cameras (mostly eight) is positioned around the subject so that by computing on the inputs of the cameras combined, a three-dimensional depiction of the subject can be visualized.

\begin{figure}[t]
  \centering
  \includegraphics[width=0.7\linewidth]{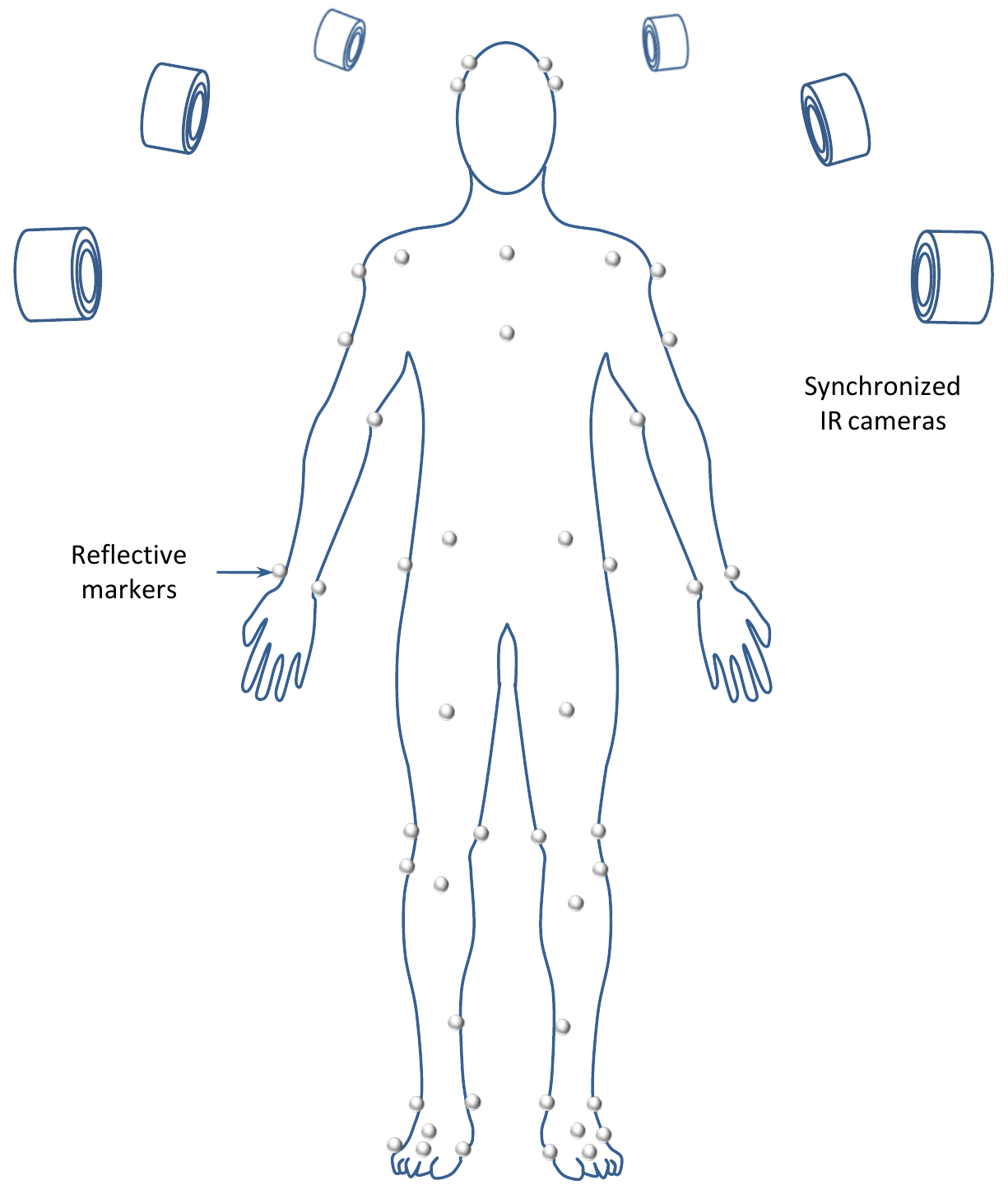}
  \caption{Simple schematic of a motion capture system}
  \label{fig:mocap}
\end{figure}

A simplified schematic of a typical MoCap system is as shown in Figure~\ref{fig:mocap}. They are of great value in clinical practice as they provide precise analysis required for treating gait related pathologies \citep{mulroy2003use, Yuan2006S67, meyns2012altered}.

\subsubsection*{RGB Camera}
\label{sec:camera}

A typical MoCap system can be very expensive and complex to set up. Two-dimensional RGB cameras, on the other hand, are cheap to obtain and do not require a complicated setup process as associated with MoCap systems. Since they lack the expressiveness of a fully functional MoCap system, their application for clinical use is limited. Nevertheless, due to their unobtrusive and pervasive nature, they find their application extensively in gait identification. 

\cite{vasconcelos2015human} proposed a method to extract the articulation points from a human gait from a single RGB camera. These points together formed a point distribution model (PDM)\citep{cootes1992training} which would hence be helpful for a model-based processing of human motion during gait. 100 contour points were approximated using the silhouette border, and 13 articulation points were labelled on the image manually. This labelling would facilitate the construction of an Active Shape Model (ASM) \citep{cootes1995active}. The ASM would combine the PDM shape model along with the grey information of associated points to be able to generalize to new images in the dataset. The CASIA-B was used for experimentation. By training the ASM using only 14 subjects along four directions, the system was able to approximate the landmark points to unseen images in the dataset.

\subsubsection*{Microsoft Kinect}
\label{sec:kinect}

The main components of the Kinect is an IR projector, an RGB camera, and an IR camera. Details of its function can be found in \cite{kinect}. The concept behind depth perception involves the analysis of structured light, a patented technique by PrimseSense \citep{freedman2014distance}. The infrared projector projects a structured image called a speckle pattern. Objects in range reflect back the speckle pattern. When this pattern is viewed away from the angle of projection, parallax occurs. The parallax would aid the IR camera in depicting the distance of the objects in view from the device. By combining the image feed from the RGB camera, depth information inferred from the IR camera and machine learning the Kinect can produce 3D model of the object it sees. It is much less expensive than a MoCap system and is more flexible than a simple RGB camera. Several techniques are proposed to extract gait features from a Kinect camera \citep{Naka_H, Mentiplay20152166, Xu2015145}. It is much more effective in gait-based recognition \citep{Chattopadhyay20159, Kastaniotis2015}, and it can even be utilized for clinical gait analysis \citep{Nocent2013e161}.

\Needspace{5\baselineskip}
\section{Soft Biometrics}
\label{sec:soft-biometrics}

Soft biometrics are characteristic that can apply to a certain group of individuals. Some of the prominent human soft biometrics include ethnicity, gender, age, height, weight, eye colour, scars and marks, and voice accent \citep{nandakumar2009soft}. These properties are considered `soft' because they do not distinguish an individual from all others. For instance, there can be more than one person who possesses a given height and weight, and it is possible to mimic the accent of another person. Out of the above, the first four can be identified through gait alone. Prominent research on soft biometrics focus on age estimation and gender classification \citep{hu2011gait}.  

\subsection{Age Estimation}
\label{sec:age-estimation}

For an average human, the muscle fibres and muscle coordination deteriorate in the later stages of his/her life. This process affects the gait of the aged human. The probable gait-related symptoms that can be exhibited at the elderly stage are the loss of stability, momentary imbalance, and tendency to fall \citep{senden2014influence}. The degree to which any of these symptoms occur can differ from one person to another but are indicators of old age nonetheless. A person's gait can also be an indicator of the level of maturity. One can easily tell the gait of an adolescent from that of an adult even when height is not a factor.

\cite{lu2010gait} studied the difference in the age-related biometric component of gait across both genders using a subset of the USF dataset consisting of 79 subjects with ages ranging from 19 to 59 years. GEI \citep{man2006individual} was adopted for feature extraction\footnote{The GEI is explained in detail in Section~\ref{sec:gei-based}} and multi-label guided subspace was applied to reduce the dimensions while the $k$-Nearest Neighbour ($k$NN) algorithm was used for the classification. The age information was encoded as a sequence of 8 binary labels, one of which was also the gender. One $k$NN was trained for each of the labels in the sequence. The label sequence was decoded to obtain the estimation of the age of the subject. The mean absolute error was reported to be approximately 5.6 years.

\cite{makihara2011gait} conducted a much more extensive experimentation for age estimation wherein the gait instances from 1728 subjects with the ages between 2 to 94 years were recorded. The Gaussian Process Regression was considered to be a highly efficient face-based age estimation algorithm \citep{rasmussen2006gaussian} and it was adopted by \cite{makihara2011gait} for gait-based age estimation. They have concluded that the performance of frequency domain features and the GEI features are equal with a mean absolute error of 8.2 years.

\subsection{Gender Classification}
\label{sec:gender-class}

Gender is found to be the most popular soft biometric that can be derived from human gait. The first major work on gait-based gender recognition was done by \cite{huang2007gender}. The silhouette of each frame in the gait cycle was segmented up into seven regions, and an ellipse was fit to each of them. The geometric parameters of the ellipses collectively become the feature set of the gait instance. The similarity measures between male and female instances were trained using an SVM classifier to form the recognition system. A similar method of segmenting was suggested by \cite{li2008gait}. Only this time, the GEI of the gait instance was segmented instead of the individual silhouettes. The GEI template was split into 6 regions whose boundaries were obtained from the analysis of the training instances. A simple pixel-wise similarity measure was adopted to form the feature space which was reduced through independent component analysis (ICA) and classified through SVM. \cite{yu2009study} also split the GEI to multiple components based on body parts. Additionally, they assigned weights to each region which was then fed to SVM for classification. The weights signify the effectiveness of gender discriminability to the respective component.

\cite{hu2011gait} adopted an entirely different method for spatiotemporal feature extraction to compiling gait templates such as the GEI. The process involves extracting shape features from multiscale grids. Their method reports the best performance till date. As the previous methods were constrained in terms of the angle of view (mostly at the sagittal angle), \cite{lu2014human} have designed a method that is said to work in arbitrary view angles. They propose the cluster-based averaged gait image (C-AGI) as an alternative to the GEI to cope with multiple walking directions. Sparse representation metric learning and classification \citep{wright2009robust} was used for gender prediction.

Apart from the studies conducted with standard video cameras, there were also research done with three-dimensional data in connection with gender identification from gait. \cite{igual2013robust,kastaniotis2013gait} used Microsoft Kinect \citep{kinect} to get depth information along with the RGB visual information. As the data obtained was computationally large, both required the use of extensive feature reduction before classifying with SVM. \cite{flora2015improved} claim to improve the recognition accuracy with the help of motion capture information. The work in this thesis focuses on the data obtained from traditional cameras as it is possible to obtain optimal classification rates without the use of depth information. Furthermore, algorithms designed for conventional two-dimensional cameras have a much greater scope as they can be readily augmented with legacy surveillance framework with minimal modification.

\section{Hard Biometrics}
\label{sec:hard-biometrics}

\begin{figure}[t]
  \centering 
  \includegraphics[width=0.8\linewidth]{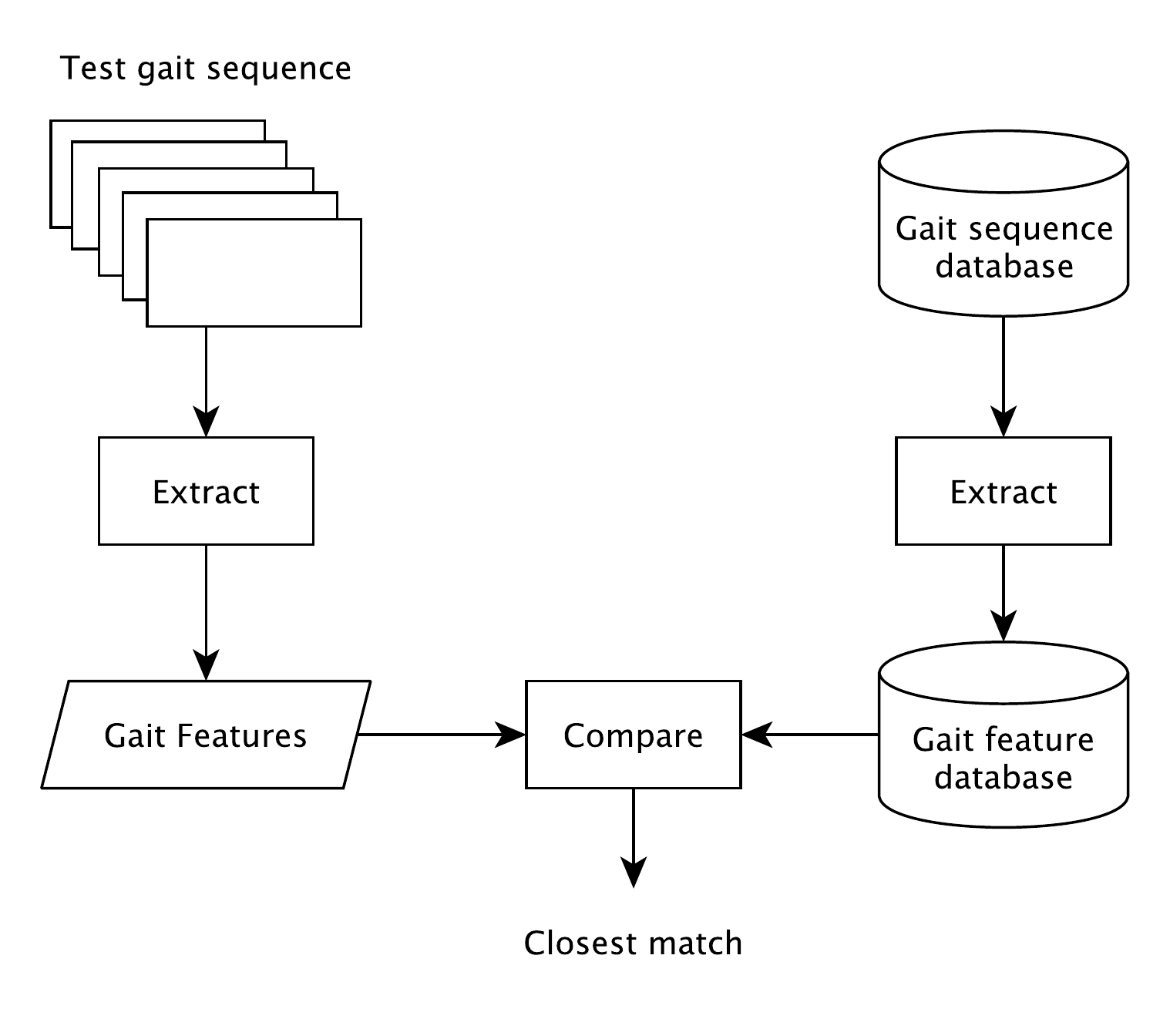}
  \caption{Simplified gait recognition system}
  \label{fig:gaitrecog}
\end{figure}

Gait is characteristic to one's individuality it is considered to be an accepted form of unobtrusive biometrics \citep{boyd2005biometric}. Hence, in theory, it does not require the cooperation or the awareness of the individual being observed. Some researchers still argue that gait itself as a soft biometric characteristic \citep{li2017deepgait,Arora2015}, most researchers today have proven that a gait signature can be established as a hard biometric. A simplified template for a gait recognition system is as shown in Figure~\ref{fig:gaitrecog}. A gait sequence is a temporal record of a person's gait. 
These are usually translated into a features database after some means of preprocessing and feature extraction. When the test gait sequence is given for identification, the system refers the gait feature database and returns the closest match (if any) according to some criteria.

The approaches for gait recognition can be broadly classified into two types: model-based and model-free. Due to its simplicity and efficiency in use, model-free methods are more prevalent. Model-free techniques can be further categorized as template-based and non-template methods.

Many benchmark datasets are available to compare the performance of
one algorithm with another. Recent literature  commonly use
the USF Gait Challenge dataset and 
CASIA-B. 
The aforementioned and the remaining available datasets for gait biometrics will be
detailed in Section~\ref{sec:datasets}.

The recognition process involves the use of an established machine learning algorithm such as the HMM (used in \cite{Zongyi_L, kale2002framework, liu2006improved, Zhang_R, Nguyen2014}). Some apply different techniques such as canonical analysis \citep{Huang_PS,Foster_JP}, spatiotemporal correlation \citep{Murase_H}, and DTW \citep{tanawongsuwan2001gait,lam2011gait,Choudhury_SD}. A few others devise newer techniques for closeness representation. All others employ the nearest neighbour classifier \citep[recent examples:][]{Xing2015, Arora2015}.

\subsection{Model-free Techniques and Early Approaches}
\label{sec:spatio}

The vast majority of model-free techniques tend to have a strong reliance on spatiotemporal analysis of silhouettes of the individual during gait. A spatiotemporal analysis takes into account the variation in the spatial domain with respect to that in the time domain. So when this is applied to gait recognition, the analyses involves the observation of the spatial locations of body parts and their movement in different stages in time.

Almost all model-free methods have background subtraction and silhouette extraction as the first step. Background subtraction is a simple method in which the change in pixel values between one frame and the successive frame is observed to bring out only the objects that are seen in motion. From these objects, the moving human silhouette can be extracted. The result of background subtraction is usually binarized in which the moving object seems to be white and the background is black, or vice versa in some cases. The novelty mainly lies in how the features are extracted. 

\subsubsection*{Earlier model-free methods}
\label{sec:earlier-methods}

The earliest known spatiotemporal gait recognition techniques started in the late 80s. \citep{niyogi1994analyzing} proposed to recognize gait at a sagittal angle with the subject walking frontoparallel. It modelled the human gait in the form of a set of spatiotemporal snakes \citep{kass1988snakes} from the slices of the moving parts of the human contour along the time domain. 
The recognition accuracy they have obtained with 26 image sequences across five human subjects reaches up to 83\%. \cite{little1998} introduced the concept of optical flow in gait recognition. In principle, the points in the image sequence that vary with time tend to oscillate periodically during the subject's gait. By observing the optical flow of these points, the $m$ time varying scalars can be produced. The phases of oscillations from these scalars are used to represent the gait instance. They reached a higher recognition rate of up to 95\% by testing their technique with seven instances for each of six human subjects.

An earlier version of template matching method was proposed by \cite{Huang_PS}. They used the Eigenspace transformation (EST), as adopted by \cite{Murase_H}, to convert the gait taken as a sequence of images to a template called the `eigengait'. On top of this, canonical space transformation was applied with generalized Fisher linear discriminant function to separate the classes boundaries required for prediction. However, they also seemed to use the small dataset used by Little and Boyd for their application to show a questionable accuracy of 100\%. 

 
Though the above methods show attractive recognition rates, all the methods proposed at those times before suffered one major drawback: their accuracies were biased to their samples which were too small when considering the application as a biometric. The advent of the DARPA's HumanID programme in the break of the second millennium led to a major development in gait biometric research. Bigger datasets with more variations challenged researchers to design better algorithms.

\subsection{Model-free Template-based Methods}
\label{sec:gei-based}

A \textit{gait template} can be obtained as the result of transforming the sequence of silhouette images taken from a gait video to a single image that holds the composition of the motion-related features of the sequence. Some of the notable template-based gait recognition methods in literature are listed in Table~\ref{tab:template-based}.

\cite{Hayfron_A} assessed the symmetry of the extracted silhouette using a generalized symmetry operator. 
The contours obtained from the silhouette sequence were used to produce a symmetry map. Euclidean distance between Fourier descriptors was used as a similarity measure for gait recognition. They reported a CCR of 97.3\% for $k=1$ and 96.4\% $k=3$ using the nearest neighbour classifier.

\cite{wang2003automatic} proposed a unique method to recognize gait by analysing the contours of the silhouettes. The shape of the contour of a given silhouette sequence was converted to a template with the use of Procrustes shape analysis. Different exemplars were created for each viewpoint. They tried three types of nearest neighbour algorithms, viz., NN, $k$NN, and ENN (NN with class exemplar), in which ENN provided the best results for gait recognition.

Experimental results of \cite{cuntoor2003combining} suggests that DTW and HMM can be combined to produce a better gait recognition result. DTW was used to align the motion of the arms and legs to normalize the phase of gait while HMM was used to define the leg dynamics.

\begin{landscape}
  \begin{table}
    \centering
    \renewcommand{\arraystretch}{1.1}
    \caption{Template-based Methods for Gait Recognition}
    \resizebox{\linewidth}{!}{
    \begin{tabular}{rcllll}
      \toprule S.No. & Year & Study & Technique & Plane & Dataset \\
      \midrule
      \begin{filecontents*}{Data/GR-T.csv}
Study,Year,Type,Technique,Plane,Dataset
Niyogi et al. \nocite{niyogi1994analyzing},1994,TS,Fitting of spatiotemporal snakes,Sagittal,$5s \sim 26$ seq.
Murase \& Sakai \nocite{Murase_H},1996,TS,Parametric eigenspace transformation,Sagittal,$7s \times 10i = 70$ seq.
P. Huang et al. \nocite{Huang_PS},1999,TS,Canonical and eigenspace transformation,Sagittal,UCSD
L. Wang et al. \nocite{wang2003automatic},2003,TS,Procrustes shape analysis,Multiview,CASIA-A
S. Sarkar et al. \nocite{sarkar2005humanid},2005,TS,Gait Baseline,Sagittal,{USF, CMU}
Ju Han \& Bir \nocite{man2006individual},2006,TS,Gait Energy Image (GEI),Sagittal,{USF v1.7, v2.1}
T. Lam et al. \nocite{Lam_T},2007,TS,MSCT and SST,Sagittal,{USF v1.7, v2.1, SOTON}
X. Yang et al. \nocite{Yang_X},2008,TS,Enhancing GEI through dynamic regions,Sagittal,USF v2.1
Kusakunniran et al. \nocite{Worapan2009auto},2009,TS,Weighted Binary Pattern (WBP),Sagittal,{CMU, CASIA-A,C}
Bashir et al. \nocite{bashir2010gait},2010,TS,Gait Entropy Image (GEnI),Sagittal,{CASIA-B, SOTON}
E. Zhang et al. \nocite{zhang2010active},2010,TS,Active Energy Image (AEI),Sagittal,{CASIA-B, CASIA-C}
T. Lam et al. \nocite{lam2011gait},2011,TS,Gait Glow Image (GFI) using optical flow,Sagittal,USF v2.1
Zheng et al. \nocite{zheng2011robust},2011,TS,View transformation model using GEI,Multiview,CASIA-B
C. Wang et al. \nocite{wang2012human},2012,TS,Chrono-Gait Image (CGI),Sagittal,{USF, SOTON, CASIA-B}
N Liu et al.\nocite{liu2013robust},2013,TS,GEI-based Multiview Subspace Representation,Multiview,{CASIA-B, CMU-MoBo}
Dupuis et al. \nocite{dupuis2013feature},2013,TS,GEI subsetting using random forests,Multiview,CASIA-B
Hofmann et al. \nocite{tum_gaid_paper},2014,TK,Depth Grad. Hist. En. Img. (DGHEI) with audio,Sagittal,TUM-GAID
Aqmar et al. \nocite{aqmar2014gait},2014,TS,Gait Fluctuation Image (GFlucI),Sagittal,{OU-ISIR-D, OU-ISIR-LP}
P. Arora et al. \nocite{Arora2015},2015,TS,Gait Information Image (GII) features,Sagittal,{OU-ISIR, SOTON, CASIA-B}
X. Xing et al. \nocite{Xing2015},2015,TS,Canonical correlation analysis of GEI,Multiview,CASIA-B
P Yogarajah et al. \nocite{Yogarajah20153},2015,TS,Joint sparsity model and $l1$-norm`,Sagittal,CASIA-B
S.D. Choudhury et al \nocite{choudhury2015robust} ,2015,TS,View-invariant multi-scale  gait recognition,Multiview,CASIA-B
Zhao et al. \nocite{Zhao2015},2016,TS,Walking Path Image (WPI),Multiview,CASIA-B
I. Rida et al. \nocite{rida2016human},2016,TS,GEI segmentation using group lasso motion ,Multiview,CASIA-B
K. Shiraga et al. \nocite{shiraga2016geinet},2016,TS,GEINet -- CNN on GEI  ,Multiview,OU-ISIR-LP
C. Li et al. \nocite{li2017deepgait},2017,TS,DeepGait -- VGG-D on GEI with Joint Bayesian,Multiview,OU-ISIR-LP
Z. Wu et al. \nocite{wu2017comprehensive},2017,TS,{Ensemble of CNN, GEI and temporal features},Multiview,{OU-ISIR-LP, CASIA-B}
\end{filecontents*}
\csvreader[late after line=\\]{Data/GR-T.csv}
      {Study=\study, Year=\y, Type=\typ, Technique=\tech, Plane=\pl, Dataset=\ds}
      {\thecsvrow. & \y & \study & \tech & \pl & \ds} \bottomrule
    \end{tabular}}
    \label{tab:template-based}
  \end{table}
\end{landscape}

The efforts of the University of South Florida \citep{sarkar2005humanid} has brought forth a new revolution to gait recognition by compiling the largest dataset of its time, the USF Gait Challenge. It was categorized to 12 challenge probe sets for experimentation and a gallery set for training. A simple baseline algorithm was developed to facilitate users of the dataset to compare the performance of gait algorithms effectively. The algorithm involves the use of the Tanimoto similarity measure to gauge the similarity between two silhouettes which is given by $Sim(p,q) = |p \cap q|/|p \cup q|$. Here, $p$ and $q$ are two binarized images where each image is represented as an ordered set of pixel values. 
The correlation between the silhouette similarities provided the measure of closeness used for the recognition step. Further details of the USF Gait Challenge dataset is provided in Section~\ref{sec:datasets}.

\begin{figure}[t]
  \centering
  \includegraphics[width=\linewidth]{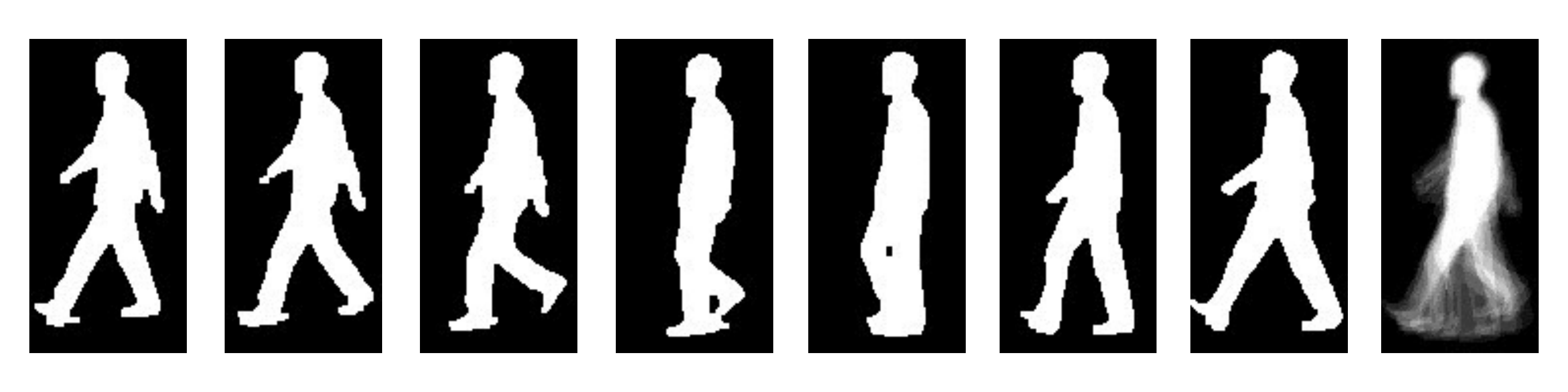}
  \caption{The Gait Energy Image of a gait sequence}
  \label{fig:GEIseq}
\end{figure}

The most notable form of silhouette-based gait recognition techniques use the production of a Gait Energy Image (GEI) template \citep{man2006individual} from the gait cycle. Technically, the GEI shows how energy is dissipated spatially through the stages of the gait cycle. It is so prevalent in literature such that silhouette based methods that are published after its time (2006) can be classified either as GEI-based or non-GEI-based. The GEI is created by superimposing the pixels of silhouette sequence of a given gait by summing the values and averaging them resulting in the output as a grey-level image of proportional pixel intensities.
Hence the final stage boils down to image comparison of the test GEI with the GEIs in the gait database. An example of a GEI obtained from a subject's gait from the CASIA-B dataset along the sagittal plane is as shown in Figure~\ref{fig:GEIseq}. The features are learned based on Multiple Discriminant Analysis (MDA)~\citep{duda2001pattern} after passing through Principle Component Analysis (PCA)~\citep{jolliffe2002PCA} for dimensionality reduction. The Euclidean distances towards the class centres with respect to the features provide the closeness measure for recognition. Experimental results of \cite{man2006individual} show recognition rates averaging around 55.64\% over all probes of the USF Gait Challenge v2.1 database, but normal recognition rates reach from 90\% to 100\%.



Although it was a a significant breakthrough in gait biometrics, the GEI had one major setback, namely, covariate factors. When the subject, for instance, wears an overcoat, the CCR can drop to less than 50\%. Many studies then published their version of an energy image adopted from the core principles of the GEI to overcome this drawback. They differ mainly in the collation operation -- the operation used to combine the silhouette sequence to a single image. Some of the notable ones are given as follows.

\begin{itemize}[topsep=0pt, noitemsep, leftmargin=*]
\item Gait Flow Image (GFI) incorporates optical flow \citep{lam2011gait}
\item Enhanced GEI considers only dynamic regions \citep{Yang_X}
\item Gait Fluctuation Image (GFlucI) highlights temporal irregularities in gait \citep{aqmar2014gait} 
\item Chrono-Gait Image (CGI) maintains temporal information through colour \citep{wang2012human}
\item Motion History Image (MHI) for activity recognition \citep{ahad2012motion} was adapted for gait recognition \citep{Nguyen2014}
\item Gait Entropy Image (GEnI) incorporates Shannon's entropy function for each pixel \citep{bashir2010gait}
\item Active Energy Image (AEI) takes the average of each frame-to-frame difference \citep{zhang2010active}
\item Depth Gradient Histogram Energy Image (DGHEI) combines depth information from Kinect \citep{hofmann20122}
\item Gait Information Image (GII) \citep{Arora2015} incorporates Hanman-Anirban entropy function \citep{hanmandlu2011content}
\end{itemize}

Instead of coming up with new types of templates, it would be wiser to take only a subset of the gait template that is resilient to covariate factors. \cite{dupuis2013feature} were among the first to propose this model. They formulated a single mask through the ranking of pixel features using the Random Forests classifier. Their panoramic gait recognition (PGR) algorithm used pose estimation for view prediction. \cite{choudhury2015robust} designed a method named view-invariant multiscale gait recognition (VI-MGR) which applied Shannon's entropy function to only the lower limb region of the GEI. The sub-region selection was later modified by \cite{rida2016human} by automating this segmentation procedure with a process known as group lasso of motion (GLM). Their approach to the problem has shown significant improvement in the covariate recognition accuracy.

Recently, the success of deep learning in the fields of image and activity recognition has motivated its application to gait recognition. The basic feature representations of these methods are also gait templates. \cite{shiraga2016geinet} employed a CNN (Convolutional Neural Network) to define the degree of dissimilarity between two GEI templates, i.e., a probe GEI and a gallery GEI. Deep learning requires a lot of training samples, so they used the largest gait dataset publicly available which is the OU-ISIR-LP dataset. This method, named GEINet, produced an accuracy of 95\% for same angle and 80-95\% for cross angle performance. \cite{li2017deepgait} proposed the DeepGait design to outperform GEINet using the VGG \cite[Visual Geometry Group,][]{vgg} deep convolution model and Joint Bayesian model for view invariance. Using the same dataset, the DeepGait achieved gait identification rates of up to 98\% and its cross-view identification accuracy range from 88 to 98\%. A more thorough experimentation using CNN was conducted by \cite{wu2017comprehensive} on both OU-ISIR-LP and CASIA-B datasets. They used three different CNN configurations for gait identification. Their experimentation revealed that the ensemble of the networks with GEI and additional temporal features gave the best accuracy. Though their cross-view normal recognition performance became the state-of-the-art, their accuracy for covariate factors are relatively low. The average CCRs with SetB ranged from 80 to 90\%, and SetC ranged from 60 to 75 across view angles.

As attractive as deep learning sounds, its integration with legacy camera networks is not so simple due to its computational complexity. Moreover, the experiments of \cite{choudhury2015robust} and \cite{rida2016human} revealed that it is possible to attain close to ideal gait recognition accuracy without the need for a deep learning framework.

\subsection{Model-free Non-template Methods}
\label{sec:nontemplate-based}

Though found to be efficient in practice, not all silhouette-based methods in literature involve the production of a template image. Notable non-template methods are shown in Table~\ref{tab:non-template}. \cite{Foster_JP} have claimed to have attained recognition rates above 75\% by
running experiments on 114 subjects of the SOTON video gait dataset. Their
method monitors the temporal changes in the areas of the clipped gait window
segmented by masked sectors. Using these time-varying area metrics, they
formulate a feature vector for recognition.


In the work of \cite{Nikolaos_VB}, accounted the average distance from the centre of the silhouette. Each silhouette was represented as a feature vector which was composed of a sequence of angular transforms made on segmented angular slices $\Delta\theta$. 
In the USF gait challenge dataset, a period in the gallery (reference) is not equal to that of the probe sequence (test). Hence, a technique called linear time normalization was utilized to make the feature matrix of each probe and gallery sequence comparable by compensating for the difference in the sequence length. The same method was used in their next paper \citep{Boulgouris} but with a different feature extraction technique. In this work, the body as depicted in the silhouette was segmented into body components. The components were ranked with according to their proportional relevance during the comparison operation.

\cite{Zongyi_L} promoted the fusion of face and gait biometrics. 
The exemplars, in this case, are 
obtained from 
specified stances analogous to the formally depicted gait cycle. These stance frames were modelled using a population EigenStance-HMM; a method that was extended from their previous technique, population HMM \citep{liu2006improved}. 
A given gait silhouette sequence can be segregated into these stance models by $k$-means clustering with Tanimoto distance as the distance measure. A cyclic Bakis variant of HMM was modelled for the gait recognition. This gait recognition algorithm was used in combination of face recognition using Elastic Bunch Graph Matching (EBGM) based on Gabor features to attain a much higher performance over the Gait Challenge baseline algorithm when compared using the toughest sets of the USF Gait Challenge dataset.

A gait recognition that analyses both shape and motion was proposed by \cite{Choudhury_SD}. The gait period here was depicted as ten phases. Spatiotemporal shape features are obtained from these phases in the form of Fourier descriptors. The silhouette contour at each step was segmented by fitting ellipses. The similarity was then calculated by utilizing Bhattacharyya distance between the ellipse parameters taken as features. DTW was applied to compare leading knee rotation with relevance to arm swing pattern over a gait cycle. DTW was used to counteract the effects of walking speed, clothing, shadows, and hair styles.

\begin{landscape}
  \begin{table}
    \centering
    \renewcommand{\arraystretch}{1.2}
    \caption{Non-template Methods for Gait Recognition}
    \resizebox{\linewidth}{!}{
    \begin{tabular}{rcllll}
      \toprule S.No. & Year & Study & Technique & Plane & Dataset \\
      \midrule
      \begin{filecontents*}{Data/GR-N.csv}
Study,Year,Type,Technique,Plane,Dataset
J. Little \& J. Boyd \nocite{little1998},1998,NS,Phase difference of optical flows,Sagittal,UCSD
Johnson \& Bobick \nocite{johnson2001multi},2001,NS,Static body parameters,Multiview,Georgia Tech
A. Kale et al. \nocite{kale2002framework},2002,NS,Frame to exemplar distance (FED),Sagittal,{CMU, UMD}
J. P. Foster et al. \nocite{Foster_JP} ,2003,NS,Temporal change in segmented areas,Sagittal,SOTON: $28s \times 4i = 112$
L. Wang et al. \nocite{wang2003silhouette},2003,NS,PCA-based eigenspace transformation,Multiview,CASIA-A
A. Kale et al. \nocite{kale2004identification},2004,NS,FED with outer contour width,Sagittal,{CMU, USF, UMD}
R. Wang et al. \nocite{wang2004framework},2004,NS,Joint-tracking via MAP hypothesis,Multiview,UMD
N. V. Boulgouris et al. \nocite{Nikolaos_VB} ,2006,NS,Linear time normalisation,Sagittal,USF
Zongyi Liu \& S. Sarkar \nocite{liu2006improved},2006,NS,Population HMM,Sagittal,{USF, CMU, UMD}
N. V. Boulgouris et al. \nocite{Boulgouris} ,2007,NS,Matching of body components,Sagittal,USF
Zongyi Liu \& S. Sarkar \nocite{Zongyi_L},2007,NS,Population EigenStance-HMM,Sagittal,{USF, CMU}
Choudhury \& Tjahjadi \nocite{Choudhury_SD},2013,NS,Shape and motion analysis of contours,Sagittal,USF
Nguyen et al. \nocite{Nguyen2014},2014,NS,HMM of motion history image (MHI),Sagittal,CASIA-A
Chin P. Lee et al. \nocite{Lee2015},2015,NS,Transient binary patterns (TBP),Sagittal,{CASIA-B, OU-ISIR-D, CMU}
Zeng \& Wang \nocite{zeng2016view},2016,NS,Periodic deformation of human gait shape,Multiview,{CASIA-B, CMU-MoBo}
Muramatsu
 et al.\nocite{muramatsu2016view},2016,NS,Frequency domain features of joint-subspace,Multiview,OU-ISIR-LP 
\end{filecontents*}
\csvreader[late after line=\\]{Data/GR-N.csv}
      {Study=\study, Year=\y, Type=\typ, Technique=\tech, Plane=\pl, Dataset=\ds}
      {\thecsvrow. & \y & \study & \tech & \pl & \ds} \bottomrule
    \end{tabular}}
    \label{tab:non-template}
  \end{table}
\end{landscape}

\subsection{Model-based Techniques}
\label{sec:joint-traj-meth}

While basic spatiotemporal methods give cues as to how the body position vary with accordance to time as a whole, it would be much more accurate to do a spatiotemporal analysis on all articulation points separately. The implementation of that sort falls under the category of joint trajectory or model-based methods. That is, the trajectory of each joint is tracked live and analysed as individual components; efforts are made to model the human structure accurately. Table~\ref{tab:model-based} list some of the model-based gait recognition literature.

Bulb markers can be attached to certain points on the body considered necessary for gait analysis such as ankles, knees, hands, elbows, shoulders and torso. When observed from a camera with a low exposure, only the bulb illumination can be perceived. This method facilitates an easier and a more accurate analysis of gait through computer vision. \cite{tanawongsuwan2001gait} implemented this by strapping 18 human subjects with 16 markers at appropriate locations and projected their gait at a sagittal angle. The sensors were able to reconstruct a mobile skeletal structure recovered from the joints. From this data, they were able to assess the articulation points over time. The variance in time was normalized by applying DTW, and the recognition was based on the nearest neighbour algorithm to produce a modest recognition rate of 73\%.

{
A study by \cite{geradts2002use} was conducted to extract gait-related parameters from all three planes -- frontal, transverse and sagittal -- from surveillance cameras. 11 human subjects participated in the experiment and involved the use of 11 bulb markers fitted to the necessary points to each subject. They were able to observe various parameters from step length, cycle time and cadence to joint angles and spatial positioning. After a simple analysis of variance (ANOVA) on the gait parameters extracted, it seemed like the foot angle exhibited the most variance and then the time average hip joint angles followed by
\unskip\parfillskip 0pt \par}

\begin{landscape}
  \begin{table}
    \renewcommand{\arraystretch}{1.2}
    \caption{Template-based Methods for Gait Recognition}
    \resizebox{\linewidth}{!}{
    \begin{tabular}{rclllll}
      \toprule S.No. & Year & Study & Type & Technique & Plane & Dataset \\
      \midrule
      \begin{filecontents*}{Data/GR-M.csv}
Study,Year,Type,Technique,Plane,Dataset
Rawesak T. \& Bobick \nocite{tanawongsuwan2001gait} ,2001,Marker,Time-normalized joint trajectories,Sagittal,Georgia Tech
Z. J. Geradts et al. \nocite{geradts2002use},2002,Marker,Multi-planar gait parameter extraction,Multiview,$11s \times 2i = 22$ seq.
Rawesak T. \& Bobick \nocite{tanawongsuwan2003perf} ,2003,Marker,Speed-invariant gait parameters,Sagittal,$15s \times 36i = 540$ seq.
R. Zhang et al. \nocite{Zhang_R},2007,Silhouette,Five-link biped human model,Sagittal,{USF, CMU}
M. Goffredo et al. \nocite{Goffredo2008},2008,Silhouette,Geometric marker-less joint estimation,Multiview,CASIA-B
P. Chattopadhyay et al. \nocite{Chattopadhyay20159},2015,Kinect,Combining depth and silhouette features,Frontal,$29s \times 8i = 232$ seq.
D. Kastaniotis et al. \nocite{Kastaniotis2015},2015,Kinect,{Euler angles, dissimilarity space mapping},Oblique,{UPCVGait}
\end{filecontents*}
\csvreader[late after line=\\]{Data/GR-M.csv}
      {Study=\study, Year=\y, Type=\typ, Technique=\tech, Plane=\pl, Dataset=\ds}
      {\thecsvrow. & \y & \study & \typ & \tech & \pl & \ds} \bottomrule
    \end{tabular}}
  \label{tab:model-based}

  \vspace{1em}
  Model-based features can either be extracted through 3D cameras with the subject fitted with IR or bulb \textit{markers}, articulation points inferred from \textit{silhouettes}, or depth information from Microsoft \textit{Kinect}.
  \end{table}
\end{landscape}

\noindent the step length. Hence these are considered to be the best parameters to be used for recognition. However, their research concluded that the gait analysis cannot be used for identification at that time (the year 2002).


\cite{tanawongsuwan2003perf} later produced a study on the recognition of gait in different speeds of walking, but this time, they resorted to a more comfortable reflective suit for the articulation point signal extraction. The experimentation is described in \cite{tanawongsuwan2003study}. A 12-camera VICON MoCap system was used for the 3D motion analysis. The result based on 15 human subjects concluded that a positive linear correlation could be observed between cadence and speed, and a negative exponential correlation could be observed between stride time and speed.

The methods described so far require the use of complicated hardware for better accuracy. They either require reflectors, bulb markers, or magneto sensors to be fitted on to the points of interest of the human subject to gather the point-light information during his/her gait. Due to their nature, these methods are not practically feasible for a biometric application. It is to note that the concept of being an `unobtrusive' means of biometrics is lost here.

Not all model-based techniques, however, are impractical in application. \cite{Zhang_R} show that it is possible to extract a five-link biped human model from a two-dimensional video feed to produce a model-based gait recognition system. 
The Metropolis-Hastings algorithm \citep{metroHastings} was used for the feature extraction. 
An innovative yet simple method for locating the articulation points of the lower limb joints was implemented by \cite{Goffredo2008}. By making smart estimates on where to initialize the points, the point-light data for the hip, both knees and ankles were extracted.
This method was able to extract these points over multiple views as provided by the CASIA-B video database. By recording the temporal changes of these points, a profile recorded could be recognized with the help of the $k$NN algorithm.

A standard video feed would provide a two-dimensional data for processing. When added with depth-based information, more accurate conclusions can be drawn to aid recognition. Microsoft Kinect provides this functionality. The Kinect was used by \cite{Chattopadhyay20159} and \cite{Kastaniotis2015} by facilitating three-dimensional data flow for a more natural and efficient biometric gait recognition. Methods using the Kinect were found to be more successful than using silhouettes alone for model-based analysis.

\subsection{Covariate Factors}
\label{sec:covariate-factors}

Covariate factors are intra-class variations that inhibit the effectiveness of biometric gait recognition by confounding the features that can be observed \citep{boyd2005biometric}. The following briefly explains each factor with relevant research on how they deviate the parameters of gait.

\noindent\textbf{Walking surface}. Studies show that when the surface is unstable such as a wet surface, the walking speed, toe-in angle and step length are significantly reduced to retain control over balance \citep{menant2009effects}. However, when walking on other irregular surfaces like grass, foam or studded with small obstacles, walking speed can be maintained with a variable cadence and a longer stride length \citep{menz2003acceleration}. Even two regular surfaces such as vinyl and carpet have a significant difference \citep{RozinKleiner2015147}. In slippery surfaces, reductions can also be observed in stance duration, load supporting foot, normalized stride length \citep{cham2002changes}.

\noindent\textbf{Footwear}. When the footwear is considerably different, so is the gait of the individual. This aspect especially concerns high heel users. To maintain their stability, people wearing high heels require more control over their centre of mass \citep{Chien20141045} and tend to have longer double support times \citep{menant2009effects}. Recent studies show that habitually shod walkers and habitual barefoot walkers exhibited a significant gait difference when switching their footwear in terms of their stride length and cadence \citep{franklin2015barefoot}.
 
\noindent\textbf{Injury}. When any portion of the lumbar region or lower limb is injured, the individual naturally adopt an antalgic gait. The individual walks in a way so as to avoid the pain caused by the injury. This style of walking restricts the range of motion of the associated limb leading to a deviation in the usual gait proportional to the magnitude of the injury. In-depth research on how an injury can affect gait was provided by \cite{wang2012biomechanical}.

\noindent\textbf{Muscle development}. The development of muscles gives a different range of control over the parts of the body that affect gait. The sheer mass of the  developed muscles alters the centre of mass at the associated mobile limb as well as the body itself. The shift depends on the difference in mass. The change in the centre of mass can modify the inclination of pressure required for proper stability \citep{lee2006detection}. An extensive study on the correlation between muscle mass with gait performance \citep{beaudart2015correlation} proves that muscle mass directly relates to gait speed, especially in the case of geriatric patients.
 
\noindent\textbf{Fatigue}. When the individual is subjected to fatigue, the stability of the concerned gait decreases while a noticeable increase in variability of gait is exhibited \citep{vieira2016effects}. The time taken for the recovery towards normal gait depends on the extent of exertion applied to the individual so as to get to a fatigued state and the individual's stamina. This aspect was observed from Ashley Putnam's thesis \citep{putnam2013effects} in which a study with army cadet treadmill protocol was conducted to analyse the effect of exhaustion on gait mechanics and possible injury. Cadets ran till exhaustion and had their gaits observed. The resulting gait had both inter- and intra-variable vertical stiffness in the lower limbs.

\noindent\textbf{Training}. When the individual is subjected to some form of physical training, it is possible that her/his gait is also subjected to change. This change can be evident as a result of military training, prolonged load condition, prolonged use of particular footwear and athletic training.

\noindent\textbf{Extrinsic control}. Humans have an ability to control their gait to an extent so as to differ from their usual gait. A person can mince walk, and depending on how self-aware the individual may be she/he can walk with a swagger or strut, a brisk walk or tip-toe. Another matter to note is the level of awareness the individual has of his/her surroundings. Hence, individuals tend to alter their gait to the extent that is determined by their self-control. This concept explains how members of the army can synchronize gait during a march.

\noindent\textbf{Intrinsic control}. Some elements can control a person's gait in such a way that the individual is sometimes unaware of the change that takes place. The best example of this case is the emotional response or mood of the individual: state of happiness, sorrow, anger or any other emotion strong enough to make an impact on one's gait. The variability can range from subtle to significant and can vary from one person to another. Related studies are provided in \cite{montepare1987identification,venture2014recognizing} and \cite{schneider2014show}.

\noindent\textbf{Age}. Although the factor may not contribute to change in gait over a short period, it certainly does influence gait to a large extent. Ageing, in general, causes musculoskeletal and neuromuscular losses. To compensate for these losses, the individual makes certain adjustments which can be observed in the individual's gait \citep{MonizPereira2012S229}.

\noindent\textbf{Clothing}. While the change of clothing does not necessarily modify the gait for slight differences in weight, it might, however, show changes in the associated silhouettes. This change would affect a major portion of gait recognition algorithms that depend on the spatial configuration of silhouettes. However, a greater change in the weight of the clothing, such as a winter suit, has a higher probability of affecting the gait itself.

\noindent\textbf{Load}. The effect of load can significantly influence gait. In a loaded condition such as wearing a backpack, the individual is subjected to a higher weight in addition to his/her body weight. The foot exerts higher pressure during plantarflexion to regulate locomotion generating a greater ground reaction force than the unloaded condition \citep{Castro201541}. Apart from the pressure applied, the body must cope with the change in balance for a stable gait \citep{mummolo2016computational}. The load can also be asymmetrical, such as a wearing a handbag, cross-bag or shoulder bag, or carrying a suitcase. In this case, a greater difference in the pelvic rotation is observed \citep{hyung2016influence}. The body shifts the pelvic movement so as to counteract the imbalance caused by the load.

\subsection{View Invariance}
\label{sec:view-invariance}

In addition to clothing and carrying conditions, the view angle is found to be the most important covariate factor that affects gait recognition performance \citep{zeng2016view, huang2008human, cilla2012probabilistic, liu2013robust}. The range of features that can be extracted can wary widely between the angles of observation. There are essentially two types of view-invariant gait recognition models: view transformation model (VTM) and view-preserving model (VPM).


\noindent\textbf{VTMs} \citep{kusakunniran2010support, kusakunniran2012gait, Zhao2015} transform the probe sequence's angle to match with that of the gallery sequence. The VTM methods may differ in the measures used to gauge the transformation accuracy \citep{muramatsu2016view}. However, a significant level of error is inevitable in VTM-based gait recognition \citep{nini2011, dupuis2013feature}.


\noindent\textbf{VPMs} consider multiple views as part of the gallery itself. This process incorporates the view information within the feature set for the extraction of relevant view-invariant gait features. Various methods can be employed to facilitate this. Examples include varying width vectors \citep{zeng2016view}, Grassmann manifold \citep{connie2016grassmannian}, geometric view estimation \citep{jia2015view}, and spatiotemporal feet positioning \citep{verlekar2016view}. A variant of VPM involves extraction of view-independent features through multi-view training and then use a single gallery view for testing \citep{nini2011,tang2017robust}.

\section{Datasets}
\label{sec:datasets}

With the outbreak of a substantial amount of algorithms to analyse gait there also comes a need to compare them. A standard gait dataset would be able to serve means so as to benchmark such algorithms especially in the case of biometric application. The categories of datasets cover a wide range based on the needs of the gait analyst from lightweight datasets to large-scale databases. An overview of the gait databases described here is given in Table~\ref{tab:dataset}.

\begin{table}[!h]
\centering
  \caption{Datasets for Biometric Gait Analysis}
  \label{tab:dataset}
  \resizebox{\linewidth}{!}{
  \renewcommand{\arraystretch}{1.4}
  \begin{tabular}{
    p{0.4\linewidth} 
    c 
    c 
    r 
    r 
    r 
    l 
    } 
    \toprule
    Database & Year* & Env & View  & Sub. & Seq. & Other covariates \\
    \midrule
    \begin{filecontents*}{Data/Datasets.csv}
Database,Year*,Env,View,Covariates,Sub,Seq,Sequences
UCSD \newline \cite{little1998},1998,FO,1,,6,42,$6s \times 7i = 42$
MIT AI \newline \cite{lee2002gait},2001,FI,4,{Time},24,194,$24s \sim 194$
CMU MoBo \newline \cite{cmu_mobo},2001,TI,6,{Speed(2), load(2), inclination(2)},25,600,$25s \times 4i \times 6v = 600$
HID Georgia Tech \newline \cite{gtech},2001,FIO,3,{Time},20,$>$280,$20s \sim 280+$
HID-UMD Dataset 1 \newline \cite{umd},2001,FO,4,,25,100,$25s \times 4v = 100$
HID-UMD Dataset 2 \newline \cite{umd},2001,FO,2,Time(2),55,220,$55s \times 2v \times 2i = 220$
CASIA Dataset A \newline \cite{wang2003silhouette},2001,FO,3,,20,240,$20s \times 4i \times 3v = 240$
SOTON Small DB \newline \cite{hidgait},2002,FI,4,{Load(4), clothing(3), footwear(6), speed(3)},12,n/a,$12s$
SOTON Large DB \newline \cite{soton_paper},2002,FTIO,2,{Terrain(3)},114,2128,$114s \sim 2128$
USF Gait Challenge \newline \cite{sarkar2005humanid},2002,FO,2,{Terrain(2), footwear(2), load(2), time(2)},122,1870,$122s \sim 1870i$
CASIA Dataset B \newline \cite{yu2006framework},2005,FI,11,{Clothing(2), load(2)},124,13640,$124s \times 11v \times 10i = 13640$
CASIA Dataset C \newline \cite{tan2006efficient},2005,FO,1,{Speed(3), load(2)},153,612,$153s \times 4i = 612$
CASIA Dataset D \newline \cite{zheng2011evaluation},2009,FI,1,,88,880,$88s$
OU-ISIR Speed Transition \newline \cite{ouisir_st},2007,FTI,1,Speed shift,179,n/a,$179s$
OU-ISIR Large Population \newline \cite{ouisir_lp},2009,FI,2,,4007,7842,$4007s \sim 7842$
TUM-IITKGP \newline \cite{hofmann2011gait},2010,FI,2,{Load(2), occlusion},35,840,$35s \times 6c \times 2v \times 2i = 840$
OU-ISIR Inertial Sensor \newline \cite{ouisir_is},2011,FI,,Inclination(3),744,3468,$744s \sim 3468$
SOTON Temporal \newline \cite{matovski2012effect},2011,FI,12,{Time(5), clothing(2), footwear(2)},25,26160,$25s \sim 2180i \times 12v = 26160$
OU-ISIR Treadmill - A \newline \cite{ouisir_tm},2012,TI,1,Speed(9),34,612,$34s \times 9v \times 2i = 612$
OU-ISIR Treadmill - B \newline \cite{ouisir_tm},2012,TI,1,Clothing(32),68,2176,$68s \times 32c = 2176$
OU-ISIR Treadmill - C \newline \cite{ouisir_tm},2012,TI,25, ,200,5000,$200s \times 25v = 5000$
OU-ISIR Treadmill - D \newline \cite{ouisir_tm},2012,TI,1,Gait fluctuations,185,370,$185s \sim 370$
TUM-GAID \newline \cite{tum_gaid_paper},2012,Fi,1,{Load(2), footwear(2), time(2), clothing(2)},305,3370,$305s \sim 3370$
\end{filecontents*}
\csvreader[late after line=\\]{Data/Datasets.csv}
    {Database=\nam, Year*=\yr, Env=\ev, View=\vw, Covariates=\cvar,
    Sub=\sub, Seq=\seq}
    {\nam & \yr & \ev & \vw  & \sub & \seq & \cvar}
    \bottomrule
    \multicolumn{7}{p{1.5\linewidth}}{
     * The year indicates the year of recent release, not the year of its
     initial release. SOTON released its first version in 1996 and its
     first temporal set in 2003. OU-ISIR Treadmill datasets were
     iteratively developed from 2007-12. 

                                             \vspace{0.5em}
  Env Legends: F - static flooring, T - treadmill, I - indoor, O -
  outdoor.

                                             \vspace{0.5em}
  n/a: not available; values are either unconfirmed or not published
}                                         
  \end{tabular}
  }
\end{table}

When DARPA launched the HumanID programme in the year 2000, many institutions joined and released their first version of the database in the year 2001. Institutions that released their dataset for public usage include MIT, CMU, SOTON, Georgia Tech, UMD, USF, and CASIA. Nearly all of the institutions whose databases are described in this section are associated with this programme. The programme ended in 2004 due to privacy issues \citep{hidgait}, but the databases compiled as a result is still publicly available from the institutions that developed them.
The choice of the database depends on the use-case. This thesis involves the design of biometric algorithms that are resilient to multiple views.
The datasets with the most views are CASIA-B and SOTON-Temporal.
However, CASIA-B has 124 subjects while SOTON-Temporal has only 25. Furthermore, most literature published in view-invariant recognition systems also use the CASIA-B for their evaluation. Using the same dataset with similar training and testing conditions would facilitate a fairer comparison. Due to better balance between covariate factors, the number of subjects and the number of instances per subject, the CASIA-B dataset is selected as a benchmark dataset for the experimental validation of the proposed algorithms.

\begin{figure*}
  \centering
  \includegraphics[width=0.99\linewidth]{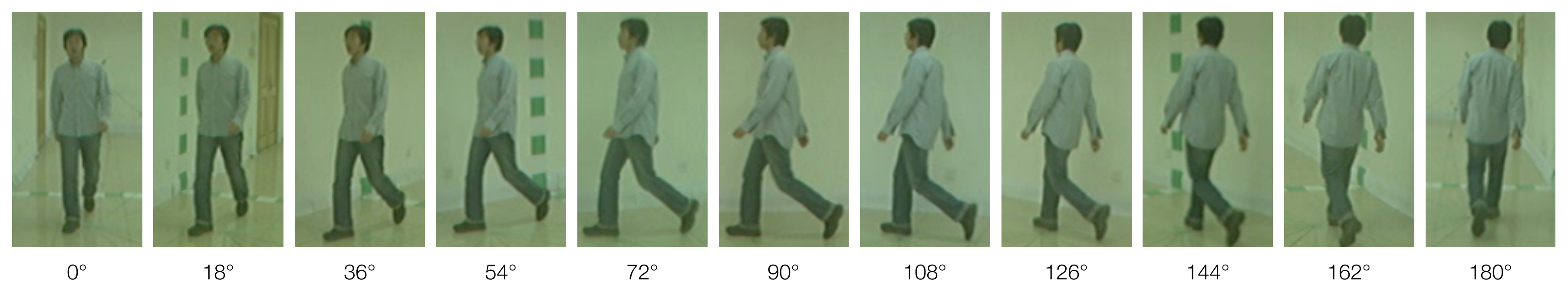}
  \caption{Cropped frames from CASIA Dataset B}
  \label{fig:casia-b}
\end{figure*}

The CASIA datasets are compiled by the Institute of Automation, Chinese Academy of Sciences. CASIA has developed four public gait databases till date. The development started in 2001 at the National Laboratory of Pattern Recognition (NLPR). In addition to the video, the silhouettes for each of their datasets are freely available for download. With 124 individuals ($s$) each performing 10 gait instances ($i$) observed in 11 simultaneous view angles ($v$), the total number of sequences for the CASIA-B dataset would be
\[ 124s \times 10i \times 11v = 13640 \text{ sequences}\]
The 10 instances can be split up as SetA containing 6 normal instances, SetB with 2 instances carrying a bag, and SetC consisting of 2 instances wearing a coat. Along with an additional video of bare background, the total number of videos becomes 15004 (approximately 17.4 GB). The background is mostly plain light green with two stage markings covering both the wall and the floor \citep{yu2006framework}.  This is the most commonly used multiview database in literature. Figure~\ref{fig:casia-b} shows a depiction of the dataset for a given time instant from each of the 11 angles.

\section{Summary}
\label{sec:ch2-summary}

The biometric features of gait can be observed in a variety of ways wherein inertial sensor-based and computer vision-based are most researched in literature. Biometrics, in general, can be categorized as either hard or soft. Gender is found to be the most popular soft biometric in gait analysis; age estimation is the next most common soft biometric. Hard biometrics consider the problem of recognition (identification) and authentication (verification).

Biometric feature extraction can be classified as either model-free or model-based. Model-free approaches analyse the gait structure as a whole while model-based methods analyse joint-trajectories (usually in three-dimensions). Model-free methods can be further classified as either template-based or non-template. Template-based methods compile a single image called a gait template while non-template blurs between model-free and model-based methods in 2D space. Template-based methods are more successful than non-template methods in recent literature and have more practical scope than model-based methods.

Covariate factors increase the difficulty of gait recognition. Gait datasets mostly focus on difference in clothing conditions, carrying conditions, footwear and view angles. CASIA-B is currently the most commonly used multi-view gait dataset.

\chapter{GENDER CLASSIFICATION THROUGH POSE-BASED VOTING} 
\label{ch:gender} 

\section{Introduction}
\label{sec:pbv-intro}

Gender classification has several applications. In surveillance systems, the prior identification of gender can substantially reduce the search space required to find a specific person among many in a video feed \citep{yu2009study}. The classification could help analyse the relationship between gender and product preference in customer-market studies \citep{jones2017gender}. The knowledge of one's gender can provide natural human-computer interaction in robotics \citep{ramey2014morphological}. Many methods were suggested to investigate the spatiotemporal patterns of the gait sequence to infer the gender of the associated subject \citep{Prakash2016}. \cite{yu2009study} claim that an effective gender identification system can exceed the accuracy of human-based observation.

\begin{figure}[t]
  \centering
  \includegraphics[width=\linewidth]{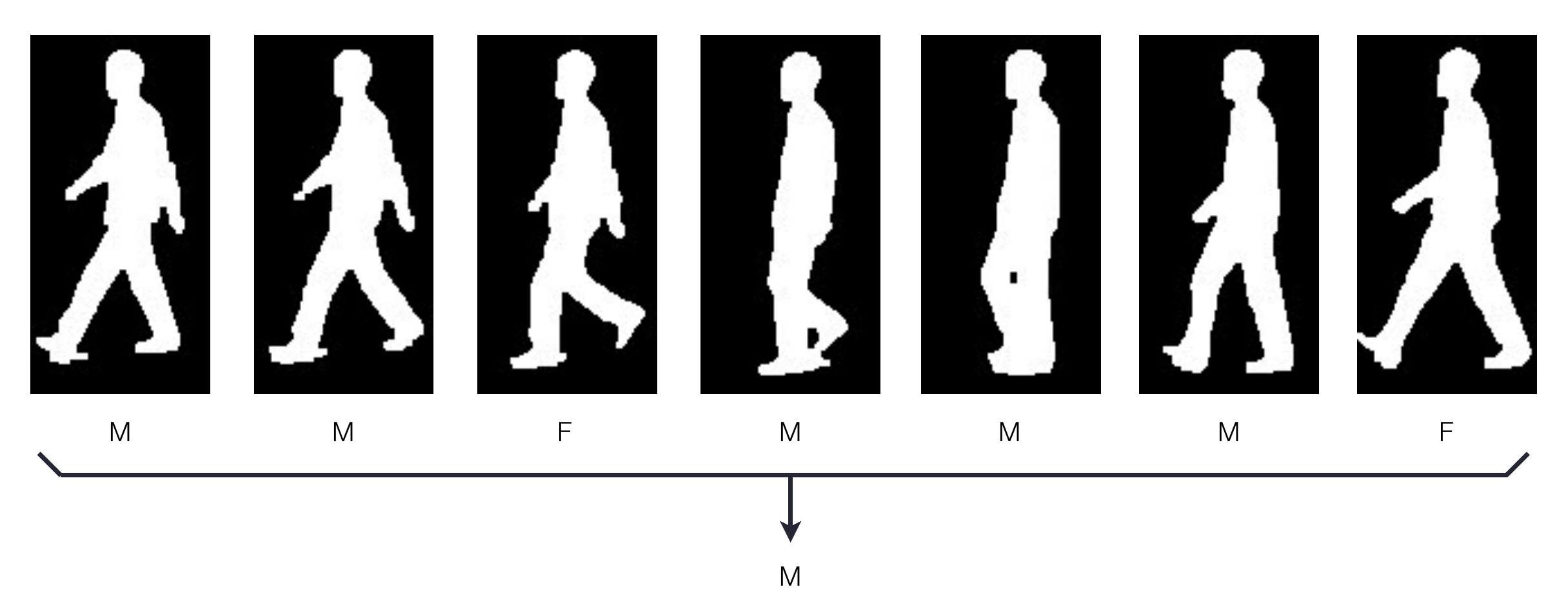}
  \caption{PBV illustration}
  \label{fig:pbv}
\end{figure}

In this chapter, we delve into a gait-based gender recognition algorithm that employs a technique that delineates the gait instance as a sequence of poses or frames. The system predicts the gender of each frame that constitutes the gait instance and decides whether the subject is male or female based on majority voting. This method is called pose-based voting (PBV) as illustrated in Figure~\ref{fig:pbv}. The intuition behind this approach relies on the fact that humans tend to assume certain poses at each part of the gait cycle that would reflect their innate gender. The classifier tries to make a prediction based on the aggregate of the poses it records. Thus, the temporal element of gait is not processed by the classifier but just the spatial features. By removing the reliance on temporal features, the system would effectively deal with partially occluded gait cycles and temporary occlusion.

Two feature extraction techniques are investigated in this method: elliptic Fourier descriptors (EFD) \citep{kuhl1982elliptic}, and consolidate vector of row-column summation (RCS). The EFD is supported by its rotation invariance property while the RCS is more simple yet effective to implement. Both of these methods are extensively studied with the widely used CASIA-B dataset, and the results are compared to the state-of-the-art approaches through recommended test conditions.

\section{State of the Art}
\label{sec:pbv-art}

The gait instance is given as input to the system in the same manner as it is done for gait recognition and authentication. That is, the video feed is converted to frames through which silhouettes are extracted. The silhouettes are pure binary colour coded with black and white. The usual norm is then to extract only a single gait cycle from the entire sequence and make the prediction based on the patterns observed from that period.

\cite{hu2011gait} developed the current state of the art in gender classification based on gait.  The process involves dividing the normalized silhouettes to $2 \times 2$ and $4 \times 4$ cells and the fitting of ellipses \citep[as in][]{little1998}. The extracted features are then modelled using MCRF along with a combination of complicated operations including the Karhunen-Loeve transformation \citep{dony2000karhunen} and Xie and Beni's (XB) index \citep{xie1991validity}. Their implementation was tested on the CASIA-B dataset to achieve an accuracy of 98.39\%.

\cite{nguyen2017gender} implemented a method to detect gender with no more than a single video frame through a CNN. However, CNNs are known to be computationally expensive in operation. As the application in focus concerns gait, more frames are available per instance. Analysing each frame of a gait sequence with a simpler algorithm can yield an equal or better result than taking only a single frame with a complex algorithm.

Most existing implementations adopt the SVM classifier for the gender recognition problem especially when a gait template like the GEI is used \citep[see][]{li2008gait, yu2009study, igual2013robust}. Some relatively new methods include the mixed continuous random field (MCRF) by \cite{hu2011gait}, sparse representation-based classification by \cite{lu2014human}, and CNN by \cite{nguyen2017gender}.

\section{Method}
\label{sec:pose-based-voting}

\begin{figure}
  \centering
  \includegraphics[width=\linewidth]{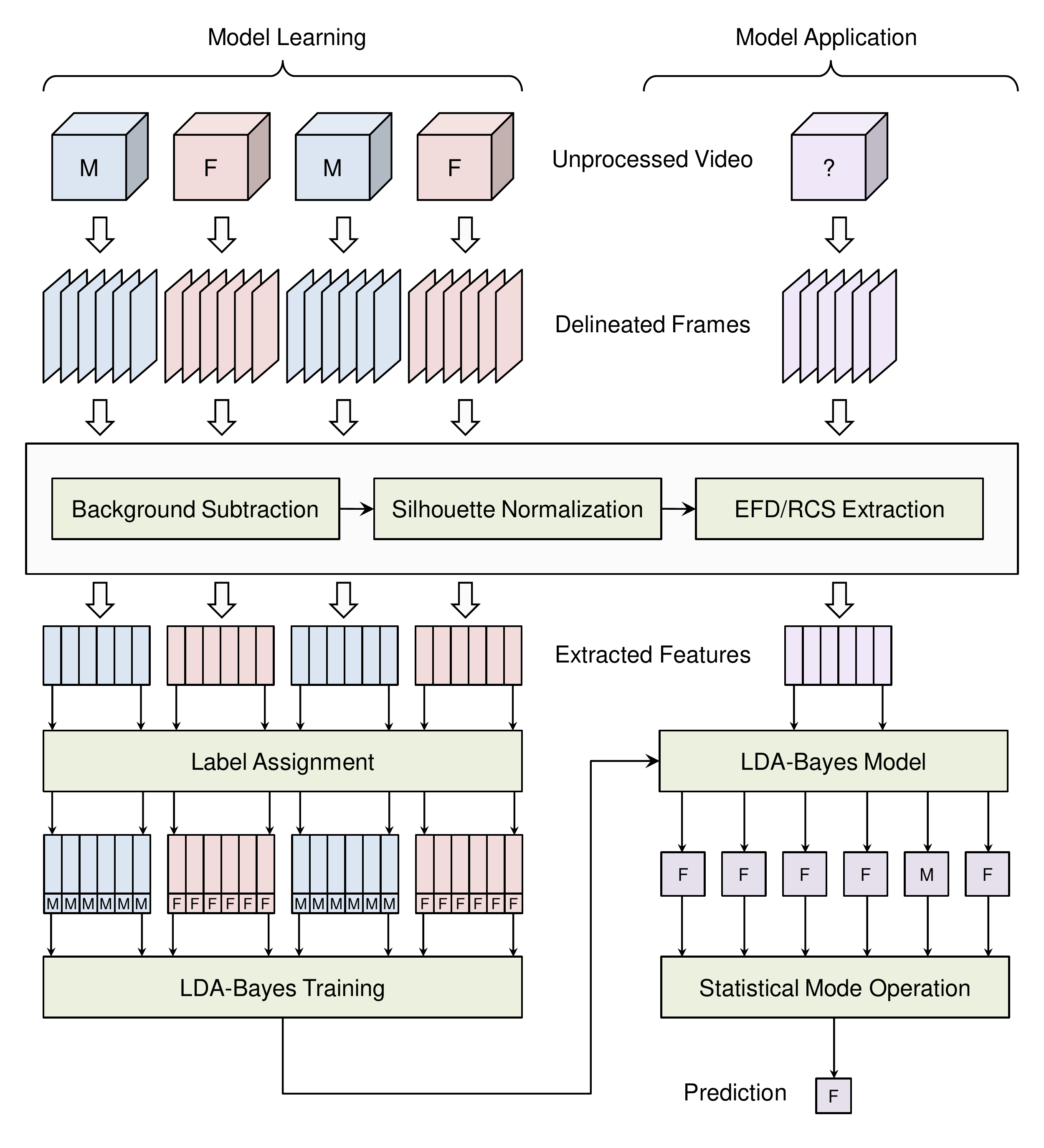}
  \caption{Schematic illustration of the PBV gender recognition system 
  }
  \label{fig:pbv-arch}
\end{figure}

The schematic flow of the proposed system is illustrated in Figure~\ref{fig:pbv-arch}.  The videos containing the gait instances are delineated into frames through which silhouettes are obtained and binarized. The forthcoming sections discuss two different feature extraction techniques, EFD and RCS. Instead of taking a spatiotemporal template of the entire gait sequence, each silhouette is considered to be a separate instance to the LDA-Bayes classifier. The classifier makes predictions on each frame of the test video. The statistical mode (majority vote) of the aggregate predictions become the final prediction of the system.

\newpage
\subsection{Elliptic Fourier Descriptors}
\label{sec:efd}

Chain encoding is a technique that is used to provide a compressed representation of image contours \citep{freeman1974computer}. \cite{kuhl1982elliptic} presented that elliptic properties of Fourier coefficients can be exploited to characterize a closed contour. The resulting features are called elliptic Fourier descriptors. The major advantage of this method is that the contours that are reconstructed are said to be independent of the starting point of the contour. Furthermore, the EFD has a rotation and scale invariance property that describes the image with the same set of coefficients irrespective of the orientation and size of the object in two-dimensional space.

\begin{figure}
  \centering
  \subfloat[$N=1$]{\includegraphics[height=0.25\textheight]{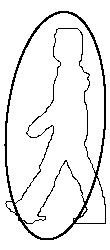}%
    \label{fig:s1}}
  \hfil
  \subfloat[$N=2$]{\includegraphics[height=0.25\textheight]{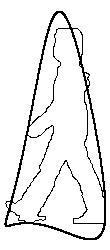}%
    \label{fig:s2}}

  \vspace{2em}
  
  \subfloat[$N=4$]{\includegraphics[height=0.25\textheight]{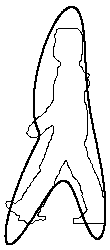}%
    \label{fig:s4}}
  \hfil
  \subfloat[$N=6$]{\includegraphics[height=0.25\textheight]{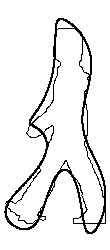}%
    \label{fig:s6}}
  \hfil
  \subfloat[$N=8$]{\includegraphics[height=0.25\textheight]{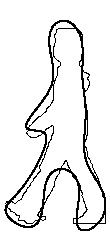}%
    \label{fig:s8}}

  \vspace{2em}
  
  \subfloat[$N=12$]{\includegraphics[height=0.25\textheight]{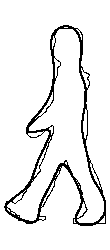}%
    \label{fig:s11}}
  \hfil
  \subfloat[$N=16$]{\includegraphics[height=0.25\textheight]{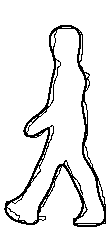}%
    \label{fig:s15}}
  \caption{Reconstruction of silhouette contours using EFD}
  \label{fig:efd}
\end{figure}

Figure~\ref{fig:efd} depicts the reconstruction of the EFD in different levels of harmonics extracted from a human silhouette. The real silhouette is superimposed with the reconstructed $N$-harmonic representations of the chain code for reference.

The chain code V of a contour consisting of K points is
\[ V = a_1a_2a_3...a_K \]
where $a_i$ can take the values from 0 through 7 representing the 8 possible
pixel directions taken anticlockwise (although the contour is traversed
clockwise). The required time to travel across the link $a_i$ is
\[\Delta t_i = 1 + \Bigg(\frac{\sqrt{2}-1}{2}\Bigg)(1 - (-1)^{a_i})\]
The required time to plot up to $p$ chain links is given by
\[t_p = \sum_{i=1}^p \Delta t_i \]
and $T=t_K$ is considered to be the chain's basic period.

\noindent The differences in the projections of $x$ and $y$ are
\begin{align}
  \Delta x_i &= \text{sgn}(6-a_i)\text{sgn}(2-a_i),\nonumber\\
  \Delta y_i &= \text{sgn}(4-a_i)\text{sgn}(a_i)\nonumber
\end{align}
where sgn is a sign function given by
\begin{equation*}
  \text{sgn}(w) = 
  \begin{cases}
    1 & \text{if}\ w > 0,\\
    0 & \text{if}\ w = 0,\\
    -1 & \text{if}\ w < 0.
  \end{cases}
\end{equation*}
The Fourier series for the $x$ and $y$ projections can be expanded as follows.
\begin{align}
  x(t) & = A_0 + \sum_{n=1}^N a_n\cos\frac{2n\pi t}{T} + b_n\sin\frac{2n\pi
         t}{T} \label{pbv-eq:1}\\
  y(t) & = C_0 + \sum_{n=1}^N c_n\cos\frac{2n\pi t}{T} + d_n\sin\frac{2n\pi
         t}{T} \label{pbv-eq:2}
\end{align}
where
\begin{align}
  a_n & = \frac{T}{2n^2\pi^2}\sum_{p=1}^K\frac{\Delta x_p}{\Delta t_p}
        \Bigg[\cos\frac{2n\pi t_p}{T} - \cos\frac{2n\pi t_{p-1}}{T}\Bigg] \nonumber\\
  b_n & = \frac{T}{2n^2\pi^2}\sum_{p=1}^K\frac{\Delta x_p}{\Delta t_p}
        \Bigg[\sin\frac{2n\pi t_p}{T} - \sin\frac{2n\pi t_{p-1}}{T}\Bigg]
        \nonumber\\
  c_n & = \frac{T}{2n^2\pi^2}\sum_{p=1}^K\frac{\Delta y_p}{\Delta t_p}
        \Bigg[\cos\frac{2n\pi t_p}{T} - \cos\frac{2n\pi t_{p-1}}{T}\Bigg] \nonumber\\
  d_n & = \frac{T}{2n^2\pi^2}\sum_{p=1}^K\frac{\Delta y_p}{\Delta t_p}
        \Bigg[\sin\frac{2n\pi t_p}{T} - \sin\frac{2n\pi t_{p-1}}{T}\Bigg] \nonumber
\end{align}

Equations \ref{pbv-eq:1} and \ref{pbv-eq:2} are expanded up to $N$
harmonics\footnote{The derivation of the elliptic Fourier descriptors for closed
  contours can be found in detail in the work
  by \cite{kuhl1982elliptic}}. In the
generalized equations, $N=\infty$. The value of $N$ determines the complexity of
fit. The DC components $A_0$ and $C_0$ are given by
\begin{align}
  A_0 &= \frac{1}{T}\sum_{p=1}^K \frac{\Delta x_p}{2\Delta t_p}
        \bigg(t_p^2 - t_{p-1}^2\bigg) + \xi_p\big(t_p - t_{p-1}\big), \nonumber\\
  C_0 &= \frac{1}{T}\sum_{p=1}^K \frac{\Delta y_p}{2\Delta t_p}
        \bigg(t_p^2 - t_{p-1}^2\bigg) + \delta_p\big(t_p - t_{p-1}\big) \nonumber
\end{align}
where
\begin{align}
   \xi_p &= \sum_{j=1}^{p-1}\Delta x_j
        - \frac{\Delta x_p}{\Delta t_p}\sum_{j=1}^{p-1}\Delta t_j , \qquad \xi_1=0, \nonumber\\
   \delta_p &= \sum_{j=1}^{p-1}\Delta y_j
        - \frac{\Delta y_p}{\Delta t_p}\sum_{j=1}^{p-1}\Delta t_j ,  \qquad \delta_1=0 \nonumber
\end{align}

For each harmonic level, four more coefficients are generated, $a_n$, $b_n$, $c_n$, and $d_n$. These parameters are used as features for the training. Thus, the EFD provides a total of $4N+2$ features per sample for a fixed value of $N$.

\subsection{Row-Column Summation}
\label{sec:rcs}

\begin{figure}[t]
  \centering
  \includegraphics[width=0.75\linewidth]{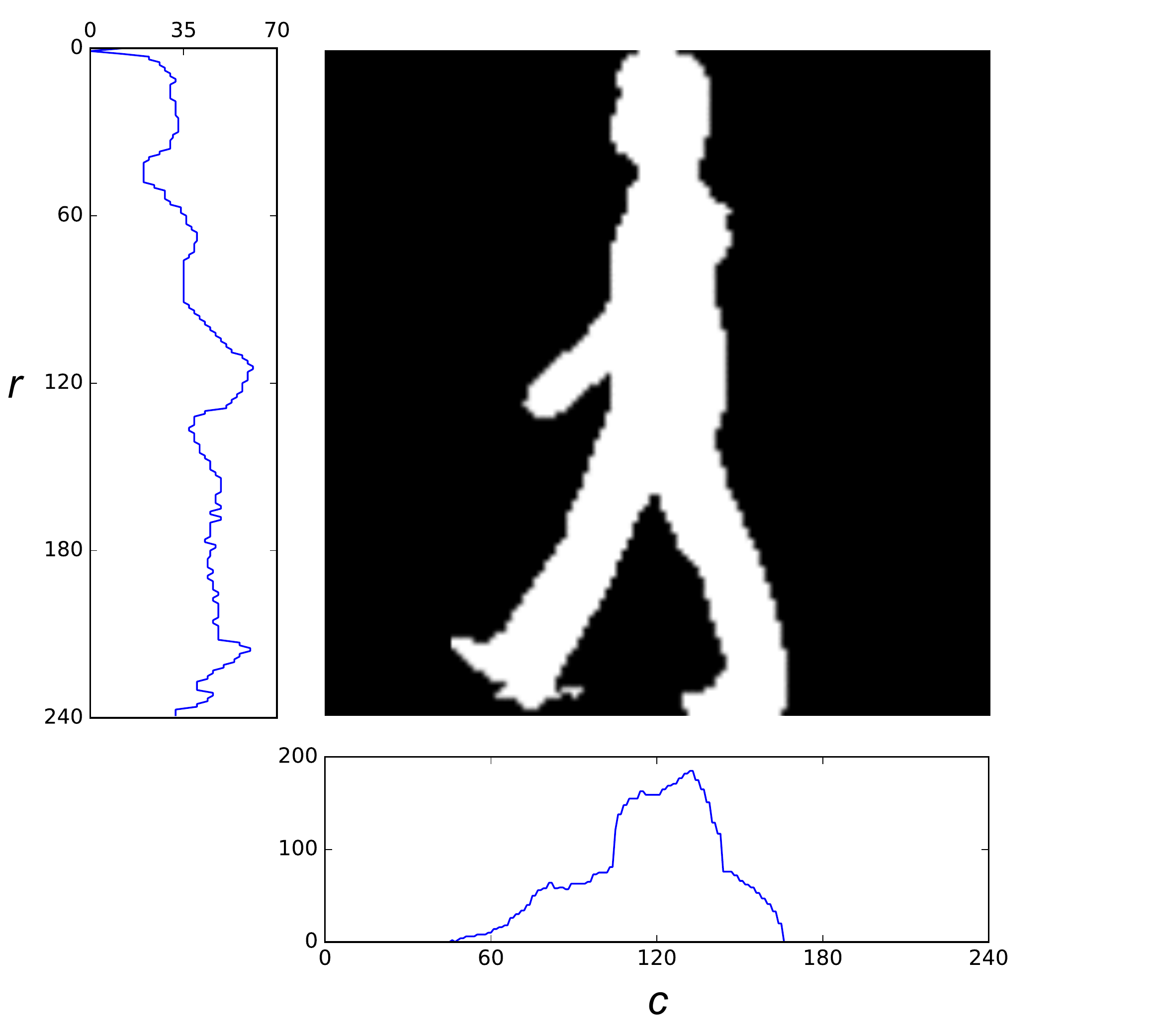}
  \caption{Extraction of RCS features from a single silhouette}
  \label{fig:rcs}
\end{figure}

One way to extract both horizontal and vertical features of a silhouette would be to take the count of white pixels in each of $x$ and $y$ axes.  An illustration of this is given in Figure~\ref{fig:rcs}. The figure depicts the number of white pixels in each row $r$ and column $c$. The properties of the binarized silhouette are exploited to facilitate the vectorized version of this operation.

Let $r_i$ denote the count of white pixels at row $i$, and $c_j$ be number of white pixels in column $j$. The row and column vectors are given by
\[ \vec{r} = \big(r_1, r_2, r_3, ..., r_n\big) \qquad \text{and} \qquad 
  \vec{c} = \big(c_1, c_2, c_3, ..., c_m\big)\] where $n$ and $m$ are taken as 240 in this
implementation. The number of white pixels of a row of a binarized image can be taken as the sum of all values in that row. Thus, $\vec{r}$ contains the row-wise sum of the silhouette. Similarly, $\vec{c}$ stores the column-wise sum. Together, these vectors describe the pose of the silhouette in the image. The concatenation of both these vectors, $\vec{r}~^\frown\vec{c}$, can be utilized as the feature set of the image and are called the row-column summation or RCS.

Note that each silhouette is taken to be a square rather than a rectangle with the height greater than the width. This size makes sure that the entire space of movement freedom is covered by the subject. According to the anatomy of an average human, the arm-span is approximately equal to the height \citep{creed1986leonardo}. This notion is also adopted by \cite{zheng2011robust} when generating a public dataset of GEI from the CASIA-B dataset.

\subsection{Pose-Based Voting}
\label{sec:proposed}

The initial step of nearly any gait-based biometric system is silhouette extraction which is achieved through background subtraction. In cases where the environment is highly dynamic, algorithms like the one suggested by \cite{panda2016detection} can be employed. They use a modified version of the colour difference histogram through fuzzy c-means clustering to eliminate the excessive noise in the background. As the dataset in this study (CASIA-B) is composed only of a static background, a slightly primitive foreground detection routine is applied. Further description of the dataset is provided in Section~\ref{sec:datasets}. In this step, the difference between the background frame $b$ and the current frame at instance $t$ would give the silhouette of the moving object at frame $t$ \citep{rosebrock2016practical}. A standard camera can introduce noise between successive frames even in a static environment. Hence, a Gaussian blur operation applied before the difference is computed to eliminate the subtle noise.

\begin{algorithm}[t]
  \begin{spacing}{1.3}
  \caption{PBV training routine}\label{alg:train}
  \begin{algorithmic}
    \Statex \textbf{Preconditions:}
    \Statex \hspace{1.5em} \textit{vList}: list of gait video instances 
    \Statex \hspace{1.5em} \textit{gLabels}: corresponding vector of gender labels 
    \Procedure{PBVtrain}{$\mathit{vList}, \mathit{gLabels}$}
    \State $X \gets \text{EmptyMatrix}$
    \State $y \gets \text{EmptyVector}$
    \For{each index $i$ in \textit{vList}} 
       \State $\mathit{silhouettes} \gets
       \text{BackgroundSubtract}(\mathit{vList}[i])$
       \State $\mathit{normSilhouettes} \gets
       \text{Normalize}(\mathit{silhouettes})$
       \State $\mathit{sampleSet} \gets \text{ExtractFeatures}(\mathit{normSilhouettes})$
       \State $\mathit{label} \gets \mathit{gLabels}[i]$
       \For{each \textit{sample} in \textit{samplesSet}}
          \State Append \textit{sample} to $X$
          \State Append \textit{label} to $y$
       \EndFor
    \EndFor
    \State $L \gets \text{LDAFit}(X, y)$
    \State $X \gets L.\text{transform}(X)$
    \State $M \gets \text{MGBayesFit}(X, y)$
    \State $\mathit{model} \gets [L, M]$
    \State \textbf{return} \textit{model}
    \EndProcedure
  \end{algorithmic}
\end{spacing}

\end{algorithm}

Each silhouette extracted is clipped and scaled to standard proportions. This procedure is considered to be normalization \citep{zheng2011robust}. Each sample contains a complete human silhouette with the dimensions $240 \times 240$ pixels and binarized.
A pseudocode of the training process is given in Algorithm~\ref{alg:train}. The feature vectors are extracted from each silhouette and are labelled to form the training set. This study compares the effectiveness of both EFD and RCS feature extraction methods. The feature space is then transformed through Linear Discriminant Analysis (LDA) and trained through a multivariate Gaussian Bayes' rule (MGBayesFit). The model thus produced is composed of a set of parameters that are used to make the gender prediction during the testing phase. These include the transformation vectors from LDA and the probabilities from Bayes' rule.

The LDA transformation maximizes inter-class distance while minimizing the intra-class distance \citep{duda2001pattern}. LDA is more commonly used in gait recognition algorithms along with principal component analysis \citep{rida2016human}. As the number of features is already sufficiently small for processing, an unclassified feature reduction like PCA would not be necessary.

The transformed data is passed through a multivariate Gaussian model of Bayes' rule \citep{hastie2005elements} for classification. This is not a na\"ive Bayes implementation where a product of multiple univariate normal distributions are taken as the likelihood. The Bayes classifier is given by
\[\Pr(y \peq k \mid x) = \frac{\Pr(x \mid y \peq k) \cdot \Pr(y \peq k)}{\Pr(x)}\] where the likelihood term, $\Pr(x \mid y \peq k)$, is calculated with $X$ distributed as a multivariate normal distribution, $X\sim\mathcal{N}(\mu_k,\Sigma_k)$.

\begin{algorithm}[t]
  \begin{spacing}{1.3}
  \caption{PBV testing routine}\label{alg:test}
  \begin{algorithmic}
    \Statex \textbf{Preconditions:}
    \Statex \hspace{1.5em} \textit{instance}: single gait video instance
    \Statex \hspace{1.5em} \textit{model}: trained PBV model
    \Procedure{PBVpredict}{$\mathit{instance}, \textit{model}$}
    \State $\mathit{silhouettes} \gets \text{BackgroundSubtract}(instance)$
    \State $\mathit{normSilhouettes} \gets \text{Normalize}(\mathit{silhouettes})$
    \State $\mathit{sampleSet} \gets \text{ExtractFeatures}(\mathit{normSilhouettes})$
    \State $[L, M] \gets \mathit{model}$
    \State $\mathit{predictions} \gets \text{EmptyList}$
    \For{each \textit{sample} in \textit{samplesSet}}
       \State $x_t \gets L.\text{transform}(sample)$ 
       \State $p \gets \text{MGBayesPredict}(x_t)$
       \State Append $p$ to \textit{predictions}
    \EndFor
    \State $\mathit{prediction} \gets \text{ModeStat}(\mathit{predictions})$
    \State \textbf{return} \textit{prediction}
    \EndProcedure
  \end{algorithmic}
\end{spacing}

\end{algorithm}

The salient part of the algorithm lies in the testing routine (Algorithm~\ref{alg:test}). The preprocessing step is analogous to the training sequence where each silhouette is normalized and has its feature extracted (with either EFD or RCS). Once each silhouette is classified, the statistical mode of the separate predictions become the final prediction of the system. This event can be thought of as each prediction trying to elect for a specific gender which can assume the pose it depicts and the final selection being the majority vote of the individual silhouettes.

A specific experiment is conducted to test the handling of occlusion which leads to the partial observation of gait cycles. In this setup, a complete gait cycle of each training instance is given to both the existing and proposed PBV method for learning. Then, during testing, partial gait cycles were given ranging from 10\% to 100\% of the gait cycle pertaining to the test gait instance. The existing method would attempt to compile a gait template and then process it for classification. The PBV method, on the other hand, judges the gender based on the votes currently accumulated.

\section{Experimentation \& Evaluation}
\label{sec:results--evaluation}

The Python programming framework is used for all simulations in this thesis. Classification and feature transformation routines are applied with the help of the Scikit-Learn package \citep{scikit-learn}. All graphs were plot using the Matplotlib \citep{matplotlib}.

\subsection{Dataset Configuration}
\label{sec:pbv-dataset}

All experiments have been conducted using the video instances from the CASIA-B gait database \citep{yu2006framework} as it is currently the most common database to evaluate gender recognition. It holds videos containing the gait of 93 males and 31 females in three different covariate factors: normal, carrying a bag, and wearing a coat. The dataset and testing conditions are formulated exactly as prescribed by \cite{hu2011gait} with 31 female subjects and 31 randomly selected male subjects following a 31-fold cross validation evaluation. All six instances of normal walk, without bag or coat, observed at the sagittal angle were utilized for the selected subjects. The total number of video instances hence used are $62\times 6=372$. As this is a two-class categorization problem, the equal proportion of both classes avoids bias in the classification results. A 31-fold cross-validation testing scheme is used for evaluation. Each fold is disjoint and contains all videos of one male and one female subject.

\subsection{Number of Harmonics for EFD}
\label{sec:nselect}

\begin{figure}
  \centering
  \includegraphics[width=0.7\linewidth]{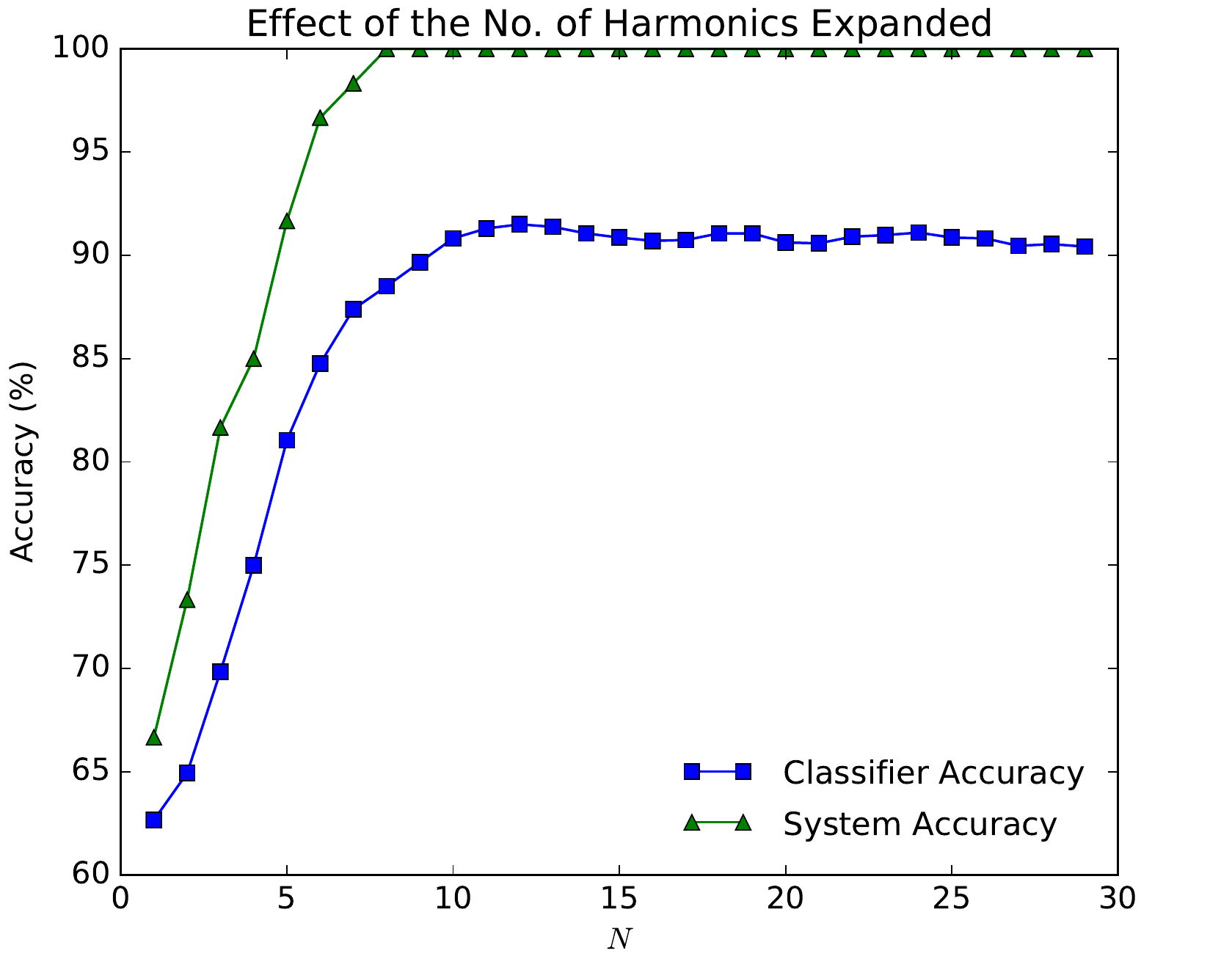}
  \caption{Effect of the number of harmonics on the system accuracy
  }
  \label{fig:nselect}
\end{figure}

The code for EFD was extended from Alessandro Ferrari's implementation \citep{ferrari2017bacterial}. The complexity of the Fourier descriptors is determined by the number of harmonics expanded, $N$. This value is empirically chosen in accordance with the effect it has on the system accuracy by conducting a standard testing-training split with varying values of $N$. The training test composes of 25 folds while the remaining 6 folds form the testing set. This configuration is similar to the test conditions used by \cite{huang2007gender}. The outcome is presented in Figure~\ref{fig:nselect}. The classifier accuracy is the CCR of the LDA-Bayes classifier, i.e., the proportion of silhouettes correctly classified to the total number of silhouettes in the test set. The system accuracy is the CCR of the PBV framework, i.e., the fraction of video instances correctly classified. One can clearly visualize the impact of the majority vote on the system. At the point where $N=8$, the system accuracy stabilizes while the classifier's CCR slowly increases. $N$ is set to 12 to safely avoid probable bias in the selection of the training and testing folds.

\subsection{Performance in Partial Gait Cycles}
\label{sec:partial-cycle}

Gait is composed of multiple phases which occur as repeated sequence of events. A gait cycle is a period between the occurrence of any two of the same gait event in succession \citep{whittlegait}. Almost all of the algorithms reported till date require the silhouettes of a complete gait cycle for an optimal CCR. The proposed framework would facilitate adequate performance even at partially observed gait cycles. A simple technique is employed to detect gait cycles to test this hypothesis.

\cite{sarkar2005humanid} estimated gait cycles by tracking the count of white pixels in the lower half the silhouettes. The pixel count would produce an oscillation signal when plotted. The silhouettes between any three troughs or crests of the signal would yield a complete gait cycle. However, the signal extracted could potentially contain noise which made the detection process difficult in certain cases. \cite{hu2011gait} suggested using variation in locally linear embedding (LLE) coefficient to overcome this issue. On the contrary, to provide a simpler yet effective approach for gait period estimation, a variant of the method used by \cite{sarkar2005humanid} in conjunction with smoothing using the Savitzky-Golay filter \citep{press1996numerical} is applied.

\begin{figure}
  \centering \subfloat[Unprocessed oscillation]{\includegraphics[width=\linewidth]{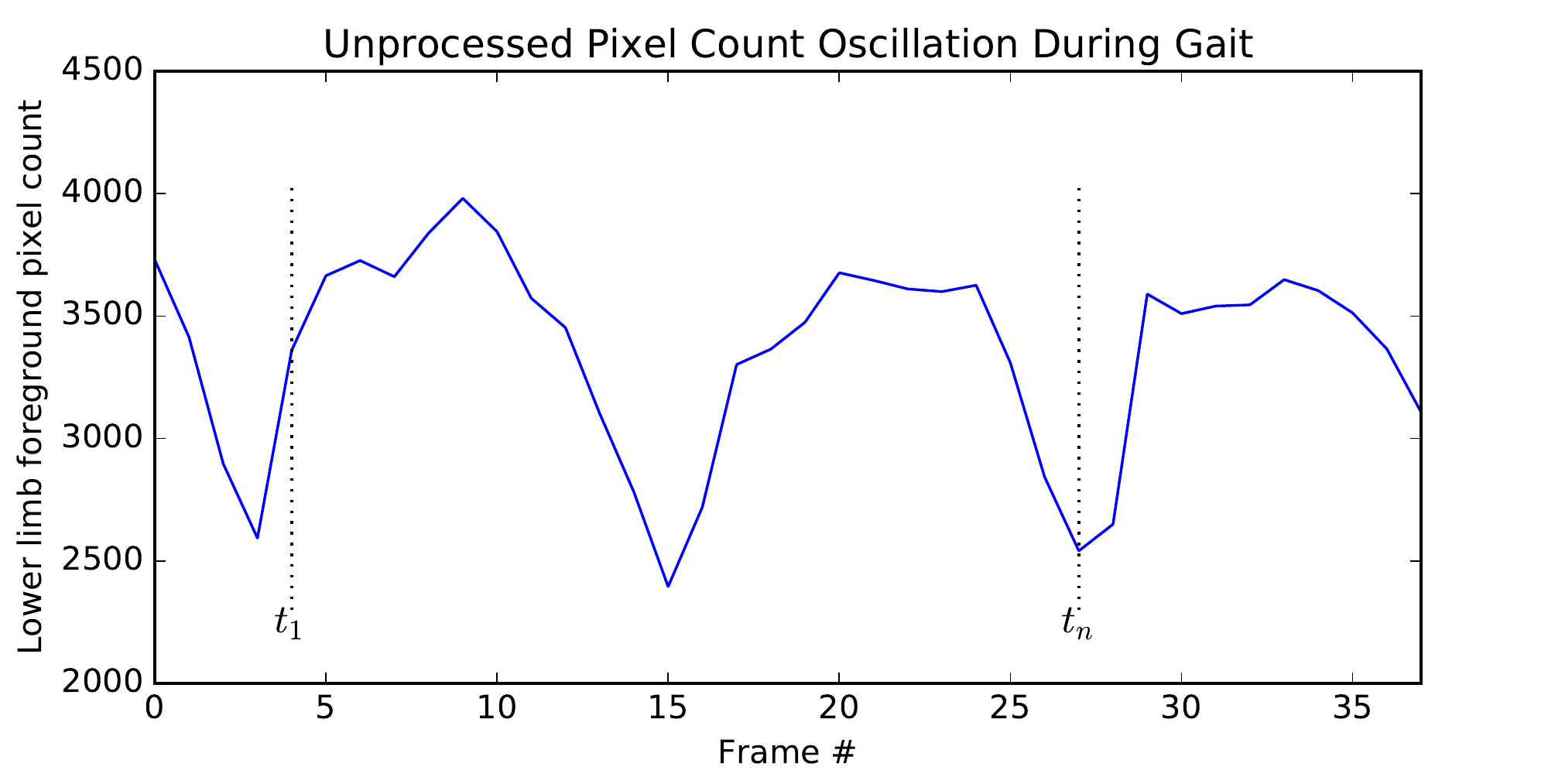}%
    \label{fig:UnprocOsc}}

  \vspace{1cm}
  
  \subfloat[Smoothed oscillation]{\includegraphics[width=\linewidth]{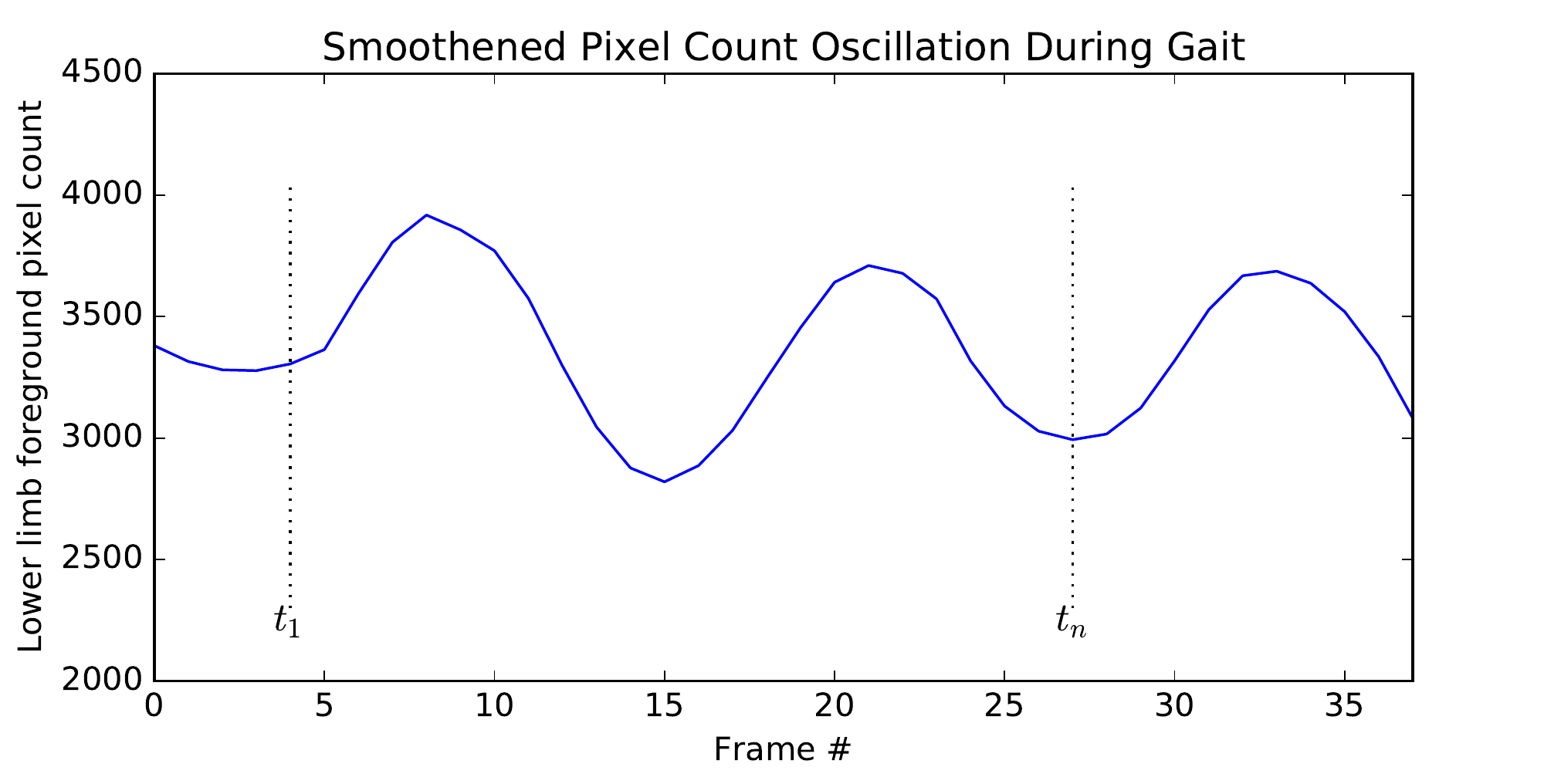}%
    \label{fig:SmoothedOsc}}

  \vspace{1cm}
  
  \caption{Flow of lower limb pixel count during gait
  }
\end{figure}

In this implementation, the white pixel count is taken below the half-way point from the probable knee region till the bottom end of the silhouette to minimize the noise that could occur. The oscillation produced can be observed as shown in Figure~\ref{fig:UnprocOsc}. Savitzky-Golay filter is applied (SciPy implementation by \cite{scipy}) to smooth the distortions in the signal. This technique keeps track of the adjacent points to approximate the position of the current point that would fit with the least error according to the least squares principle. The resulting curve would look almost like a sine curve as shown in Figure~\ref{fig:SmoothedOsc}. The silhouettes between three observable troughs are taken to compose a single complete gait cycle.

The intuition behind selecting three troughs is that at the first trough, both feet are together with one on the floor and the other performing midswing. The second trough denotes the half-cycle where the latter leg is stable while the former is at midswing. The full cycle completes at the third trough when all phases of gait are encompassed.

The GEI is used as a benchmark to represent the characteristic of algorithms that require a complete cycle \citep{li2008gait, yu2009study, lu2014human}. The classifier that is based on the GEI is trained in a similar fashion to those in literature with pixel intensities as attributes. The features are reduced through LDA and classified with SVM.

The experimental setup involves testing both the PBV and the GEI-based framework in increasing proportions of the gait cycle from partially observed to completely observed. Learning is applied as usual with the availability of the whole of the training video. Then, during testing, partial gait cycles were given ranging from 10\% to 100\% of the gait cycle pertaining to the test gait instance. The GEI-based method would attempt to compile a gait template with the available silhouettes and then process it for classification while the PBV follows the voting method as described in Section~\ref{sec:proposed}. The 31-fold cross-validation accuracy is recorded at each step. To this end, the objective is to prove that when LDA is applied, both SVM and Bayes' rule perform similarly. Hence, both SVM and Bayes are also compared in the analysis with both GEI and PBV methods. SciKit Learn package \citep{scikit-learn} was used for LDA and SVM. Moreover, the experiment is repeated for 3 different random combinations of male subjects (as there are only 31 female subjects) and the final scores are averaged at each stage.

\begin{figure} \centering
  \includegraphics[width=0.9\linewidth]{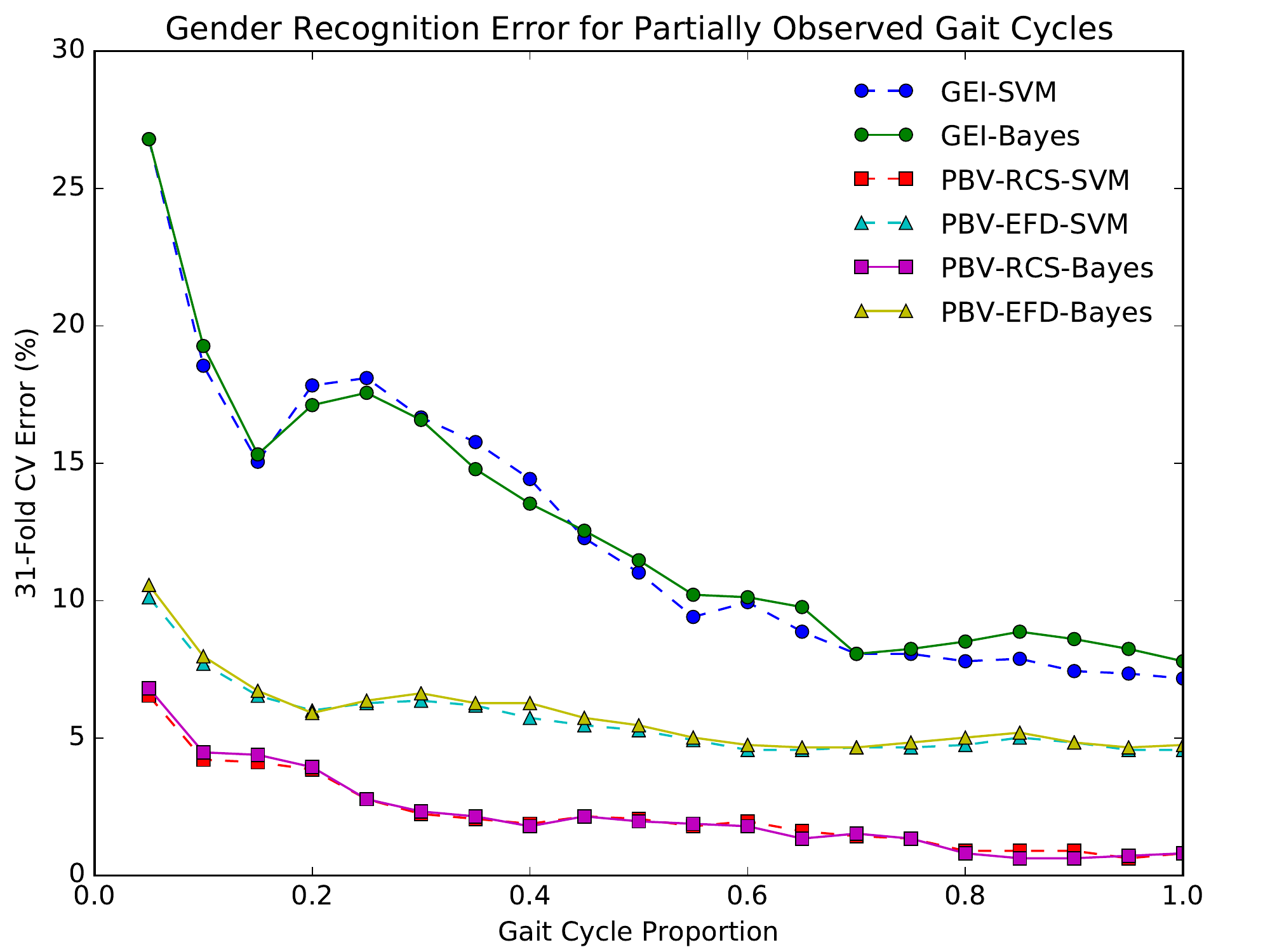}
  \caption{Gender recognition in partial occlusion}
  \label{fig:partial}
\end{figure}

The plots in Figure~\ref{fig:partial} clearly show that the proposed PBV method outperforms the standard template-based method. The difference is more pronounced at lower proportions of observed gait periods than at the complete cycle. Also, it can be observed that both SVM and Bayes' rule perform almost equally with both GEI and PBV frameworks as both are preceded with LDA transformation. Within PBV, the RCS features provide a much more robust CCR compared to the EFD feature set.

\Needspace{5\baselineskip}
\subsection{Overall Performance}
\label{sec:performance}

\begin{table}[t]
  \centering
  \caption{Gender Recognition Performance Comparison}
  \label{tab:result}
  \vspace{1em}
  \renewcommand{\arraystretch}{1.3}
    \begin{tabular}{l p{1cm} r}
      \toprule
      Method && CCR(\%) *\\
      \midrule
      \cite{chen2009multilinear,chen2010distance} && 93.28\\
      \cite{yu2009study} && 95.97\\
      \cite{hu2010combining} && 96.66\\
      \cite{hu2011gait} && 98.39\\
      \cite{lu2014human} && 98.00 \\
      Proposed PBV-EFD && 94.89\\
      Proposed PBV-RCS && 100.00\\
      \bottomrule
    \end{tabular}

    \vspace{1em} * 31-fold cross validation with 31 males and 31 females.
\end{table}

The performance of the overall system along with that of the existing implementations are compared in Table~\ref{tab:result}. The proposed PBV system gives an ideal result with the RCS feature scheme and a moderately good result with the EFD feature set. The RCS relies only on the spatial features as it incorporates both the body shape and pose of the subject that enables it to characterize the gender of the subject under observation. Note that the PBV accuracy that is tabulated is greater than that in Figure~\ref{fig:partial}. This phenomenon is because, during the overall test, the system considers all possible silhouettes of the video instance; not just that of a single gait cycle. Each video contains slightly more than a single gait cycle while some may even contain up to two. The more silhouettes that are extracted, the more electors there would be participating in the voting scheme leading to a higher probability that the outcome would be correct.

The harmonics were recomputed for higher orders up to 40 and tested again to affirm the performance of EFD, yet the same accuracy was obtained. Though the EFD may seem a little inaccurate compared to the other state of the art, one can hypothesize that its full potential can be realized through its rotation and scale invariance property. Unlike the RCS and the other methods in the literature, the EFD is said to generate the same set of descriptors for any orientation of the image. This property would be useful in surveillance systems where the camera orientation can vary, causing the silhouettes extracted to seem rotated in 2-D space.

\section{Summary}
\label{sec:pbv-summary}

The state-of-the-art methods for gait-based gender recognition extract spatiotemporal features from the gait sequence and project the entire gait sequence as a single instance. On the contrary, the proposed PBV projects each frame of the sequence as an instance. The PBV scheme aggregates the predictions of each silhouette of a single gait sequence based on a majority vote to give the final gender prediction.

Existing implementation of gait-based gender recognition show promising results but do not consider the problem of temporary occlusion. The proposed PBV approach yields an ideal gender recognition accuracy and is robust to temporary occlusion of sequences of a gait video instance.

Two distinct feature representations were used in PBV, namely, EFD and RCS. After a pass through LDA, Bayes' rule can provide close to equal performance to SVM. The results of both existing models and proposed model were compared based on the CASIA-B gait dataset to show that PBV with RCS outperforms the state-of-the-art methods.


\chapter{GAIT RECOGNITION THROUGH GENETIC TEMPLATE SEGMENTATION}
\label{ch:gait-recog} 

\section{Introduction}
\label{sec:gts-intro}

Gait recognition is the process of identifying a person through only their gait patterns. Although the term should be technically gait identification, the term gait recognition is much more pronounced in literature. As discussed in Section~\ref{sec:hard-biometrics}, there are both model-free and model-based methods to do this task. Model-based methods \citep{bouchrika2007model, Goffredo2008, Zhang_R, yam2004automated} attempt to track the dynamic changes in the articulation points during gait and hence require intense computational effort. The model-free approach is the most preferred as it captures the gait patterns without this requirement.

Recent methods adopt the use of gait templates as they are simple to implement yet highly effective in practice with the GEI \citep{man2006individual} being the most popular one among them. The GEI quickly became the most successful method for multi-view gait recognition. Its major drawback was its weakness to covariates like clothing and load carrying which could adversely affect its performance. Many similar methods followed aiming to mitigate this weakness with their implementation of gait templates. Notable templates include the AEI ~\citep{zhang2010active} and the GEnI \citep{bashir2010gait}. With a slight trade-off in normal walk gait recognition, these new templates were able to produce a better recognition accuracy over the clothing and carrying covariates in gait.

In this chapter, a VPM is devised that can be applied to any gait template for gait identification. To refine the templates themselves, a method is proposed to automate its segmentation process with the use of the genetic algorithm (GA). These segments depict the optimal regions of the gait template that can be used to obtain the best recognition result at any visually affecting covariate factor. As a design choice, Bayes' rule is used instead of the widely adopted $k$NN.

\section{State of the Art}
\label{sec:gts-art}

Research suggests that using a proportion of the gait template that is least affected by the covariate factors would improve the overall accuracy of the gait recognition system. \cite{dupuis2013feature} formulated a single mask through the ranking of pixel features using the Random Forests classifier. Their panoramic gait recognition (PGR) algorithm uses pose estimation for view prediction. \cite{choudhury2015robust} designed a VPM named view-invariant multiscale gait recognition (VI-MGR) which applied Shannon's entropy function to the lower limb region of the GEI. The sub-region selection was later modified by \cite{rida2016human}  automating this segmentation procedure with a process known as group lasso of motion (GLM). Their approach to the problem has shown significant improvement in the covariate recognition accuracy.

Though the following implementations do not concern view-invariance or covariate factors, their aspects add to the motivation of this approach.  \cite{jia2015view} have shown how incorporating the head and shoulder mean shape (HSMS) along with the Lucas-Kanade variant of the gait flow image (GFI) \citep{lam2011gait} greatly improves recognition accuracy. The genetic algorithm \citep{Goldberg} was previously used by \cite{yeoh2014genetic} to optimize the selection of model-based gait parameters and also by \cite{tafazzoli2015genetic} for the selection of superimposed contour features.

\newpage
\section{Method}
\label{sec:gts}

An overview of the method is illustrated in Figure~\ref{fig:gts-arch}. The first step is to extract the gait template (such as the GEI) from the video that contains the gait sequence. After which the database is split into two disjoint sets -- tuning set and evaluation set. The tuning set is fed to the GA to formulate the segments for optimal performance. Only those segments are extracted from the evaluation set to test the final accuracy of the system.

\subsection{Gait Template Extraction}
\label{sec:gait-templ-extr}

The model-free template representation of gait composes the feature set for this study. All gait templates are produced in a similar procedure to the one given below. Silhouettes in here are obtained through background subtraction and encoded in grayscale.
\begin{enumerate}[topsep=0pt, noitemsep, leftmargin=*]
\item Extract only the silhouettes of the subject during a single gait cycle.
\item The silhouettes are center-aligned and scaled to a standard size; 240 x
  240 in this case.
\item The standardized silhouettes for a given gait sequence are merged through
  a collation process to generate the gait template.
\end{enumerate}

The silhouettes are usually binarized composing only of pure white (1) and black (0) pixels before collation. The characteristic difference between one type of template to another lies in the collation process adopted to compile the template image. The final image that is produced after the collation contain pixel values varying between 0 to 1. These values are scaled to 0 to 255 for grayscale representation.

\begin{landscape}
\begin{figure}
  \centering
  \includegraphics[width=0.8\linewidth]{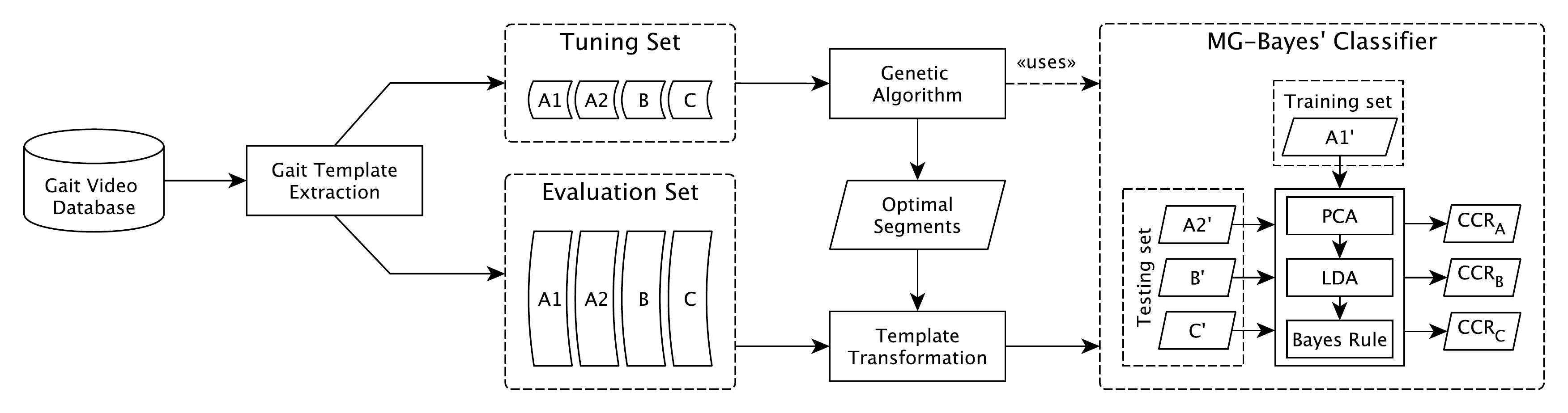}
  \caption{The complete flow of the proposed method}
  \label{fig:gts-arch}
\end{figure}

\begin{figure}
  \centering
  \includegraphics[width=0.8\linewidth]{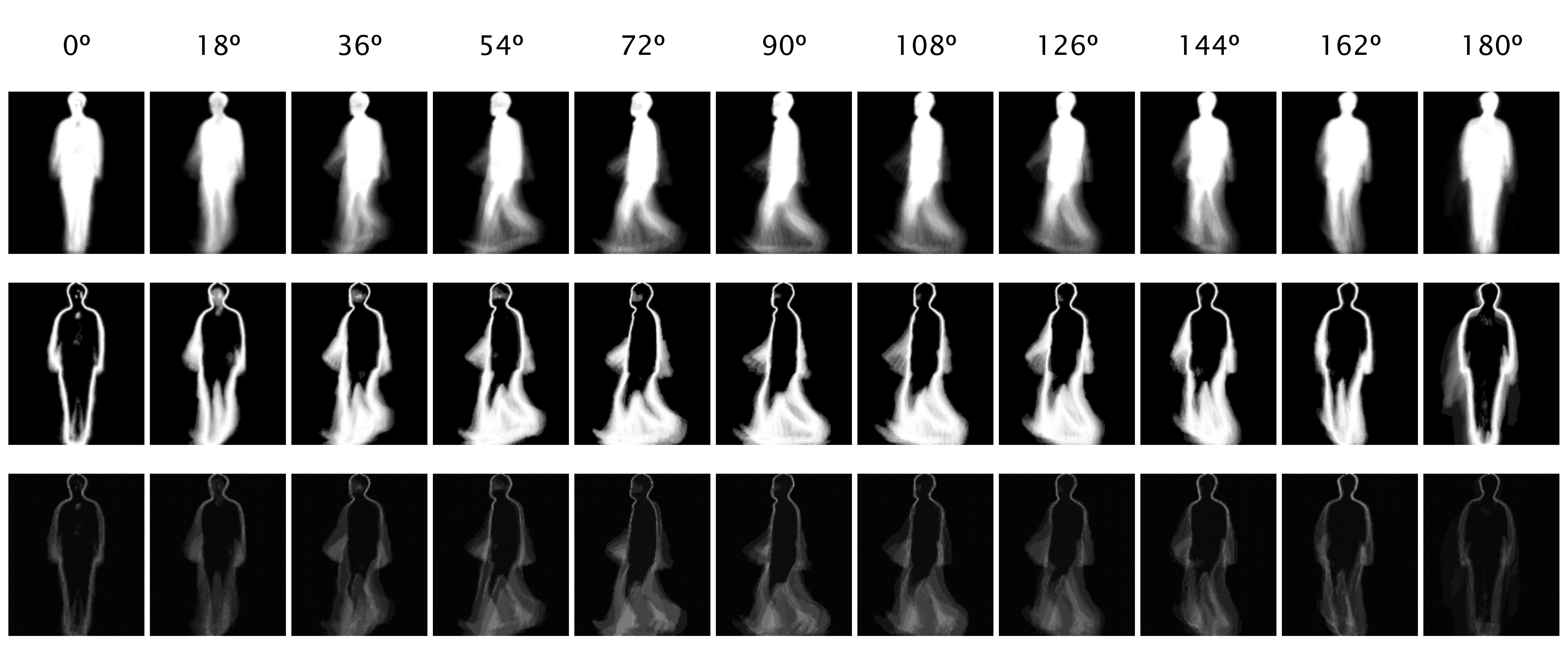}
  \caption{Gait templates taken from all 11 angles of CASIA-B of the same subject}
  \label{fig:templates}
\end{figure}
\end{landscape}

If $N$ is the number of frames in the subset, and $B_t$ is the normalized silhouette at frame $t$, then the GEI \citep{man2006individual} template of the gait instance is collated as
\[G_\text{GEI}(x,y) = \frac{1}{N}\sum_{t=1}^NB_t(x,y)\]
Similarly, the collation operation for the AEI \citep{zhang2010active} is given as follows assuming $B(0)= [0]$ (black frame).
\[G_\text{AEI}(x,y) = \frac{1}{N}\sum_{t=1}^N\big(B_t(x,y)-B_{t-1}(x,y)\big)\]
The GEnI \citep{bashir2010gait} template is given by
\[G_\text{GEnI}(x,y) = -z\log z - (1-z)\log (1-z)\]
where $z=p_1(x,y)$ is the probability of the pixel $(x,y)$ being white in the given subset of frames from $B_1$ through $B_N$. Samples of the above templates are illustrated in Figure~\ref{fig:templates} for different covariate conditions of CASIA-B.

\subsection{Learning Model}
\label{sec:gait-recog}

The gait template that is extracted is of the dimensions $240\times240$ which yields the total of 57600 pixels. Each of these pixels functions as a feature for recognition. Feeding an excessively large amount of features directly to a machine learning algorithm would not be advisable due to following reasons:
\begin{enumerate}[topsep=0pt, noitemsep, leftmargin=*, label=(\alph*)]
\item Most of these features may not correlate with the objective function, which in this case is gait recognition.
\item Many of the required features may be linearly correlated and hence are redundant in the feature set.
\item Selecting only the most useful features can considerably speed up the training time of the model.
\end{enumerate}

Hence the features require reduction before being processed by a machine learning algorithm. Feature reduction algorithms can either be supervised or unsupervised. Unsupervised feature reduction algorithms transform the given feature space of unlabelled  instances while the supervised alternative requires labelled instances for feature reduction.

For this application, the features are preprocessed by Principal Component Analysis (PCA) followed by a multi-class Linear Discriminant Analysis (LDA). The Multi-class LDA, also referred to as Multiple Discriminant Analysis (MDA)~\citep{duda2001pattern}, is a supervised dimensionality reduction method that would maximize inter-class distance while minimizing intra-class distance.  PCA~\citep{jolliffe2002PCA} is an unsupervised dimensionality reduction algorithm that projects the given features to feature space that corresponds to the highest variance. The use of PCA yields a net positive effect on the performance of the classifier concerning both processing time and accuracy. This type of MDA with PCA is sometimes referred to as Canonical Discriminant Analysis (CDA) \citep{rida2016human}.

\begin{figure}
  \centering
  \includegraphics[width=\linewidth]{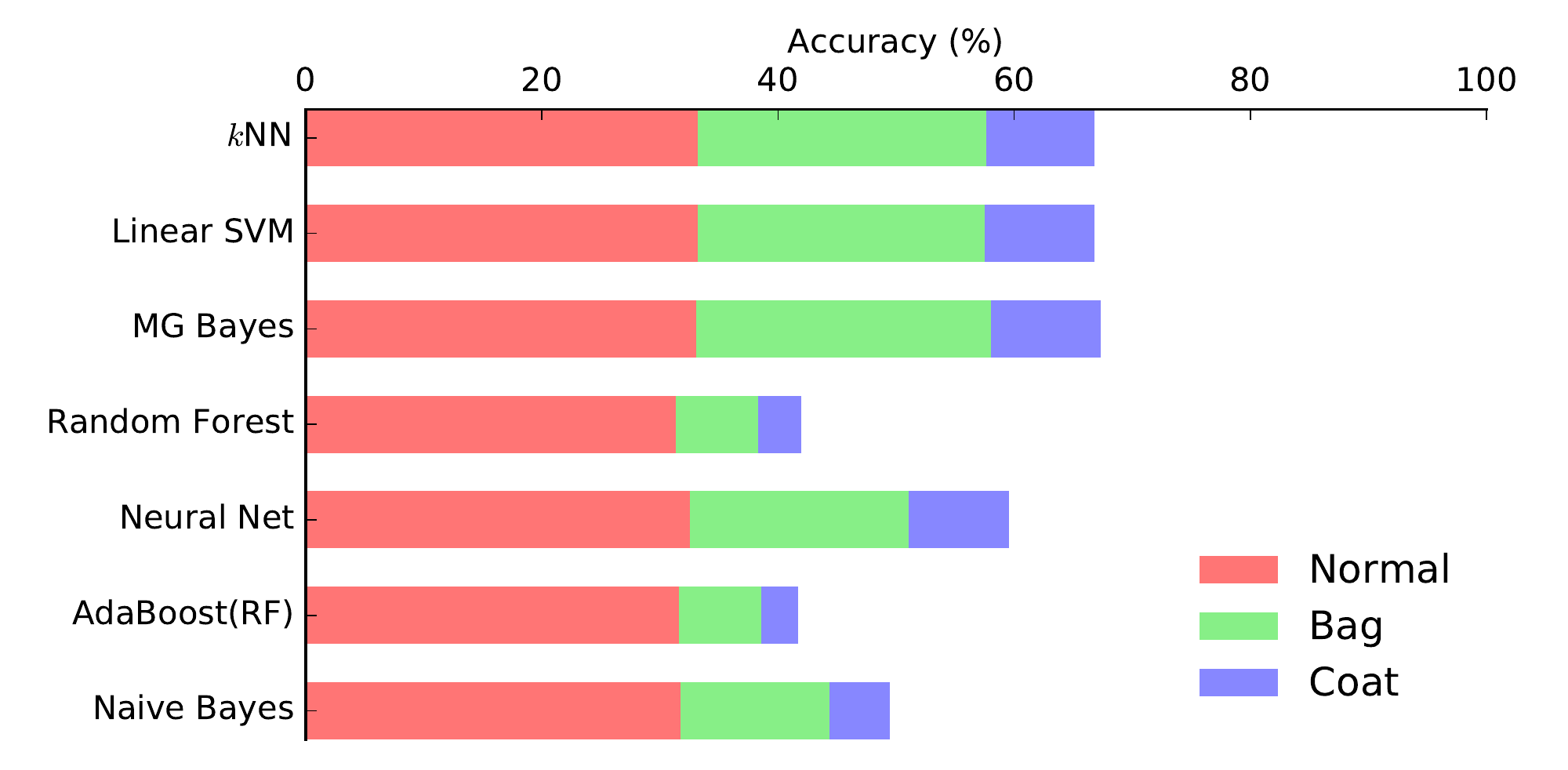}

  \vspace{1em}
  \caption{Performance comparison of classifiers using the GEI}
  \label{fig:classifier-comp}
\end{figure}

$k$NN is the standard classifier used for gait recognition. This study explores Bayes' rule as an alternative to $k$NN. The performance of these two classifiers along with a few other compatible classifiers on the GEI template is depicted in Table~\ref{tab:classifier-comp} and graphically illustrated in Figure~\ref{fig:classifier-comp}. The parameters of all the classifiers have been tuned to obtain the best possible performance. \cite{hastie2005elements} provides the detail for each of the algorithms applied. The Bayes' rule performs marginally better than $k$NN which perform equally well as the SVM with linear kernel.

\begin{table}
  \centering
  \caption{Classifier Performance Comparison using GEI on CASIA-B}
  \label{tab:classifier-comp}
  \renewcommand{\arraystretch}{1.2}
  \vspace{1em}
  
  \begin{tabular}{l *{5}{c}}
    \toprule
    Classifier & Normal & Bag & Coat & Mean & Std \\
    \cmidrule(lr){1-1}
    \cmidrule(lr){2-4}
    \cmidrule(lr){5-6}
    $k$NN & 99.60 & 73.39 & 27.42 & 50.40 & 32.50 \\
    Linear SVM & 99.60 & 72.98 & 27.82 & 50.40 & 31.93 \\
    MG Bayes & 99.19 & 75.00 & 27.82 & 51.41 & 33.36 \\
    Random Forest & 93.95 & 20.97 & 10.89 & 15.93 &  7.13 \\
    Neural Net & 97.58 & 55.65 & 25.40 & 40.52 & 21.38 \\
    AdaBoost(RF) & 94.76 & 20.97 &  9.27 & 15.12 &  8.27 \\
    Naive Bayes & 95.16 & 37.90 & 15.32 & 26.61 & 15.97 \\
    \bottomrule
  \end{tabular}
  
\end{table}

Although it is well adapted to many problems, Random Forest (RF) \citep{breiman2001random} does not apply well the problem of gait recognition. Due to its potential in remote sensing \citep{ulc}, the RF's parameters were optimized and further improved with AdaBoost\footnote{The details of AdaBoosted Random Forest and its optimization is explained in the paper \cite{ulc-ebe}}. However, applying AdaBoosted Random Forest does not make an improvement either. This is because Random Forest is not as much effective when the number of instances per class is small, especially when the system contains highly dependent features. In this case, the training set only contains four video samples for each subject, i.e., four instances per class. This would mean that the RF does not have enough samples to learn the pattern well.

From the above discussion, we can arrive at a decision to use the Bayes' rule for the classification step of the gait recognition algorithm. Let $x$ be an instance composed of features $[x_1, x_2, x_3, ..., x_n]$ and $y$ be the discrete outcome, then Bayes' rule is given by
\[\Pr(y \peq k \mid x) = \frac{\Pr(x \mid y \peq k) \cdot \Pr(y \peq k)}{\Pr(x)}\]
where $\Pr(x \mid y \peq k)$ is called the likelihood. The likelihood can be either na\"ive or multivariate. In the na\"ive Bayes model, the likelihood is the product of the probability density functions $f$ of multiple univariate Gaussian distributions like so
\[ \Pr(x \mid y \peq k) = \prod_{i=1}^nf(x_i \mid \mu_k^{(i)}, \sigma_k^{(i)})\]
The complete Bayes' model, on the contrary, employs multivariate Gaussian probability density as follows.
\begin{align}
  \Pr(x \mid y \peq k) &= f(x \mid \mu_k,\Sigma_k) \nonumber\\
                       &= \frac{1}{(2\pi)^n|\Sigma_k|^{\frac{1}{2}}}\exp\Big(-\frac{1}{2}(x-\mu_k)'\Sigma_k^{-1}(x-\mu_k)\Big) 
\end{align}
For LDA, $\Sigma_k=\Sigma~\forall k$, where $\Sigma$ is the covariance matrix of the whole transformed set of gallery features \citep{hastie2005elements}. The results in Figure~\ref{fig:classifier-comp} clearly show that the multivariate model performs much better than the na\"ive model.

\subsection{Genetic Template Segmentation}
\label{sec:chromosome}

\begin{figure}[b]
  \centering
  \includegraphics[width=0.4\linewidth]{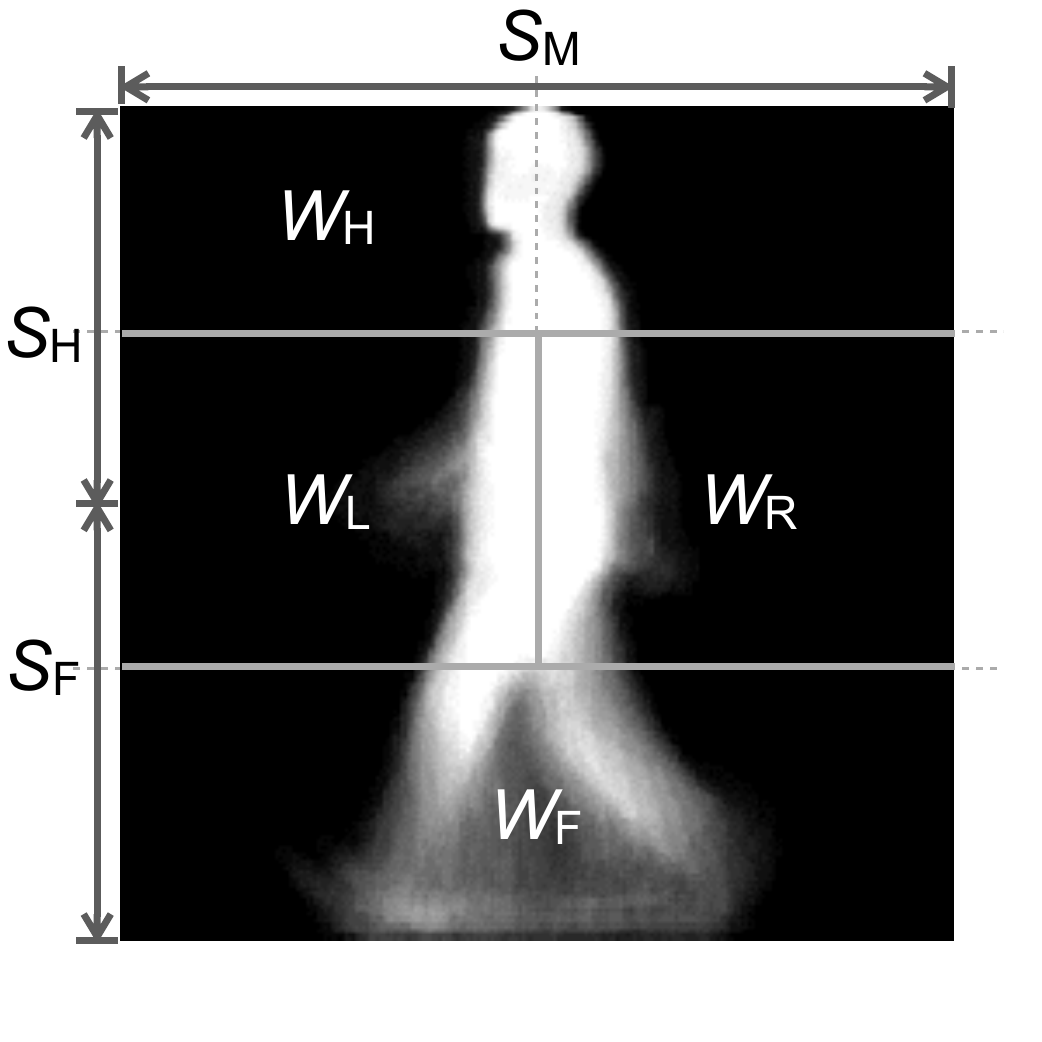}
  \caption{Parameters optimized by the genetic algorithm}
  \label{fig:genparams}
\end{figure}

The boundary selection process is automated through GA to find the optimum boundary to segment the gait template before the actual training process. The gait template is to be split into four segments, viz., head portion H, leg portion F, mid-left section L and mid-right section R. The parameters to be optimized are the split points which divide these sections and a weight bit per region to decide whether the respective region should be included in the training as shown in Figure~\ref{fig:genparams}. This process is used to produce a masking template for each view angle.
The chromosome structure for the genetic optimization is given as
\[[S_\text{H}, S_\text{M}, S_\text{F}, W_\text{H}, W_\text{L}, W_\text{R}, W_\text{F}]\]
The variables denoted $S_i$ are split variables that determine the boundary for the region to segment and is represented by 8 bits each. $S_\text{H}$ defines the line between the head portion and the midsections; $S_\text{F}$ determines the split between the midsections and leg region; $S_\text{M}$ divides the two midsections. If $d$ is the decimal equivalent of the 8 bits used to represent the split variables, then its value can be decoded as
\[S_i = \text{min}_i + (\text{max}_i-\text{min}_i) \times d_i/255\]
where min$_i$ and max$_i$ are the minimum and maximum possible values for the variable $S_i$. The variables $W_i$ are binary variables that determine whether the segment is included for training, 1 indicates inclusion while 0 represents masking. The total size of chromosome hence becomes 28 bits.

A set of subjects with all covariates included is used as a tuning set to determine boundary locations for segmentation. The fitness function evaluates the hypothesis generated by the chromosome against the tuning set to produce a fitness measure. The three covariates considered here is A: normal walk, B: carrying a bag and C: clothing condition. If the fitness measure is simply set to the average of the accuracy of the three covariate sets, then the GA would make a significant trade-off on the normal walk sequence to maximize the overall accuracy. This was experimentally observed to at 90\% while the state-of-the-art approaches produce accuracies of above 95\% \citep{rida2016human}. The fitness measure, $F$ for a given chromosome, $h$, is calculated as
\[F(h) = \Big(\frac{1}{2}\text{CCR}_\text{A}(h) 
  + \frac{1}{6}\text{CCR}_\text{B}(h) 
  + \frac{1}{3}\text{CCR}_\text{C}(h)\Big)^2 \]
where CCR$_K$ represents the CCR for the corresponding covariate $K$. Giving equal weights to each of the CCR$_k$ causes a trade-off in normal condition performance leading to an accuracy of 95.6\% which is among the lowest of the normal CCR (refer Table~\ref{tab:ccr}). Thus, the highest priority was given to CCR of the normal setting, CCR$_\text{A}$, to compete with the state of the art. In most approaches, clothing conditions pose the greatest challenge to template-based recognition systems. Hence the accuracy pertaining to the clothing condition, CCR$_\text{C}$, was given the next highest weight after normal setting to boost its accuracy on par with the carrying condition, CCR$_\text{B}$. These priority weights were assigned empirically.

The elitist selection variant of the generation propagation is used for this implementation of the GA \citep{baluja1995removing}. That is, the chromosome corresponding to the highest fitness of a generation T$_n$ is made sure to be propagated to the next generation T$_{n+1}$. The GA is set to follow a uniform crossover with probability 0.6, a single bit mutation probability of 0.03 and populates 20 chromosomes per generation. The optimization runs for 15 generations although convergence was mostly attained before the 8\textsuperscript{th} generation during experimental observation.

\begin{figure}[t]
  \centering
  \includegraphics[width=\linewidth]{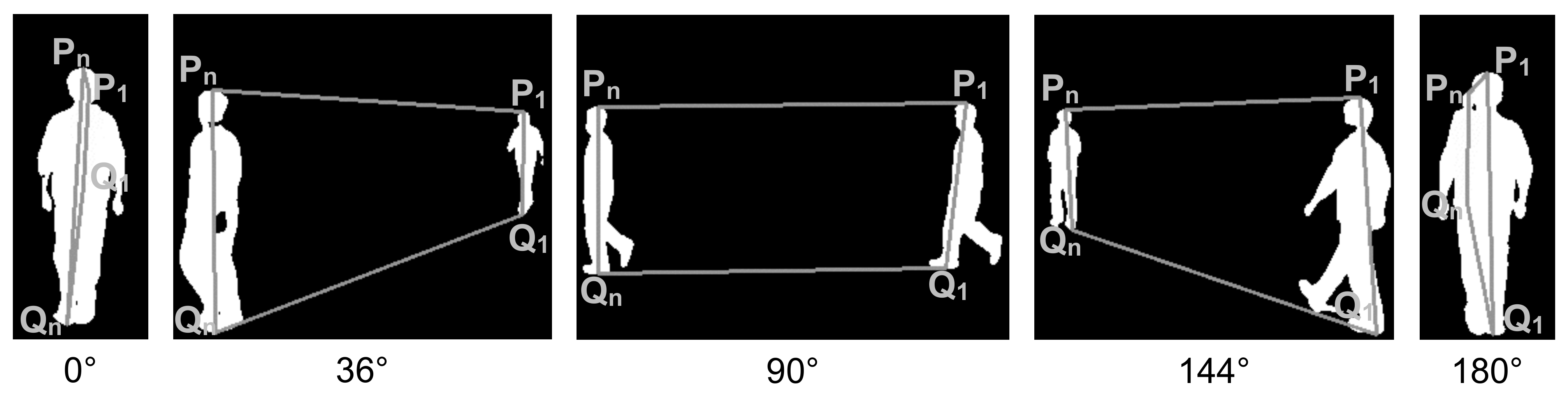}
  \caption{Estimating viewpoints from different views}
  \label{fig:viewest}
\end{figure}

\subsection{View Estimation}
\label{sec:view-est}
Under the assumption that the subjects walk in a straight line for verification, the first and last visible silhouettes, $S_1$ and $S_n$, are taken into consideration. Let $P_1$ and $Q_1$ be the topmost and bottom-most points of $S_1$ as illustrated in Figure~\ref{fig:viewest}.  Similarly, $P_n$ and $Q_n$ denote the topmost and bottom-most points of $S_n$. Let $m_P$ and $m_Q$ be the slopes of the lines $P_1P_n$ and $Q_1Q_n$ respectively. These two slopes alone form the features required to train the view-estimation classifier with the view as output labels. To reduce the number of cases, the sequence is passed through a simple check to verify whether the angle lies in the coronal plane ($0\degree$ or $180\degree$). If the last silhouette overlaps the first, then the viewpoint is determined to be at $0\degree$ and the direct opposite for $180\degree$. If both of these cases fail, then the angle should be one among those other than the two in the coronal plane.

\newpage
\section{Experimentation \& Evaluation}
\label{sec:gts-results-evaluation}

\subsection{Dataset Configuration}
\label{sec:gts-datas-conf}

The CASIA dataset B is the benchmark gait database used for the experimental validation. The dataset includes six instances of normal walk (SetA), two instances of walking while carrying a bag (SetB) and two instances of walking while wearing an overcoat (SetC) of 124 individuals. SetA is split to SetA1 containing four instances and SetA2 for the remaining. Only SetA1 is used for training. 24 subjects were randomly selected from the CASIA-B dataset to participate in the tuning set. These subjects were removed from the gallery for the evaluation phase just as in the work of \cite{jia2015view}.

\subsection{Sagittal View Performance}
\label{sec:sagitt-view-perf}

\begin{figure}
  \centering
  \includegraphics[width=1.0\linewidth]{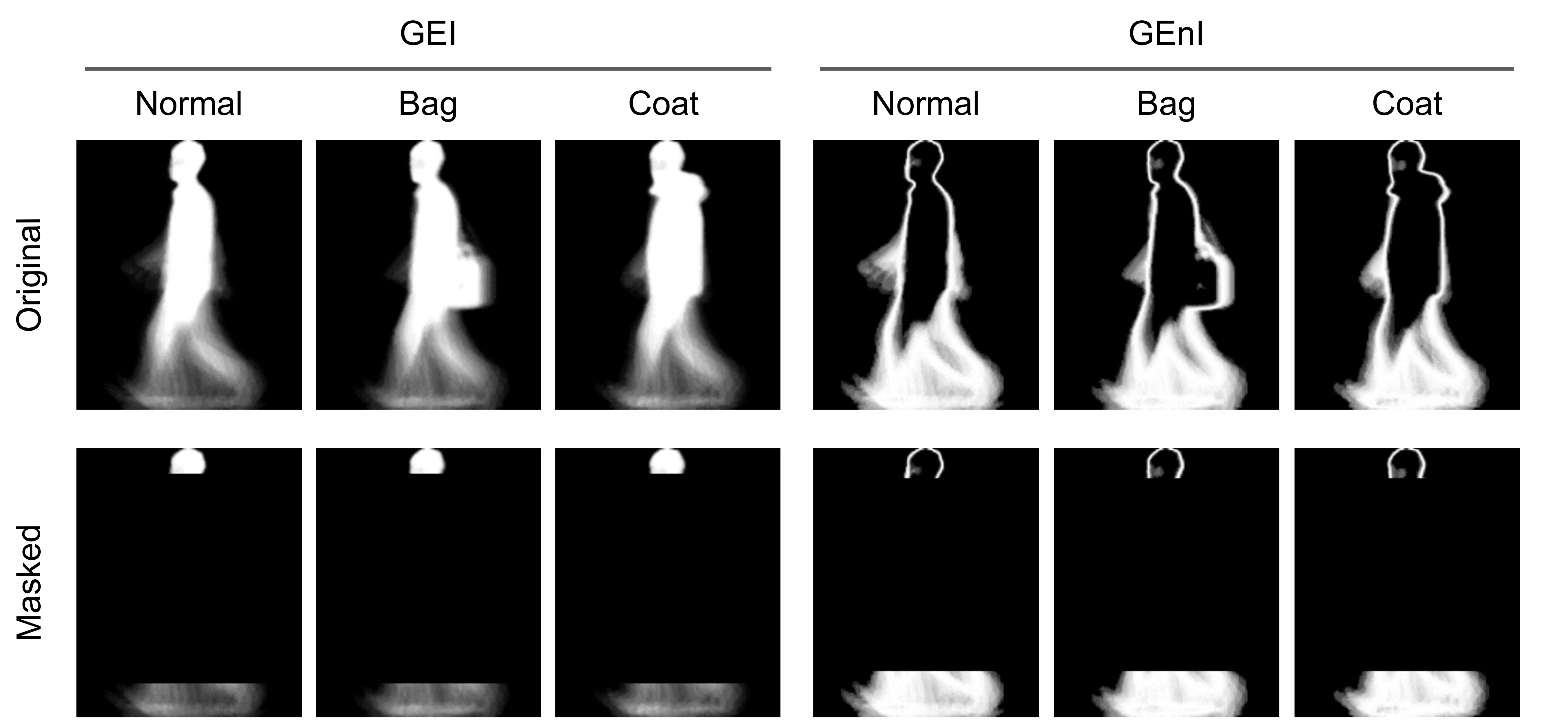}
  \caption{Template transformation using the GTS hypothesis}
  \label{fig:gtsmask}
\end{figure}

The experiments were first executed under the sagittal angle, 90$\degree$ view, to focus on the effect of carrying and clothing covariates. The GEI, GEnI, and AEI were used as the base templates. The templates before and after GTS appear as shown in Figure~\ref{fig:gtsmask}. The performance of the proposed GTS is compared against that claimed by other approaches in Table~\ref{tab:ccr}.

\begin{table}
  \centering
  \caption{CCR(\%) of Different Algorithms on
      CASIA Dataset B at $90\degree$ view}
  \label{tab:ccr}
  \renewcommand{\arraystretch}{1.2}
  \centering
  
  \begin{tabular}{c p{3cm} *{5}{c}}
    \toprule
    Year & Method & Normal & Bag & Coat & Mean & Std \\
    \cmidrule(lr){1-2}
    \cmidrule(lr){3-5}
    \cmidrule(lr){6-7}
    2006 & Han and Bhanu \nocite{man2006individual} & 99.60 & 57.20 & 23.80 & 60.20 & 37.99 \\
    2010 & Bashir et al. \nocite{bashir2010gait} & \textbf{100.0} & 78.30 & 44.00 & 74.10 & 28.24 \\
    2013 & Dupuis et al. \nocite{dupuis2013feature} & 98.43 & 75.80 & 91.86 & 88.70 & 11.64 \\
    2014 & Kusakunniran \nocite{kusakunniran2014attr} & 94.50 & 60.90 & 58.50 & 71.30 & 20.13 \\
    2015 & Arora et al. \nocite{Arora2015} & 98.00 & 74.50 & 45.00 & 72.50 & 26.56\\
    2015 & Yogarajah et al. \nocite{Yogarajah20153} & 97.60 & 89.90 & 63.70 & 83.73 & 17.77 \\
    2016 & Rida et al. \nocite{rida2016human} & 98.39 & 75.89 & 91.96 & 88.75 & 11.59 \\
    2017 & GEI with GTS & 98.00 & \textbf{95.50} & \textbf{93.00} & \textbf{95.50} &  \textbf{2.50} \\
    2017 & GEnI with GTS & 97.00 & 95.00 & 91.00 & 94.33 &  3.06 \\
    2017 & AEI with GTS & 89.50 & 85.50 & 77.50 & 84.17 &  6.11 \\
  \bottomrule
  \end{tabular}%
\end{table}

The upper portion of the gait template segmented by the GA chose only the head of the subject and neglected the shoulders as opposed to what was selected by \cite{jia2015view}. The GA detected that the shoulder metric would lead to a considerable loss in accuracy while wearing an overcoat and hence chose $S_\text{H}$ a little before the shoulder region.

It is evident from the previously reported results in Table~\ref{tab:ccr} that the clothing condition is the most challenging covariate leading to a lesser CCR. Clothing conditions cause a greater change in the subjects' silhouettes. As template-based methods rely on spatiotemporal changes of the silhouettes during gait, the recognition performance is adversely affected. A more efficient performance is attained when the regions that have an impact on such covariates are masked out. The arm-swing constraints imposed by the weight of the clothing and the carrying condition would compromise the accuracy at the midsection. As speculated, the mid-left and mid-right sections were ignored in the optimal hypothesis generated by the GTS for every angle and for each type of gait template. Note that the segmented GEI has a much smaller lower (foot) section due to the greater effect of the covariates on the GEI template. The area permitted by the mask is 25.2\% of the total template area; neglecting the constant features, only 8.4\% of the feature space is utilized. Nevertheless, the GEI masked with GTS outperforms the existing methods.

Genetic algorithm is known to have a tendency to give suboptimal results. There comes a requirement to tune the parameters after the genetic algorithm converges. The outcome of the GTS shows that only two parameters are variable: $S_\text{H}$ and $S_\text{F}$. That is, weight bits are optimally assigned as
\[[W_\text{H}, W_\text{L}, W_\text{R}, W_\text{F}]=[1,0,0,1]\]
This assignment leaves $S_\text{M}$ irrelevant as both mid-sections are ignored. These two variables can be sequentially optimized starting with $S_\text{F}$ with a fixed $S_\text{H}$ and then $S_\text{H}$ with the optimized $S_\text{F}$. This process is also followed using the tuning set for validation.

\subsection{View-Invariant Performance}
\label{sec:view-invar-perf}

\begin{figure}[t]
  \centering
  \includegraphics[width=\linewidth]{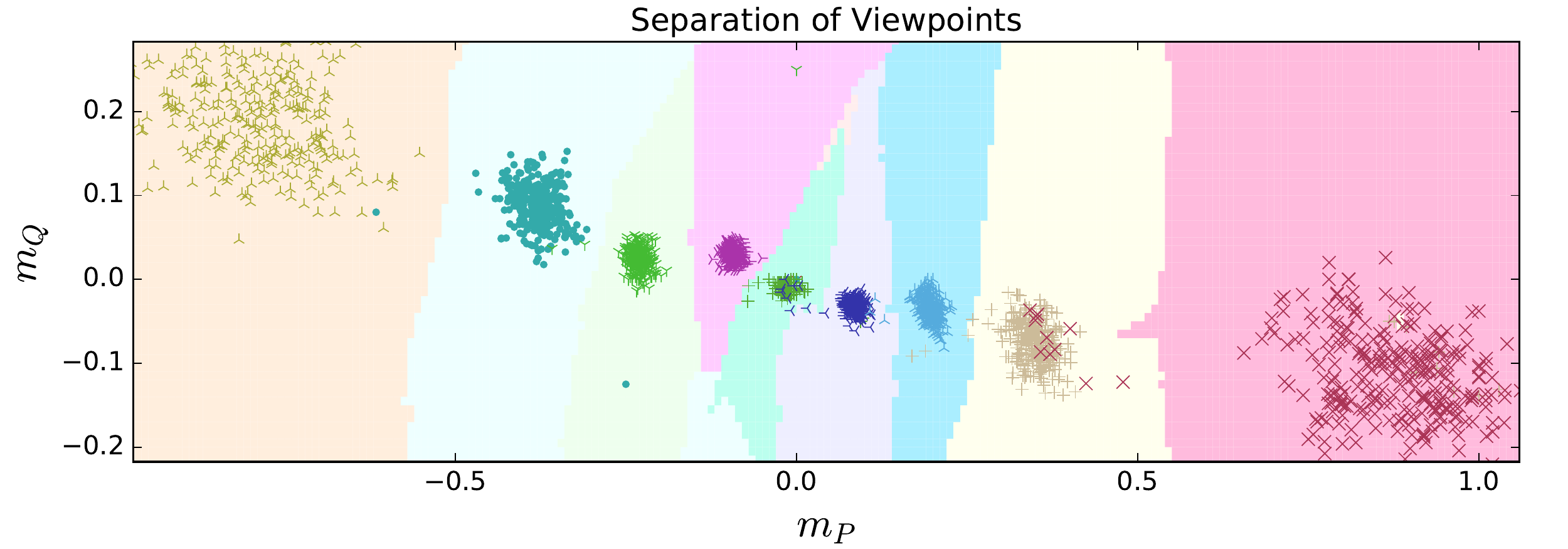}
  \caption{Boundary separation of the view estimator}
  \label{fig:view-sep}
\end{figure}

The GTS is applied to generate one masking template for every angle using the tuning set. The tuning set is also used to train the view estimator. The evaluation set is separated into gallery and probe sets. After which, 11 LDA-Bayes classifiers are trained (one for each view angle) using the gallery set. The angle of each instance of the probe set is predicted with the view estimator. The boundary separation of the view estimator is depicted in Figure~\ref{fig:view-sep}. The instance is then passed to the appropriate view-specific classifier for the identity prediction. Note that each angle set also has its own CDA transformation. PCA is set to retain 99\% of data variance. This resulted in retaining a different number of eigenvectors for each angle for a given template. The numbers range from 123 to 181 for GEI, 147 to 181 for GEnI, and 95 to 147 for AEI. The number of features through LDA is at most one less than the number of classes, which is 99 in this case.

\begin{figure*}
  \centering
  \subfloat[]{\includegraphics[width=0.48\linewidth]{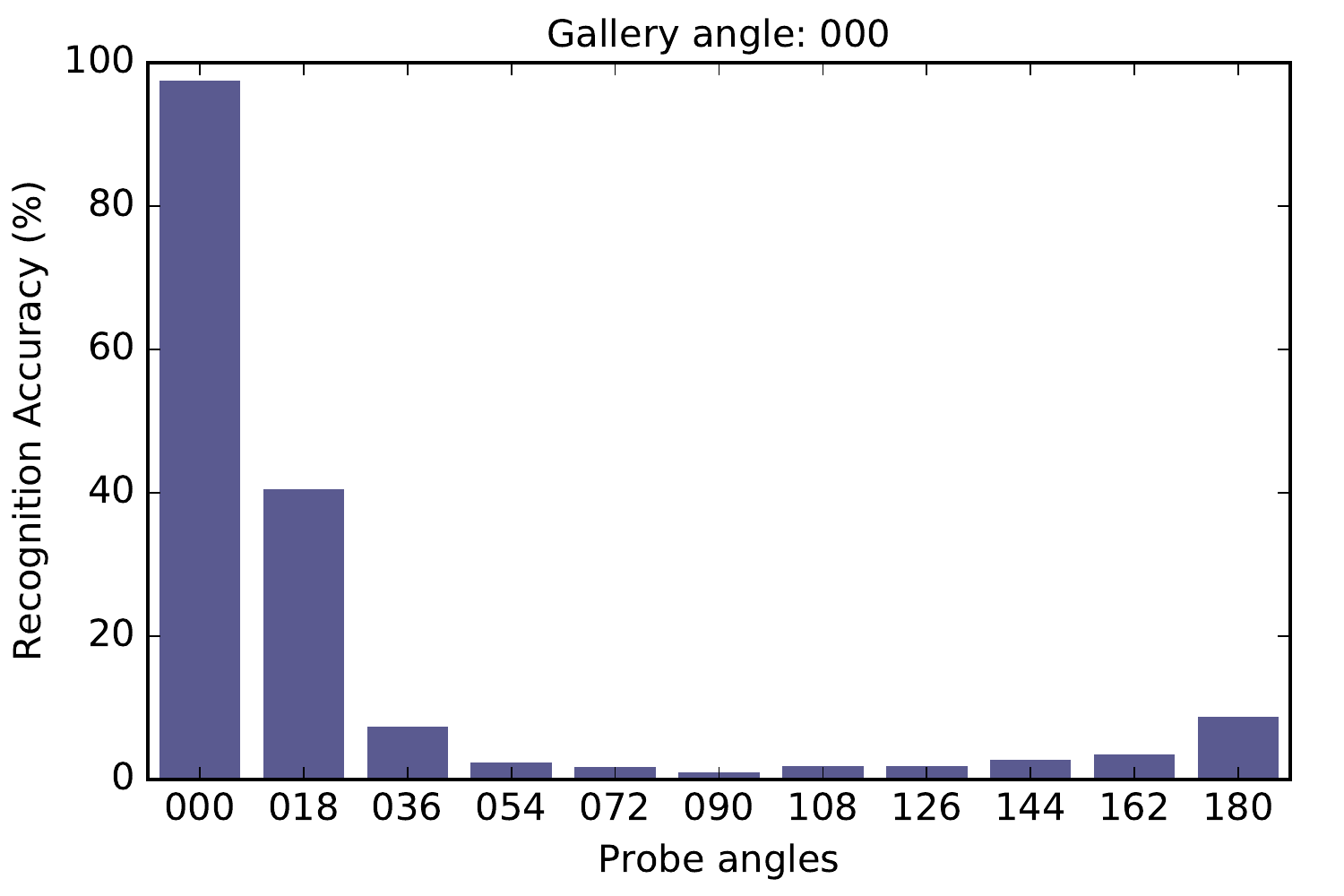}%
    \label{fig:gts-000}}
  \hfil
  \subfloat[]{\includegraphics[width=0.48\linewidth]{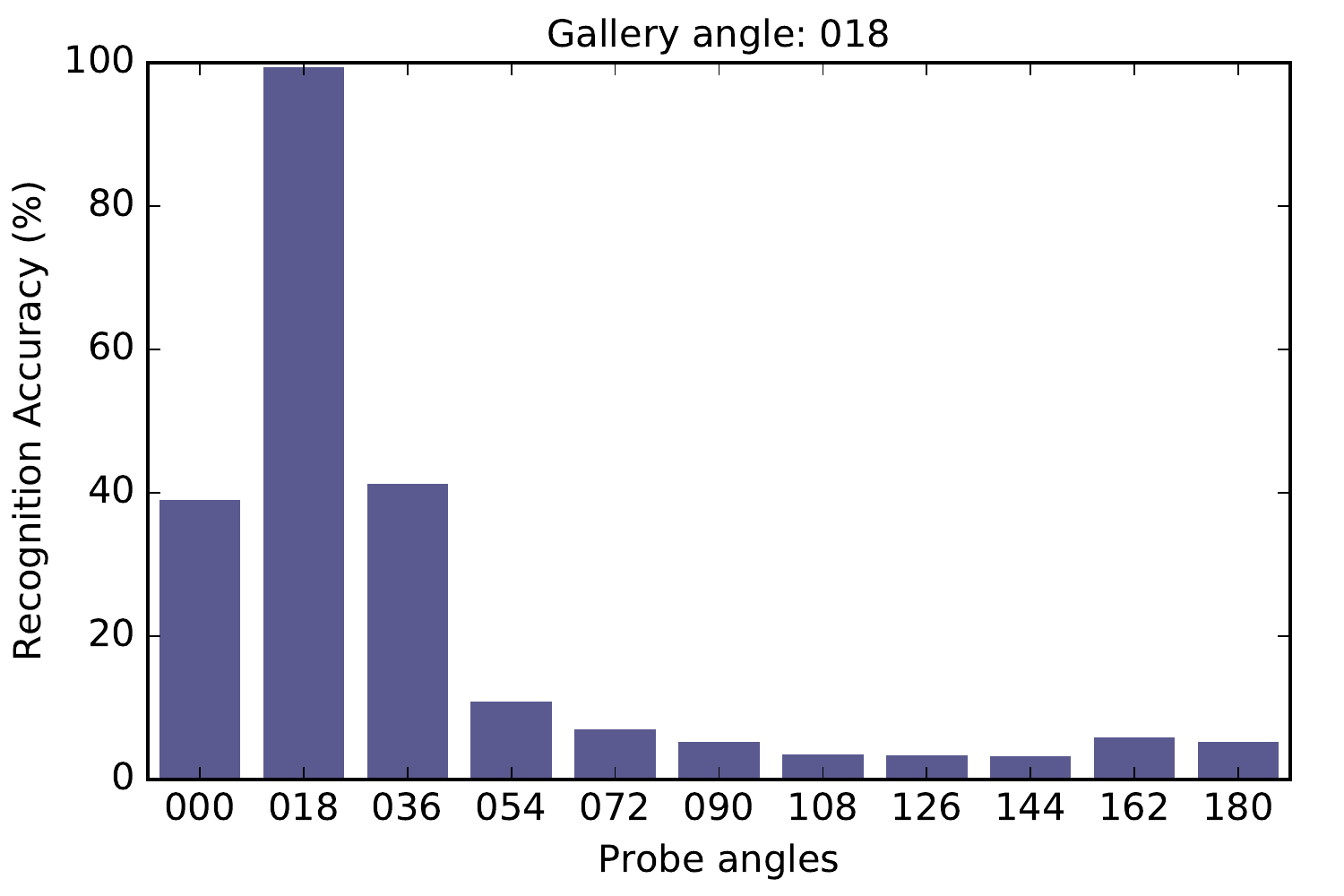}%
    \label{fig:gts-018}}
  
  \subfloat[]{\includegraphics[width=0.48\linewidth]{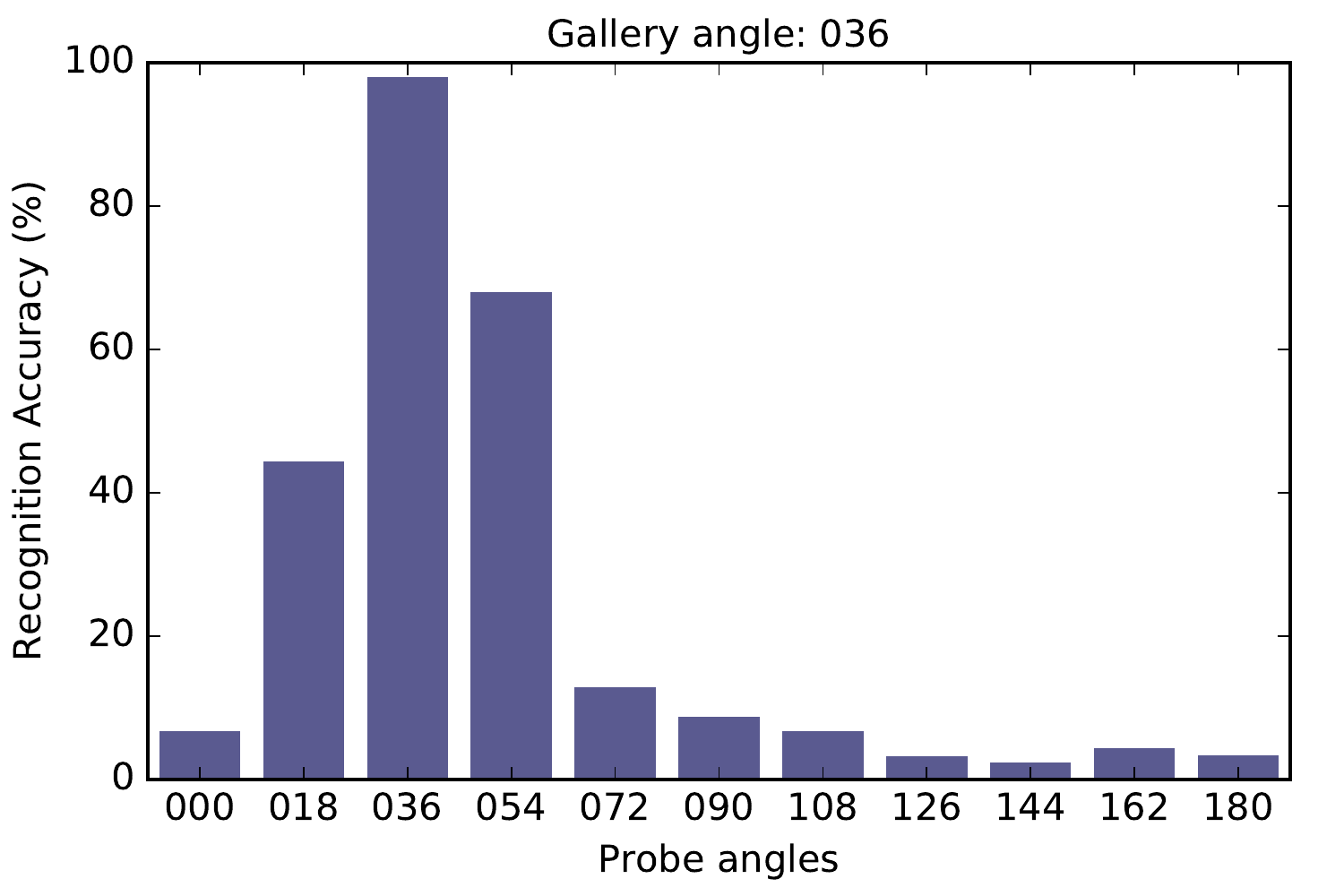}%
    \label{fig:gts-036}}
  \hfil
  \subfloat[]{\includegraphics[width=0.48\linewidth]{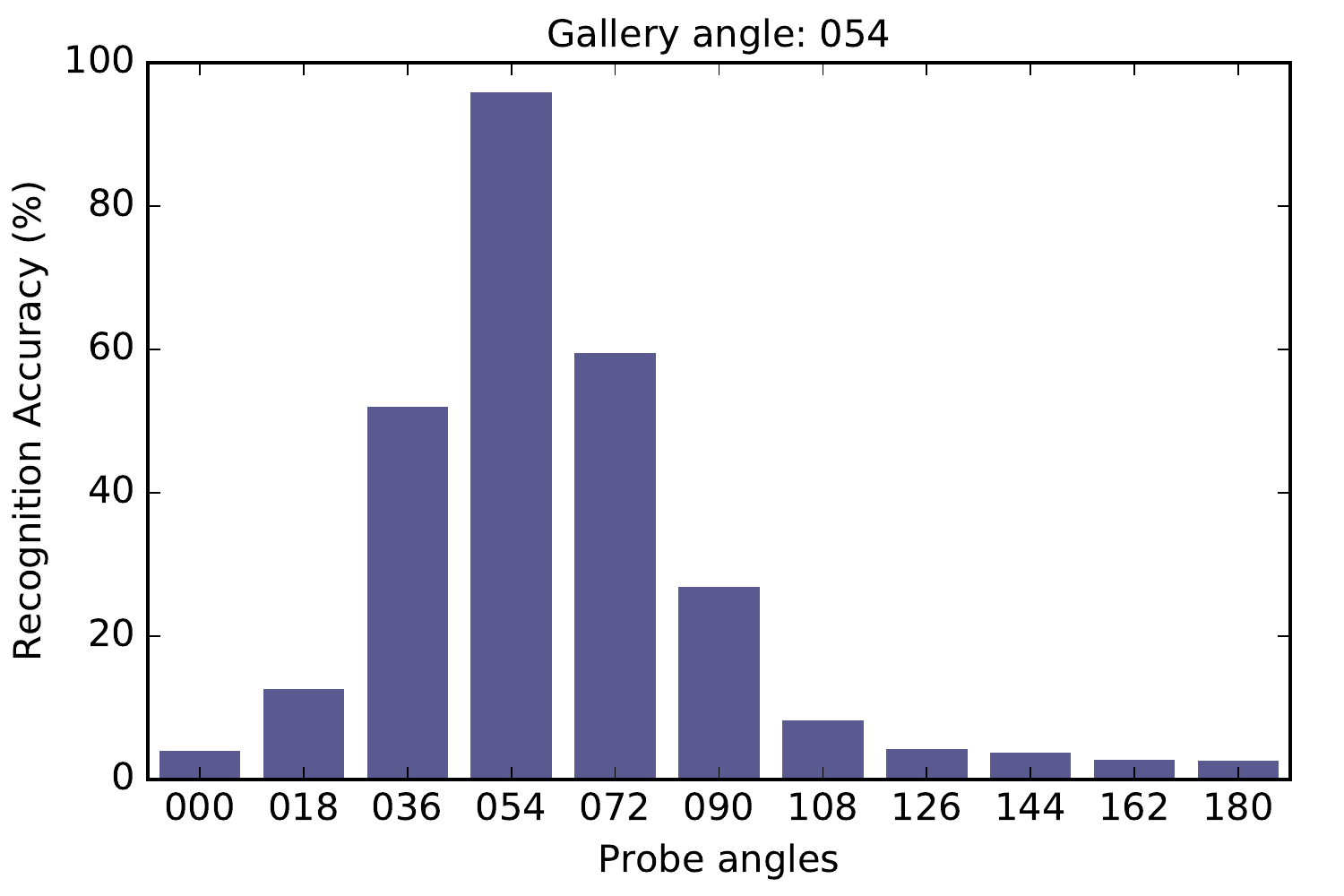}%
    \label{fig:gts-054}}
  
  \subfloat[]{\includegraphics[width=0.48\linewidth]{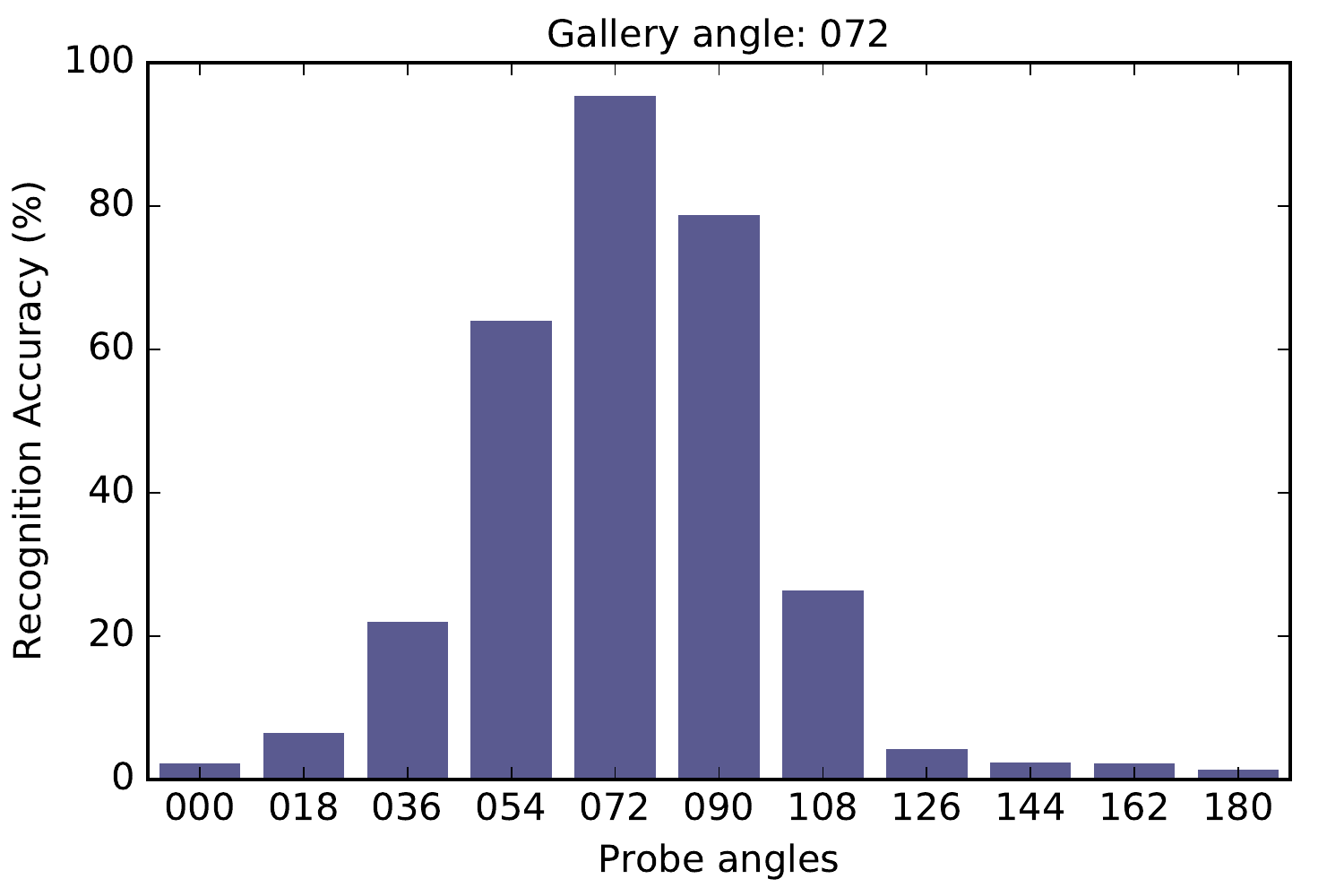}%
    \label{fig:gts-072}}
  \hfil
  \subfloat[]{\includegraphics[width=0.48\linewidth]{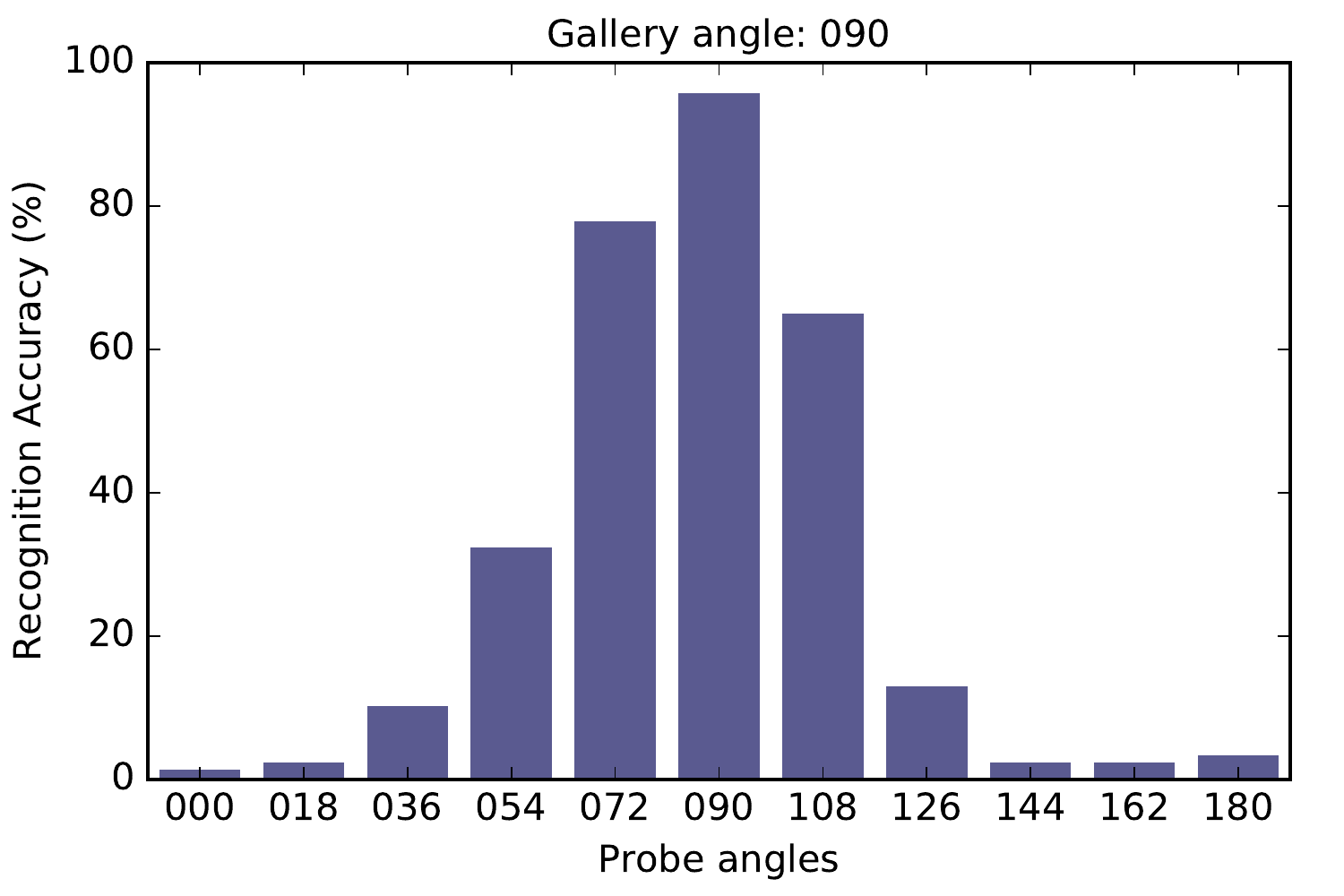}%
    \label{fig:gts-090}}
  \caption{GTS performance on angles (Part-1)}
  \label{fig:all-angle-acc1}
\end{figure*}

\begin{figure*}
  \centering
  \subfloat[]{\includegraphics[width=0.48\linewidth]{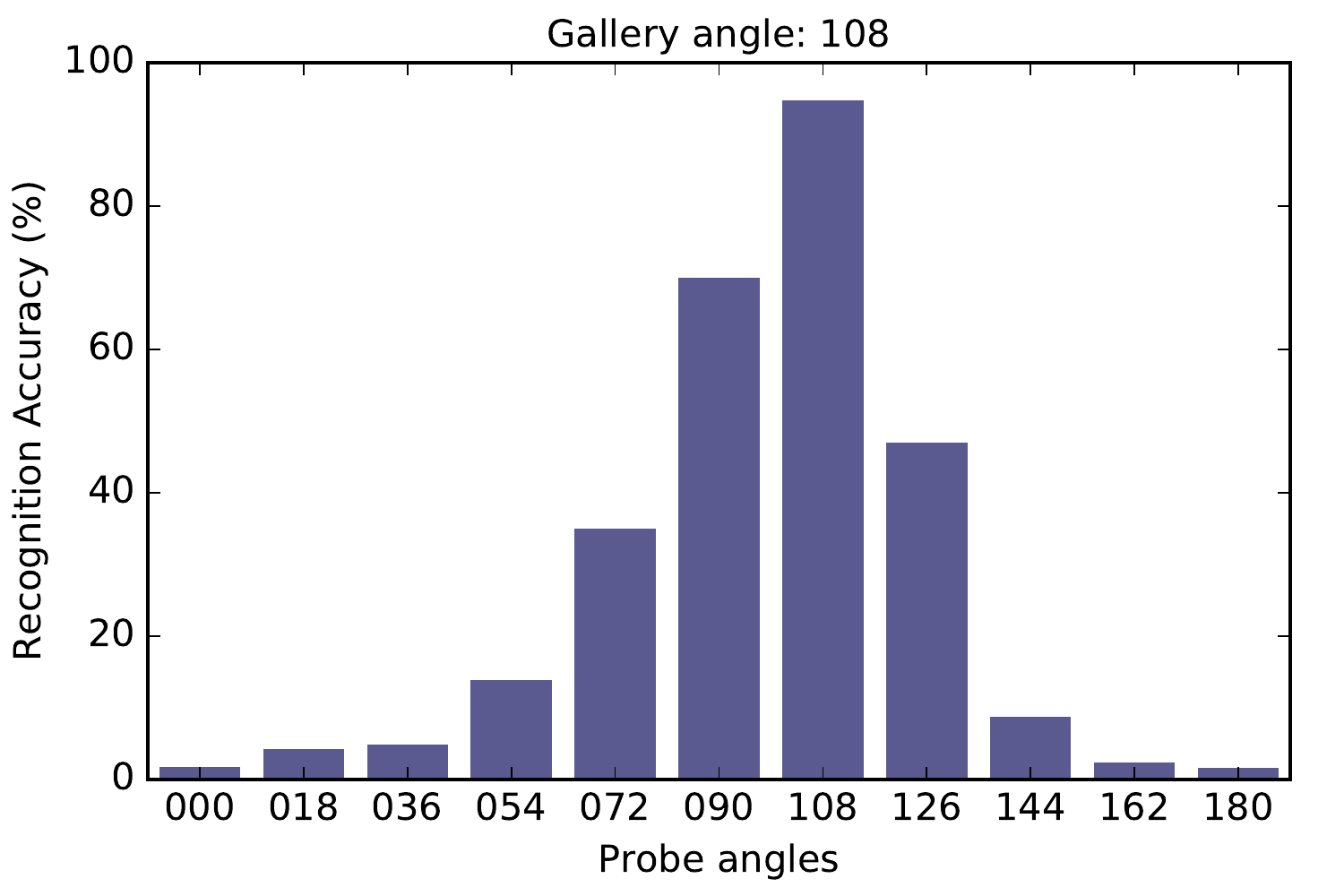}%
    \label{fig:gts-108}}
  \hfil
  \subfloat[]{\includegraphics[width=0.48\linewidth]{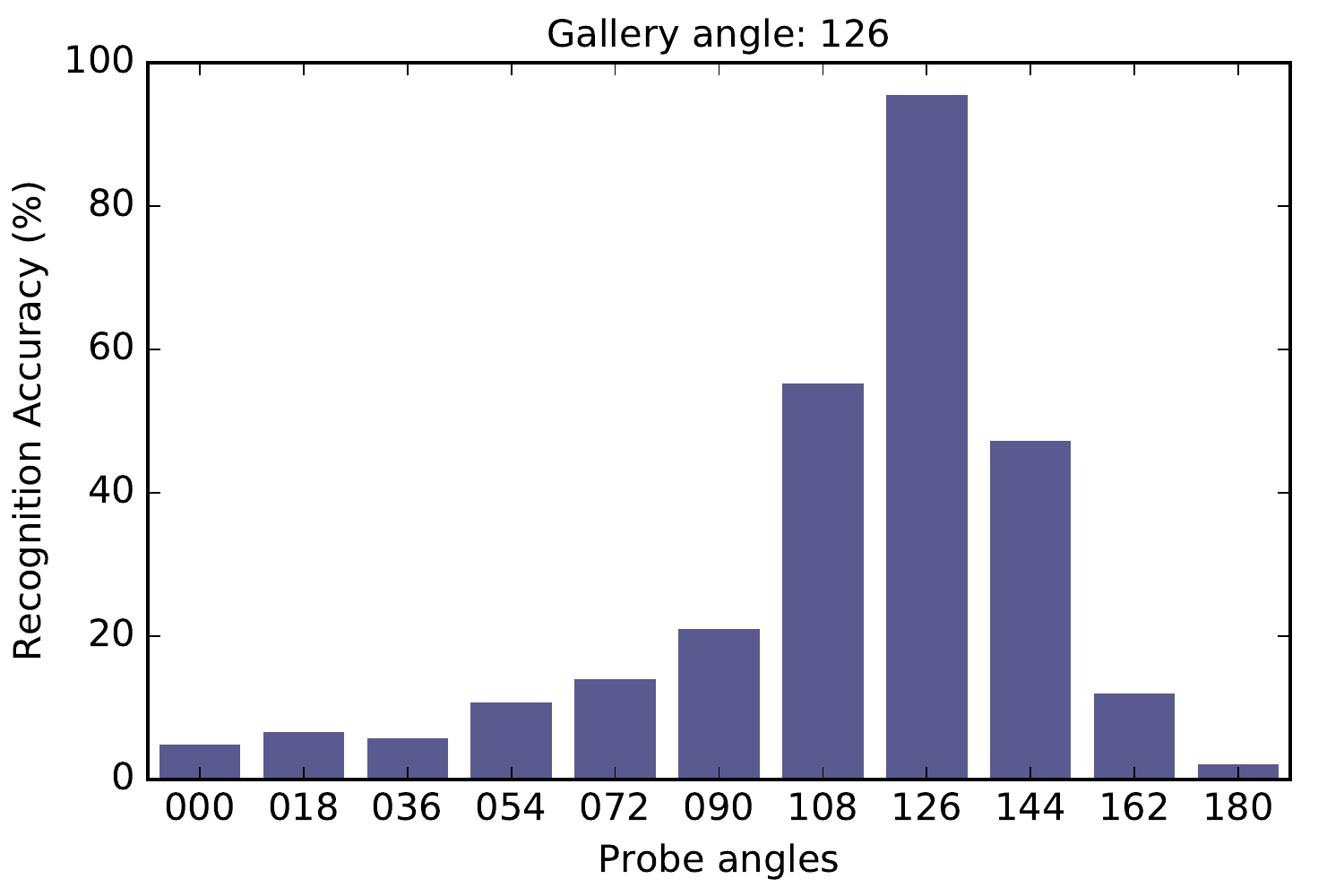}%
    \label{fig:gts-126}}

  \subfloat[]{\includegraphics[width=0.48\linewidth]{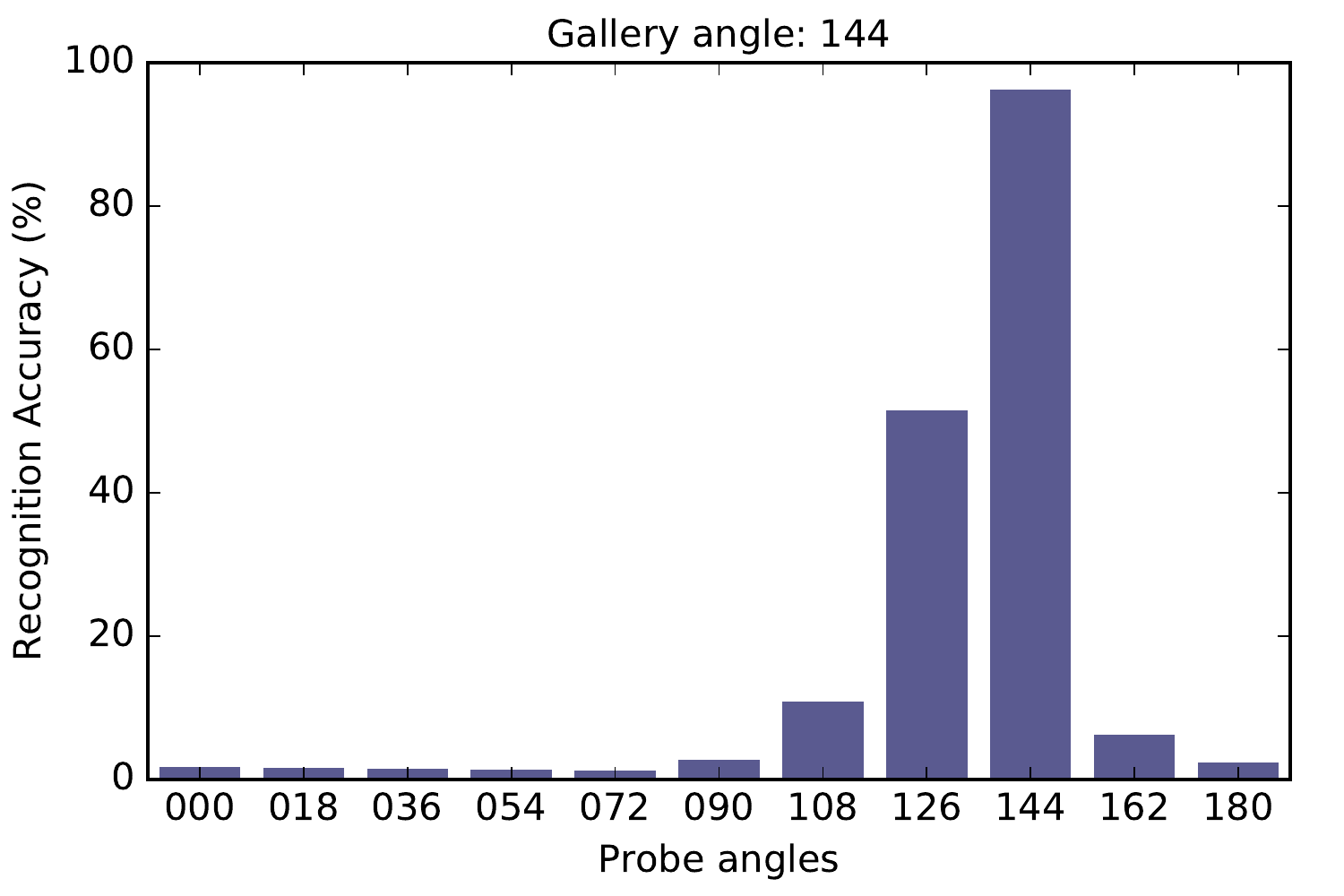}%
    \label{fig:gts-144}}
  \hfil
  \subfloat[]{\includegraphics[width=0.48\linewidth]{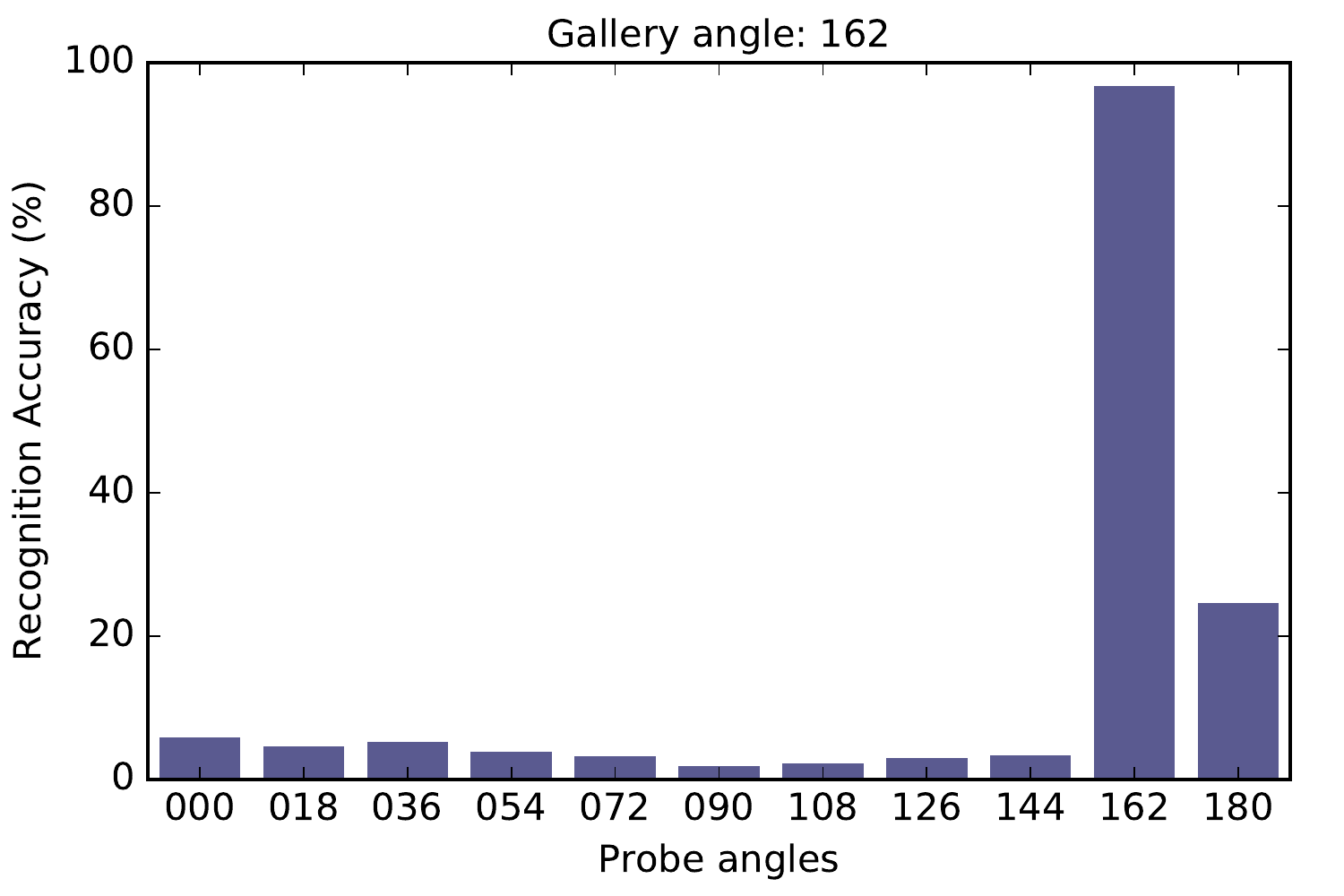}%
    \label{fig:gts-162}}
  
  \subfloat[]{\includegraphics[width=0.48\linewidth]{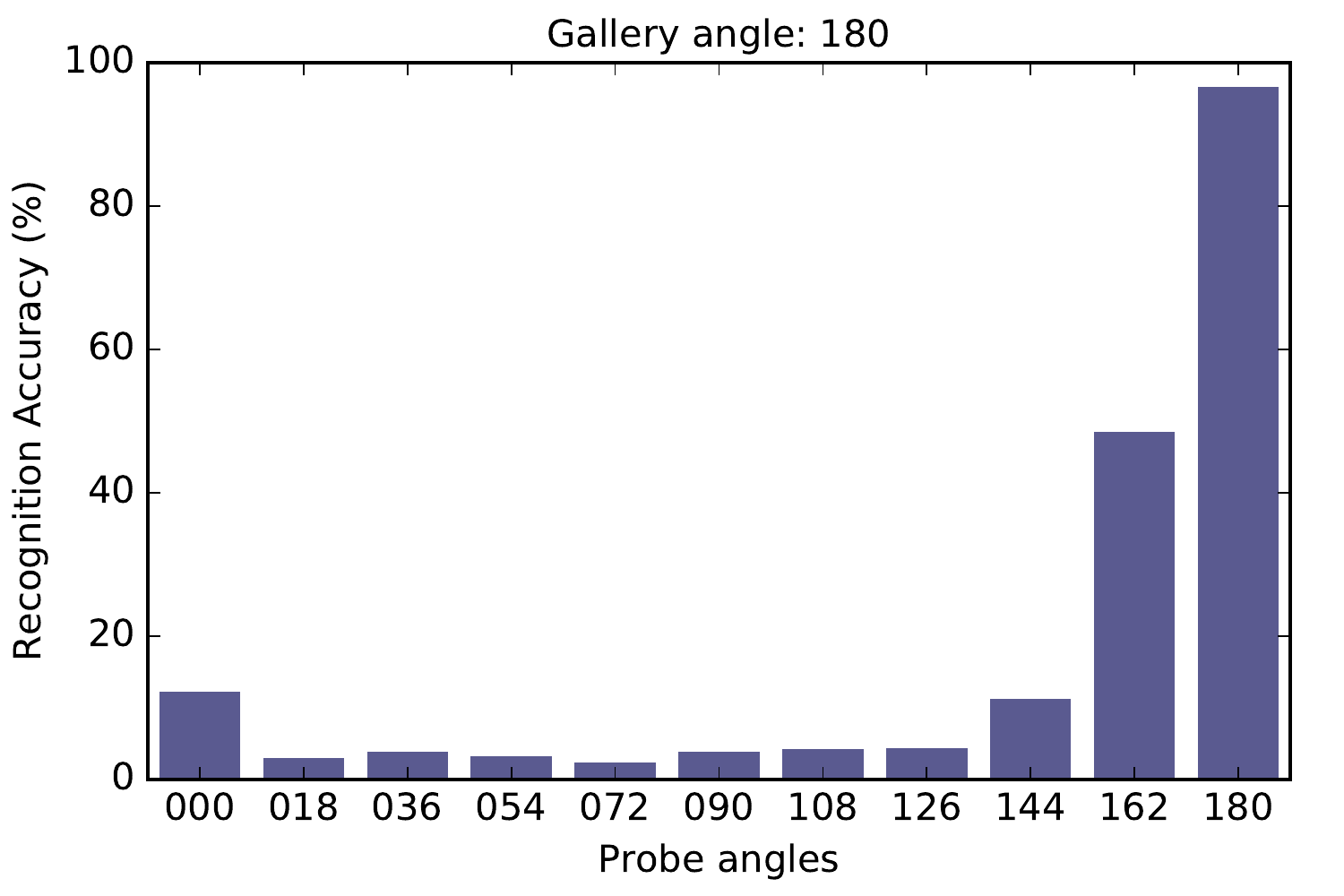}%
    \label{fig:gts-180}}
  \hfil
  \subfloat[]{\includegraphics[width=0.48\linewidth]{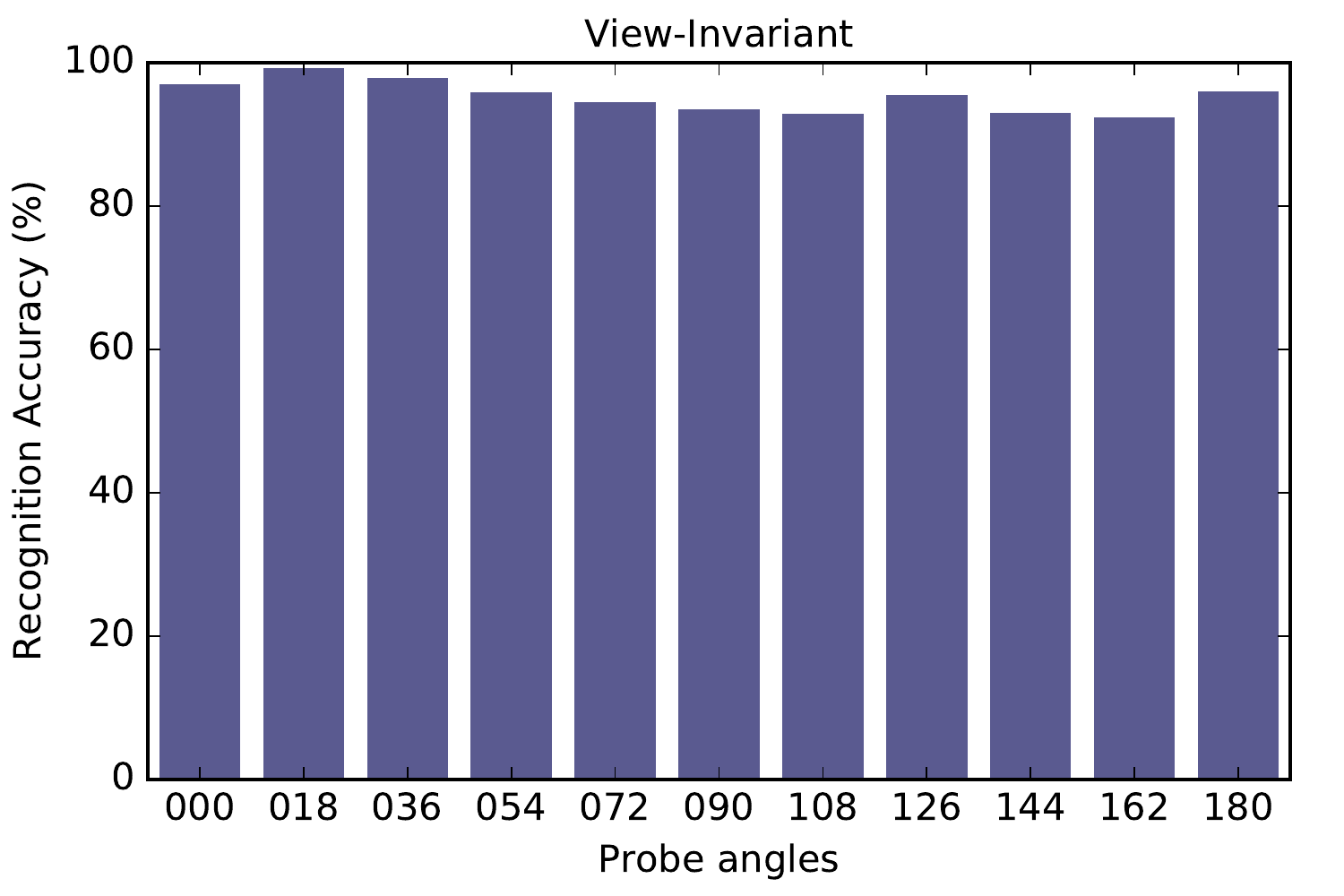}%
    \label{fig:gts-vim}}
  \caption{GTS performance on angles (Part-2)}
  \label{fig:all-angle-acc2}
\end{figure*}

The accuracy of each of the 11 classifiers is shown in Figure~\ref{fig:all-angle-acc1} and Figure~\ref{fig:all-angle-acc2}. Each graph depicts the CCR of the given classifier learned with labelled samples from a specific gallery angle and tested on all probe angles. The results show that the system can, to a degree, handle the neighbouring views as well but with lesser accuracy. The neighbouring view angle performance works especially well for angles closer to the sagittal view. On a closer examination of the boundary separation graph in Figure~\ref{fig:view-sep}, we can notice that the only possible error the view estimator can make is to mistaken the current view with the neighbouring views. Hence, even if the neighbouring view model is selected to predict the instance, the system can mitigate the error to a certain level. This phenomenon becomes apparent in Figure~\ref{fig:all-angle-acc2}(f) when the error is tested along with the view estimator to select the appropriate models for the test probe instances.

\begin{figure}
  \centering
  \includegraphics[width=\linewidth]{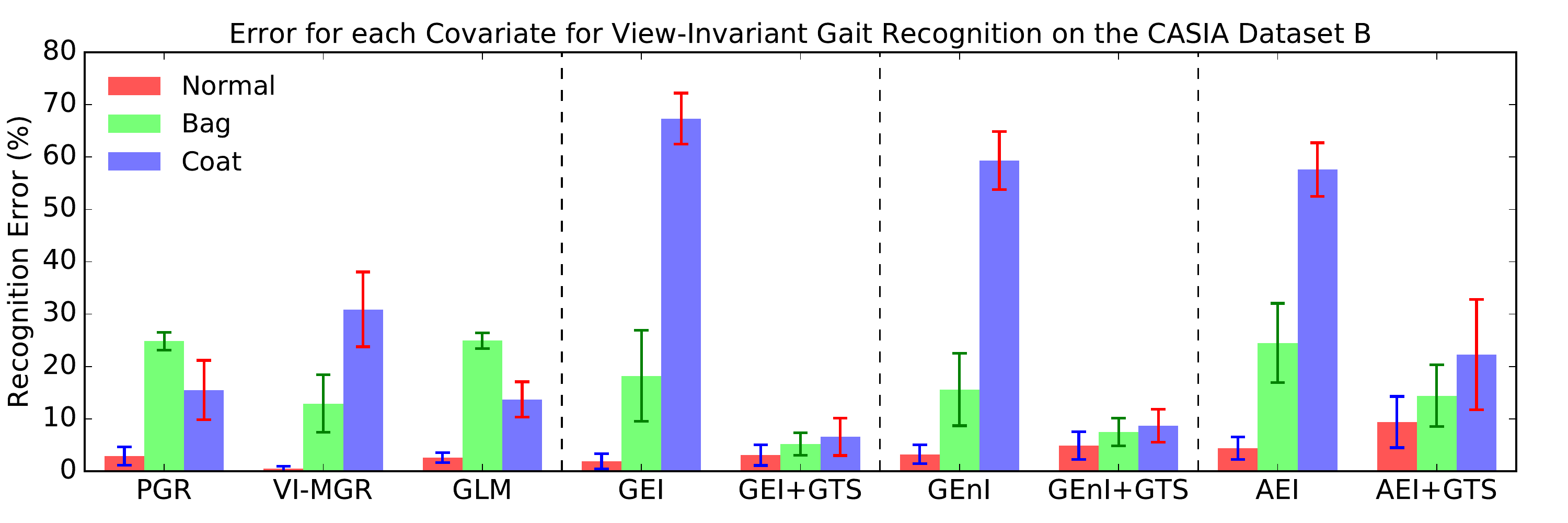}
  \caption{Recognition error for each covariate with different algorithms and
    various templates without prior knowledge of view angle}
  \label{fig:covarerror}
\end{figure}

Table~\ref{tab:ccr-angle} reports the CCR of the state-of-the-art view-invariant gait recognition methods along with the best performing template with the GTS, the GEI. All of the scores in this table have been claimed to be obtained without the prior knowledge of the actual view angle. The overall performance of the methods including the base templates taking into account all angles is provided in Table~\ref{tab:ccr-viewinvar}. Note that entries with bold font represents the best entry of that column, i.e., the maximum CCR,  mean CCR and the lowest standard deviation. The stability of the prediction capability across the intra-class variations is inversely related to the standard deviation of the respective accuracies.
Thus, a lower standard deviation infers a more stable system across covariates. Figure~\ref{fig:covarerror} compares the error associated for each covariate for different methods. Error lines depict the standard deviation of the error taken over the 11 views. It is evident that the GTS has improved the covariate performance of all of the base gait templates.

\begin{table}
  \centering
  \caption{CCR(\%) Without Prior Knowledge of View Angle}
  \label{tab:ccr-angle}
  \resizebox{\linewidth}{!}{%
  \renewcommand{\arraystretch}{1.2}
  \begin{tabular}{p{0.08\linewidth}*{12}{>{\raggedleft\arraybackslash}p{0.06\linewidth}}}
    \toprule
    Angle & $0\degree$ & $18\degree$ & $36\degree$ & $54\degree$ & $72\degree$ 
    & $90\degree$ & $108\degree$ & $126\degree$ & $144\degree$ & $162\degree$ & $180\degree$\\ 
 
    \cmidrule{2-12} 
    \rule{0em}{1em} &\multicolumn{11}{l}{(a) {PGR on GEI \citep{dupuis2013feature}}}\\
    \mbox{Normal} & 97.17 & 99.60 & 97.15 & 96.33 & 98.76 & 98.43 
                  & 97.14 & 97.57 & 97.14 & 92.97 & 96.00\\
    \mbox{Bag} & 73.15 & 74.07 & 74.70 & 76.33 & 78.49 & 75.81 
                  & 76.29 & 76.71 & 73.41 & 73.19 & 74.56\\
    \mbox{Coat} & 81.64 & 87.39 & 86.29 & 84.34 & 89.96 & 91.86 
                  & 89.50 & 85.04 & 72.24 & 78.40 & 82.70\\
    Mean & 83.99 & 87.02 & 86.05 & 85.67 & 89.07 & 88.70 
                  & 87.64 & 86.44 & 80.93 & 81.52 & 84.42\\
    \cmidrule{2-12} 
    \rule{0em}{1em} &\multicolumn{11}{l}{(b) {VI-MGR on GEI \citep{choudhury2015robust}}}\\
    \mbox{Normal} & 100.0 & 99.00 & 100.0 & 99.00 & 100.0 & 100.0 
                  & 99.00 & 99.00 & 100.0 & 100.0 & 99.00\\
    \mbox{Bag} & 93.00 & 89.00 & 89.00 & 90.00 & 77.00 & 80.00 
                  & 82.00 & 84.00 & 92.00 & 93.00 & 89.00\\
    \mbox{Coat} & 67.00 & 56.00 & 80.00 & 71.00 & 75.00 & 77.00 
                  & 75.00 & 65.00 & 64.00 & 64.00 & 66.00\\
    Mean & 86.67 & 81.33 & 89.67 & 86.67 & 84.00 & 85.67 
                  & 85.33 & 82.67 & 85.33 & 85.67 & 84.67\\
    \cmidrule{2-12} 
    \rule{0em}{1em} &\multicolumn{11}{l}{(c) {GLM
                      on GEI \citep{rida2016human}}}\\
    \rule{0em}{1.2em}%
    \mbox{Normal} & 97.97 & 98.79 & 96.37 & 96.77 & 98.39 & 97.98 
                  & 97.18 & 95.56 & 96.77 & 97.98 & 97.58\\
    \mbox{Bag} & 72.76 & 72.58 & 75.81 & 76.42 & 75.81 & 73.66 
                  & 74.60 & 76.92 & 76.11 & 75.10 & 76.11\\
    \mbox{Coat} & 80.49 & 83.47 & 85.08 & 87.85 & 91.53 & 91.07 
                  & 87.90 & 86.23 & 87.45 & 84.90 & 83.06\\
    Mean & 83.74 & 84.95 & 85.75 & 87.01 & 88.58 & 87.57 
                  & 86.56 & 86.24 & 86.78 & 85.99 & 85.58\\
    \cmidrule{2-12} 
    \rule{0em}{1em} &\multicolumn{11}{l}{(d) {Proposed 
                      GTS on GEI}}\\
    \rule{0em}{1.2em}%
    \mbox{Normal} & 98.50 & 98.98 & 99.00 & 97.00 & 97.50 & 96.00 
                  & 95.00 & 97.50 & 94.00 & 93.85 & 98.99\\
    \mbox{Bag} & 95.00 & 98.47 & 96.50 & 96.00 & 97.50 & 93.50 
                  & 93.50 & 94.00 & 92.50 & 91.33 & 94.44\\
    \mbox{Coat} & 97.00 & 99.49 & 97.50 & 94.00 & 88.00 & 90.50 
                  & 89.50 & 94.50 & 92.00 & 91.28 & 93.94\\
    Mean & \textbf{96.83} & \textbf{98.98} & \textbf{97.67} & \textbf{95.67} 
                                                                 & \textbf{94.33} & \textbf{93.33} & \textbf{92.67} & \textbf{95.33} 
                                                & \textbf{92.83} & \textbf{92.15} & \textbf{95.79}\\
    \bottomrule
  \end{tabular}%
  }
\end{table}

\begin{table}[t]
  \centering
  \caption{View-Invariant CCR Comparison}
  \label{tab:ccr-viewinvar}
\resizebox{\columnwidth}{!}{%
  \renewcommand{\arraystretch}{1.2}
  \begin{tabular}{l *{5}{c}}
    \toprule
    Method & Normal & Bag & Coat & Mean & Std \\
    \cmidrule(lr){1-1}
    \cmidrule(lr){2-4}
    \cmidrule(lr){5-6}
    PGR  \citep{dupuis2013feature} & 97.11 & 75.16 & 84.49 & 85.59 & 11.02\\
    VI-MGR \citep{choudhury2015robust} & \textbf{99.55} 
                    & 87.09 & 69.09 & 85.24 & 15.31\\
    GLM \citep{rida2016human} & 97.39 & 75.08 & 86.28 & 86.25 & 11.16\\
    Whole GEI & 98.12 & 81.77 & 32.66 & 70.85 & 34.07\\
    GEI with GTS & 96.94 & \textbf{94.9} & \textbf{93.3}
                                 & \textbf{95.05} & \textbf{1.77}\\
    Whole GEnI & 96.76 & 84.41 & 40.64 & 73.94 & 29.49\\
    GEnI with GTS & 95.11 & 92.52 & 91.32 & 92.98 & 1.94\\
    Whole AEI & 95.62 & 75.51 & 42.42 & 71.18 & 26.86\\
    AEI with GTS & 90.61 & 85.58 & 77.71 & 84.63 & 6.50\\
    \bottomrule
  \end{tabular}%
}
\end{table}

The accuracy of the view estimator plays a vital role in view-invariant recognition. The proposed view estimator is $97.77\%\pm 1.57$ accurate in finding the correct angle of the given gait sequences in contrast to the $94.43\%\pm 1.39$ proposed in \cite{dupuis2013feature}. Besides, the view-dependent classifiers are also capable of producing an applicable accuracy to neighbouring views minimizing the error of the overall recognition.

The VI-MGR shows the highest normal condition CCR, but with a substantially lower CCR for the clothing condition. The PGR and GLM perform equally well with a slight trade-off in carrying condition. The GTS with the GEI shows the best CCR in both carrying and clothing condition with minimal trade-off in normal condition resulting in a far superior overall performance.

The entire operation was also implemented with $k$NN in place of Bayes’ rule for comparison. On an average of all 11 views and 3 covariates, GTS-GEI with $k$NN (k=1) yielded an accuracy of 94.54\% which is marginally lesser than Bayes’ rule with 95.50\%.

The average time taken to run the genetic segmentation on the tuning set was less than 5 minutes on a fourth generation Intel-i7 processor with a 13.5 GB RAM. Though GA is an evolutionary procedure, the genetic segmentation operation was applied only once to study the covariate-independent regions of the gait templates. The results of this experimentation has established that only the head and feet portions of the gait templates are required for an optimal biometric performance. Hence, even when a different database is considered, the masks that are computed from CASIA-B can be applied to it provided the corresponding view and template size are the same. Should the masks be reconfigured for a different angle or resolution, the sequential optimization of $S_H$ and $S_F$ can be applied as described in Section~\ref{sec:sagitt-view-perf} without the need to execute the GA. The average time taken to run the sequential optimization operation on the tuning set using the same machine was less than 30 seconds.

  

\section{Summary}
\label{sec:gts-summary}

Template-based gait recognition methods are found to be more successful if the covariate-independent features are extracted. \cite{choudhury2015robust} selected the foot-portion based on predefined knowledge of the human-body; \cite{dupuis2013feature} selected the masking region through features selected by Random Forests classifier while \cite{rida2016human} employed GLM to do the same.

The proposed method used the genetic algorithm to automate robust region segmentation and hence termed to be genetic template segmentation. The GTS was applied on three popular gait templates, namely GEI, GEnI, and AEI. The GEI performs better than the other template representation when treated with the GTS.

The traditional gait recognition methods employ $k$NN for classification. The findings of this chapter shows that Bayes' rule can perform equal to or even better than the $k$NN classifier given that the features are preprocessed by CDA, i.e., PCA followed by LDA (MDA).

The GTS forms a separate mask for each of the possible views in the gallery. A view estimator is designed to determine the angle of reception to know which mask to use. The overall results clearly depict that the proposed GTS method outperforms the existing methods in literature.


\chapter{GAIT AUTHENTICATION USING MULTIPERSON SIGNATURE MAPPING} 
\label{ch:msm} 

\section{Introduction}
\label{sec:msm-intro}

The recognition task that is set up for most of the currently implemented systems involves the mapping of a given gait sequence to an identity corresponding to the closest match to the gallery set \citep{liang2016gait, Chattopadhyay20159, MarinJimenez2015}. This process cannot be applied directly in the case of authentication as unauthorized subjects may or may not be part of the gallery set, i.e., it can be a complete outsider unknown to the system. One is said to be authorized if the system identifies the subject as the identity the subject claims to be.

An authentication (or verification) system that relies purely on gait features and Euclidean threshold alone may not be the most secure system; an impersonator can be mapped on to any one of the authorized persons provided the gallery set is sufficiently large. That is, the level of security associated with such a system varies inversely with the number of authenticated users. To address this shortcoming, supplementary features are incorporated to improve the robustness of gait authentication systems. \cite{ntantogian2015gaithashing} express the effectiveness of combining both gait properties of the user and a physical authentication token possessed by the user for authentication.

Recent approaches on video-analytic gait authentication show excellent results, but their implementations are threshold-based which trade off a set amount of FAR (false acceptance rate) to produce an acceptable FRR (false rejection rate). The proposed Multiperson Signature Mapping (MSM) approach overcomes this drawback with a design that makes the FAR independent from the FRR. The state-of-the-art algorithms mostly prefer the $k$NN classifier where the Euclidean distance calculated from the extracted feature hyperplane is taken as the closeness measure. This chapter proves that the Bayes' rule applied over the extracted feature set provides a much better performance compared to the conventional $k$NN approach. The MSM is applied on top of template-based gait recognition algorithms to produce an efficient gait authentication system. The method is evaluated on four different gait templates including the GTS described in Chapter~\ref{ch:gait-recog}. The study analyses the performance across different clothing and carrying conditions over multiple views. Experimental results with the CASIA-B gait database depict the potential of the proposed approach.

\section{State of the Art}
\label{sec:msm-art}

Authentication forms a minority in gait biometric literature as most articles focus on recognition. Occasionally, gait authentication may be used synonymously with gait recognition \citep{Arora2015} and vice versa \citep{Nikolaos_VB}. \cite{sarkar2005humanid} was one of the earliest to suggest a method for gait authentication as well as recognition called the `baseline algorithm'. The distance between gait instances was measured through frame-wise Tanimoto similarity of corresponding silhouette sequences. Though the method was primitive in nature, it was aimed at setting a benchmark through which other algorithms can be compared along with an introduction of the ``Gait Challenge'' database. \cite{Nikolaos_VB} proposed an angular transform which is applied on time-normalized silhouette sequences. When used for authentication, their method outperformed the baseline algorithm.

The state-of-the-art framework of gait authentication involves feature extraction in the form of gait templates like the GEI \citep{man2006individual}, AEI~\citep{zhang2010active} and GEnI \citep{bashir2010gait} and finding the Euclidean distance between the test instance and the closest gallery instance of the claimed identity. If this distance falls within the set threshold, the test instance is considered authenticated otherwise rejected as an impostor.

In addition to the gait itself, the shape of the head can also provide useful information when used in conjunction with the spatiotemporal analysis of the leg motion. This notion was employed by \cite{jia2015view} to boost the verification rate. However, their study did not include the effect of covariate conditions on the system performance.

\section{Method}
\label{sec:}

\begin{figure*}
  \centering
  \includegraphics[width=0.6\linewidth]{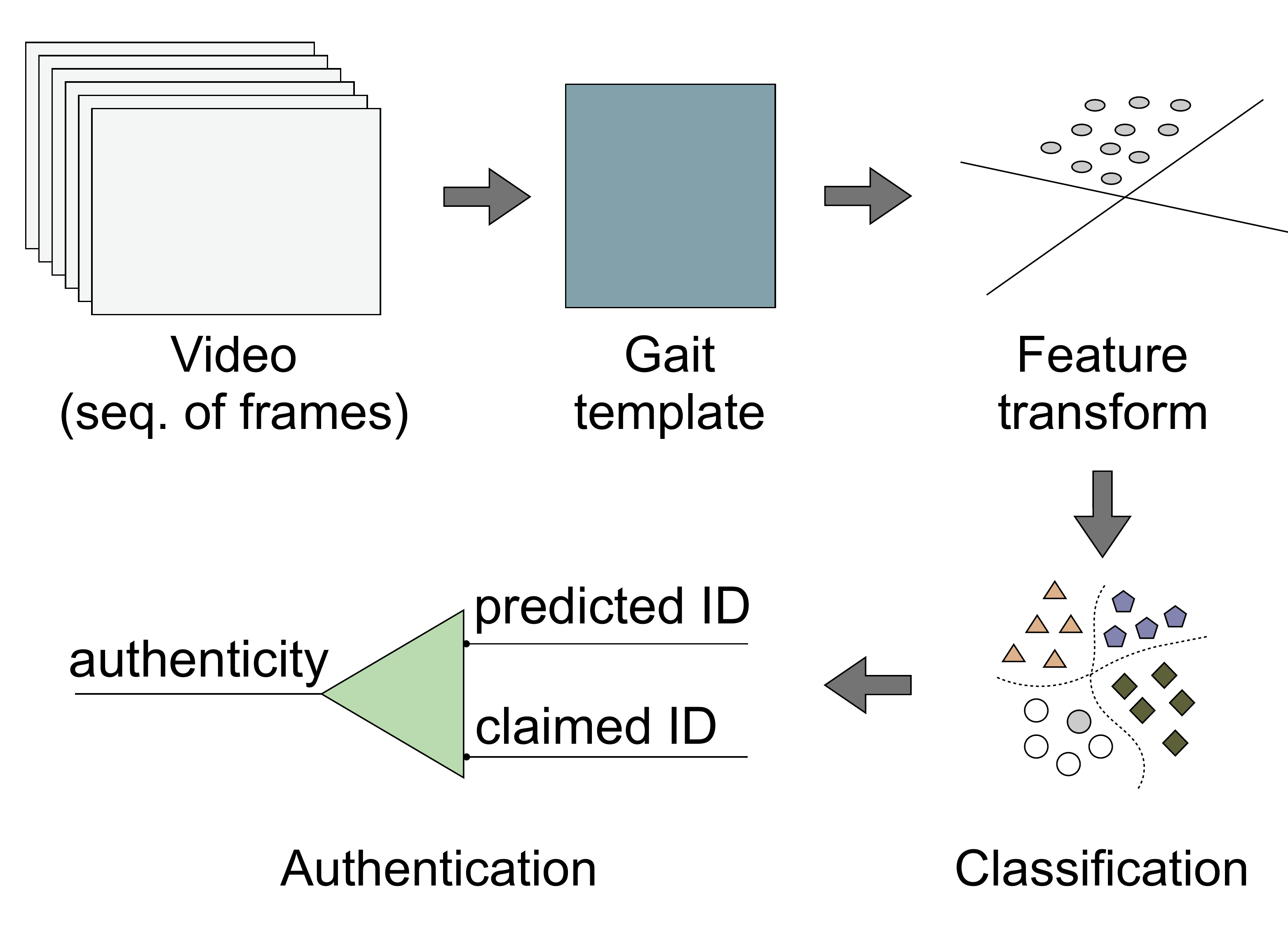}
  \caption{The basic flow of the proposed MSM framework. 
  }
  \label{fig:msm-flow}
\end{figure*}

The general outline of the proposed framework is as illustrated in
Figure~\ref{fig:msm-flow}. The input gait instance is given as a sequence of frames. The frames are cropped and are used to generate a gait template. The template features are transformed through CDA. The transformed set of features are classified with Bayes' rule. Finally, the authenticity is checked through predicted signature matching.

\subsection{Spatiotemporal Feature Extraction}
\label{sec:spatio}

Features are extracted in the form of gait templates. The elementary part of which are silhouettes extracted from the walking human subject. The templates used for this chapter includes GEI, GEnI, and AEI which are obtained as in Section~\ref{sec:gait-templ-extr}. In addition to this, the GTS as discussed in Section~\ref{sec:gts} is also used. The study concluded that the optimum accuracy is attained when the GEI is used as the base template. Therefore, throughout the rest of this chapter, the GTS refers to the GTS masking applied to the GEI gait templates.
\[G_\text{GTS}(x,y) = G_\text{GEI}(x,y) \times M_\text{GTS}(x,y)\] 
The gait recognition model is obtained when the spatiotemporal features are reduced through CDA and trained using the multivariate Gaussian Bayes classifier so that given a gait template, the model can predict the identity of the subject.



\subsection{Multiperson Signature Mapping}
\label{sec:msm}

A typical gait authentication method would transform the given instance to a feature space based on some gait template configuration, $T$. The transformation may include feature reduction techniques like PCA, LDA or a combination of both. When a subject is to be authenticated, the test instance features are transformed and compared against that of the claimed identity in the trained model through the Euclidean distance between them. If this distance is within the tolerable threshold, then the subject belonging to the instance is considered authenticated. Algorithm~\ref{alg:trad-auth} shows the general outline of such a scheme.

\begin{algorithm}[t]
  \begin{spacing}{1.3}
  \begin{algorithmic}
    \caption{Typical threshold-based gait authentication}\label{alg:trad-auth}
    \Procedure{AuthenticateGaitThresh}{$\mathit{instance}, \mathit{claimedID}$}
    \State $X_p \gets \text{transform}(\mathit{instance}, T)$
    \State $X_g \gets \text{getGalleryFeature}(\mathit{claimedID})$
    \State $\mathit{dist} \gets || X_p - X_g ||$
    \State $\mathit{authenticity} \gets (\mathit{dist} < \mathit{threshDist})$
    \State \textbf{return} authenticity
    \EndProcedure
  \end{algorithmic}
\end{spacing}
\end{algorithm}

\begin{algorithm}[t]
  \begin{spacing}{1.3}
    \begin{algorithmic}
    \caption{MSM framework for gait authentication}\label{alg:msm-auth}
    \Procedure{RecognizeGait}{$\textit{instance}, \textit{T\_model}$}
    \State $T \gets \mathit{T\_model}.\text{getConfig}()$
    \State $X_p \gets \text{transform}(\textit{instance}, T)$
    \State $\textit{predictedID} \gets \textit{T\_model}\text{.predict}(X_p)$
    \State \textbf{return} \textit{predictedID}
    \EndProcedure
    \State
    \Procedure{AuthenticateGaitMSM}{$\mathit{instance}, \mathit{claimedID}$}
    \State $\mathit{predictedID} \gets \text{RecognizeGait}(\mathit{instance}, \mathit{T\_model})$
    \State $\mathit{authenticity} \gets (\mathit{predictedID} == \mathit{claimedID})$
    \State \textbf{return} \textit{authenticity}
    \EndProcedure
  \end{algorithmic}
\end{spacing}

\end{algorithm}

A gait recognition model classifies any given instance, regardless of whether registered or unregistered, to one among the trained identities. The MSM gait authentication framework converts the gait recognition model to one that can be used for gait authentication. Two inputs are passed through the system: the test gait sequence and the claiming identity reference. The classifier would label the test sequence to one of its identities based on the input gait features alone and would consider the instance `authenticated' only if the output label matches the claimed identity reference as illustrated in Algorithm~\ref{alg:msm-auth}. Although it seems simple, this technique is highly effective in practice.

The object \textit{T\_model} encapsulates the classifier trained for the template $T$ as well as the necessary configurations to extract template $T$ and corresponding PCA and LDA transformation parameters for dimensionality reduction. The primary strength of this technique is derived from the correct classification rate of the recognition module and population of registered authenticated users in the system.

A theoretical derivation of the FRR and FAR of the system justifies the applicability of the proposed framework over the existing one. Let $k$ be a random number that represents the identity trained in the system. Let $C$ and $I$ be discrete random variables that denote the identity predicted by the classifier and the actual identity of the subject respectively. The true acceptance rate or precision $p$ of the authentication system can be represented by Bayes' theorem as
\begin{equation}
  \label{eq:1}
  p = \Pr(C\peq k \mid I\peq k) = \frac{\Pr(I\peq k \mid C\peq k) \cdot
    \Pr(C\peq k)}{\Pr(I\peq k)}
\end{equation}

The data provided for training for a biometric system does not contain any skew, i.e., the training instances are provided per subject are equally distributed during both training and testing phase. This condition would make the probabilities
\begin{align}
  \label{eq:p}
  \Pr(C\peq k) & = \Pr(I\peq k) \nonumber\\
  \implies  p & = \Pr(I\peq k \mid C\peq k) 
\end{align}

By definition, $\Pr(I\peq k \mid C\peq k)$ is the CCR of the recognition system. The FRR of the authentication system can hence be derived as follows.
\begin{align}
  \label{eq:frr}
  \text{FRR} & = 1 - p \nonumber\\
             & = 1 - \Pr(I\peq k \mid C\peq k) \nonumber\\
             & = 1 - \text{CCR}
\end{align}

Assume that any unauthorized identity is equally likely to be accepted as one of $n$ authorized identities trained in the system (false acceptance). This assumption would lead to the FRR being represented as
\begin{equation}
\label{eq:far} \text{FAR} = \frac{1}{n}
\end{equation}
The resulting average error rate of the authentication system can hence be calculated as
\begin{align} \text{AER} & = \frac{\text{FRR}+\text{FAR}}{2} \label{eq:aer1} \\ & = \frac{1 - \text{CCR} + \frac{1}{n}}{2} \label{eq:aer2}
\end{align}

Equation~\ref{eq:aer2} accounts for the theoretical error of the authentication system. This will be experimentally verified in the next section with the empirical error through Equation~\ref{eq:aer1} the next section.
The strength of this technique lies in the FAR (Equation~\ref{eq:far}). This would
mean that the FAR is inversely proportional to the number of members registered
in the system. 

\subsection{Two-pass Variation of MSM for GTS}
\label{sec:two-pass-variation}

The fitness function in the GTS template is designed to have a minor trade-off in normal walk accuracy to minimize errors associated with the covariate factors. The fitness function was initially given by
\[ F(h) = \Big(\frac{1}{2}\text{CCR}_\text{A}(h) + \frac{1}{6}\text{CCR}_\text{B}(h) + \frac{1}{3}\text{CCR}_\text{C}(h)\Big)^2\] where CCR$_i$ is the accuracy for covariate condition $i$; A being normal walk, B for carrying condition and C for clothing variation.  However, the GEI provides the best normal walk CCR compared to all known templates but with the worst covariate CCR. To bring the best of both templates, a two-pass variant of the MSM framework is proposed in this study.


The two-pass MSM would require predictions from two templates. If the claimed identity of the subject matches any one of the two predictions, then the subject is considered authenticated. This technique will bring the FRR of the system for each covariate factor down to the least of that pertaining to the given templates. This implementation is essentially an ensemble in which one template classifier would cover the probable misclassifications of the other.  To maximize the effectiveness of this method, the GTS masks are recomputed to maximize the covariate recognition accuracy without the special regard for the normal walk accuracy. The new fitness function weighs each covariate equally as given by the following equation.
\[ F(h) = \Big(\text{CCR}_\text{A}(h) 
  + \text{CCR}_\text{B}(h) 
  + \text{CCR}_\text{C}(h)\Big)^2\]

\section{Results \& Discussion}
\label{sec:msm-results}

\subsection{Dataset Configuration}
\label{sec:msm-dataset}

The dataset is formulated from the CASIA-B gait database \cite{yu2006framework} to evaluate the robustness against covariate factors. It is composed of gait sequences from 124 subjects. This experimentation makes use of all its covariates: normal walk (SetA), walk while carrying a bag (SetB), and walk while wearing an overcoat (SetC). The set of the first four instances of normal walk, SetA1, is used for training and the set of the other two instances, SetA2, is used for testing. All 11 views in the database are considered. Thus a total of $124 \times (6+2+2) \times 11 = 13640$ gait sequences are used for experimentation.

For the setup for authentication, both authorized and unauthorized members are required. Hence, 100 randomly selected subjects are assumed as authorized members forming the member set D$_1$. The remaining 24 subjects form the member set D$_2$ and are considered unauthorized. In the authentication testing session, members of D$_1$ use their own id number as their claiming identity and each member of D$_2$ randomly assume an id of D$_1$ as their claiming identity. In practice, to keep the process unobtrusive, the subject can provide the claimed identity for authentication in the form of an NFC (near-field communication, e.g., as part of the employee's ID card). The assignment of false identities is done subject to the constraint that each member of D$_2$ shall assume at least one but no more than two IDs of D$_1$ so that there is a total of 100 forged attempts. This mapping ensures that there is an equal number of positives (genuine members) and negatives (impostors) for each testing session and each covariate.


Instances of SetA1 $\cap$ D$_1$ alone is used for the training phase. During the test for authentication, all instances of SetA2, SetB and SetC are used which includes both D$_1$ and D$_2$. This leads to a total of 400 test cases (200 positives and 200 negatives) for each covariate set for a given view angle.

\subsection{Threshold Method vs. MSM}
\label{sec:thresh-vs-msm}

The conventional gait authentication system uses a nearest neighbour model that contains the transformed features of all members of the training set that pertains to the registered subjects. The transform that is used here is also CDA, i.e., PCA followed by LDA. Applying the same feature transformation for both threshold-based and MSM frameworks facilitate in an effective comparison between them.

\begin{figure}
  \centering
  \includegraphics[width=0.85\linewidth]{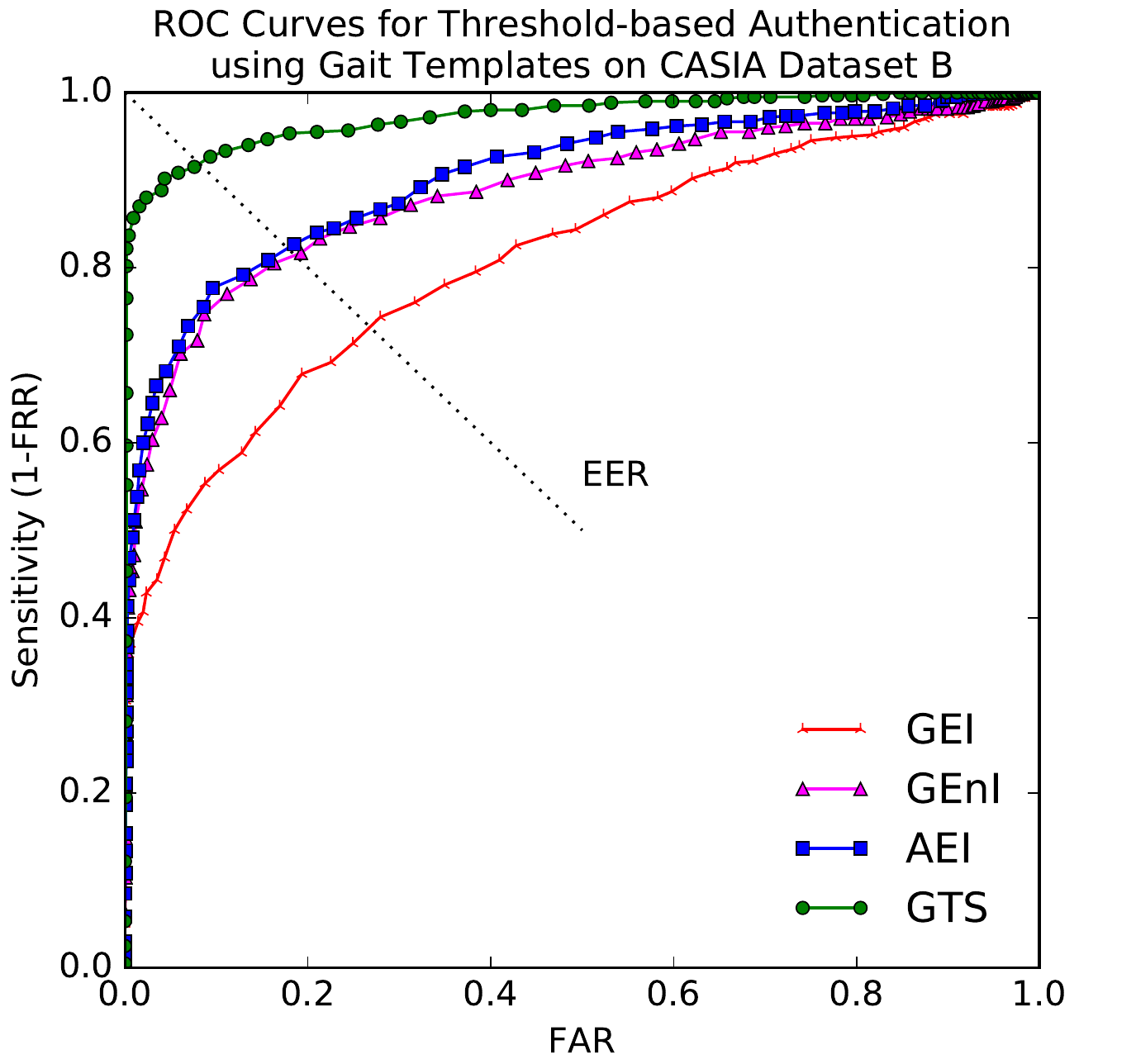}
  \caption{ROC curves of threshold-based verification using different gait templates on the
    CASIA-B dataset at the sagittal angle}
  \label{fig:roc-all}
\end{figure}

The traditional method involves fixing a threshold which controls the trade-off between the FAR and FRR. It is evaluated using ROC curves where the verification rate (sensitivity; $1-\text{FRR}$) is plotted against the FAR. The performance of the gait templates over the threshold-based framework is shown in Figure~\ref{fig:roc-all} with the error rates as averages across all three covariates: SetA2, SetB and SetC. As all templates discussed in this article show superior normal walk performance, the average accuracy over all covariates is considered to plot the ROC curves so that the difference in the overall performance can be illustrated.

The usual convention is to compare the point where the FAR and FRR becomes the same and is known as the EER: equal error rate. The MSM framework does not include a variable threshold to obtain an EER. Hence to compare the performance of MSM and threshold-based methods, the optimum AER is calculated. This value corresponds to the combination of FRR and FAR which yields the minimum possible AER for the specific threshold-based authentication scheme. The optimal AER would not exceed the EER of the same system.

\begin{figure}
  \centering
  \includegraphics[width=0.87\linewidth]{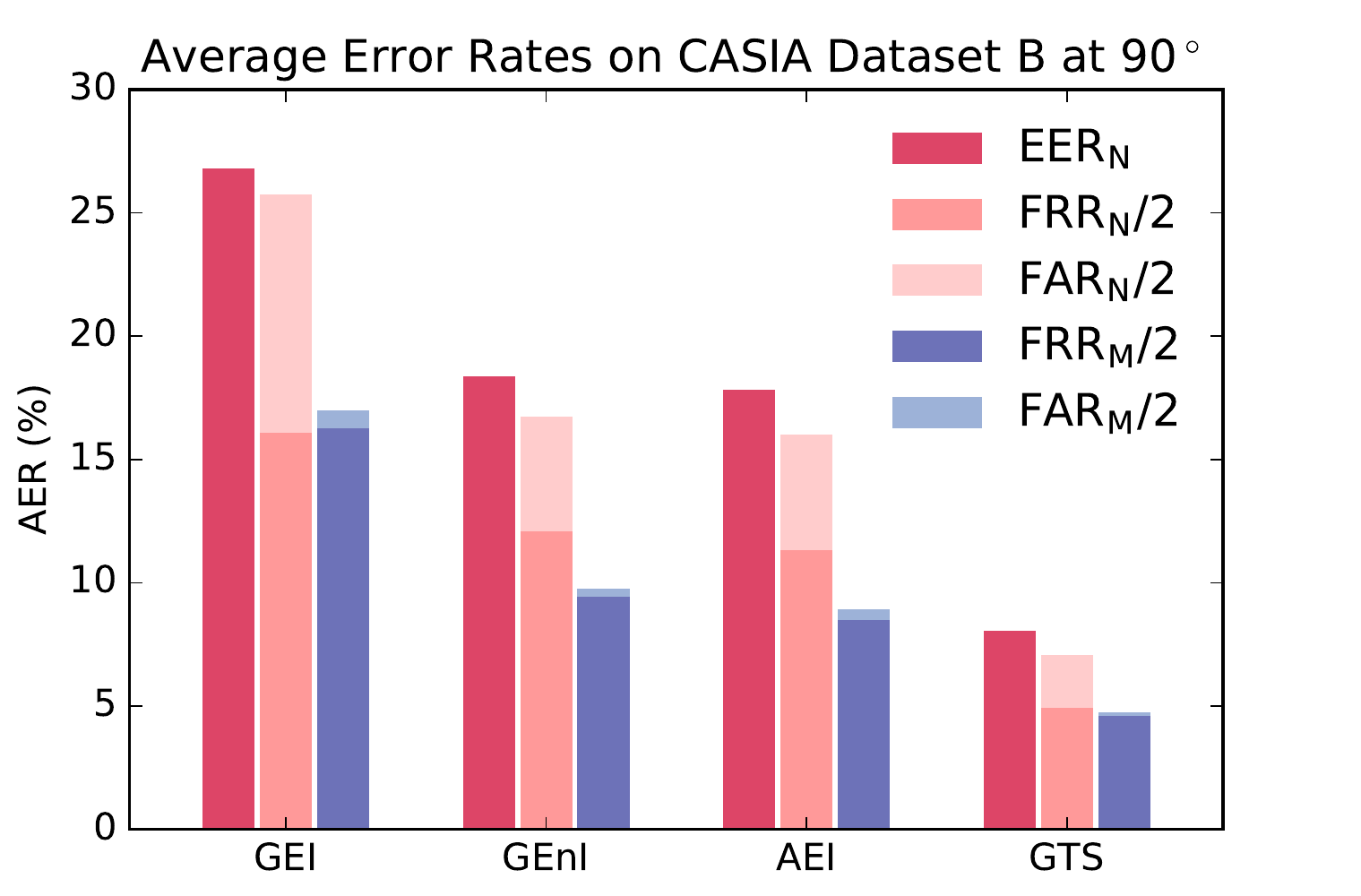}
  \caption{Comparison of threshold-based and MSM frameworks}
  \label{fig:aer-msm}
\end{figure}

In the MSM framework, for a given gait template, a PCA-LDA-Bayes classifier is with the features of the registered members. The classification was done with the help of Scikit-Learn package in Python \citep{scikit-learn}. The performances of both threshold-based and MSM for the four mentioned templates are as shown in Table~\ref{tab:thres-vs-msm} and graphically compared in Figure~\ref{fig:aer-msm}. From these results, it can be observed that the classification through Bayes' rule has minimized the FRR. The verification rate ($1-\text{FRR}$) is found to be equal to the CCR of the classifier as depicted in Equation~\ref{eq:frr}. Also, the FAR is plummeted to an average of 1\% across all templates proving Equation~\ref{eq:far}. That is, the number of authentic members registered (trained) in the system is $n=100$ leading to an FAR of $1/100=0.01$. This implies that AER as given in Equation~\ref{eq:aer2} holds true for the MSM framework.

\begin{table}[t]
  \caption{Performance of the threshold-based and MSM frameworks on the CASIA-B
    dataset}
  \label{tab:thres-vs-msm}
  
  \begin{center}
    \renewcommand{\arraystretch}{1.1}
    \begin{tabular}{l *{4}{r} c *{3}{r}}
      \toprule
      \multirow{2}{*}{Template } & \multicolumn{4}{c}{Threshold-based}
      && \multicolumn{3}{c}{MSM-based} \\
      \cmidrule{2-5} \cmidrule{7-9}
                                 & EER & FRR* & FAR* & AER && FRR & FAR & AER \\
      \midrule
      GEI & 26.79 & \textbf{32.17} & 19.33 & 25.75 && 32.50 & \textbf{ 0.75} & \textbf{16.62}\\
      GEnI & 18.38 & 24.17 &  9.33 & 16.75 && \textbf{18.83} & \textbf{ 0.33} & \textbf{ 9.58}\\
      AEI & 17.83 & 22.67 &  9.33 & 16.00 && \textbf{17.00} & \textbf{ 0.42} & \textbf{ 8.71}\\
      GTS &  8.04 &  9.83 &  4.33 &  7.08 && \textbf{ 9.17} & \textbf{ 0.17} & \textbf{ 4.67}\\
      \bottomrule
    \end{tabular}
  \end{center}

 \vspace{1em} 
 \small
 
 * The FAR and FRR of the threshold-based method are obtained by tuning the
  threshold so as to minimize its AER. All FAR mentioned here is the average of
 both type 1 and type 2 FAR.
\end{table}

\subsection{Types of FAR}
\label{sec:types-far}

There are fundamentally two types of impostors who try to gain access to the system. Type 1 impostors are those who are unknown to the system and claim the identity of a member who is registered in the system. Type 2 impostors are those whose real identities are registered in the system but attempt to claim the identity of another member of the same system. The test set for type 2 impostors consists of the IDs of each member of SetA2 $\cap$ D$_1$ shuffled such that none would receive their original ID. This leads to an additional 200 negative samples for each covariate factor.

Type 2 errors are close to negligible in the MSM framework, provided the base classifier used for the recognition be sufficiently accurate. For a type 2 impostor to succeed, his/her identity true identity must be mispredicted by the base classifier, and the mispredicted value should match the false identity that he/she claims to be. In theory, if the misprediction rate of the classifier is given by $1~-~\text{CCR}$, and there are $n$ members registered in the system, the probable type two error is $(1~-~\text{CCR})/n$. This effect can be observed in Table~\ref{tab:far}. In a threshold-based system, a type 2 impostor is treated no different from a typical type 1 impostor as the same Euclidean threshold plays its role. However, the design of the MSM framework makes type 2 impostors close to non-existent.

\begin{table}[t]
  \caption{Comparison based on the type of FAR}
  \label{tab:far}
  \centering
  
  \begin{center}
  \renewcommand{\arraystretch}{1.2}
  \begin{tabular}{ll rr c rr}
      \toprule
      \multirow{2}{*}{Templates} && \multicolumn{2}{c}{Threshold-based} &
      &  \multicolumn{2}{c}{MSM-based} \\
      \cmidrule{3-4} \cmidrule{6-7}
                                 && Type 1 & Type 2 && Type 1 & Type 2 \\
      \midrule
      GEI && 15.17 & 23.50 &&  \textbf{1.00} &  \textbf{0.50} \\
      GEnI &&  6.67 & 12.00 &&  \textbf{0.67} &  \textbf{0.00} \\
      AEI &&  9.00 &  9.67 &&  \textbf{0.67} &  \textbf{0.17} \\
      GTS &&  6.00 &  2.67 &&  \textbf{0.33} &  \textbf{0.00} \\
      \bottomrule
    \end{tabular}
  \end{center}

  \small \vspace{1 em} The values are the mean of the respective errors over all three covariate factors.
\end{table}

\Needspace{5\baselineskip}
\subsection{Covariate Factors and GTS-2P}
\label{sec:covar-fact}

\begin{figure}
  \centering
  \includegraphics[width=0.88\linewidth]{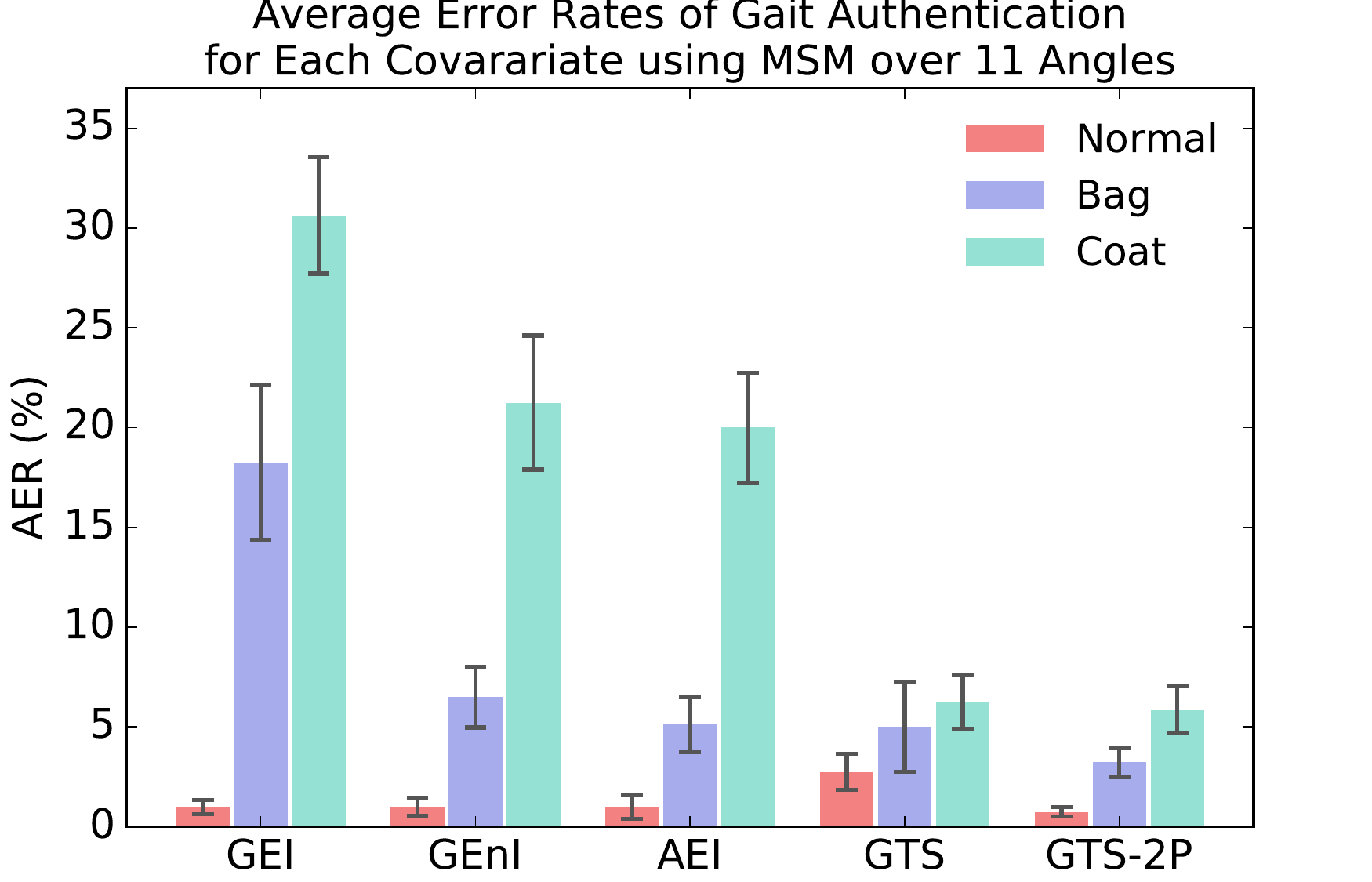}
  \caption{AER of each covarate on the CASIA-B dataset}
  \label{fig:aer-covar}
\end{figure}

Covariate factors play a vital role in both recognition and authentication. Figure~\ref{fig:aer-covar} depicts the AER for each template observed through the MSM framework. Clothing variations are known to decrease the performance of gait biometric system substantially \citep{MarinJimenez2015}. This phenomenon can be noted in all three of the base templates, GEI, GEnI and AEI. The GTS was designed to circumvent this problem but with a little trade-off in normal condition error. This setback of the GTS is nullified when coupled with the GEI in the two-pass MSM variation, GTS-2P.

In GTS-2P, two classifiers are used, one trained with the GEI templates and one with the GTS templates. When one or both of these templates predict an ID that matches the claimed ID of the subject, then the subject is assumed accepted. As stated in Section~\ref{sec:msm}, the true acceptance rate or verification rate is treated as the precision of the classifier considering `accept' to be a positive label. When the final prediction of the ensemble is given by the logical OR of the input binary classifiers, then the precision shall be no lesser than any of the classifiers in the ensemble. That is, the FRR of each covariate can either be the minimum of the two classifiers or even lesser than both. This effect is what can be observed in Figure~\ref{fig:aer-covar}; the normal AER is slightly better in GTS-2P ($0.75\pm 0.24\%$) than that of the GEI ($1.0 \pm 0.35\%$). Similarly, the GTS-2P coat AER ($5.87\pm 1.2\%$) is marginally better than that of the original GTS ($6.25 \pm 1.34\%$). However the improvement observed in the carrying condition (bag) is much more pronounced in GTS-2P ($3.25 \pm 0.71\%$) than the original GTS ($5 \pm 2.27\%$). This is because the GTS masks were recomputed to favor the all conditions equally for the two pass variation. The difference in the normal and clothing conditions seem negligible due to the pitfall with multiple passes which increases the FAR. For each pass, the impostor subject is given an extra chance to slip through the system. Thus, increasing the number of passes by two theoretically doubles the FAR of the system. This is also the reason why the number of passes was limited to just two in the study and we did not consider formulating an ensemble that merges the predictions of all possible templates. As the theoretical FAR of the one-pass MSM was only 1\%, the increase towards an FAR of 2\% does not weigh against the significant decrease in FRR, and in terms, the overall AER. That is, the AER of MSM on the original GTS is $4.67\pm 1.5\%$ while that of GTS-2P is $3.29\pm 0.72\%$.

\Needspace{5\baselineskip}
\subsection{Spoofing Attacks and Limitations}
\label{sec:final-remarks}

The strength of gait biometrics is often questioned. It is considered possible for people to imitate the body language of another and hence gait as well. This assumption drives the argument that gait authentication systems are vulnerable to spoofing attacks which lead to the following questions:
\begin{enumerate}[topsep=0pt, noitemsep, label=(\alph*)]
\item To which extent can one's gait be spoofed in a biometric system?
\item Can spoof attacks increase the theoretical FAR of the MSM? 
\end{enumerate}

To test a system for authentication, both genuine and impostor gaits are required. The impostors ought to ideally attempt to impersonate the gait of the identities they claim. The public datasets available do not conform to this requirement as of yet, so methods in literature adopt random assignment of genuine identities to random impostors of the same dataset. Hence, those who wish to study the effect of spoofing are required to compile their dataset with true impersonators who try to imitate the gait of genuine members they claim.

A natural gait is a walking style adopted by a person to balance the body and at the same time provide locomotion. Many factors come to play when one's natural gait is observed:
\begin{enumerate}[topsep=0pt, noitemsep, label=(\alph*)]
\item Physical proportions
\item Weight distribution
\item Metabolism
\end{enumerate}

\begin{figure}[t]
  \centering
  \includegraphics[width=0.9\linewidth]{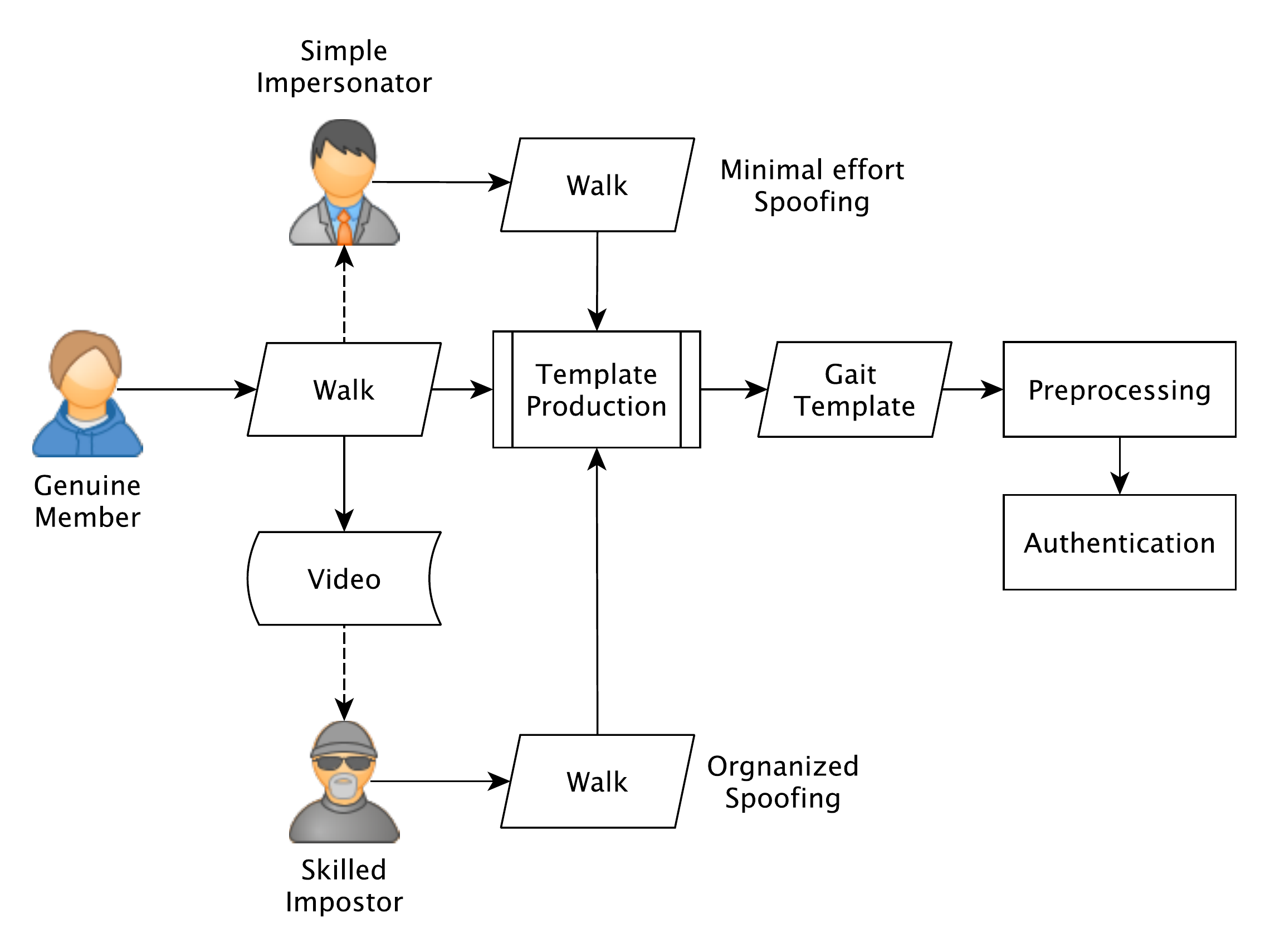}
  \caption{Simple spoofing scenario in gait authentication}
  \label{fig:spoofing}
\end{figure}

Apart from the above there are also covariate factors (Section~\ref{sec:covariate-factors}). To completely spoof one's gait, the attacker should conform to at least the physical proportions of the target concerned, and match the weight distribution to a degree. A simple scenario of spoofing is shown in Figure~\ref{fig:spoofing}.

\cite{gafurov2007spoof} were the first to do a comprehensive analysis on spoofing attack on gait authentication. Their study included 100 members and was based on wearable sensors. The results concluded that gait imitation with minimal effort shows negligible change in the EER. Despite that, those with an in-depth knowledge of their target's gait and with closer body proportions are found pose a significant threat to the authentication system. \cite{mjaaland2010walk} studied gait spoofing with 50 subjects and found that each individual had a \textit{plateau} which is the limit to which they can mimic the natural gait of another person. Their study concluded that spoofing one's gait is a highly non-trivial task.

However, the experiments of both \cite{gafurov2007spoof} and \cite{mjaaland2010walk} included only a single sensor attached to the hip (belt) of the subjects studied. The findings of \cite{geradts2002use} shown that the foot angle is a much greater indicator of gait than the hip angle. Unless the system draws data explicitly from a sensor attached to the feet, the foot angles cannot be inferred.

A video gait template-based system inherently records the foot angle of the subjects under observation. Nevertheless, as most appearance-based templates rely largely on the shape of the subjects, attackers with similar build and clothing conditions as the target can spoof the system \citep{hadid2013improving}.

\subsubsection*{Possibility of spoofing in a biometric system}

The extent to which one's gait can be spoofed lies in the design of the gait biometric system. If the algorithm is largely based on shape features, the video-based gait biometric system is vulnerable to spoofing attacks.  This weakness is only a concern if the attacker is of the same build and trained specifically to mimic the gait of the target subject. On the contrary, the system is more secure if it gives a greater weight to spatiotemporal features. Note that this notion does not hold for soft biometrics where shape features play a vital role in the objective function. \cite{mjaaland2010walk} observed that the more the subjects practice to mimic the gait of their target, the more that some tend to underperform in their attempt. The reduction in spoofing efficiency results from arrhythmic patterns adopted by the attacker due to excessive caution.

\subsubsection*{Effect of spoof attacks on the theoretical FAR of the MSM}

As far as the GTS features are concerned, the answer to the question, \textit{``Can spoof attacks increase the theoretical FAR of the MSM?''}, is no. This is because the GTS template covers only the head shape and inclination and the spatiotemporal features of the foot region. Thus the reliance on shape features are minimal and nullifies the effect of spoofing attacks. A person may be able to mimic the arm swing and stride length and body inclinations to a certain limit (as determined by the plateau), but one cannot accurately reproduce the foot angle of a natural gait of another human. Moreover, the CDA feature transformation maps the input instances to confined regions. For an attacker, this mapping is unpredictable. Even a small deviation in some of the features of the instance may cause wider deviations in the CDA feature space. This means that the attacker's gait instance can be mapped to any one of the regions formed by the $n$ trained identities with equal probability. Thus the theoretical FAR of the MSM algorithm still remains $1/n$.

\subsubsection*{Limitation of the MSM}

The proposed MSM is proven to be a much better framework for gait authentication than the conventional threshold-based authentication methods. FAR rising from type 2 imposters are close to zero and the type 1 FAR scales inversely to the system population ($n$). However, when the system consists of only a few members, its greatest strength becomes its weakness. This would mean that the impostor has a much greater chance to be mapped to the person whom he claims to be. In such situations where the number of authentic subjects is minimal, it is recommended that the members who are not part of the authority also have their gait signatures registered in the system, but they need not be granted the same access possessed by the authorized registered members. This will ensure that the system population is large enough to maintain an acceptable FAR.

\section{Summary}
\label{sec:msm-summary}

Gait recognition is a mapping of input instance to an identity whereas gait authentication is the problem of verifying whether the given instance belongs to the claimed identity. The state-of-the-art implementations of gait authentication use thresholds based on Euclidean nearest neighbour distance. A Euclidean threshold is empirically set depending on the trade-off between FAR and FRR.

The proposed method, MSM, eliminates the need of a threshold by implementing a random signature mapping technique for a system of multiple subjects. In MSM, a genuine gait instance is more likely to be mapped to the claimed identity while an impostor is more liable to be mapped on to a different identity. The MSM outperforms the Euclidean NN systems over the same template features and preprocessing steps provided the system population be sufficiently large.

A gait authentication system can become vulnerable to spoofing attacks if the algorithm imposes a higher dependency on shape-based features. However, when the GEI-GTS features are employed, the reliance on shape-based features is reduced and a greater weight is given to temporal deviations of the foot angle. Moreover, the foot angle is found to be the best discriminative parameter for a gait biometric system according to \cite{geradts2002use}.

As the MSM derives its strength from the system population, its efficiency will decrease if the number of registered members in the system is considerably low. Artificially increasing the system population by registering new members who do not claim access would theoretically increase the performance of MSM.





\chapter{GAIT AUTHENTICATION USING BAYESIAN THRESHOLDING}
\label{ch:mgb} 

\section{Introduction}
\label{sec:mgb-intro}

Although the MSM has proven its worth as a much more reliable authentication framework over the conventional methods, it suffers from a drawback when the system population is smaller. For instance, the results in previous chapter is based on a dataset with a system population of registered members as $n=100$, this caused the FAR to be $1/n = 0.01$, which is 1\%. If the system population is small as 10, then the FAR would be as bad as 10\%, which means that there is a chance of one in ten that the impostor can get mapped on to the region of the claimed identity.  This brings the need to an even better framework that is effective even when the number of registered members is low.

This chapter projects the extension to the MSM using probability as a threshold, namely, Bayesian posterior probability, but it does not rely on the Euclidean distance threshold like the existing implementations. That is, once the features are extracted, their posterior probability with respect to the claimed identity is calculated. If this probability is higher than the assigned threshold probability, the test instance is said to be authenticated. This threshold would provide the means to separate genuine members from impostors even if they get mapped on to the same region. This method is compared with the MSM with identical conditions to illustrate the difference in performance between them. Just like the MSM, the Bayesian thresholding method is also a paradigm that can be applied to any feature representation.

\section{Mapping of Instances to Regions}
\label{sec:mgb}

\begin{figure}[b!]
  \centering
  \includegraphics[width=0.85\linewidth]{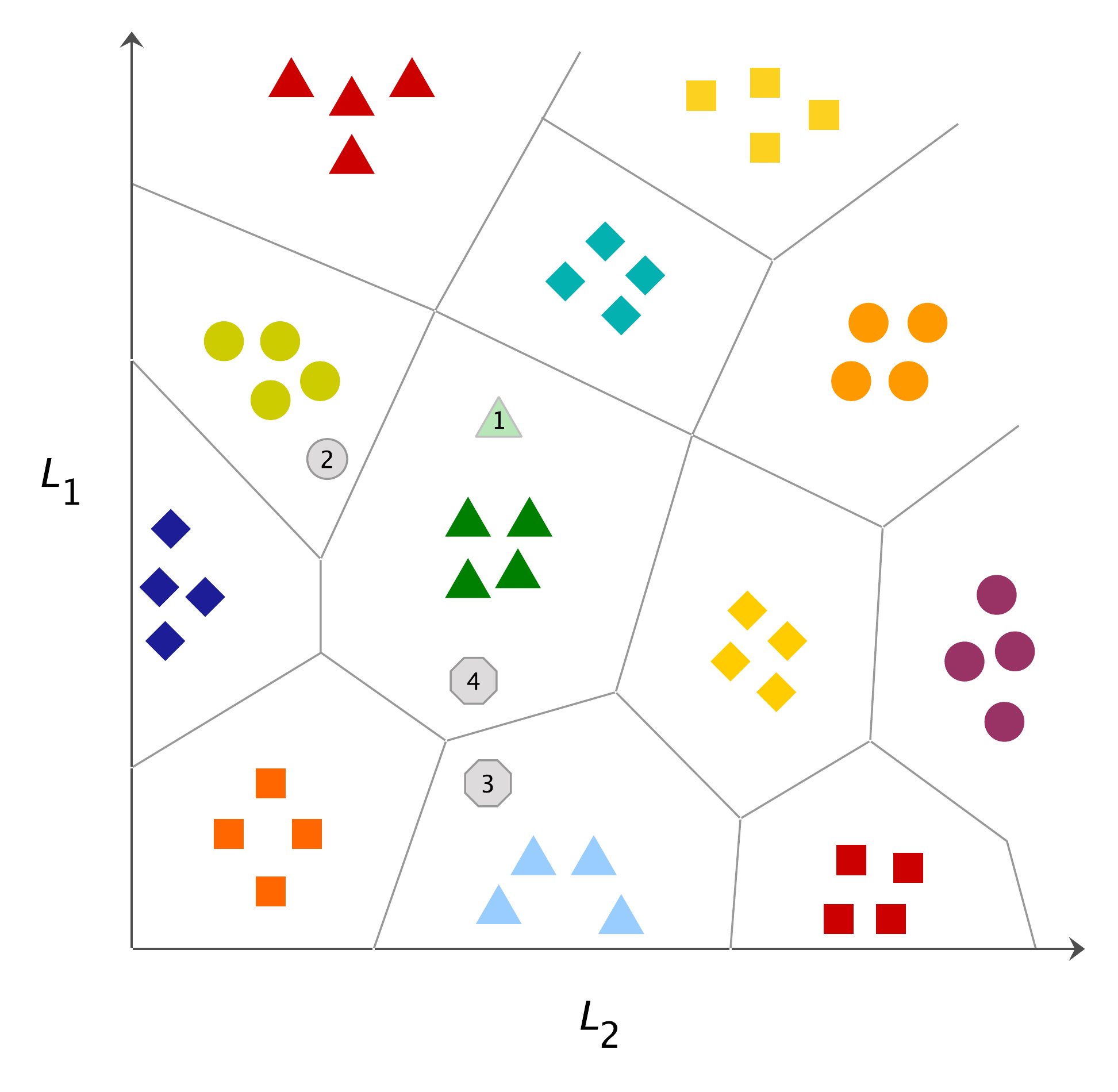}
  \caption{Mapping of instances through discriminant analysis}
  \label{fig:auth-mapping}
\end{figure}

LDA transforms the feature space of the instances so that the instances of the same class is packed closer together. This compaction creates clusters that are separated from each other with well-defined boundaries. These boundaries cannot be easily visualized as the feature space spans to multiple dimensions. For instance, the result of LDA after PCA (i.e., CDA) for the GTS (in Section~\ref{sec:gts-results-evaluation}) reached up to 99 discriminant features for each gait template.

An illustration of the LDA mapping is shown in Figure~\ref{fig:auth-mapping}. Only two discriminant features, $L_1$ and $L_2$, are included in this example for a simple illustration. Assume that this is the transformation that resulted from applying CDA to the GEI-GTS template representation of SetA1 of the CASIA-B dataset used for the training phase. Each item in the plot represents a gait instance and there are four instances per person registered in the system. The lines depict the boundaries formed as a result of this feature transformation.

When a gait instance of a registered member is provided as input, the transformed instance is more likely to be mapped to the region that confines the features that correspond to the member. In case 1 of Figure~\ref{fig:auth-mapping}, the authentic (genuine) member is mapped to his respective region, the claimed identity matches the identity of the region mapped, and hence considered authorized.

Let us then consider the case of a type 2 impostor, that is, a registered member claiming the identity of another member. A type 2 impostor (case 2 in Figure~\ref{fig:auth-mapping}) is a registered member who tries to pose as another in the system. As a registered member, a type 2 impostor would already have a signature region in the discriminant feature space. Hence he/she is more likely to be mapped to his/her respective region. Therefore, the system would recognize that this member is an impostor as the claimed identity will not match the identity returned (according to MSM).

Consider the case of a type 1 impostor. In a system where the population of registered members is considerably large (greater than 100), the adversary who attempts to claim the identity of a known user is more likely to be mapped to a random location in the discriminant feature space (case 3). In this case, the claimed and returned identities would not match and the impostor would be detected by the MSM system. However, in a system where lesser number of members are registered, then it is more likely for the adversary to be mapped on to the region that correspond to the claimed identity (case 4). In such a case, the MSM system would fail.

The solution to this problem proposed in this chapter would be to use Bayesian probability as a threshold. Bayesian probability would be more robust than the Euclidean nearest neighbour threshold as Bayes' rule is optimal when used in conjunction with LDA \citep{hamsici2008bayes}. This will make sure that the system performance is not adversely affected by the system population.


\section{Method}
\label{sec:method}

\begin{figure}
  \centering
  \includegraphics[width=1.0\linewidth]{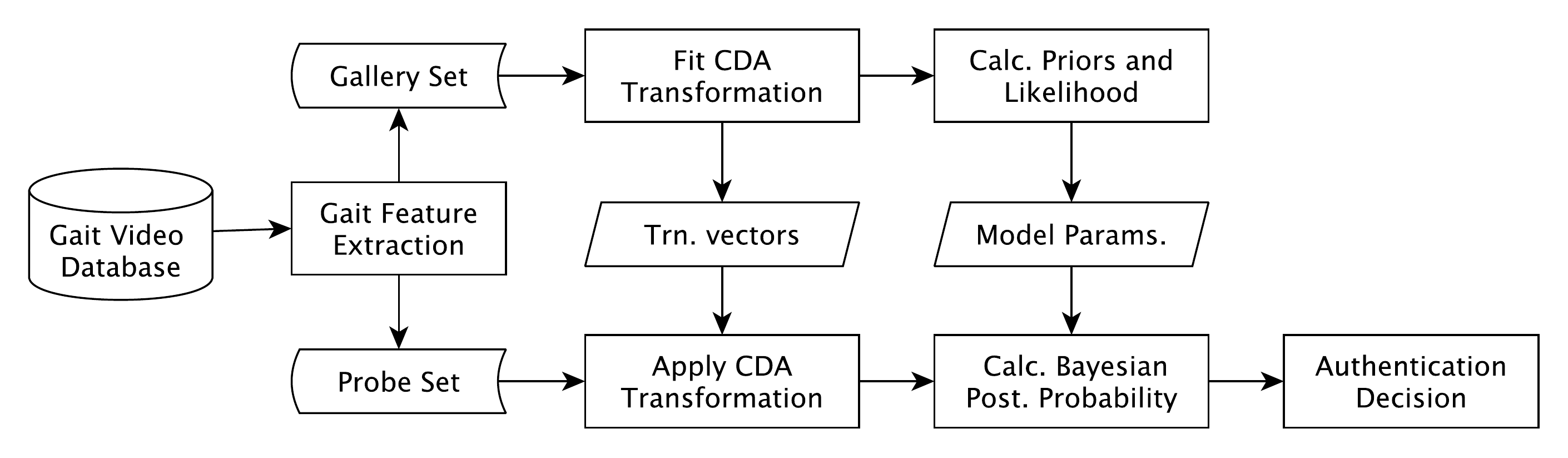}
  \caption{The basic flow of the proposed framework}
  \label{fig:arch}
\end{figure}

A simple outline of the proposed method is as shown in Figure~\ref{fig:arch}. The dataset is composed of gait feature templates which are divided into gallery and probe sets. The gallery proceeds through feature reduction after which the priors and likelihood of the feature space are calculated. The transformation vectors produced by the feature reduction process are used to apply the same on the probe set. The probe subjects are authenticated by the inferring the posterior probability from the calculated priors and likelihood.

\subsection{Gait Feature Extraction}
\label{sec:feature}

The same gait template feature representations explained in Section~\ref{sec:spatio} is also used in this chapter for experimentation. The feature templates include GEI, GEnI, AEI, and GTS applied to GEI. Samples of the above templates for a single subject are illustrated in Figure~\ref{fig:templates} for different covariate conditions of the CASIA-B database. 

\begin{figure}
  \centering
  \includegraphics[width=0.87\linewidth]{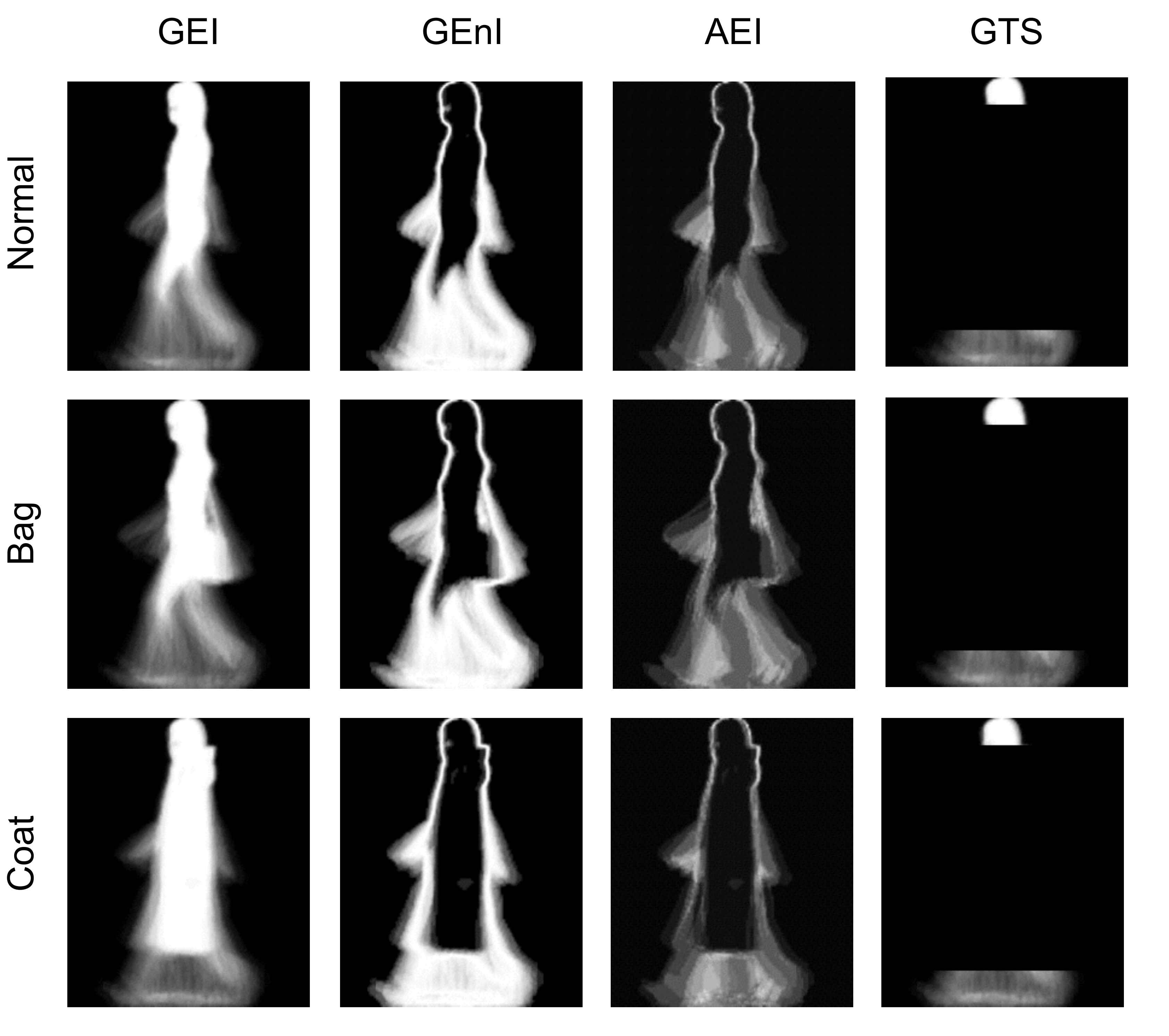}
  \caption{Sample gait templates for each intra-class variation of CASIA-B}
  \label{fig:templates}
\end{figure}

\subsection{Bayesian Thresholding}
\label{sec:model}

A classifier based on the probabilistic generative model models the posterior probability for all classes of the system \citep{bishopProbModel}. Examples of generative models include HMM, Bayes classifier, and Markov random fields.  Although the problem at hand is not a classification problem \textit{per se}, this property can be exploited to utilize this classifier as a tool for authentication. The Bayes classifier is one such model that works efficiently with LDA.

Let $d$ be the total number of features (dimensions) in a feature vector and $y_i$ be the identity for a subject, $i$. A subject with gait features $x=[x_1,x_2,...,x_d]$ who is to be authenticated ought to conform to the identity claimed $y_k$ with a sufficiently high posterior probability $\Pr(y_k \mid x)$. This relation is given by the Bayes' rule as follows.
\[\Pr(y_k \mid x) = \frac{\Pr(x \mid y_k) \cdot \Pr(y_k)}{\sum_i\Pr(x \mid y_i) \cdot \Pr(y_i)}\]
The subject is said to accepted only when $\Pr(y_k \mid x) > \theta_p$, where $\theta_p$ is the threshold probability ($0<\theta_p<1$) that is to be empirically estimated. The likelihood, $\Pr(x\mid y_k)$, is calculated through a multivariate Gaussian distribution $\mathcal{N}(\mu_k,\Sigma_k)$. The expression for the probability density function of the Gaussian likelihood is given in Section~\ref{sec:gait-recog}. 

It is theoretically sufficient to use only the likelihood as the probability threshold in place of the posterior probability. However, it is experimentally observed that this causes the decision boundary to be extremely sensitive making locating the optimal threshold difficult. Hence using just the likelihood would not be applicable for the purpose of authentication.

\begin{algorithm}[t]
  \begin{spacing}{1.3}
  \begin{algorithmic}
    \caption{BT-based gait authentication}\label{alg:mgb-auth}
    \Procedure{AuthenticateGaitBT}{$\mathit{instance}, \mathit{claimedID}$}
    \State $X_p \gets \text{transform}(\mathit{instance}, T)$
    \State $\mathit{prob} \gets \Pr(\mathit{claimedID} \mid X_p)$
    \State $\mathit{authenticity} \gets (\mathit{prob} > \theta_p)$
    \State \textbf{return} authenticity
    \EndProcedure
  \end{algorithmic}
  \end{spacing}
\end{algorithm}

A simple description of the Bayesian thresholding (BT) for gait authentication is depicted in Algorithm~\ref{alg:mgb-auth}. In a nutshell, the supplied gait instance is transformed according to a template configuration, $T$. The posterior probability, \textit{prob} (based on the Gaussian likelihood) of the claimed identity given the transformed feature instance, $X_p$ is calculated. The instance is considered authentic only when this value is greater than $\theta_p$. Thus the threshold probability, $\theta_p$ can be thought of as the minimum posterior probability required to reject the given instance.

\subsection{Two-pass Variation of BT for GTS}
\label{sec:mgb-2p}

The performance of the GEI is ideal in normal conditions while the GTS-GEI is optimal in covariate conditions. Hence, when the predictions of both of these templates are taken into account, a better authentication system can be obtained. Consider an ensemble consisting of two authentication systems, S$_1$ and S$_2$, where S$_1$ uses a model which is trained with the GEI templates and that of S$_2$ is trained with the GTS-GEI templates (GEI with GTS masking). The test gait instance is considered authenticated as its claimed identity if any one of the systems, S$_1$ or S$_2$, accepts it. That is, the instance is rejected only if it is rejected by both systems of the ensemble. This process is called the two-pass variation for the GTS (GTS-2P). The steps are illustrated in Algorithm~\ref{alg:2pbt}. Both $\theta_p$ and $\theta_q$ are empirically set thresholds.

\begin{algorithm}[t]
  \begin{spacing}{1.3}
  \caption{Two-pass BT for gait authentication}\label{alg:2pbt}
  \begin{algorithmic}
    \Statex \textbf{Preconditions:}
    \Statex \hspace{1.5em} S$_1$ is trained with $T_1$ configuration
    \Statex \hspace{1.5em} S$_2$ is trained with $T_2$ configuration
    \Procedure{AuthenticateGaitBT\_2P}{\textit{instance},\textit{claimedID}}
    \State $X_{p} \gets \text{transform}(\mathit{instance}, T_1)$
    \State $\mathit{prob}_p \gets \Pr(\mathit{claimedID} \mid X_{p})_{S_1}$
    \State $X_{q} \gets \text{transform}(\mathit{instance}, T_2)$
    \State $\mathit{prob}_q \gets \Pr(\mathit{claimedID} \mid X_{q})_{S_2}$
    \State $\mathit{authenticity} \gets (\mathit{prob}_p > \theta_p)~~\mathtt{OR}~~(\mathit{prob}_q > \theta_q)$
    \State \textbf{return} \textit{authenticity}
    \EndProcedure
  \end{algorithmic}
  \end{spacing}
\end{algorithm}

The two-pass variation for BT is slightly different to the one explained in Section~\ref{sec:two-pass-variation} for MSM. The MSM does not have any threshold to tune so the application is straight-forward. In BT, each template representation has a different threshold range. Therefore, a common threshold cannot be used to tune the error rates of the authentication system. The conclusion drawn from Section~\ref{sec:covar-fact} shows that though this operation decreases the FRR of the system, it increases the FAR. This result entails that the two-pass variation combines not only the strengths of both systems, but also their weaknesses. As the increase in FAR is inevitable in this variation, it is minimized for each individual system such that the consolidate FAR of the ensemble is optimal. The thresholds for both S$_1$ and S$_2$ are tuned towards 1\% FAR for an easier comparison with MSM\footnote{The theoretical FAR of the MSM system is also 1\% for system population, $n=100$.}. 

\section{Results \& Evaluation}
\label{sec:bt-results}

\subsection{Dataset Configuration}
\label{sec:bt-dataset}

The dataset configuration used in the previous chapter assumed a simple scenario where impostors claim random registered identities. In this section, a slightly different configuration is adopted to clearly study the strength of the Bayesian thresholding system compared to the MSM.

As in Section~\ref{sec:msm-dataset}, the members of the dataset is divided into two sets, D$_1$ and D$_2$. Member set D$_1$ assumes 100 authorized members where the remaining 24 are assumed as unauthorized members D$_2$. This time, each member ($m$) of D$_2$ tries all possible identities of D$_1$ for type 1 impersonation. As for type 2, each member of D$_1$ claims the identity of every other member of D$_1$ other than the identity owned. Just as before, only instances (say $i$) of SetA1 $\cap$ D$_1$ is used for training and the rest for testing. For each covariate, the sum of positive cases is
\[ 2i \times 100m \times 1\text{ ID} = 200\]
while that of negative cases for type 1 attackers and type 2 attackers are respectively
\[ 2i \times 24m \times 100\text{ ID} = 4800\] 
\[ 2i \times 100m \times 99\text{ ID} = 19800\]
This formulation is specifically done to study the effect of FAR in detail. 

\subsection{Performance Comparison with NN Threshold}
\label{sec:bt-vs-nn}

Conventional authentication models find the Euclidean distance in the feature space between the test instance $x_t$ and the instance $x_k$ of the claiming identity, $k$. Usually, $x_c$ can either be
\begin{enumerate}[topsep=0pt, noitemsep, leftmargin=*, label=(\alph*)]
\item A point on the multidimensional plane that represents the mean of the gallery instances of subject $k$ \citep{sarkar2005humanid}
\item An exemplar instance of $k$ \citep{kale2004identification}
\item The instance of $k$ which is closest to $x_t$, i.e., the nearest neighbour \citep{matovski2012effect}
\end{enumerate}
The third case is used for benchmark comparison for this study. This is the same case used to compare the Euclidean threshold method with the MSM in the previous chapter.

\begin{figure*}
  \centering
  \subfloat[]{\includegraphics[width=0.48\linewidth]{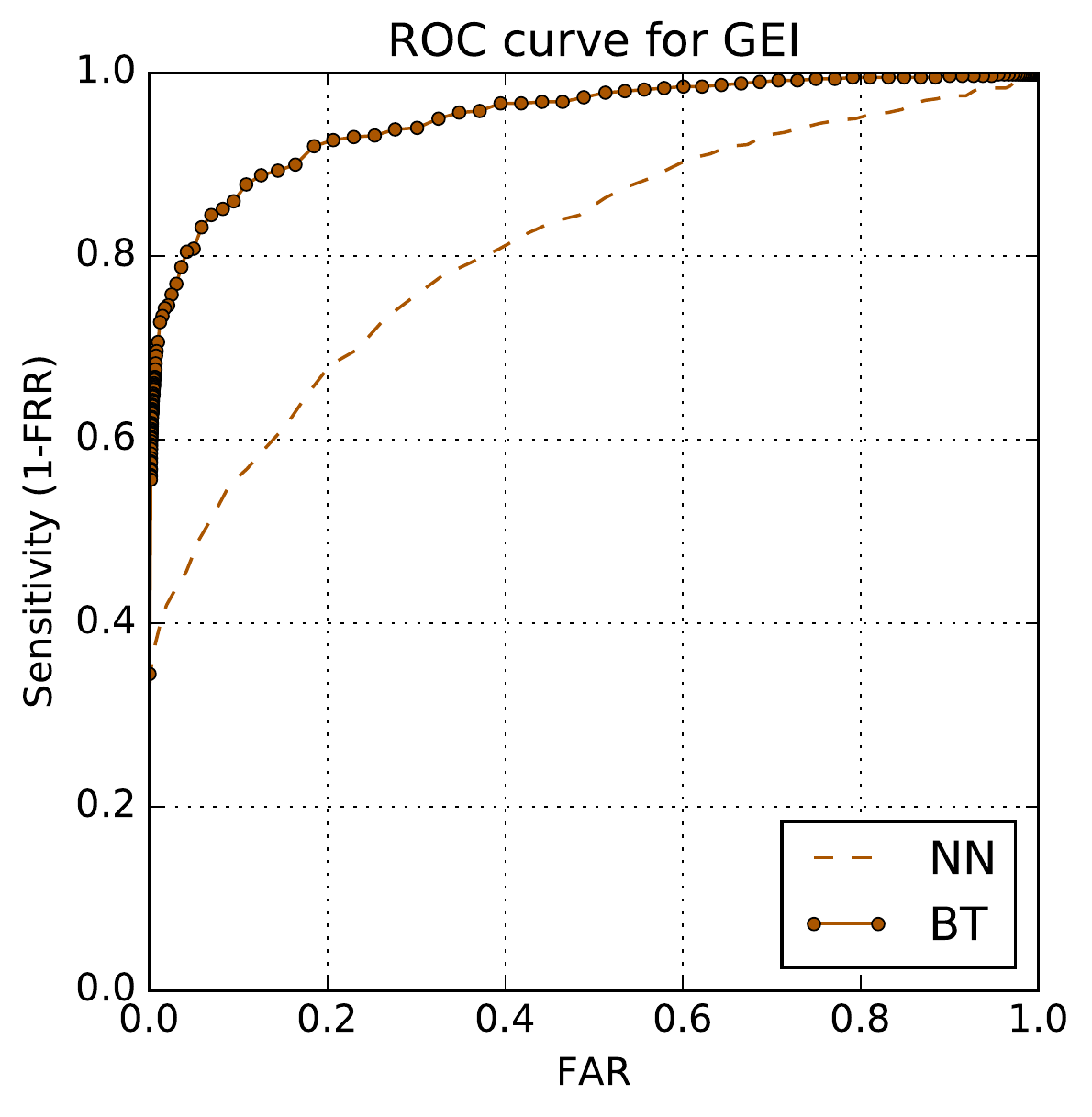}%
    \label{fig:roc-gei}}
  \hfil
  \subfloat[]{\includegraphics[width=0.48\linewidth]{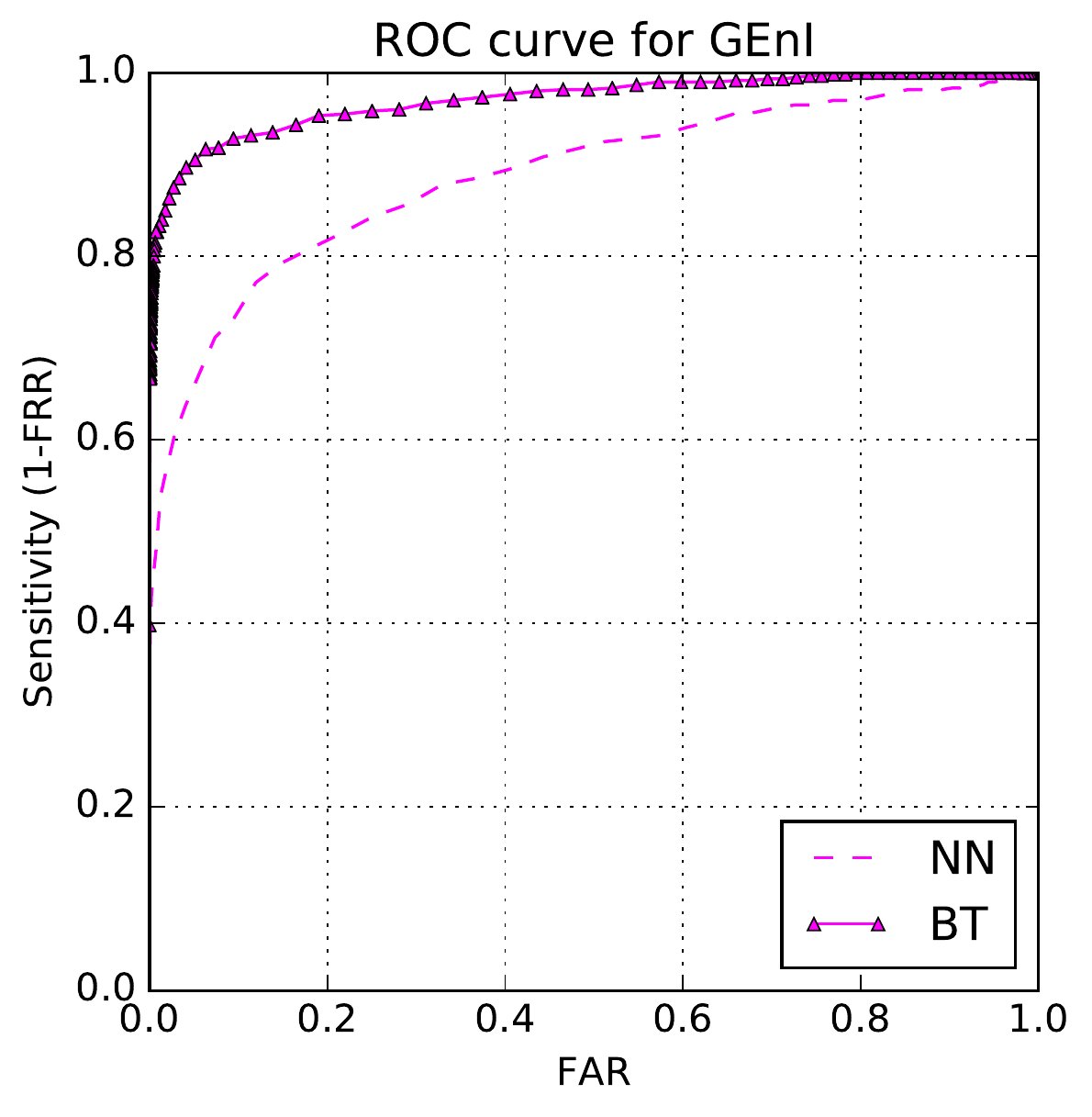}%
    \label{fig:roc-geni}}

  \vspace{1em}
  \subfloat[]{\includegraphics[width=0.48\linewidth]{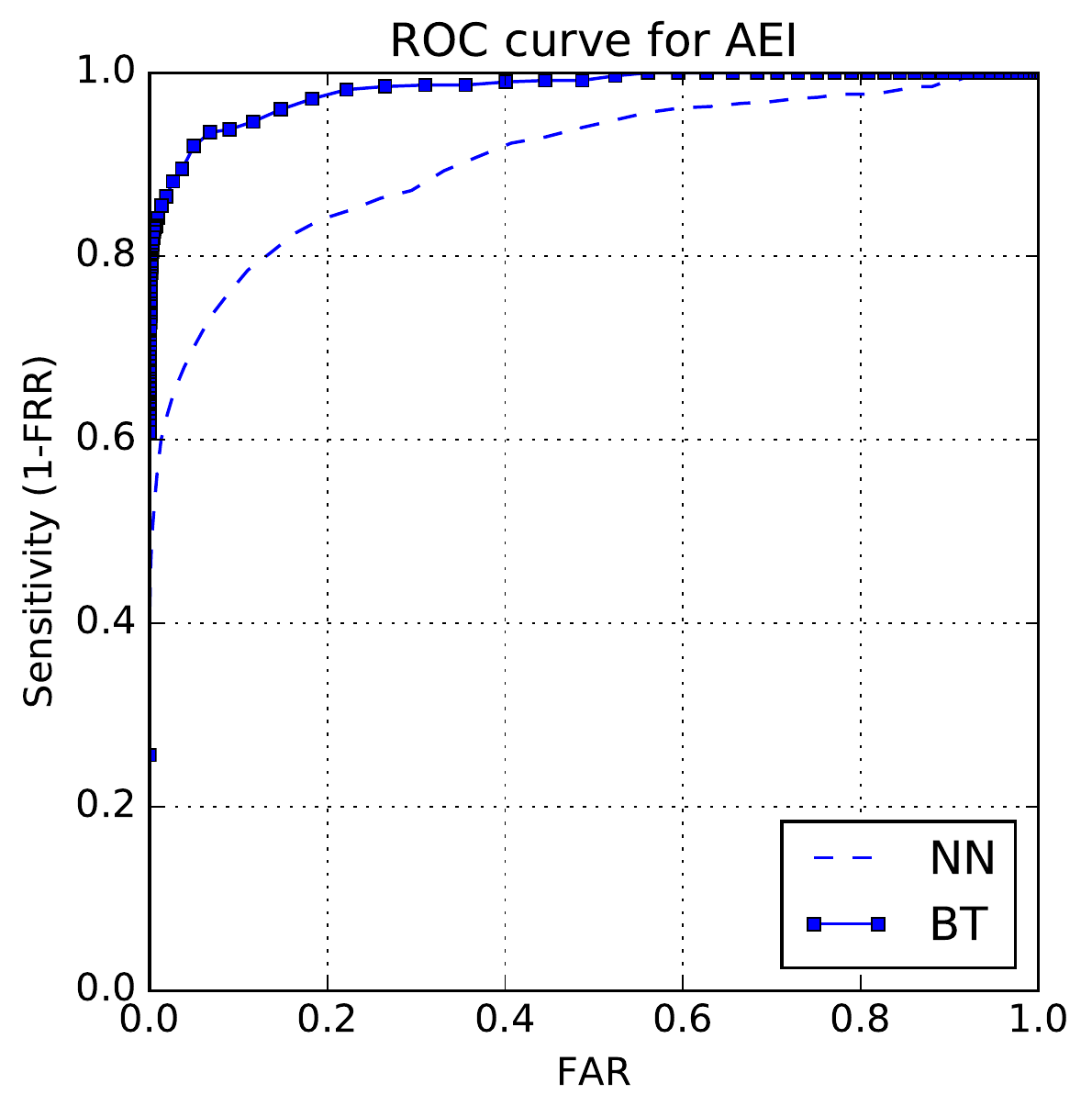}%
    \label{fig:roc-aei}}
  \hfil
  \subfloat[]{\includegraphics[width=0.48\linewidth]{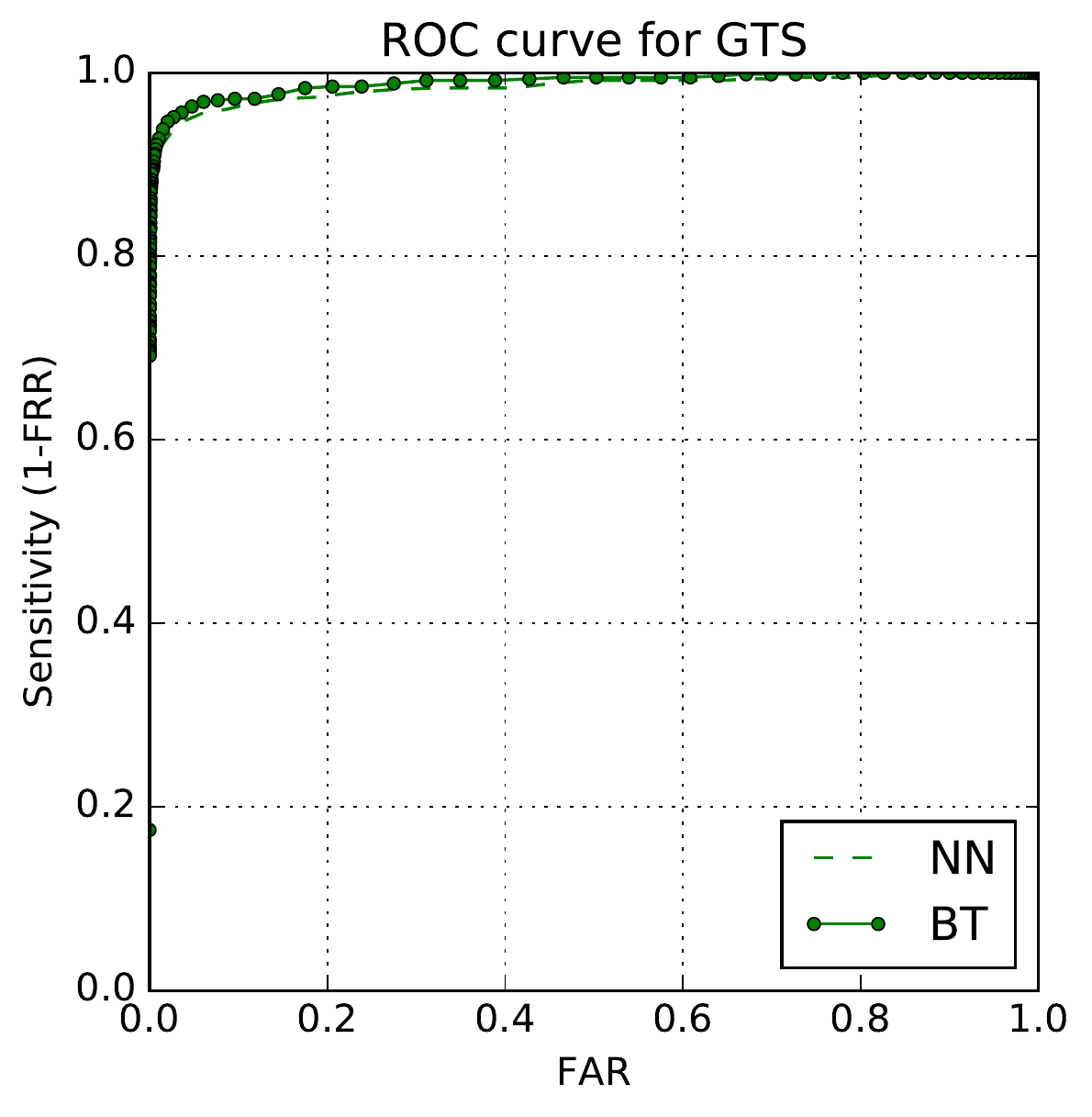}%
    \label{fig:roc-gts}}
  \caption{ROC curves of BT vs NN for each gait template representation at the sagittal angle (90\degree)}
  \label{fig:roc-nn-bt}
\end{figure*}

The improvement in performance between the NN and BT framework can be easily noticed in their respective ROC curves. The ROC curves for each template for both NN and BT appear as shown in Figure~\ref{fig:roc-nn-bt}. The error rates shown are the average across all three covariates of the CASIA-B database: normal, bag, and coat. The greater area under the ROC curve for the BT shows that the proposed method performs considerably better than the de facto Euclidean thresholds. This effect is much more pronounced for GEI, GEnI and AEI than the GTS as the covariate-resilient feature set of GTS causes the ROC curves of NN and BT to almost overlap. Even so, on a closer inspection, the area under the ROC curve of BT is still greater than that of NN for the GTS template as well.

\begin{figure*}
  \centering

  \subfloat[]{\includegraphics[width=0.42\linewidth]{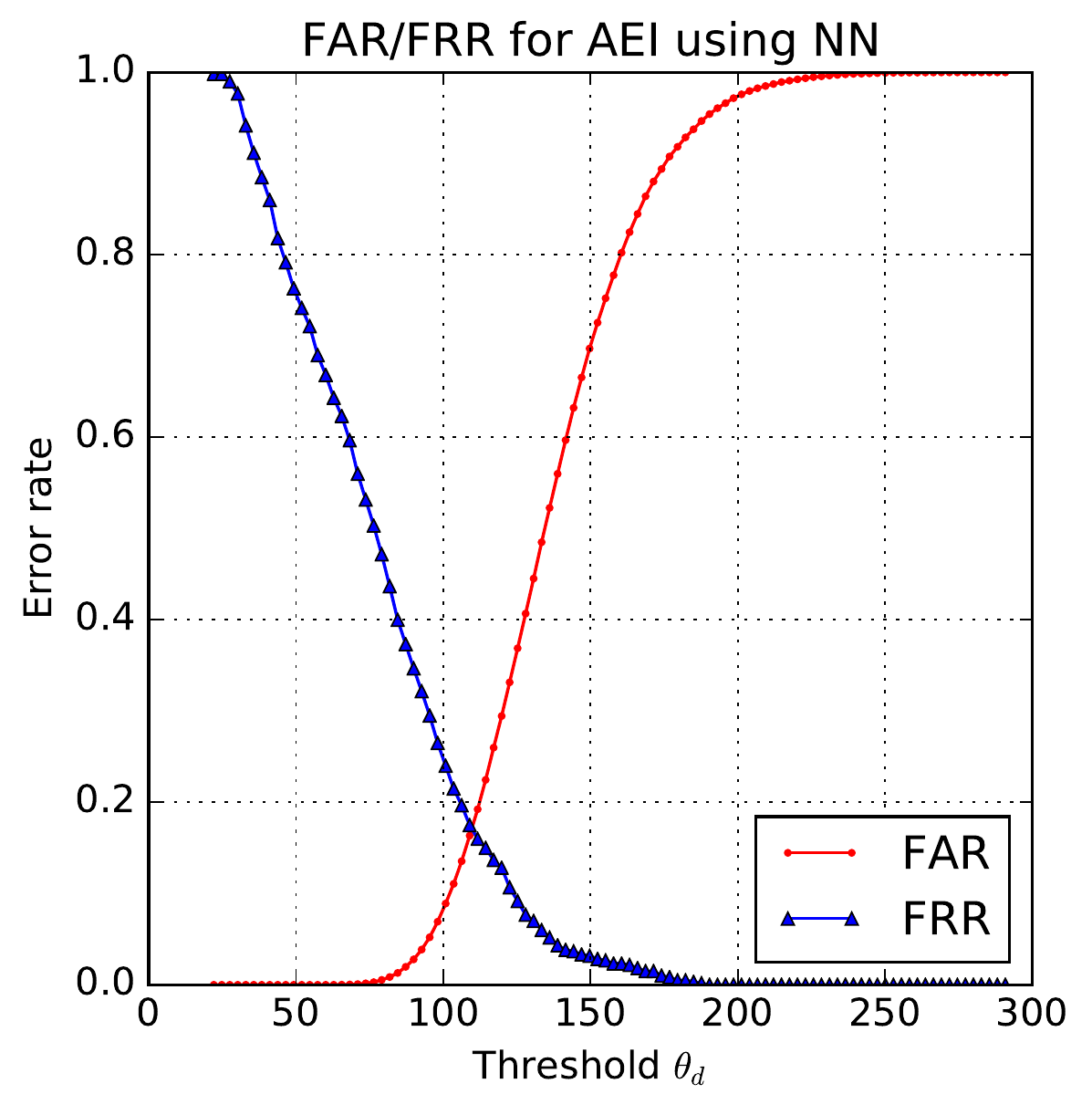}%
    \label{fig:er-aei-nn}}
  \hfil
  \subfloat[]{\includegraphics[width=0.42\linewidth]{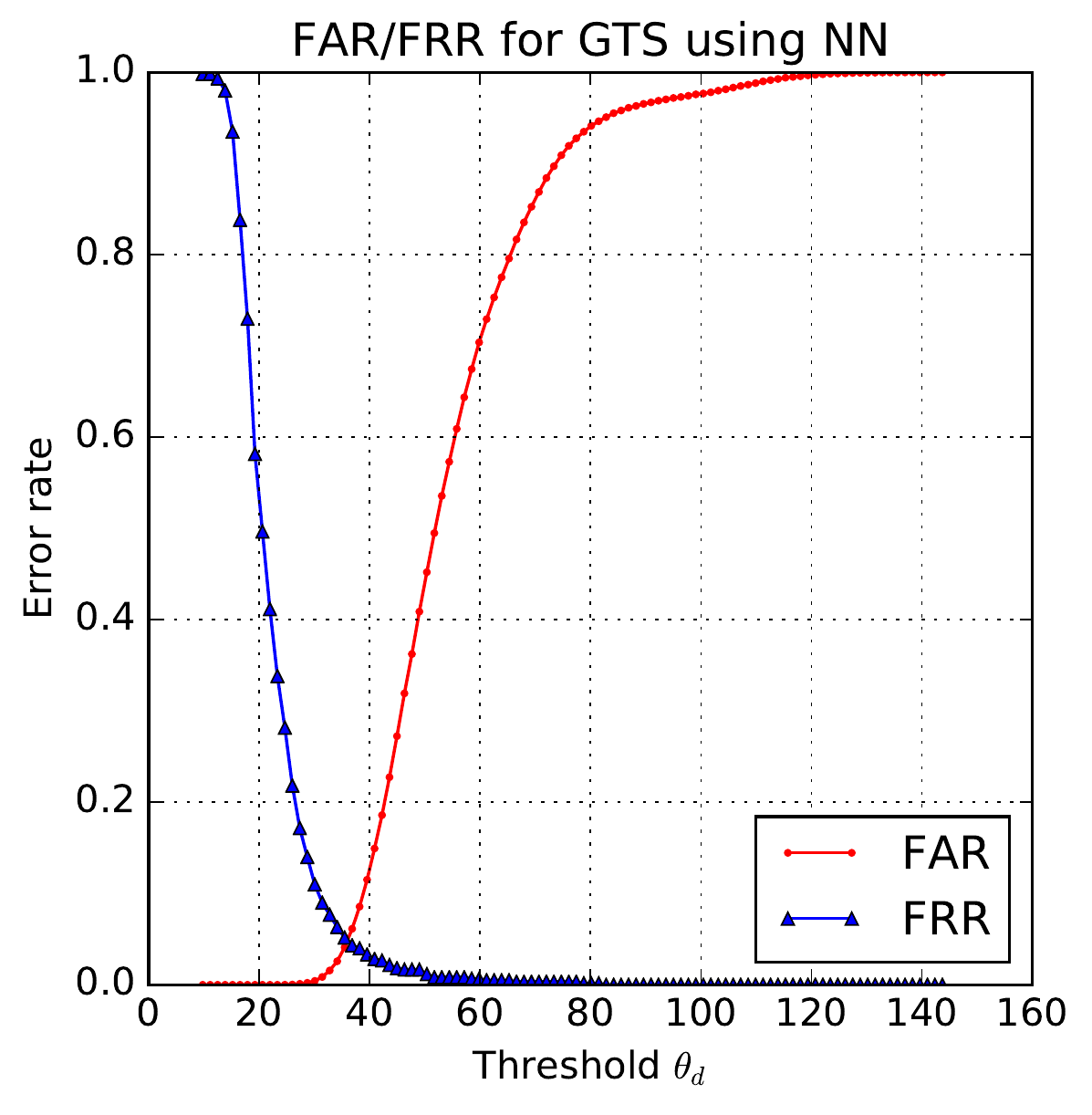}%
    \label{fig:er-gts-nn}}

    \subfloat[]{\includegraphics[width=0.42\linewidth]{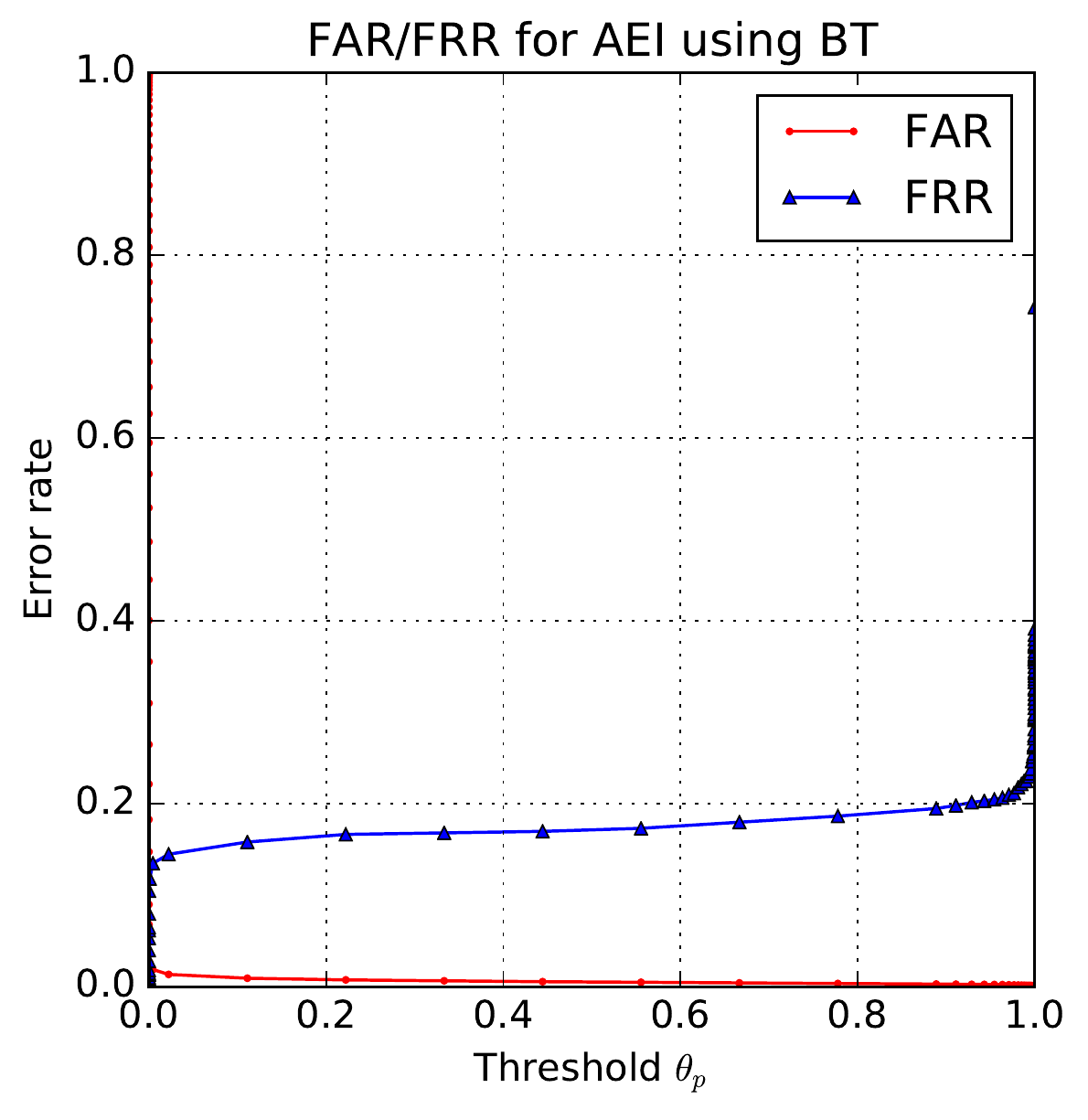}%
    \label{fig:er-aei-bt}}
  \hfil
  \subfloat[]{\includegraphics[width=0.42\linewidth]{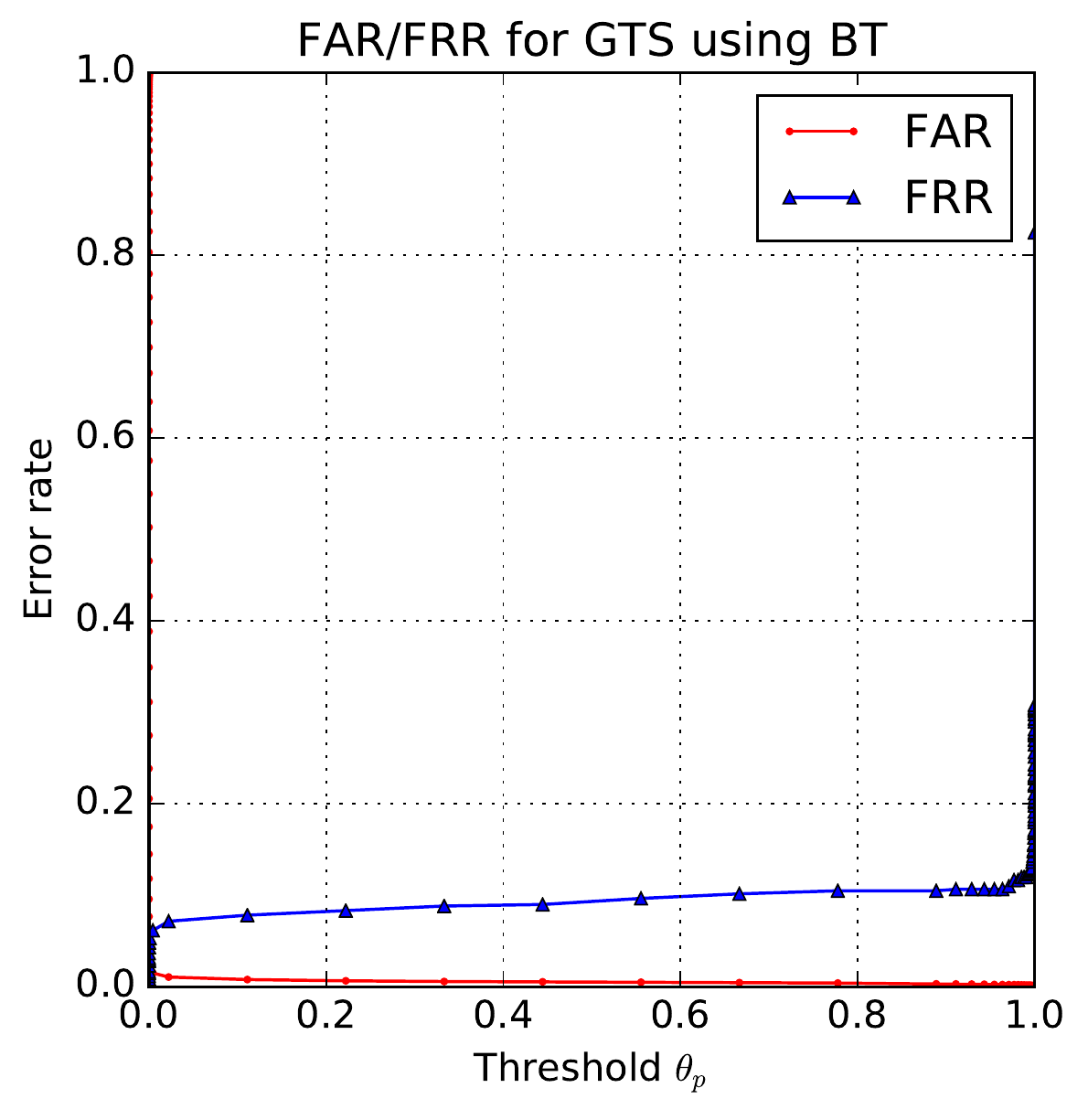}%
    \label{fig:er-gts-bt}}

    \subfloat[]{\includegraphics[width=0.42\linewidth]{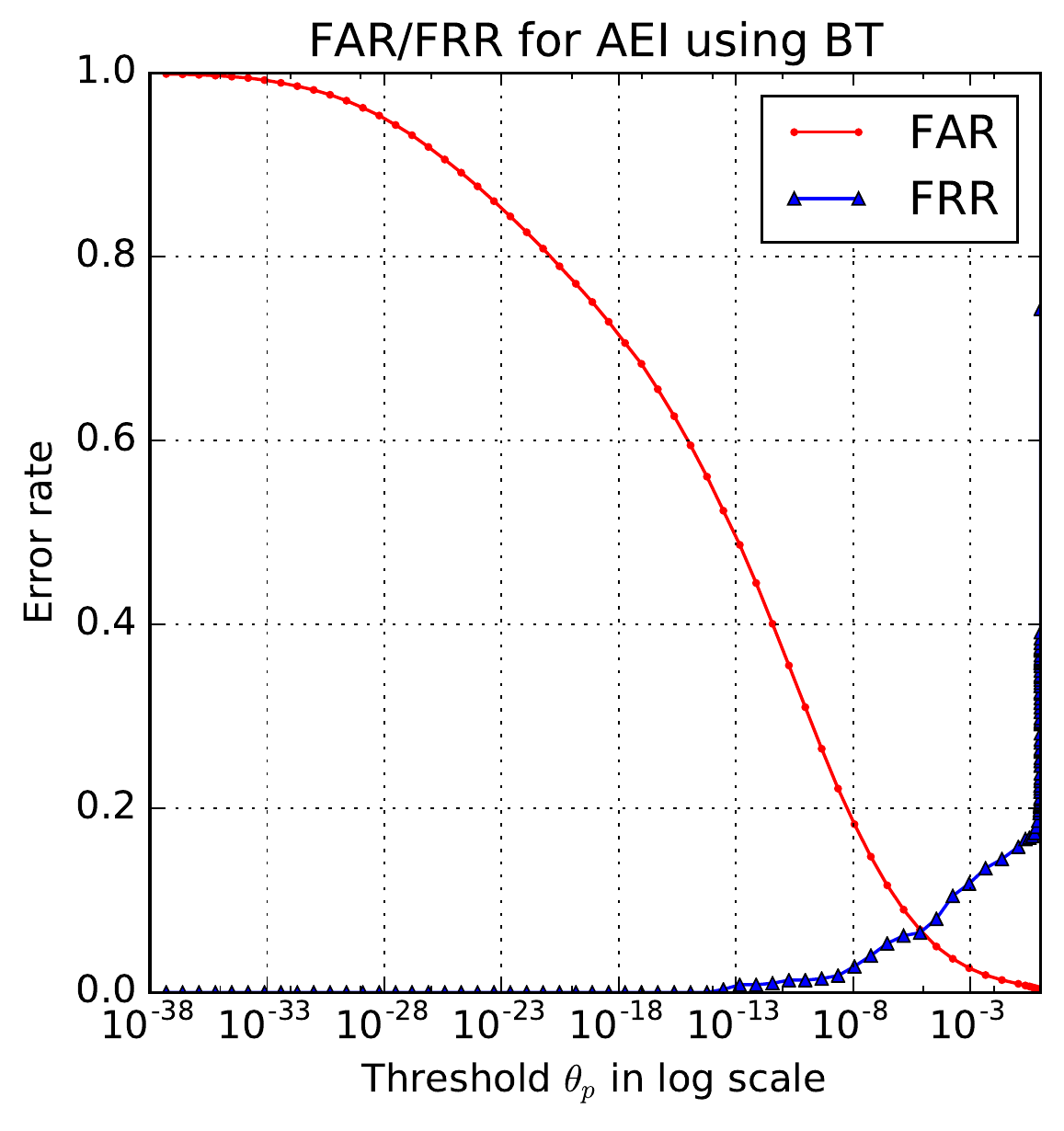}%
    \label{fig:er-aei-bt-log}}
  \hfil
  \subfloat[]{\includegraphics[width=0.42\linewidth]{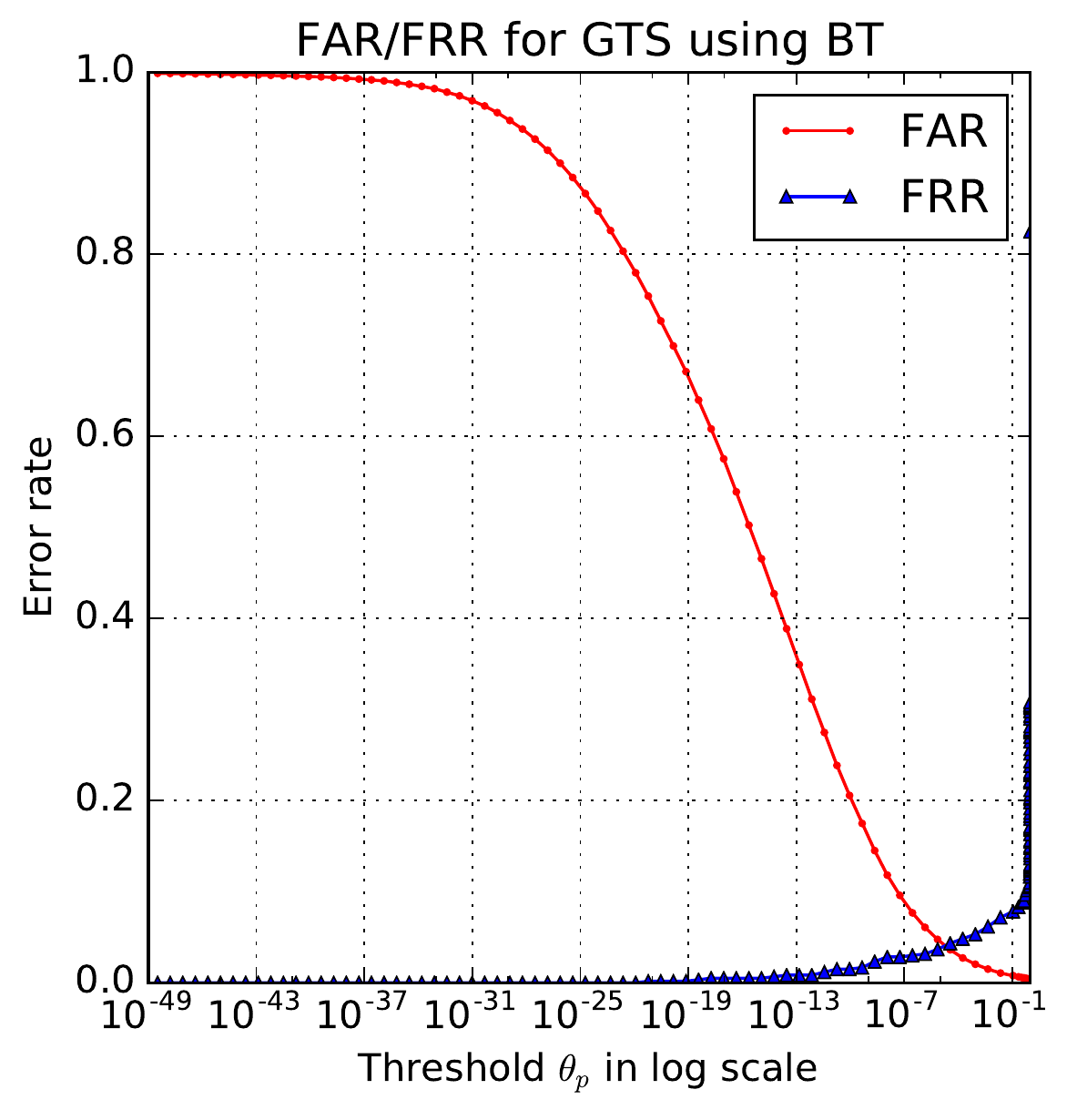}%
    \label{fig:er-gts-bt-log}}

  \caption{FAR and FRR curves of BT and NN for AEI and GTS at the sagittal angle (90\degree)}
  \label{fig:er-nn-bt}
\end{figure*}

Unlike MSM, the BT has a tunable threshold which facilitates it to have an EER. The EER can be obtained through intersection of their respective FAR and FRR curves. Let us limit our discussion to the AEI and the GTS\footnote{GTS in here refers to the GTS mask applied to the GEI template} templates as they are the best performing templates in this experiment.  Figures~\ref{fig:er-nn-bt}(a) and \ref{fig:er-nn-bt}(b) depicts the change in FAR and FRR with the Euclidean threshold using NN. As the distance threshold, $\theta_d$, grows system becomes more lax allowing more intruders to slip through leading to an increase in the FAR. However, a greater distance threshold minimizes the FRR. On the contrary, a smaller distance threshold makes the system stricter in evaluation thus reducing the FAR while increasing the FRR.

The error curves for the BT in Figures~\ref{fig:er-nn-bt}(c) and \ref{fig:er-nn-bt}(d) shows that the FAR becomes negligible immediately after an extremely small change in the threshold probability $\theta_p$. There is also a minor margin of difference between the point where the slope of FRR seems to saturate to the point where it reaches 100\%. The effect of probability as a threshold is so sensitive that the EER cannot be easily visualized at a linear scale. To get a better perception of the exact EER, the threshold response can be viewed in log scales as shown in Figures~\ref{fig:er-nn-bt}(e) and \ref{fig:er-nn-bt}(f). Notice the difference in the slopes of the curves to that of the NN threshold. A greater probability threshold, $\theta_p$, in here would make the authentication system stricter causing the reduction in FAR and increase the FRR as a consequence. The opposite effect can be observed for a smaller value of $\theta_p$.

\begin{figure}
  \centering
  \includegraphics[width=0.8\linewidth]{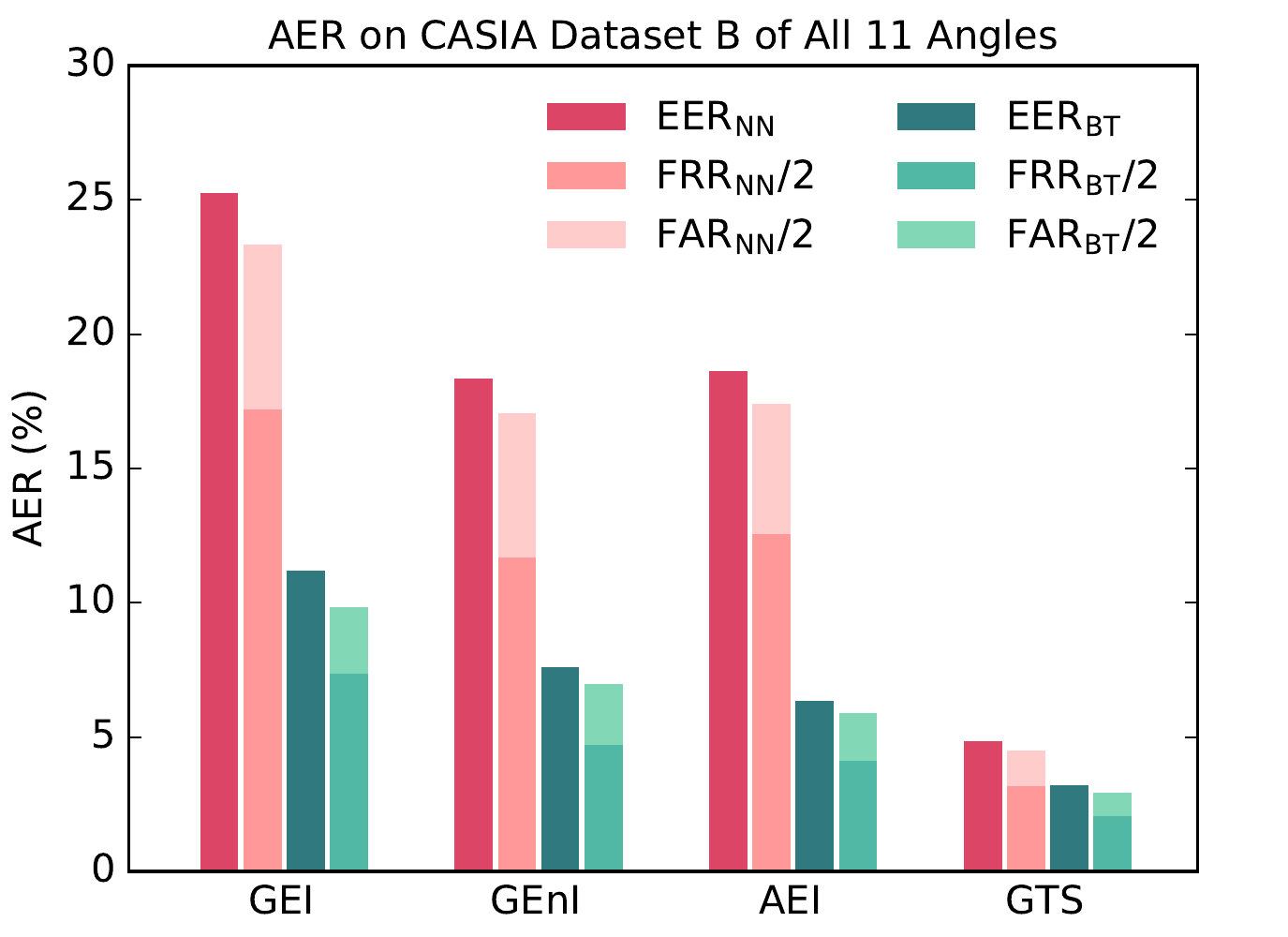}
  \caption{Comparison of Euclidean-NN and BT methods. The FAR is the average of Type 1 and Type 2 FARs.}
  \label{fig:aer-mgb}
\end{figure}

The performance of each method is compared through their minimum observable AER in addition to their EER. The results in Figure~\ref{fig:er-nn-bt} includes only the performance of the sagittal angle. The experiment is repeated for all 11 angles of the CASIA-B dataset. The results are averaged and depicted in Figure~\ref{fig:aer-mgb} and Table~\ref{tab:nn-vs-bt}. The FAR and FRR are obtained by tuning the threshold so as to minimize its AER. The values are the mean of the respective errors over all three covariate factors averaged over the 11 views of the CASIA-B dataset. We could infer that BT significantly outperforms the existing Euclidean-NN-based implementation for authentication. 

The FAR depicted so far in this section is the average of both type 1 and type 2 FARs. The types of FAR is separately evaluated and compared in Table~\ref{tab:far}. The BT has a better type 1 FAR and far superior type 2 FAR compared to that of NN. However, the type 1 FAR of both NN and BT using the GTS template is marginally equal. Thus, the proposed BT exceeds in performance in comparison with NN over all the metrics in the evaluation.

\begin{table}[t]
  \begin{center}
    \caption{Nearest Neighbour vs. Bayesian Thresholding Framework}
    \label{tab:nn-vs-bt}

    \renewcommand{\arraystretch}{1.1}
    \begin{tabular}{l *{4}{r} c *{4}{r}}
      \toprule
      \multirow{2}{*}{Template } & \multicolumn{4}{c}{Nearest Neighbour} &&
                                                                            \multicolumn{4}{c}{Bayesian Threshold}\\
      \cmidrule{2-5} \cmidrule{7-10}
                                 & EER & FAR & FRR & AER && EER & FAR & FRR & AER \\
      \midrule
      GEI & 25.25 & 12.31 & 34.37 & 23.34 && \textbf{11.20} &  4.94 & 14.71 & \textbf{ 9.82} \\
      GEnI & 18.36 & 10.69 & 23.41 & 17.05 && \textbf{ 7.61} &  4.58 &  9.41 & \textbf{ 6.99} \\
      AEI & 18.61 &  9.69 & 25.14 & 17.41 && \textbf{ 6.35} &  3.52 &  8.24 & \textbf{ 5.88} \\
      GTS &  4.85 &  2.64 &  6.37 &  4.51 && \textbf{ 3.21} &  1.72 &  4.12 & \textbf{ 2.92} \\
      \bottomrule
    \end{tabular}
  \end{center}
  
 \vspace{1em}
\end{table}

\begin{table}[t]
  \centering
  \caption{Comparison Based on the Types of FAR at Optimum AER}
  \label{tab:far}
  
  \renewcommand{\arraystretch}{1.1}
  \begin{tabular}{l rr c rr}
    \toprule
    \multirow{2}{*}{Template } & \multicolumn{2}{c}{Nearest Neighbour} &&
                                                                          \multicolumn{2}{c}{Bayesian Threshold}\\
    \cmidrule{2-3} \cmidrule{5-6}
                               & Type 1 & Type 2 && Type 1 & Type 2 \\
    \midrule
    GEI & 12.87 & 11.75 && \textbf{ 8.16} & \textbf{ 1.72} \\
    GEnI & 12.19 &  9.19 && \textbf{ 7.72} & \textbf{ 1.43} \\
    AEI & 10.97 &  8.41 && \textbf{ 6.09} & \textbf{ 0.95} \\
    GTS &  \textbf{2.79} &  2.50 && 3.11 & \textbf{ 0.33} \\
    \bottomrule
  \end{tabular}

  \vspace{1em}
\end{table}

\begin{table}[h!]
  \centering
  \caption{Average FRR of MSM and BT (at FAR=1\%) Taken Over 11 Angles}
  \label{tab:msm-vs-bt-frr}

  \renewcommand{\arraystretch}{1.1}
  \begin{tabular}{l r r}
    \toprule
    Template & FRR$_\text{MSM}$ & FRR$_\text{BT}$ \\
    \midrule
    GEI & 26.99 & \textbf{24.16} \\
    GEnI & 18.89 & \textbf{16.20} \\
    AEI & 15.95 & \textbf{14.01} \\
    GTS &  7.32 & \textbf{ 5.37} \\
    \bottomrule
  \end{tabular}
\end{table}

\subsection{Performance Comparison with MSM}
\label{sec:bt-vs-msm}

\begin{figure}[t]
  \centering
  \includegraphics[width=0.6\linewidth]{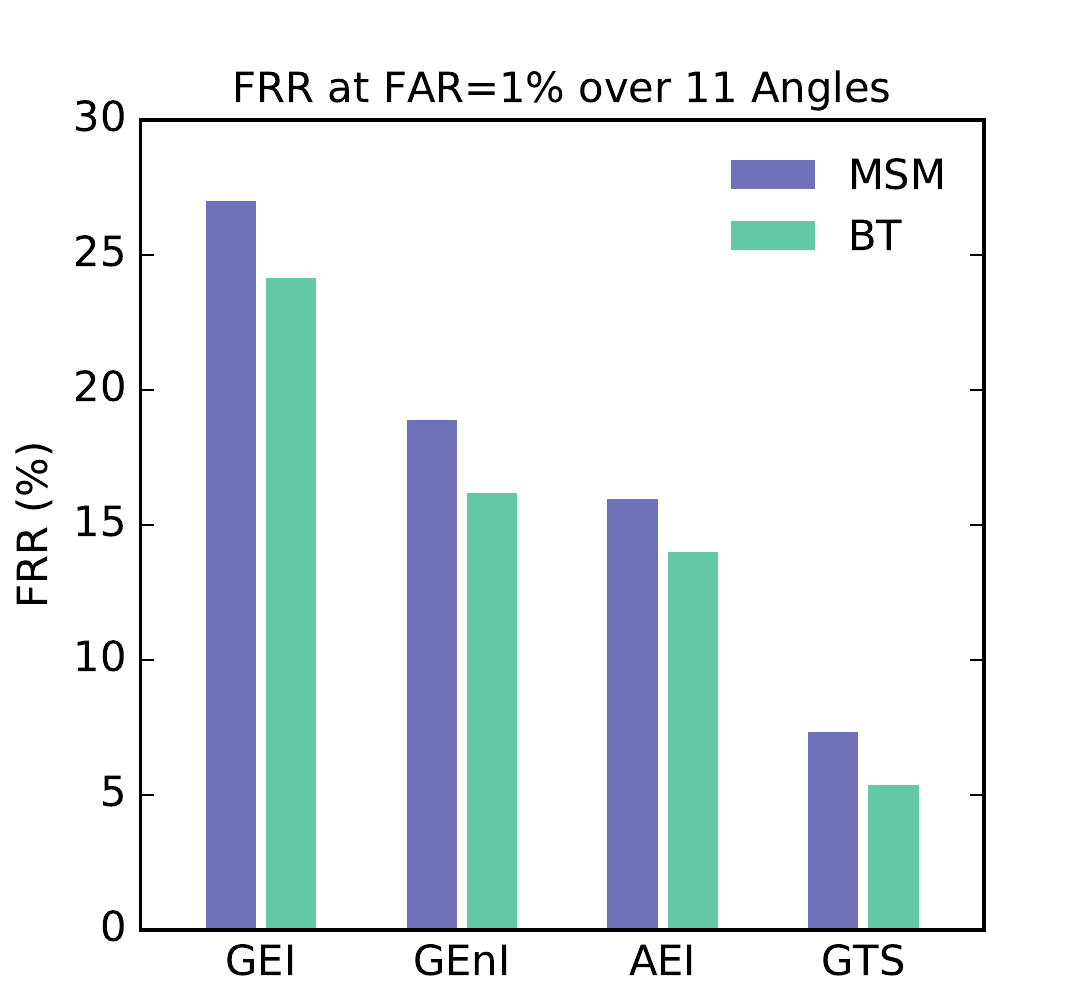}
  \caption{BT vs. MSM}
  \label{fig:frr-bt-msm}
\end{figure}

The FAR of the MSM system is $1/n$ where $n$ is the number of genuine members whose gait signatures are registered in the system. This expression was experimentally verified in Section~\ref{sec:thresh-vs-msm}. To compare the performance of BT against that of MSM, the threshold, $\theta_p$, ought to be tuned to match the fixed FAR of the MSM. The total number of unique genuine members in the training set are 100, which means that the FAR of MSM is $1/100=1\%$. Therefore, $\theta_p$ is adjusted such that the FAR of BT is also $1\%$. The metric to be compared in this section is the FRR of both methods. The FRR of both MSM and BT was obtained in this experimental setting over all 11 angles of the CASIA-B dataset. The resulting average for each template is as shown in Table~\ref{tab:msm-vs-bt-frr} and illustrated as a bar graph in Figure~\ref{fig:frr-bt-msm}. We can clearly observe that the FRR of BT is much lesser than that of MSM for all of the templates in the experiment.

\begin{figure}
  \centering
  \includegraphics[width=\linewidth]{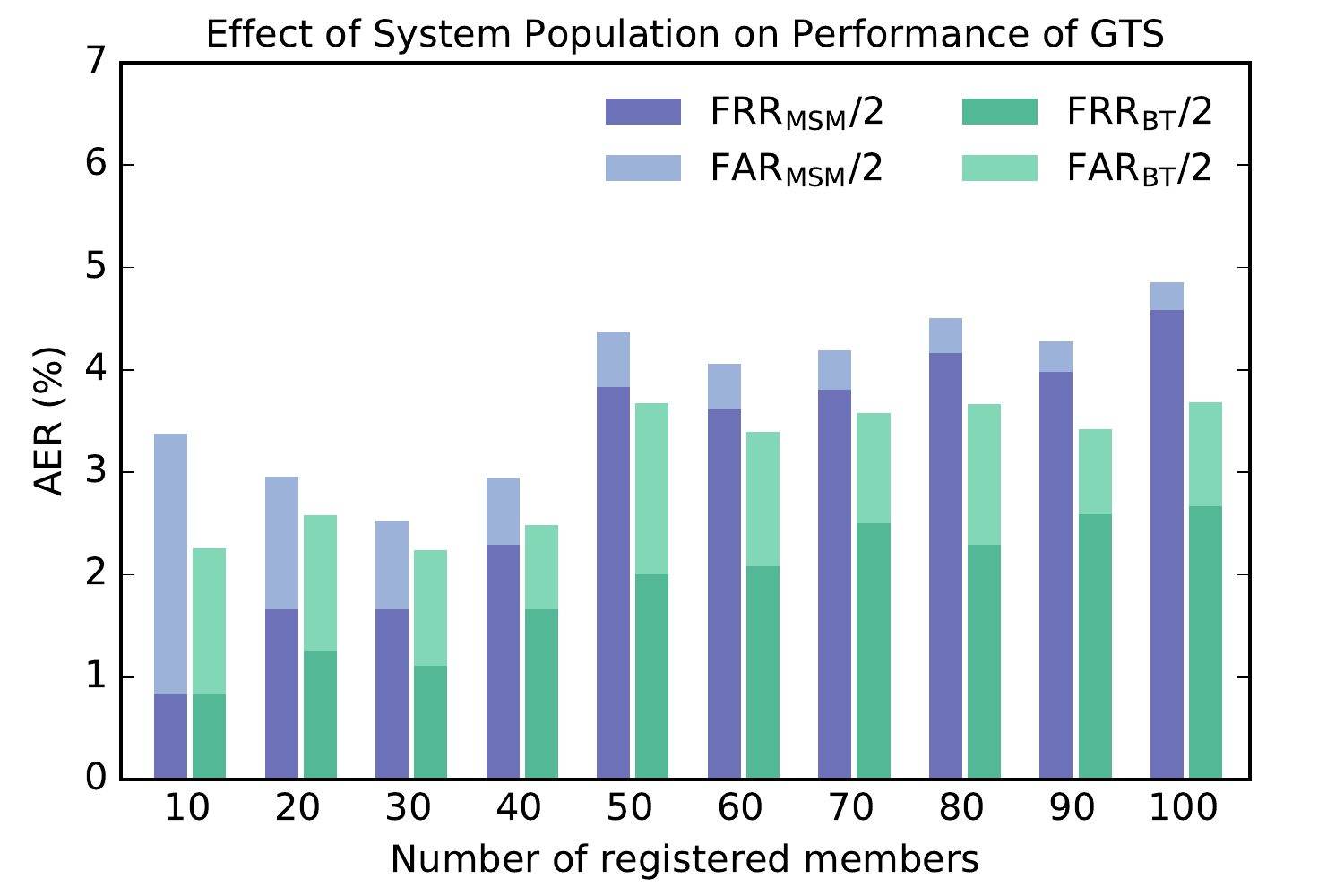}
  \caption{Effect of System Population on Authentication Error (GTS)}
  \label{fig:gts-sp}
\end{figure}

\begin{figure}
  \centering
  \includegraphics[width=\linewidth]{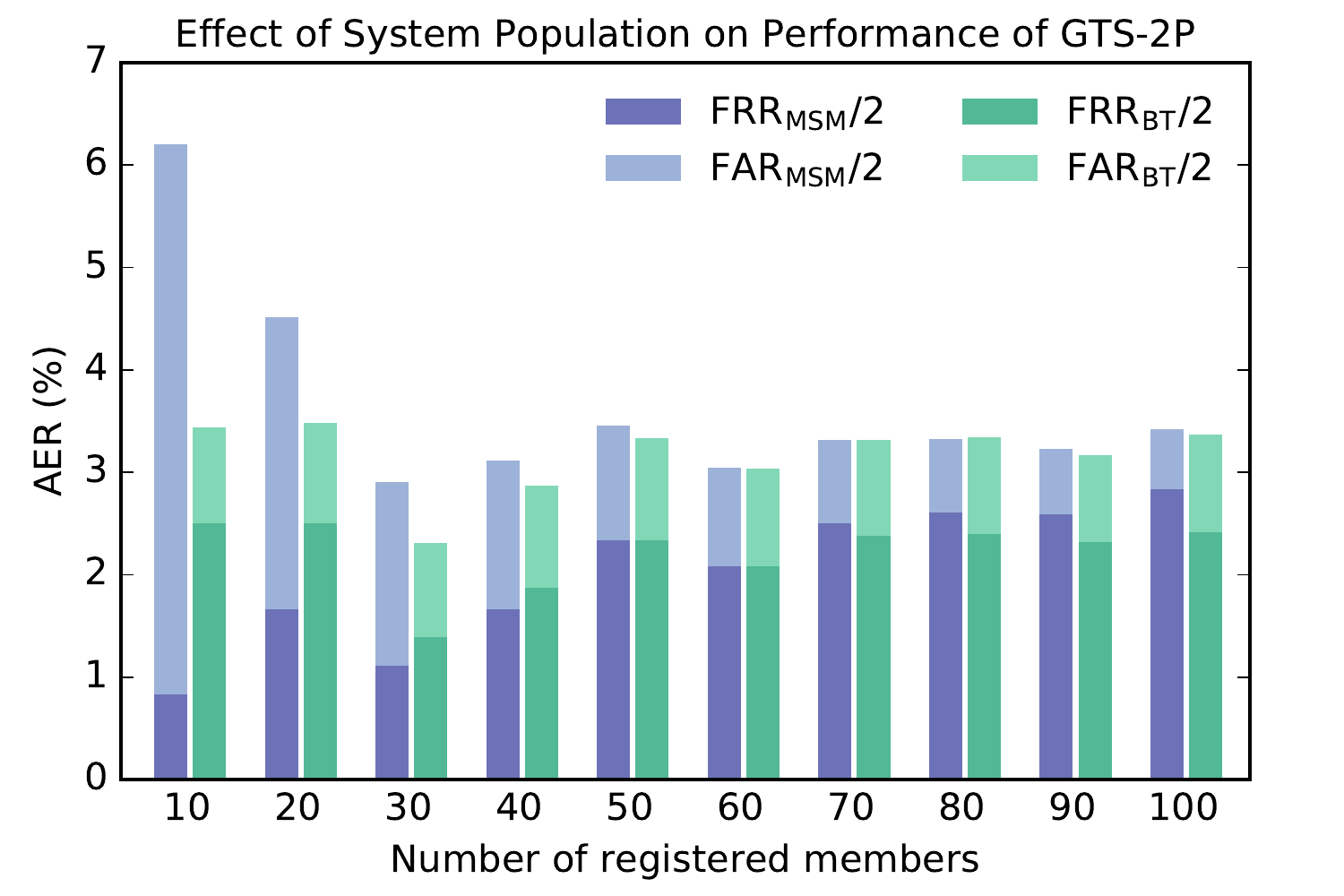}
  \caption{Effect of System Population on Authentication Error (GTS-2P)}
  \label{fig:gts-sp-2p}
\end{figure}

The change in system population can largely influence the performance of an authentication system. The main drawback that was identified in MSM is that its performance decreases with smaller population. To verify whether or not the BT shares the same weakness, their error rates are compared with different sizes of system population,~$n$. For this test, the $\theta_p$ that corresponds to the  minimum AER is used for BT. The outcome is depicted in Figure~\ref{fig:gts-sp} and Table~\ref{tab:gts-sp}. The BT method exhibited a smaller AER for all values of $n$ in the experiment. Initially, the FAR of MSM is smaller than that of BT. However, as expected, the FAR of the MSM increased with the decrease in $n$. The experimentation for the GTS-2P (Figure~\ref{fig:gts-sp-2p} and Table~\ref{tab:gts-sp-2p}) gave intriguing results where this effect is much more adverse.

\begin{table}
  \centering
  \caption{ Effect of System Population on Authentication Error (GTS)}
  \label{tab:gts-sp}
  
  \vspace{0.5em}
  \renewcommand{\arraystretch}{1.1}
  \begin{tabular}{r c *{3}{>{\centering\arraybackslash}p{0.07\linewidth}}
    c *{3}{>{\centering\arraybackslash}p{0.07\linewidth}}}
    \toprule
    \multirow{2}{*}{$n$ } && \multicolumn{3}{c}{MSM} &&
                                                                      \multicolumn{3}{c}{Bayesian Threshold}\\
    \cmidrule{3-5} \cmidrule{7-9}
                          && FRR & FAR & AER && FRR & FAR & AER \\
    \midrule
    10 &&  1.67 &  5.09 &  3.38 &&  1.67 &  2.85 & \textbf{ 2.26} \\
    20 &&  3.33 &  2.59 &  2.96 &&  2.50 &  2.66 & \textbf{ 2.58} \\
    30 &&  3.33 &  1.72 &  2.53 &&  2.22 &  2.26 & \textbf{ 2.24} \\
    40 &&  4.58 &  1.31 &  2.95 &&  3.33 &  1.63 & \textbf{ 2.48} \\
    50 &&  7.67 &  1.08 &  4.37 &&  4.00 &  3.34 & \textbf{ 3.67} \\
    60 &&  7.22 &  0.89 &  4.06 &&  4.17 &  2.63 & \textbf{ 3.40} \\
    70 &&  7.62 &  0.77 &  4.19 &&  5.00 &  2.17 & \textbf{ 3.58} \\
    80 &&  8.33 &  0.68 &  4.51 &&  4.58 &  2.75 & \textbf{ 3.67} \\
    90 &&  7.96 &  0.60 &  4.28 &&  5.19 &  1.65 & \textbf{ 3.42} \\
    100 &&  9.17 &  0.55 &  4.86 &&  5.33 &  2.04 & \textbf{ 3.68} \\
    \bottomrule
  \end{tabular}
\end{table}

\begin{table}
  \centering
  \caption{ Effect of System Population on Authentication Error (GTS-2P)}
  \label{tab:gts-sp-2p}
  
  \vspace{0.5em}
  \renewcommand{\arraystretch}{1.1}
  \begin{tabular}{r c *{3}{>{\centering\arraybackslash}p{0.07\linewidth}}
    c *{3}{>{\centering\arraybackslash}p{0.07\linewidth}}}
    \toprule
    \multirow{2}{*}{$n$ } && \multicolumn{3}{c}{MSM} &&
                                                                      \multicolumn{3}{c}{Bayesian Threshold}\\
    \cmidrule{3-5} \cmidrule{7-9}
                          && FRR & FAR & AER && FRR & FAR & AER \\
    \midrule
    10 &&  1.67 & 10.74 &  6.20 &&  5.00 &  1.88 & \textbf{ 3.44}\\
    20 &&  3.33 &  5.70 &  4.52 &&  5.00 &  1.97 & \textbf{ 3.49} \\
    30 &&  2.22 &  3.59 &  2.91 &&  2.78 &  1.84 & \textbf{ 2.31} \\
    40 &&  3.33 &  2.89 &  3.11 &&  3.75 &  1.98 & \textbf{ 2.87} \\
    50 &&  4.67 &  2.25 &  3.46 &&  4.67 &  2.01 & \textbf{ 3.34} \\
    60 &&  4.17 &  1.92 &  \textbf{3.04} &&  4.17 &  1.91 & \textbf{3.04} \\
    70 &&  5.00 &  1.64 &  \textbf{3.32} &&  4.76 &  1.87 & \textbf{3.32} \\
    80 &&  5.21 &  1.44 & \textbf{3.33} &&  4.79 &  1.90 &  3.35 \\
    90 &&  5.19 &  1.26 &  3.22 &&  4.63 &  1.70 & \textbf{ 3.17} \\
    100 &&  5.67 &  1.17 &  3.42 &&  4.83 &  1.90 & \textbf{ 3.37} \\
    \bottomrule
  \end{tabular}
\end{table}

The decrease in FRR is a change that can be observed in both MSM and BT systems in Figure~\ref{fig:gts-sp}. This is because when lesser members are registered in the system, there are lesser classes to discriminate as each member is viewed as a class in the discriminant analysis. The reduction in classes allows for a simpler boundary definition to discriminate each member.

In the GTS-2P, the two template representations employed are GEI for $S_1$ and the GEI-GTS for $S_2$. Since the adversary has two chances to enter into the system, the probability of him/her successfully impersonating a registered member is a little less than the FAR of $S_1$ and $S_2$ combined. The thresholds for $S_1$ and $S_2$ were both set to reflect $1\%$ FAR. Thus, the FAR of GTS-2P is almost $2\%$. Looking at the effect of $n$ in GTS-2P, the AER of BT and MSM is almost identical when $n$ ranges from 60 to 100 members. At population values lesser than 60,  the MSM began to show slightly lesser FRR. However, BT started to outperform MSM in terms of AER when the FAR of MSM became significantly large.

\subsection{Discussion}
\label{sec:discussion}

The sharp changes in sensitivity are mainly due to the compaction of classes by LDA-Bayes' model. The posterior probability based on the multivariate Gaussian model encompasses the interrelation between each feature. The boundary thus formed between each class is well-defined according to the features of the training set \citep{hamsici2008bayes}. The discriminating nature of LDA causes the model to map each probe instance to a region confined by these boundaries. An instance that pertains to an authorized entity shall be mapped to the region of the corresponding identity with a sufficiently high probability which minimizes FRR, and in the same time, brings the type 2 FAR close to zero. Consider the  case of an unauthorized instance claiming the identity of an authorized user; the probability $q$ that the model maps this instance to the region of the claimed identity is $1/n$ where $n$ is the total number of uniquely authorized subjects. For our case, $n=100$ which implies that $q=1\%$. This causes the separation of most unauthorized subjects from their claimed identity with a significantly large probability thus greatly reducing the FAR. Furthermore, if the instance is mapped to the region claimed, it should also be close enough to the other instances in that region for it to be considered genuine which brings the FAR closer to 0\% for a sufficiently large value of~$n$.

The GTS-2P has proved to perform significantly better than its single template counterpart for MSM (for $n<50$). However, as far as the BT method is concerned, it does not give a significant improvement in performance. Though in BT, the experimental error rates of GTS-2P is marginally smaller than that of the single template system, the difference is not compelling enough to establish its use over the simpler BT framework. 

The method proposed is neither a feature extraction technique nor a new classifier but a novel authentication paradigm making it flexible to cope with future advancements. This would mean that BT can be applied to any feature representation along with any generative probabilistic classifier for a given authentication task. The method can also be integrated to view estimators  to enable a view-invariant authentication operation.

\section{Summary}
\label{sec:mgb-summary}

The BT is a gait authentication framework designed to overcome the weakness of the MSM framework. BT uses probability as a threshold in contrast to the Euclidean distance threshold used by standard NN-based authentication systems. Once the features are transformed through CDA, the posterior probability of the Bayes' rule can act as the threshold for the decision function. The input instance is accepted as the identity claimed if it conforms to that identity with a sufficiently high posterior probability. This limit is determined by empirically set threshold, $\theta_p$.

Unlike MSM, the FAR of BT is not adversely affected by a reduction in the number of registered members, i.e., system population. The BT can be applied to any system population with a considerably low AER. The two pass variation can also be applied to BT similar to MSM, however, it does not provide a significant increase in performance.

The proposed framework BT can be applied to any type of gait authentication system regardless of the feature representation used. The experimental results show that it is far superior to the de facto NN distance method and the MSM framework.


\chapter{CONCLUSIONS AND FUTURE WORK}
\label{ch:conclusion}





To this end, the objective of this thesis has been to develop robust algorithms to employ gait in both soft and hard biometrics. Each of the proposed methods was implemented and evaluated with the widely used CASIA-B dataset, and compared with prescribed test conditions in literature and found to outperform their existing counterparts. The conclusions drawn from the study with regard to the proposed methods are as follows.

A novel approach for gait-based gender recognition was proposed called Pose-Based Voting. While existing approaches for this problem assess the entire gait as a single spatiotemporal instance, the proposed method views each frame of the video as individual instances. The existing gender recognition techniques make use of only a single gait cycle regardless of the length of the gait sequence fed as input. On the other hand, the accuracy of PBV increases with the length of the given gait sequence as it makes use of all of the frames containing the gait of the subject under observation. This technique also has the ability to cope with possible occlusion and partially observed gait cycles with minimal loss in overall prediction accuracy.

Two different feature representations were proposed in this dissertation for gender recognition through gait, viz., Elliptic Fourier descriptors and row-column sum vectors. EFD is a chain encoding technique used in computer graphics. It was selected for this problem as it can describe the shape of any closed contour with a set of equations. The coefficients of these equations can be encoded as a feature vector of a specified length for any input shape. The EFD performed well for gender recognition along the sagittal plane, but was not well enough to match the performance of the state-of-the-art solutions. However, a simpler row-column sum was found to be more effective in representing the horizontal and vertical structural features of the subject, thus outperforming the state of the art. 

The outcome of the PBV experiment showed that it is possible to neglect the temporal aspect of gait while analysing soft biometric information like gender. This was achieved with the aggregation of predictions. Individually, each frame by itself may not suffice to provide a precise classification of gender. However, with a set of frames, a better consensus can be arrived by taking prediction of each frame into account. The study has also proved that Bayes' rule could be used as an alternative to the widely used SVM when the feature set is reduced through LDA to provide identical results.

Many gait templates have been proposed over the past decade with each successive template addressing some weakness of the previously established templates for gait recognition. Recent advancements suggested to split templates to extract only the relevant features that are invariant to covariate factors such as clothing style and carrying conditions. However, such segmentation measures mostly relied on the predefined knowledge of the human body. To achieve the peak performance of any gait template, a novel segmentation technique, called the GTS, was proposed to find the optimal regions of the gait template. The boundary selection parameters of this method were handled by the genetic algorithm while CDA was used for feature extraction. The mask thus produced made the template robust to covariate factors.

In practical conditions, a person in an open space can be observed at any angle. The slope-based view-estimator designed in this thesis was found be more efficient while being computationally inexpensive. It is interesting to note that just two variables is sufficient to estimate the view of observation. This technique is found to be more accurate than the existing implementations with complicated steps for the same objective. Coupled with the view estimator, the GTS has proven to be a highly efficient view-invariant gait recognition model. 

The MSM is a novel framework designed for authentication to overcome the trade-off between FAR and FAR that exists in the de facto Euclidean distance-based threshold methods. The algorithm extends a gait recognition model to be used for authentication. To simply put, if the recognition system returns the identity as claimed by the subject, then he/she is said to be accepted as a genuine user. This technique relies on the strength of the recognition system and the number of genuine  members registered in the system, i.e., the system population. The FAR of this framework scales inversely with the system population and the FRR became equal to the CCR of the base recognition system. Though the MSM framework was proven to be successful, its performance is suppressed in lower system population.

To overcome the weakness of MSM, a new paradigm was designed to use the Bayesian posterior probability as a threshold in place of the NN-based Euclidean threshold. The resulting model exhibited a significant improvement in performance. Unlike the Euclidean threshold, the Bayes' rule with the multivariate Gaussian likelihood function takes the relationship between the variables considered. Moreover, Bayes' rule is found to be optimal after LDA. The CDA feature transformation applied to the gait templates ends with LDA. Hence, the Bayesian threshold performed far superior to the existing distance-based threshold methods for gait authentication. The method has also surpassed the limitation of the MSM concerning the system population.

The PBV gender recognition produces an ideal accuracy while all of the other aforementioned models -- GTS, MSM, and Bayesian thresholding -- are flexible. That is, they can work on top of any feature extraction procedure and any classifier. This would mean that even when a new template or classifier model is invented to surpass the ones discussed in this research, applying the proposed paradigms would only yield a much better performance than what it could yield with the existing framework. This characteristic would facilitate these algorithms to cope with future advancements.

The EFD is visually more expressive than the RCS. It could prove to be better than other feature representations when the subjects under consideration are observed at an oblique angle. All of the recent gait-based gender recognition algorithms published so far considers only the sagittal angle. Hence an in-depth study of gait at oblique angles ought to be conducted in future for both hard and soft biometrics to further gauge the capabilities of the EFD.

The Bayesian probability thresholding had proven itself to be far superior to the Euclidean distance-based thresholding in gait authentication. This theory can also be extended to other modes of authentication. Hence, we can test the BT method against other existing authentication algorithms such as handwritten signature verification, fingerprint verification, iris recognition, and face-based authentication.

Gait stratification is an unexplored area in gait analysis. It is a way of grouping people by means of their gait. The feature extraction for this problem would not be the same as that for a biometric. That is, the objective is not to extract the identifying characteristics but that of which can pertain to a group -- a set that blurs between hard and soft biometric features. That is, it is more general than hard biometrics and more specific than soft biometrics.  It could provide new insights on the relationship between gait and human behaviour and give rise to unseen inferences.


\cleardoublepage
\phantomsection 


\bibliographystyle{aubibstyle}
\cleardoublepage
\renewcommand\bibname{REFERENCES}

\newgeometry{tmargin=35mm,bmargin=30mm,lmargin=33mm,rmargin=33mm, bindingoffset=10mm}
 
\addcontentsline{toc}{chapter}{~~~~~~~~~REFERENCES}

\begin{spacing}{1}
\bibliography{GaitAnalysis} 
\end{spacing} 

\newpage
\clearpage
\restoregeometry
\normalsize
\chapter*{LIST OF PUBLICATIONS}
\addcontentsline{toc}{chapter}{\uppercase{~~~~~~~~~LIST OF PUBLICATIONS}}


\begin{sloppypar}
\noindent\textbf{Journal}
\begin{enumerate}[leftmargin=*]
\item Ebenezer Isaac, Easwarakumar K.S., and Joseph Isaac, 2017, `Urban Landcover Classification from Multispectral Image Data using Optimised AdaBoosted Random Forests,' Remote Sensing Letters, Taylor \& Francis, Volume 8, Issue 4, pp 350--359. Impact Factor: 1.532.

\item Ebenezer R.H.P. Isaac, Susan Elias, Srinivasan Rajagopalan, and K.S. Easwarakumar, 2017, `View-Invariant Gait Recognition Through Genetic Template Segmentation,' IEEE Signal Processing Letters. Volume 24, Issue 8, pp 1188--1192. Impact Factor: 2.528.

\item Ebenezer R.H.P. Isaac, Susan Elias, Srinivasan Rajagopalan, and K.S. Easwarakumar, `Gait Verification System through Multiperson Signature Matching for Unobtrusive Biometric Authentication,' revised version submitted to Journal of Signal Processing Systems, Springer. Impact Factor: 0.893.

\item Ebenezer R.H.P. Isaac, Susan Elias, Srinivasan Rajagopalan, and K.S. Easwarakumar, `Multiview Gait-Based Gender Classification through Pose-Based Voting,' revised version submitted to Pattern Recognition Letters, Elsevier. Impact Factor: 1.995

\item Ebenezer R.H.P. Isaac, Susan Elias, Srinivasan Rajagopalan, and K.S. Easwarakumar, `Template-Based Gait Authentication through Bayesian Thresholding,' submitted to IEEE/CAA Journal of Automatica Sinica.

\end{enumerate}
\end{sloppypar}


\addtocontents{toc}{\protect\newpage}
\addtocontents{lot}{\protect\newpage}
\addtocontents{lof}{\protect\newpage}

\end{document}